\documentclass[11pt, a4paper]{thuc3i}

\usepackage[numbers,sort&compress]{natbib}
\setheadertext{A Comprehensive Survey on Robot Manipulation}
\renewcommand{\today}{V2: 2026-08-15}

\usepackage[utf8]{inputenc} %
\usepackage[T1]{fontenc}    %
\usepackage{url}            %
\usepackage{tabularx}
\usepackage{makecell}
\usepackage{array,ragged2e,booktabs}
\newcolumntype{P}[1]{>{\RaggedRight\arraybackslash}p{#1}}
\usepackage{longtable}
\usepackage{amsfonts}       %
\usepackage{amsthm}         %

\usepackage{nicefrac}       %
\usepackage{microtype}      %
\usepackage{xcolor}         %
\usepackage{graphicx}
\usepackage{algorithm}
\usepackage{algorithmicx}
\usepackage{algpseudocode}

\definecolor{darkblue}{rgb}{0, 0, 0.5}
\hypersetup{colorlinks=true, citecolor=darkblue, linkcolor=darkblue, urlcolor=darkblue}

\usepackage{xurl}
\usepackage{soul}
\usepackage{amsmath}
\usepackage{subcaption}
\usepackage{graphicx}
\usepackage{changes}
\usepackage{amsmath,mathtools}
\usepackage{changes}
\usepackage{tocloft}
\usepackage{lipsum}
\usepackage{colortbl}
\usepackage{wrapfig}
\usepackage{xspace}
\usepackage{multirow}
\usepackage{enumitem}
\usepackage{pifont}
\usepackage{listings}
\usepackage{fontawesome5} 
\usepackage{caption}
\usepackage{makecell}
\usepackage{tabularx}
\usepackage{afterpage}

\usepackage[edges]{forest}
\usepackage[normalem]{ulem}
\usepackage{caption}
\usepackage{CJKutf8}
\usepackage{bbding}
\usepackage[most,skins,theorems]{tcolorbox}

\newcommand{\supref}[2]{\textsuperscript{\href{#1}{#2}}}

\usepackage{setspace}

\usepackage{booktabs}
\usepackage{multirow}
\usepackage{amssymb}
\usepackage{pifont}

\usepackage[edges]{forest}
\usepackage{tikz}
\usepackage{graphicx}
\definecolor{hidden-red}{RGB}{205, 44, 36}
\definecolor{hidden-blue}{RGB}{194,232,247}
\definecolor{hidden-orange}{RGB}{243,202,120}
\definecolor{hidden-green}{RGB}{34,139,34}
\definecolor{hidden-pink}{RGB}{255,245,247}
\definecolor{hidden-black}{RGB}{20,68,106}
\definecolor{purple}{RGB}{144,153,196}
\definecolor{yellow}{RGB}{255,228,123}
\definecolor{hidden-yellow}{RGB}{255,248,203}
\definecolor{tkcolor}{RGB}{224,223,255}
\definecolor{darkblue}{rgb}{0, 0.40, 0.75}

\title{\vspace{-5pt}\raisebox{-0.5em}{\includegraphics[height=1.5em]{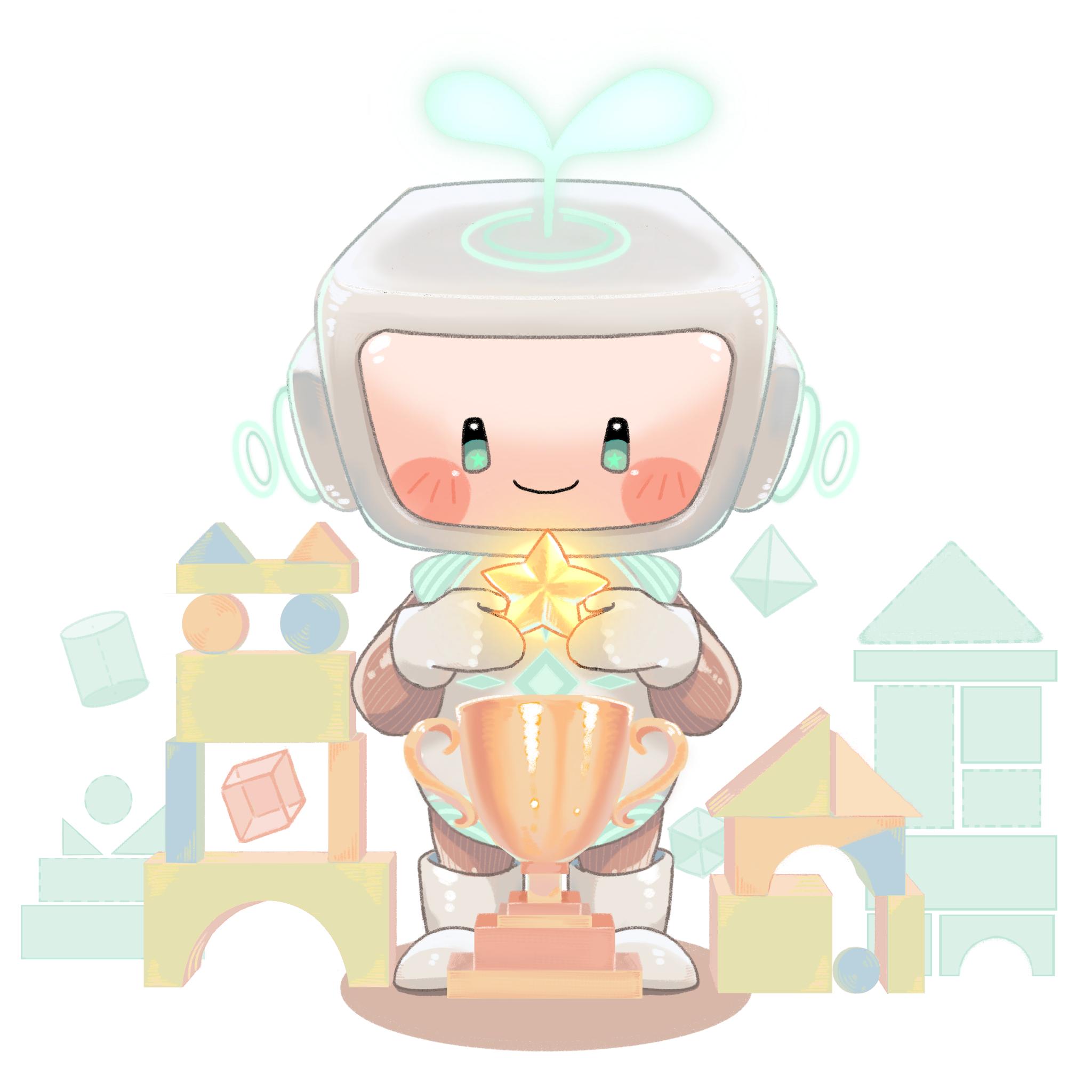}}\ \ Towards a Unified Understanding of Robot Manipulation: A Comprehensive Survey}

\author{
  Shuanghao Bai$^{1*\dagger}$ \quad Wenxuan Song$^{2*}$ \quad Jiayi Chen$^{2*}$ \quad  Yuheng Ji$^{3*}$ \quad Zhide Zhong$^{2*}$ \quad
  Jin Yang$^{1*}$ \quad Han Zhao$^{4,5*}$ \quad Wanqi Zhou$^{1*}$ \quad Wei Zhao$^{4,5*}$ \quad Zhe Li$^{6*}$ \quad
  Pengxiang Ding$^{4,5}$ \quad Cheng Chi$^{7}$ \quad Haoang Li$^{2}$ \quad Chang Xu$^{6}$ \quad Xiaolong Zheng$^{3}$ \quad Donglin Wang$^{4}$ \quad Shanghang Zhang$^{7,8}$\Envelope \quad
  Badong Chen$^{1}$\Envelope 
  \vspace{1mm} \\
    $^{1}$ Xi'an Jiaotong University,
    $^{2}$ Hong Kong University of Science and Technology (Guangzhou),
    $^3$ Chinese Academy of Sciences,\\ 
    $^4$ Westlake University, 
    $^5$ Zhejiang University,
    $^6$ University of Sydney, 
    $^7$ BAAI, 
    $^8$ Peking University
    \vspace{1mm} \\
    {\textbf{$^\dagger$ Project Lead}~~ $^*$ \textbf{Core Contributors}~~  \textbf{\Envelope~ Corresponding Authors}} \vspace{1mm} \\
  \faEnvelope[regular]~\texttt{chenbd@mail.xjtu.edu.cn} \quad \faGithub~\href{https://github.com/BaiShuanghao/Awesome-Robotics-Manipulation}{\texttt{Awesome-Robotics-Manipulation}} 
}

\begin{abstract}
Embodied intelligence has witnessed remarkable progress in recent years, driven by advances in computer vision, natural language processing, and the rise of large-scale multimodal models.
Among its core challenges, robot manipulation stands out as a fundamental yet intricate problem, requiring the seamless integration of perception, planning, and control to enable interaction within diverse and unstructured environments.
This survey presents a comprehensive overview of robotic manipulation, encompassing foundational background, task-organized benchmarks and datasets, and a unified taxonomy of existing methods.
We extend the classical division between high-level planning and low-level action modeling by broadening high-level planning to include language, code, motion, affordance, and 3D representations, while introducing a new taxonomy of low-level learning-based action modeling grounded in training paradigms such as input modeling, latent learning, and policy learning.
Furthermore, we provide the first dedicated taxonomy of key bottlenecks, focusing on data collection, utilization, and generalization, and conclude with an extensive review of real-world applications.
Compared with prior surveys, our work offers both a broader scope and deeper insight, serving as an accessible roadmap for newcomers and a structured reference for experienced researchers.
\end{abstract}

\begin{document}

\maketitle

\begin{figure}[h]
\centering
\includegraphics[width=\textwidth]{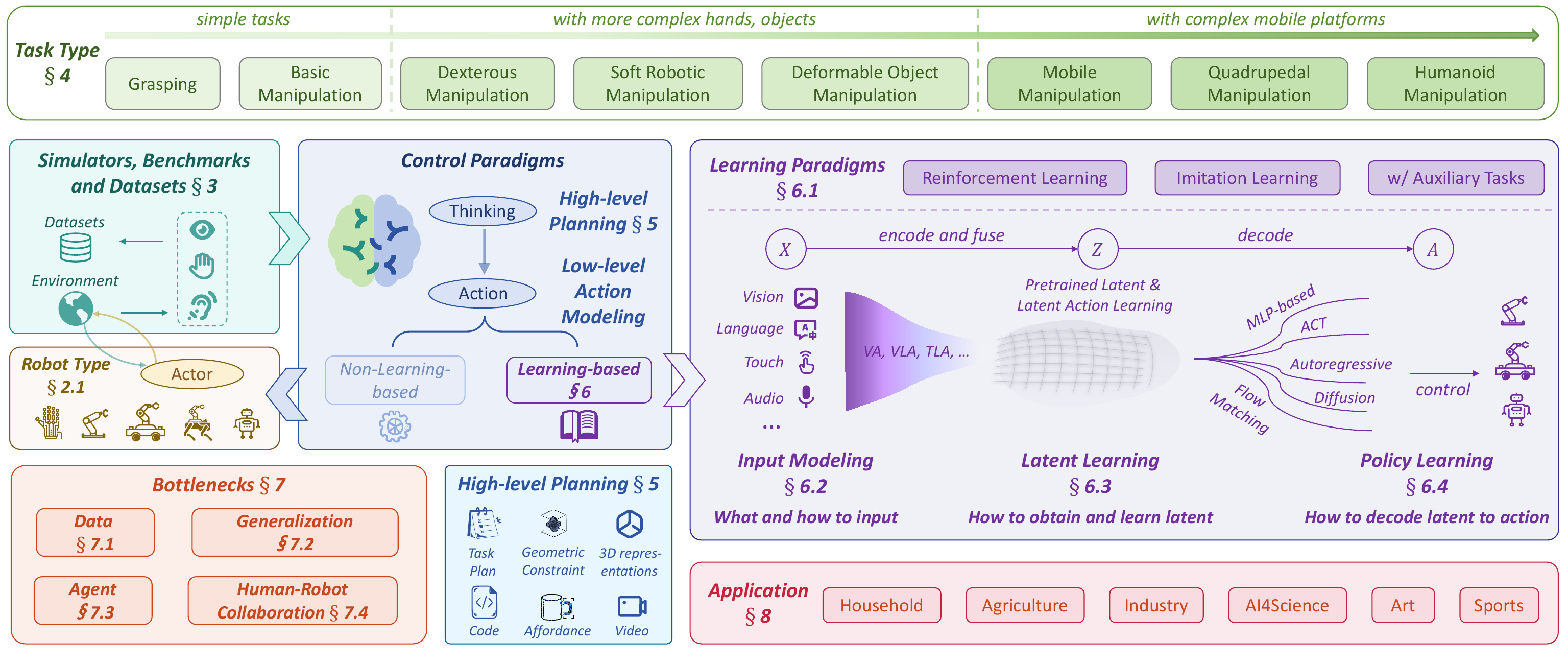}  
\caption{Overview of the survey. We first provide a broad introduction to existing benchmarks, datasets, and manipulation tasks, followed by an extensive review of representative methods with a particular focus on learning-based action modeling. We then discuss two fundamental challenges, namely data and generalization, and conclude with an overview of their diverse applications.}
\label{fig: summary}
\end{figure}

\newpage
\begingroup
\setlength{\baselineskip}{1.25\baselineskip}
\tableofcontents
\endgroup
    
\newpage


\section{Introduction}
\label{sec:intro}

In recent years, embodied intelligence has attracted increasing attention, largely driven by advances in computer vision and natural language processing, particularly the success of large-scale models. These developments have not only demonstrated remarkable machine intelligence but also offered a glimpse into the potential of artificial general intelligence (AGI). Building on this progress, large-scale language and multimodal models~\cite{brown2020language,touvron2023llama,liu2023visual} have accelerated the development and deployment of robotic systems by enhancing perceptual and semantic understanding, enabling operation in unstructured environments, and supporting natural-language task specification. Their zero-shot and few-shot generalization capabilities improve the adaptability of robotic systems, while multimodal interaction enhances usability and reduces deployment barriers in real-world scenarios.

Robot manipulation is a core and extensively studied problem in embodied intelligence, referring to a robot's ability to perceive its environment, plan task execution, generate appropriate actions, and control its effectors to physically interact with and modify the environment, such as by grasping, moving, or using objects. Its development spans classical rule-based and non-learning methods~\cite{espiau1991new,lavalle1998rapidly,hutchinson2002tutorial,kalakrishnan2011stomp} from the late twentieth century through the 2010s, followed by deep learning-based approaches~\cite{levine2016end,pinto2016supersizing}, the widespread adoption of imitation learning (IL) and reinforcement learning (RL)~\cite{duan2017one,rajeswaran2018learning}, and, more recently, the integration of large language and vision-language models into IL and RL frameworks~\cite{zitkovich2023rt,kim2025openvla}. In this survey, non-learning methods are primarily introduced as methodological background, while the main discussion focuses on data-driven and learning-based approaches.

\subsection{Survey Scope and Key Research Questions}

In this survey, we aim to provide newcomers with a concise roadmap to the development, tasks, and methods of robot manipulation, while offering experienced researchers a structured reference and a broader perspective on the field. To this end, we organize the survey around the following research questions:

\textbf{1. What benchmarks and task categories define robot manipulation today?}
We review the current landscape of benchmarks in Section~\ref{sec:benchmarks} and organize representative manipulation problems into major task categories in Section~\ref{sec:manipulation_tasks}.

\textbf{2. What methods have been proposed to address these manipulation tasks?}
Beyond basic manipulation, Section~\ref{sec:manipulation_tasks} reviews representative approaches for dexterous, deformable-object, mobile, quadrupedal, and humanoid manipulation. For these categories, we briefly introduce classical non-learning methods and place greater emphasis on learning-based approaches, including RL, IL, VLA models, and strategy-augmented methods, while highlighting the distinct methodological and embodiment-specific challenges of each domain.

Basic manipulation, in contrast, represents the most extensively studied setting and is supported by a substantially richer body of literature. We therefore provide a dedicated analysis in Sections~\ref{sec:high_level_planning} and~\ref{sec:low_level_action_modeling}, distinguishing between high-level planning methods that produce task-level plans or intermediate planning artifacts and low-level action-modeling methods that generate executable robot actions from observations and task specifications. Actuation-level control, which converts these actions into joint-level position, velocity, or torque commands, is introduced separately as part of the foundational background. Although this taxonomy is developed primarily in the context of basic manipulation, its underlying distinction between planning, action generation, and physical execution is also applicable to other manipulation settings.

\textbf{3. What are the current bottlenecks in robot manipulation?}
We identify data collection and utilization, together with generalization, as two central challenges. Section~\ref{subsec:data} reviews the evolution of robot-data collection and examines strategies for improving the efficiency and effectiveness of data utilization during training. Section~\ref{subsec:generalization_tasks} analyzes the major forms of generalization in robot manipulation and summarizes the corresponding methodological strategies.

\textbf{4. What are the practical applications of manipulation techniques beyond research?}
We survey how advances in robot manipulation are being deployed across a wide range of real-world application domains in Section~\ref{sec:applications}.

\subsection{Comparison with Previous Surveys and Contributions}

\textbf{First, compared with prior surveys that are limited in scope, our work provides a comprehensive and systematic overview of robot manipulation.}
Existing surveys typically adopt narrower perspectives. Some focus on particular task domains, such as dexterous manipulation~\cite{an2025dexterous,li2025developments}, deformable-object manipulation~\cite{blanco2024t}, mobile manipulation~\cite{thakar2023survey}, or humanoid manipulation~\cite{gu2025humanoid}. Others emphasize methodological paradigms that recur across different tasks, including vision-language-action models~\cite{ma2026survey,zhong2025survey,xiang2025parallels,li2025survey}, diffusion models~\cite{wolf2025diffusion}, and generative approaches~\cite{zhang2025generative}. A further group concentrates on specific methodological concepts, such as language-conditioned learning~\cite{zhou2023language} or object-centric representations~\cite{zheng2025survey}. Several broader embodied-intelligence surveys cover a wide range of topics but treat manipulation only as one subsection, providing insufficient depth for a systematic understanding of the field~\cite{liu2025aligning,wong2025survey}.

\textbf{Second, our survey introduces a unified taxonomy that covers the robot-manipulation pipeline more extensively than existing categorizations.}
The resulting organization provides an accessible blueprint for newcomers while offering experienced researchers a structured perspective on the relationships among different problem formulations and methodological paradigms.

Specifically, we provide a comprehensive background in Section~\ref{sec:background}, covering hardware embodiments, actuation-level control paradigms, and foundational robotic models. We introduce benchmarks and datasets organized by manipulation-task category in Section~\ref{sec:benchmarks}, present a systematic overview of representative manipulation tasks, and develop a refined methodological taxonomy in Sections~\ref{sec:high_level_planning} and~\ref{sec:low_level_action_modeling}. While prior surveys often distinguish broadly between high-level planning and low-level control, we refine this division by separating high-level planning, low-level action modeling, and actuation-level control according to the system interface and the form of the model output. We broaden high-level planning in Section~\ref{sec:high_level_planning} to encompass planning representations expressed through language, code, motion, affordances, and 3D scene structures. For low-level action modeling in Section~\ref{sec:low_level_action_modeling}, we propose a taxonomy grounded in training paradigms and further organize existing methods according to learning strategy, input modeling, latent learning, and policy learning.

In addition, we provide a detailed analysis of current bottlenecks in robot manipulation in Section~\ref{sec:bottlenecks} and introduce a dedicated taxonomy covering data collection, data utilization, and generalization. Finally, Section~\ref{sec:applications} presents a broader and more systematic review of real-world applications than those provided in previous surveys.

\textbf{Finally, based on these contributions and recent developments in the field, we identify emerging research trends and outline four promising directions for future work.}
The first concerns the development of a genuine robot brain, namely a general-purpose architecture that integrates broad perception, reasoning, action-generation, and control capabilities. The second addresses the data bottleneck, as current robot learning has not yet exhibited a reliable scaling law comparable to those observed in language and vision, largely because of the high cost of physical data acquisition and the limitations of simulation. The third concerns perception, particularly the need for richer multimodal sensing and more reliable interaction with deformable, articulated, and otherwise physically complex objects. The fourth emphasizes safe human--robot coexistence, which remains essential for the large-scale deployment of robotic systems in real-world environments.
\section{Background}
\label{sec:background}

In this section, we first introduce the hardware types commonly used in robotic manipulation (Section~\ref{subsec:hardware}). We then outline the main categories of control strategies, namely non-learning-based and learning-based approaches (Section~\ref{subsec:action_modeling}). Next, we review the robotic models widely adopted for learning-based control (Section~\ref{subsec:robotics_models}) and discuss the evaluation protocols used to assess the robotic models within these frameworks (Section~\ref{subsec:eval}).

\subsection{Hardware Platforms for Manipulation}
\label{subsec:hardware}

\begin{figure}[!t]
\centering
\includegraphics[width=0.95\linewidth]{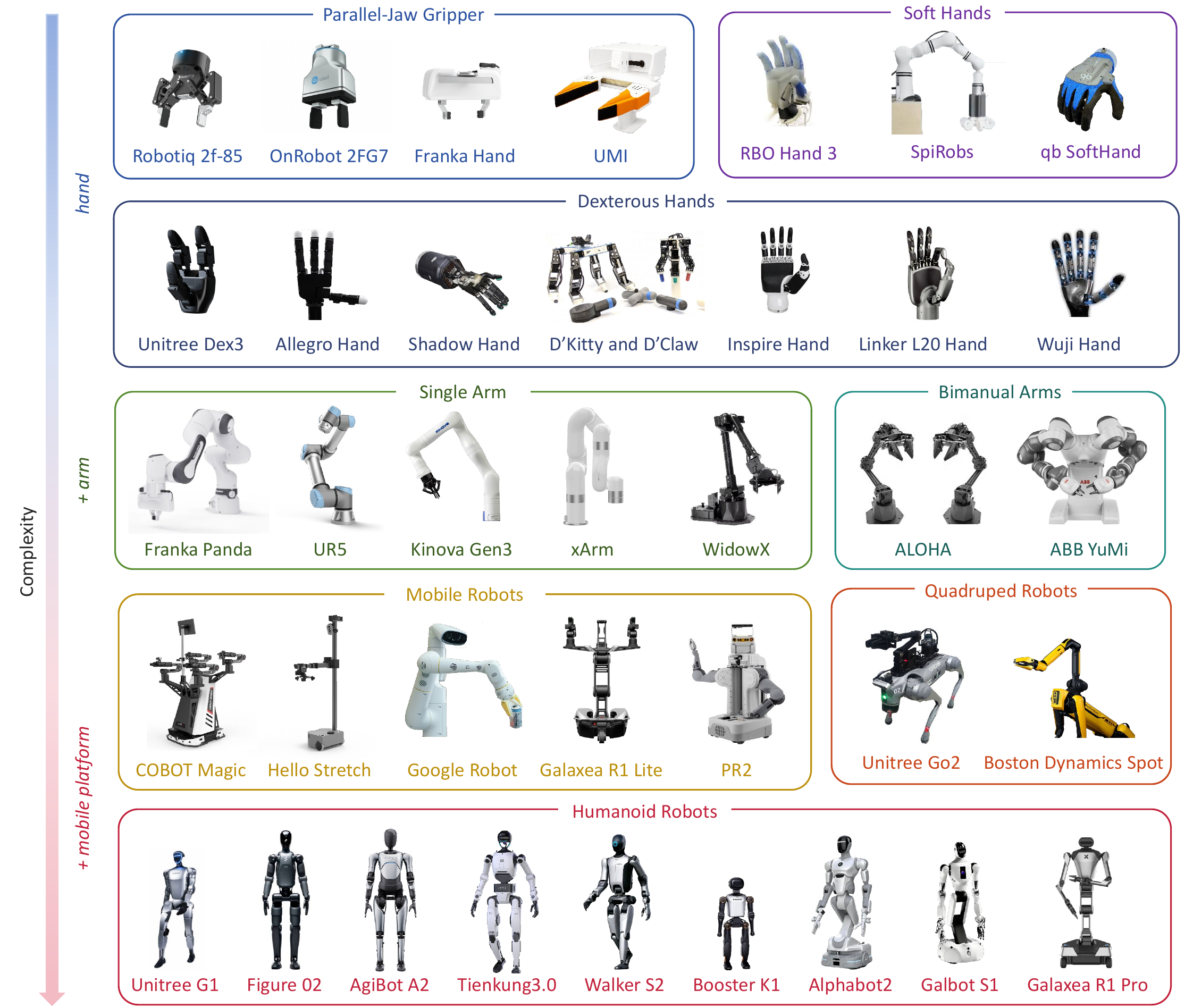}  
\caption{Overview of hardware platforms.}
\label{fig: hardware}
\end{figure}

Before introducing manipulation tasks and their corresponding methodologies, it is important to first understand the hardware systems that enable these operations. Robotic manipulation can be achieved through various embodiments composed of fundamental components such as hands, arms, and mobile platforms. Different combinations of these components define specific embodiments and their functional capabilities. For example, pairing a parallel-jaw gripper with a Franka Panda arm enables basic manipulation tasks such as pick-and-place or insertion, while integrating a Unitree G1 humanoid platform with a dexterous hand allows for humanoid-level manipulation that demands greater dexterity and coordination. We summarize the commonly used robotic hardware types in current research in Figure~\ref{fig: hardware}.

\noindent \textbf{Parallel-Jaw Grippers.}
Parallel-jaw grippers are among the most widely adopted end-effectors in robotic manipulation because their low-dimensional actuation, repeatable contact geometry, and reliable force transmission simplify control, demonstration collection, and policy learning. Representative commercial platforms include the Robotiq 2F-85\supref{https://robotiq.com/products/adaptive-grippers}{1}, OnRobot 2FG7\supref{https://onrobot.com/en/products/2fg7}{2}, and Franka Hand\supref{https://franka.de/de/franka-hand}{3}. Research-oriented designs further extend this basic morphology, including DexWrist~\cite{peticco2025dexwrist}, which augments a gripper with additional wrist dexterity, and UMI~\cite{chi2024universal}, which integrates a portable gripper and visual sensing interface for scalable real-world demonstration collection. Although these grippers provide limited in-hand dexterity, their compact action spaces and robust grasping behavior make them effective for pick-and-place, insertion, and large-scale imitation-learning experiments.

\noindent \textbf{Dexterous Hands.}
Dexterous hands employ multiple articulated fingers and higher-dimensional actuation to support contact-rich manipulation beyond simple opening and closing. Representative multi-finger end-effectors include the Robotiq 3-Finger Adaptive Robot Gripper\supref{https://robotiq.com/support}{4}, Allegro Hand\supref{https://www.allegrohand.com/}{5}, Shadow Dexterous Hand\supref{https://shadowrobot.com/}{6}, Unitree Dex3-1\supref{https://www.unitree.com/Dex3-1/}{7}, D'Claw and D'Kitty from the ROBEL platform~\cite{ahn2020robel}, Inspire Hand\supref{https://en.inspire-robots.com/the-dexterous-hands-2}{8}, Linker Hand L20\supref{https://linkerbot.cn/index}{9}, and Wuji Hand\supref{https://wuji.tech/en/hand}{10}, together with other learning-oriented designs~\cite{shaw2024demonstrating,wan2025rapid}. Their articulated structures enable object reorientation, in-hand manipulation, tool use, and coordinated multi-contact interaction. However, the increased number of controllable joints also enlarges the action space and places substantially greater demands on sensing, calibration, demonstration quality, and closed-loop policy learning.

\noindent \textbf{Soft Hands.}
Soft hands exploit material compliance, underactuated transmission, or deformable actuation to conform passively to object geometry. Representative platforms include the RBO Hand 3~\cite{puhlmann2022rbo}, Festo BionicSoftHand\supref{https://www.festo.com.cn/cn/zh/e/about-festo/research-and-development/bionic-learning-network/bionic-grippers-and-soft-robots/bionicsofthand-id_68106}{11}, SpiRobs~\cite{wang2025spirobs}, and qb SoftHand\supref{https://qbrobotics.com/product/qb-softhand-research/}{12}. In contrast to rigid dexterous hands, these systems partially transfer the burden of contact adaptation from explicit control to mechanical compliance, improving grasp robustness and interaction safety under geometric uncertainty. This adaptability is obtained at the cost of less precise kinematic modeling and more difficult estimation of contact states and deformation-dependent dynamics.

\noindent \textbf{Single Arm.}
Fixed-base single-arm manipulators remain the predominant platforms for learning-based manipulation because they provide accurate and repeatable motion within a well-defined workspace. Commonly used systems include the KUKA LBR iiwa\supref{https://www.kuka.com/en-de/products/robot-systems/industrial-robots/lbr-iiwa}{13}, Franka Emika Panda~\cite{haddadin2024franka}\supref{https://franka.de/}{14}, UR5 and UR10\supref{https://www.universal-robots.com/}{15}, Kinova Gen3\supref{https://www.kinovarobotics.com/}{16}, xArm6 and xArm7\supref{https://www.ufactory.us/}{17}, and WidowX\supref{https://www.interbotix.com/widowxrobotarm}{18}. These platforms typically provide six or seven degrees of freedom and are widely employed for behavior cloning, reinforcement learning, visuomotor control, and sim-to-real evaluation. More recently, low-cost and reproducible systems supported by open-source ecosystems such as LeRobot\supref{https://github.com/huggingface/lerobot}{19} have lowered the barrier to physical experimentation and facilitated standardized data collection across research groups.

\noindent \textbf{Bimanual Arms.}
Bimanual systems extend the workspace and contact capabilities of single-arm platforms by coordinating two manipulators. Representative systems include dual Franka Panda configurations\supref{https://franka.de/}{20}, ALOHA~\cite{fu2025mobile}, and ABB YuMi\supref{https://new.abb.com/products/robotics/robots/collaborative-robots/yumi}{21}. They support tasks such as bilateral assembly, object handover, deformable-object manipulation, container opening, and coordinated tool use. Nevertheless, bimanual manipulation introduces additional challenges associated with temporal synchronization, inter-arm collision avoidance, assignment of complementary roles, and learning coupled actions whose effectiveness depends on the coordinated behavior of both arms.

\noindent \textbf{Mobile Manipulators.}
Mobile manipulators combine one or more arms with a movable base, thereby extending manipulation beyond the bounded workspace of a fixed-base robot. Representative platforms include COBOT Magic\supref{https://www.agilex.ai/page/690aef2d5e78cfa260412cb5}{22}, Hello Robot Stretch 3\supref{https://hello-robot.com/}{23}, the Google mobile manipulation platform\supref{https://deepmind.google/blog/scaling-up-learning-across-many-different-robot-types/}{24}, Galaxea R1 Lite\supref{https://galaxea-ai.com/cn/products/R1}{25}, PR2\supref{https://github.com/pr2}{26}, and TIAGo\supref{https://pal-robotics.com/robot/tiago/}{27}. By coupling navigation, base positioning, torso adjustment, and arm motion, these platforms can perform object fetching, household assistance, long-horizon rearrangement, and human--robot interaction across extended environments. Their operational range also introduces partial observability, accumulated localization error, dynamic obstacles, and the need to coordinate manipulation and locomotion within a unified decision process.

\noindent \textbf{Quadruped Robots.}
Quadruped platforms provide agile locomotion over uneven terrain and can be equipped with articulated arms or specialized end-effectors to support mobile manipulation. Representative systems include Unitree Go2, B1, and Aliengo\supref{https://www.unitree.com/}{28}, together with Boston Dynamics Spot\supref{https://bostondynamics.com/products/spot/}{29}. Their legged morphology enables operation in environments that are inaccessible to conventional wheeled platforms, including stairs, rough terrain, and infrastructure-inspection settings. Manipulation on quadrupeds nevertheless requires tight coupling among locomotion, body stabilization, arm motion, and contact-force regulation, particularly when manipulation forces disturb the support configuration of the robot.

\noindent \textbf{Humanoid Robots.}
Humanoid and anthropomorphic whole-body platforms are designed to operate in spaces, use tools, and interact with objects originally configured for humans. Representative systems include Tesla Optimus Gen 2\supref{https://www.tesla.com/AI}{30}, Boston Dynamics Atlas\supref{https://bostondynamics.com/atlas/}{31}, Figure 02\supref{https://www.figure.ai/}{32}, 1X NEO\supref{https://www.1x.tech/neo}{33}, Unitree G1\supref{https://www.unitree.com/g1/}{34}, AgiBot A2\supref{https://www.agibot.com/products/A2}{35}, Tiangong 3.0\supref{https://kw.beijing.gov.cn/xwdt/kcyx/xwdtkjqy/202602/t20260211_4507734.html}{36}, UBTECH Walker S2\supref{https://www.ubtrobot.com/cn/humanoid/products/walker-s2}{37}, Booster K1\supref{https://www.booster.tech/booster-k1/}{38}, AlphaBot 2\supref{https://ai2robotics.com/}{39}, Galbot S1\supref{https://www.galbot.com/s1}{40}, and Galaxea R1 Pro\supref{https://galaxea-ai.com/cn/products/R1}{41}. This category increasingly encompasses both bipedal humanoids and wheeled anthropomorphic systems that retain a human-scale torso and manipulation workspace. Compared with arm-centered platforms, these embodiments expand the action space to include locomotion, balance, torso motion, whole-body contact, and coordinated use of multiple limbs. They therefore provide a promising substrate for general-purpose embodied intelligence while introducing substantial challenges in data efficiency, safety, state estimation, hierarchical control, and reproducible evaluation.

\subsection{Low-Level Action Modeling: Non-Learning and Learning-Based Paradigms}
\label{subsec:action_modeling}

In this survey, low-level action modeling broadly refers to methods that transform task goals, environment states, or observations into executable robot paths, trajectories, actions, or action sequences. This definition is based on the exposed system interface rather than the internal computational mechanism. Accordingly, low-level actions may be constructed analytically, obtained through sampling or optimization, or learned from data. They are subsequently translated into joint-level position, velocity, force, or torque commands by actuation-level controllers.

Similar to how humans may follow explicit rules, such as stopping at a red light, or acquire adaptive skills through repeated practice, such as learning to ride a bicycle, low-level robot actions can be generated through either classical non-learning methods or data-driven learning approaches. Non-learning methods offer interpretability, predictable behavior, and explicit constraint handling in well-defined settings, whereas learning-based methods provide greater adaptability and generalization in complex or uncertain environments.

\subsubsection{Non-Learning-Based Action Modeling}

Non-learning-based action modeling generates executable paths, trajectories, or control sequences using analytical construction, search, optimization, and model-based control without learning the underlying observation-to-action mapping from data. We include these methods as foundational counterparts to learning-based action models and because they are frequently combined with learned perception, planning, or policy modules. However, they are not the primary focus of the methodological taxonomy developed in this survey.

\noindent \textbf{Interpolation-Based Planning.}
Classical manipulators often employ interpolation-based planning~\cite{duysinx2006introduction,tian2004effective}, in which smooth joint-space trajectories are generated, typically offline, by fitting polynomial curves, such as cubic splines, between predefined start and goal states. These methods are computationally lightweight and straightforward to deploy, making them prevalent in repetitive and well-structured industrial tasks. However, their dependence on predefined boundary conditions provides limited adaptability to dynamic or uncertain environments, constraining their applicability in unstructured settings.

\noindent \textbf{Sampling-Based Planning.}
Sampling-based planners~\cite{elbanhawi2014sampling,kim2006sampling,schmitt2017optimal,huh2018constrained} construct feasible paths by sampling configuration space and incrementally building a graph or tree of collision-free states, rather than explicitly representing free space or solving a large global optimization problem. Canonical algorithms include Rapidly-exploring Random Trees (RRT)~\cite{kim2006sampling,huh2018constrained} and Probabilistic Roadmaps (PRM)~\cite{schmitt2017optimal}, together with asymptotically optimal variants such as RRT*. These methods scale effectively to high-dimensional configuration spaces, but the resulting paths may be suboptimal or non-smooth and therefore often require subsequent smoothing, time parameterization, and trajectory tracking.

\noindent \textbf{Optimization-Based Planning.}
Optimization-based planners formulate low-level action generation as a constrained optimization problem over paths, trajectories, or control sequences. They minimize task-specific objectives while enforcing constraints related to robot kinematics, dynamics, collision avoidance, smoothness, and task completion.

\textit{Offline optimization-based planners} solve the trajectory-generation problem before execution by optimizing the complete motion as a batch. Representative methods include CHOMP (Covariant Hamiltonian Optimization for Motion Planning)~\cite{zucker2013chomp}, TrajOpt (Trajectory Optimization)~\cite{dai2018improving}, and STOMP (Stochastic Trajectory Optimization for Motion Planning)~\cite{kalakrishnan2011stomp}. These approaches can produce smooth and collision-free trajectories, but they often require substantial computation and cannot directly adapt to unexpected changes encountered during execution.

\textit{Online optimization-based planners}, such as Model Predictive Control (MPC)~\cite{kouvaritakis2016model,bhardwaj2022storm}, repeatedly solve a finite-horizon optimization problem using the current state as the initial condition. MPC leverages a predictive model to estimate future states and generate executable control actions in a receding-horizon manner. This repeated re-optimization enables adaptation to disturbances and dynamic environments. In the taxonomy adopted here, MPC is included as a low-level action-modeling method because its exposed output is a finite sequence of executable actions or control inputs, whereas the downstream servo loop that realizes these commands is treated as actuation-level control.

\subsubsection{Learning-Based Action Modeling}

Learning-based action modeling aims to learn a policy that maps environment observations, robot states, and optional task specifications to executable actions or action sequences. Depending on the source of supervision and the interaction setting, representative paradigms include reinforcement learning and imitation learning.

Sequential decision-making for learning-based action modeling is commonly formulated as a Markov Decision Process (MDP), which provides a formal framework for modeling interactions between a robot and its environment under uncertainty. An MDP is defined as a five-tuple:
\begin{equation}
\mathcal{M}=\langle\mathcal{S},\mathcal{A},\mathcal{P},\mathcal{R},\gamma\rangle,
\end{equation}

\noindent where
\begin{itemize}
  \item $\mathcal{S}$ denotes the state space containing all possible environment and robot states;
  \item $\mathcal{A}$ denotes the action space containing the executable actions available to the robot;
  \item $\mathcal{P}(s' \mid s,a)$ denotes the transition probability of reaching state $s'$ after executing action $a$ in state $s$;
  \item $\mathcal{R}:\mathcal{S}\times\mathcal{A}\rightarrow\mathbb{R}$ denotes the reward function specifying the expected immediate reward obtained by executing action $a$ in state $s$;
  \item $\gamma\in[0,1]$ denotes the discount factor that balances short-term and long-term rewards.
\end{itemize}

This formulation provides a unified basis for learning policies that map states or observations to executable actions and underlies a wide range of robot-learning approaches, including reinforcement learning and imitation learning.

\paragraph{Reinforcement Learning (RL).}
The objective of reinforcement learning is to learn an optimal policy $\pi^{*}:\mathcal{S}\rightarrow\mathcal{A}$ that maximizes the expected cumulative discounted reward, also referred to as the return~\cite{han2023survey}:
\begin{equation}
\pi^*=\arg\max_{\pi}\mathbb{E}_{\pi}\left[\sum_{t=0}^{\infty}\gamma^t R(s_t,a_t)\right],
\qquad
a_t\sim\pi(\cdot\mid s_t).
\end{equation}

This objective provides the theoretical foundation for a broad family of algorithms, including Q-learning~\cite{watkins1992q,haarnoja2018composable}, policy-gradient methods~\cite{sutton1999policy,peters2006policy}, and actor-critic approaches~\cite{grondman2012survey,haarnoja2018soft}. According to how interaction data are obtained, RL methods can be categorized into offline RL~\cite{levine2020offline,chen2021decision}, which learns entirely from a fixed pre-collected dataset; online RL~\cite{schulman2017proximal,haarnoja2018soft}, which continually interacts with the environment to collect new experience; and offline-to-online RL~\cite{lee2022offline,nakamoto2023cal}, which initializes learning from offline data and subsequently improves the policy through online interaction.

\paragraph{Imitation Learning (IL).}
Imitation learning acquires action policies from expert demonstrations and is commonly divided into Behavior Cloning (BC), Inverse Reinforcement Learning (IRL), and Generative Adversarial Imitation Learning (GAIL).

\textbf{BC} formulates action modeling as supervised learning from expert state-action pairs, without requiring an explicitly specified reward function~\cite{torabi2018behavioral,bai2025rethinking}. Let $s_t$ denote the system state at timestep $t$, which may include the robot proprioceptive state, visual observations, language instructions, and observation histories. A policy $\pi$ maps these inputs to an action or action sequence. Given an expert demonstration dataset $\mathcal{D}_e$, the optimization objective can be written as:
\begin{equation}
\begin{aligned}
\label{func:vanilla_bc}
\pi^*=\arg\min_{\pi}\mathbb{E}_{(s_t,\hat{a}_t)\sim\mathcal{D}_e}
\left[
\mathcal{L}\left(\pi(s_t),\hat{a}_t\right)
\right],
\end{aligned}
\end{equation}
where $\hat{a}_t$ denotes the expert action label. In conventional BC, $\mathcal{L}$ is typically instantiated as cross-entropy loss for discrete action spaces and as mean squared error or $L_1$ loss for continuous action spaces.

\textbf{IRL} seeks to recover a reward function that explains the expert's behavior rather than directly imitating expert actions. By inferring the objective underlying expert demonstrations, the learned policy may generalize more effectively to unseen states and related tasks~\cite{arora2021survey,das2021model}. IRL assumes that the expert behaves approximately optimally with respect to an unknown reward function $R(s,a)$. Given expert demonstrations $\mathcal{D}_e$, the objective is to identify a reward function $R^*$ under which the expert policy $\pi_e$ achieves a higher expected return than alternative policies:
\begin{equation}
\begin{aligned}
\label{func:irl}
R^*=\arg\max_R
\mathbb{E}_{\pi_e}\left[\sum_{t=0}^{\infty}\gamma^tR(s_t,a_t)\right]
-
\mathbb{E}_{\pi}\left[\sum_{t=0}^{\infty}\gamma^tR(s_t,a_t)\right],
\end{aligned}
\end{equation}
where $\pi$ denotes a policy induced by the learned reward and $\gamma\in[0,1)$ is the discount factor. After estimating $R^*$, a reinforcement-learning algorithm is employed to obtain a policy $\pi^*$ that maximizes the expected return under the inferred reward. This separation between reward inference and policy optimization enables the model to capture objectives implicit in expert behavior rather than merely reproducing individual demonstrated actions.

\textbf{GAIL} formulates imitation learning as distribution matching between the discounted state-action occupancy measures induced by the learner and the expert, thereby avoiding explicit recovery of a task reward. The discounted occupancy measure induced by a policy $\pi$ is defined as:
\begin{equation}
\label{eq:occupancy}
\rho_{\pi}(s,a)
=
(1-\gamma)\sum_{t=0}^{\infty}\gamma^t
P\!\left(s_t=s,a_t=a\mid\pi,\mathcal{P}\right),
\end{equation}
where $\gamma\in[0,1)$ is the discount factor, $\mathcal{P}$ denotes the MDP transition kernel $\mathcal{P}(s'\mid s,a)$, and the prefactor $(1-\gamma)$ normalizes the occupancy measure over the infinite horizon. A standard adversarial objective is:
\begin{equation}
\begin{aligned}
\label{eq:gail}
\min_{\pi}\max_{D:\mathcal{S}\times\mathcal{A}\rightarrow(0,1)}
\Big\{
&\mathbb{E}_{(s,a)\sim\rho_E}\left[\log D(s,a)\right]
+
\mathbb{E}_{(s,a)\sim\rho_{\pi}}\left[\log\left(1-D(s,a)\right)\right]
\Big\}
-\lambda\mathcal{H}(\pi),
\end{aligned}
\end{equation}
where $D$ is a discriminator over state-action pairs, $\rho_{\pi}$ and $\rho_E$ denote the learner and expert occupancy measures, respectively, $\mathcal{H}(\pi)=\mathbb{E}_{(s,a)\sim\rho_{\pi}}[-\log\pi(a\mid s)]$ is the policy entropy, and $\lambda\geq0$ controls the strength of entropy regularization. At the optimal discriminator, this objective corresponds to minimizing a divergence between $\rho_E$ and $\rho_{\pi}$. In practice, the policy is updated through policy-gradient optimization using a discriminator-induced pseudo-reward, without requiring an externally specified task reward.

\subsection{Robotics Models}
\label{subsec:robotics_models}

\noindent \textbf{Vision Models.}
To perceive the environment, robotic models typically incorporate vision models to extract informative visual features. Common choices for visual encoders include models trained purely on visual data, such as the ResNet family~\cite{he2016deep}, Vision Transformers (ViT)~\cite{vaswani2017attention}, and self-supervised models like DINO~\cite{zhang2023dino}. For 3D perception tasks, point cloud-based encoders such as PointNet~\cite{qi2017pointnet} and PointTransformer~\cite{zhao2021point} are widely used. Additionally, vision-language pre-trained encoders, such as the image encoders of CLIP~\cite{radford2021learning} or SigLIP~\cite{zhai2023sigmoid}, are employed to obtain semantically enriched representations that align visual observations with linguistic inputs.
Some models leverage additional visual information, including object detection results (such as bounding boxes)~\cite{carion2020end} and visual tracking outputs~\cite{karaev2024cotracker}, to improve perception and support downstream tasks more effectively.

\noindent \textbf{Language Models.}
In recent years, especially since around 2020, language has emerged as a more natural and human-friendly modality, leading to a growing integration of language models into robotics. 
To understand human instructions, language embedding models are commonly employed for extracting textual features, such as BERT~\cite{devlin2019bert} and the language encoder of CLIP. 
With the leap from autoregressive models like GPT-2~\cite{radford2019language} to GPT-3~\cite{brown2020language}, large language models (LLMs) have demonstrated remarkable advances in text understanding and generalization. 
To further leverage the powerful generalization capabilities of LLMs~\cite{touvron2023llama, chowdhery2023palm}, many recent robotic models adopt LLMs as backbones, where textual inputs are processed through tokenization.

\noindent \textbf{Text-conditioned Vision Models and Vision-Language Models (VLMs).}
Robotic models also adopt VLMs such as LLaVA~\cite{liu2023visual}, PaLM-E~\cite{driess2023palm}, and Prism~\cite{karamcheti2024prismatic} as backbones, building upon their architectures to enhance multimodal understanding and control. 
In addition, some works leverage text-conditioned image editing~\cite{brooks2023instructpix2pix} and video generation models~\cite{liu2024sora} to guide action generation through visual imagination and goal specification.

\noindent \textbf{Vision-Action Models and Vision-Language-Action Models.}
Early approaches relied on visual servoing for control~\cite{hutchinson2002tutorial} and gradually incorporated RL or IL methods that map states or images to actions~\cite{jing2023exploring, chi2023diffusion}, giving rise to vision–action (VA) models. Over time, VA architectures have evolved from simple MLPs to more advanced diffusion-based frameworks~\cite{chi2023diffusion}, accompanied by diverse policy designs.
The concept of VLA models was introduced by RT-2~\cite{zitkovich2023rt}. 
In the narrow sense, VLAs refer to models that are fine-tuned from foundation VLMs using robotic trajectory data, enabling them to take human instructions and visual observations as input and directly generate robot actions in an end-to-end manner. 
In a broader sense, any model that takes both visual and language inputs and outputs robot actions through an end-to-end pipeline can be considered a VLA model.

\subsection{Evaluation}
\label{subsec:eval}

\noindent \textbf{Evaluation Metrics.}
The most commonly used metric for evaluating robotic performance is the success rate, which measures whether a given task is completed successfully. For long-horizon tasks~\cite{mees2022calvin}, this has been extended to metrics such as average success length, which captures the average number of consecutive tasks completed within a sequence of up to $n$ tasks. Beyond success-based measures, efficiency metrics such as task completion time and action frequency are also employed to assess how quickly and effectively a robot accomplishes a task. In RL settings, return is also widely used as an overall measure of performance.

\noindent \textbf{Model Selection.}
Evaluation strategies vary depending on the experimental setting. A common approach is to evaluate the model every $k$ epochs and select the checkpoint with the highest success rate. Alternatively, for more stable comparisons, one can average the results of the top-$n$ checkpoints from the final training phase. These strategies are frequently employed in single-task settings. In contrast, multi-task settings often adopt a simpler approach by reporting performance at the final training epoch or step, which enables more straightforward and consistent cross-task comparisons. Another widely used strategy is to select the checkpoint that achieves the highest validation performance for subsequent testing.

\section{Simulators, Benchmarks and Datasets}
\label{sec:benchmarks}

Simulators, Benchmarks, and Datasets provide the empirical foundation for advancing robotic manipulation, enabling standardized evaluation, reproducibility, and fair comparison across methods. They are crucial for assessing generalization, robustness, and scalability in data-driven models.
This section reviews key resources across six major areas: \textbf{grasping datasets} that support perception--pose learning, \textbf{single-embodiment manipulation simulators and benchmarks} that focus on a specific robotic embodiment, \textbf{cross-embodiment simulators and benchmarks} that evaluate generalization across different morphologies, \textbf{trajectory datasets} that capture multimodal robot interaction sequences, \textbf{human video datasets and video-world-model benchmarks} that provide large-scale human interaction data and evaluate predictive visual modeling for embodied control, and \textbf{embodied QA and affordance datasets} that connect perception with semantic and functional understanding.

\begin{figure}[!t]
\centering
\includegraphics[width=\linewidth]{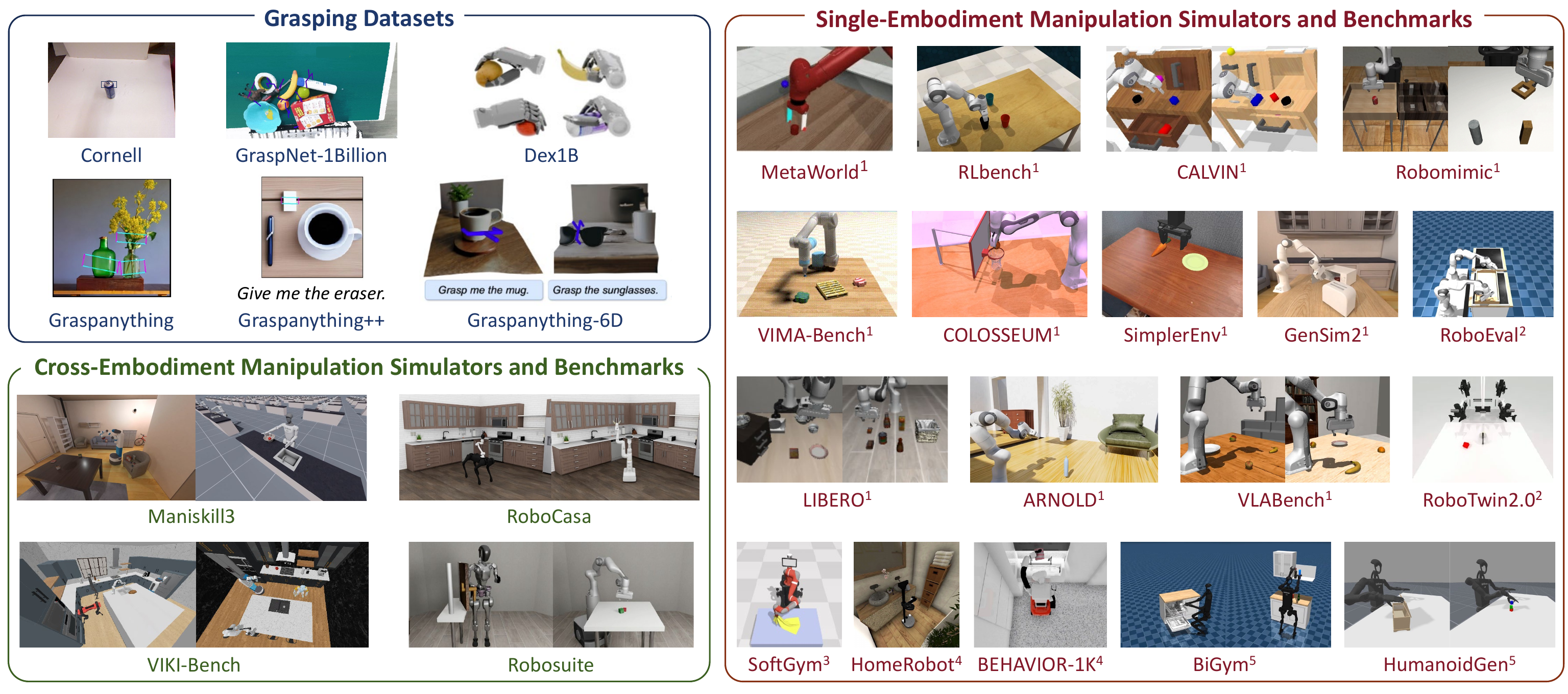}  
\caption{Overview of simulators and benchmarks.
$^1$Basic Manipulation with Single Arm, $^2$Basic Manipulation with Bimanual Arms, $^3$Deformable Object Manipulation, $^4$Mobile Manipulation, $^5$Humanoid Manipulation.
}
\label{fig: benchmark}
\end{figure}

\subsection{Grasping Datasets}
\label{subsec: grasping_datasets}

\begin{table*}[t]
\centering
\caption{Summary of grasping datasets.}
\label{tab: grasping_datasets}
\vskip -0.05in
\resizebox{\textwidth}{!}{
\begin{tabular}{lcccccccc}
\toprule
\textbf{Dataset} & \textbf{Venue} & \textbf{Grasp Type} & \textbf{Scene} & \textbf{\#Objects} & \textbf{Domain} & \textbf{Size} & \textbf{Visual Modality} & \textbf{w/ Language} \\
\midrule
Cornell~\cite{jiang2011efficient} & ICRA 2011 & rect. & single object & 240 & real & 1035 images, 8019 grasps & RGB-D & $\times$ \\
Jacquard~\cite{depierre2018jacquard} & IROS 2018 & rect. & single object & 11k & sim & 54k images, 1.1M grasps & RGB-D & $\times$ \\
GraspNet~\cite{fang2020graspnet} & CVPR 2020 & 6-DoF & cluttered & 88 & real & 97,280 images, $\sim$1.2B grasps & RGB-D & $\times$ \\
ACRONYM~\cite{eppner2021acronym} & ICRA 2021 & 6-DoF & multi-object & 8872 & sim & 17.7M grasps & RGB-D & $\times$ \\
Regrad~\cite{zhang2022regrad} & RA-L 2022 & rect. $\&$ 6-DoF & multi-object & 50K & sim & 1.02K images, 100M grasps & RGB-D & $\times$ \\
MetaGraspNet~\cite{gilles2022metagraspnet} & CASE 2022 & 6-DoF & cluttered & - & sim + real & 217k (sim) + 2.3k (real) images & RGB-D & $\times$ \\
Dexgraspnet~\cite{wang2023dexgraspnet} & ICRA 2023 & dexterous & single-object & 5355 & sim & 1.32M grasps & - & $\times$ \\
MetaGraspNet-V2~\cite{gilles2023metagraspnetv2} & TASE 2023 & 6-DoF & cluttered & - & sim + real & 296k (sim) + 3.2k (real) images & RGB-D & $\times$ \\
Grasp-Anything~\cite{vuong2024grasp} & ICRA 2024 & rect. & multi-object & 3M & syn & 1M images, $\sim$600M grasps & RGB & $\checkmark$ \\
Grasp-Anything++~\cite{vuong2024language} & CVPR 2024 & rect. & multi-object & 3M & syn & 1M images, 10M grasps & RGB & $\checkmark$ \\
Grasp-Anything-6D~\cite{nguyen2024graspanything6d} & ECCV 2024 & 6-DoF & multi-object & 3M & syn & 1M images, 200M grasps & RGB-D & $\checkmark$ \\
Dex1B~\cite{ye2025dex1b} & RSS 2025 & dexterous & single object & 6K & sim & 1B grasps & PC & $\times$ \\
GraspClutter6D~\cite{back2025graspclutter6d} & RA-L 2025 & 6-DoF & cluttered & 200 & real & 52K images, 9.3B grasps & RGB-D & $\times$ \\
RealVLG-11B~\cite{li2026realvlg} & CVPR 2026 & rect. & single to cluttered & $\sim$800 & real & 165K images, 11B grasps & RGB & $\checkmark$ \\
\bottomrule
\end{tabular}
}
\end{table*}

This subsection primarily focuses on grasp detection and generation tasks, where the goal is to predict viable grasp configurations directly from sensory inputs. While some works explore grasping as a downstream outcome of broader manipulation objectives, our discussion is centered on methods that explicitly target grasp prediction rather than those that infer grasps from manipulation trajectories or task goals.  
Data annotations are typically categorized into two formats: rectangle-based and 6-DoF-based. The rectangle-based format labels each grasp using a 5-dimensional grasp rectangle representation, defined by the center position $(x,y)$, the gripper's width and height, and the orientation angle relative to the horizontal axis. In contrast, the 6-DoF format directly annotates the end-effector’s six-degree-of-freedom pose, including position $(x,y,z)$, orientation (e.g., Euler angles or quaternions), and may also include additional information such as the approach direction and a grasp quality score.
We provide a comprehensive summary of existing grasping datasets in Table~\ref{tab: grasping_datasets}.

Grasping datasets have undergone significant evolution along several dimensions, each contributing to the advancement of the field. 
First, annotation has shifted from manual labeling to model-based automation, greatly reducing human effort and enabling scalable data generation. 
Second, the amount of annotated data has increased from small to large-scale datasets, allowing for more robust and data-driven learning. 
Third, grasp representations have evolved from simple 2D rectangles~\cite{jiang2011efficient, depierre2018jacquard, zhang2022regrad, vuong2024grasp, vuong2024language, li2026realvlg} to full 6-DoF~\cite{fang2020graspnet, eppner2021acronym, gilles2022metagraspnet, gilles2023metagraspnetv2, nguyen2024graspanything6d} and dexterous hand~\cite{wang2023dexgraspnet, lin2025unifucgrasp, zhang2024dexgraspnet, ye2025dex1b} poses, enabling a more accurate modeling of the complexity inherent in real-world grasping.
Fourth, task settings have expanded from single-object scenarios to multi-object and cluttered environments, offering more realistic and challenging conditions. 
Lastly, the input modalities have progressed from purely vision-based inputs to language-conditioned instructions, facilitating more flexible and semantically rich manipulation. 
These changes collectively support the development of generalizable and intelligent grasping systems.

\subsection{Single-Embodiment Manipulation Simulators and Benchmarks}

\begin{table*}[t]
\centering
\caption{Summary of robot manipulation simulators and benchmarks. 
All benchmarks provide proprioceptive or pose observations by default, and most include additional modalities. 
}
\label{tab: manipulation_simulator_benchmark}
\vskip -0.05in
\resizebox{\textwidth}{!}{
\begin{tabular}{lccccccccc}
\toprule
\textbf{Name} & \textbf{Venue/Year} & \textbf{Simulator} & \textbf{\#Objects} & \textbf{\#Tasks} & \textbf{\#Demos} & \textbf{Observation} & \textbf{Robot Type} & \textbf{Mani. Type} \\
\midrule
MetaWorld~\cite{yu2020meta} & CoRL 2019 & MuJoCo & 80 & 50 & - & Pose & Sawyer & Ba, Uni \\
Franka Kitchen~\cite{gupta2020relay} & CoRL 2020 & MuJoCo & 10 & 7 & 513 & Pose & Franka Panda & Ba, Uni \\
RLBench~\cite{james2020rlbench} & RA-L 2020 & CoppeliaSim & 28 & 100 & - & RGB, D, S & Franka Panda & Ba, Uni \\
Robomimic~\cite{mandlekar2022matters} & CoRL 2021 & MuJoCo & 15 & 8 & 6K & RGB, D & Franka Panda & Ba, Uni \\
Maniskill~\cite{mu2maniskill} & NeurIPS 2021 & SAPIEN & 100+ & 4 & 30K+ & RGB, D, S & Franka Panda & Ba, Uni \\
CALVIN~\cite{mees2022calvin} & RA-L 2022 & Pybullet & 28 & 34 & 40M & RGB, D & Franka Panda & Ba, Uni \\
VLMbench~\cite{zheng2022vlmbench} & NeurIPS 2022 & CoppeliaSim/\cite{james2020rlbench} & 22 & 8 & 6K+ & RGB, D, S & Franka Panda & Ba, Uni \\
Maniskill2~\cite{gumaniskill2} & ICLR 2023 & SAPIEN & 2144 & 20 & 30K+ & RGB, D, S & Franka Panda & Ba, Mo, Uni, Bi \\
VIMA-Bench~\cite{jiang2022vima} & ICML 2023 & Pybullet/Ravens & 100+ & 17 & 600K+ & RGB, D, S & UR5 & Ba, Uni \\
ARNOLD~\cite{gong2023arnold} & ICCV 2023 & Isaac Sim & 40 & 8 & 10K+ & RGB, D, S & Franka Panda & Ba, Uni \\
LIBERO~\cite{liu2023libero} & NeurIPS 2023 & MuJoCo/\cite{zhu2020robosuite} & 67 & 130 & 6.5K & RGB, D & Franka Panda & Ba, Uni \\
COLOSSEUM~\cite{pumacay2024colosseum} & RSS 2024 & CoppeliaSim/\cite{james2020rlbench} & - & 20 & 2K & RGB, D & Franka Panda & Ba, Uni \\
SimplerEnv~\cite{li2025evaluating} & CoRL 2024 & SAPIEN & - & 10 & 140K+ & RGB, D & Google Robot, WidowX & Ba, Uni \\
GenSim2~\cite{hua2025gensim2} & CoRL 2024 & SAPIEN & 200 & 100 & - & RGB, D & Franka Panda & Ba, Uni \\
GemBench~\cite{garcia2025towards} & ICRA 2025 & CoppeliaSim & 20+ & 60 & - & RGB, D & Franka Panda & Ba, Uni \\
RoboTwin~\cite{mu2025robotwin} & CVPR 2025 & SAPIEN/\cite{tao2025maniskill3} & 10+ & - & 320 & RGB, D, S & Aloha-AgileX & Ba, Bi \\
GENMANIP~\cite{gao2025genmanip} & CVPR 2025 & Isaac Sim & 10K & 200 & 200K & RGB, D, S & Franka Panda & Ba, Uni \\
VLABench~\cite{zhang2025vlabench} & ICCV 2025 & MuJoCo & 2000+ & 100 & 5K & RGB, D & Franka Panda & Ba, Uni \\
AGNOSTOS~\cite{zhou2025agnostos} & NeurIPS 2025 & CoppeliaSim/\cite{james2020rlbench} & - & 41 & 3.6K & RGB, D, S & Franka Panda & Ba, Uni \\
RoboTwin 2~\cite{chen2025robotwin} & 2025 & SAPIEN/\cite{tao2025maniskill3} & 731 & 50 & 100K & RGB, D, S & Aloha, UR5, Franka, ARX-X5 & Ba, Uni, Bi \\
ROBOEVAL~\cite{wang2025roboeval} & 2025 & - & - & 8 & 3K+ & RGB, D & Franka Panda & Ba, Bi \\
INT-ACT~\cite{fang2025intention} & 2025 & SAPIEN/\cite{li2025evaluating} & - & 50 & 60K+ & RGB, D & Franka Panda & Ba, Uni \\
RoboDojo~\cite{chen2026robodojo} & 2026 & Isaac Sim & - & 60 & 5.3K & RGB & ARX X5, Piper, Piper X & Ba, Bi \\
\midrule
TacSL~\cite{akinola2025tacsl} & T-RO 2025 & Isaac Gym & - & 3 & - & RGB, D, T & Franka Panda & Ba, Uni \\
ManiFeel~\cite{luu2025manifeel} & 2025 & Isaac Gym/\cite{akinola2025tacsl} & - & 6 & - & RGB, D, T & Franka Panda & Ba, Uni \\
\midrule
\cite{rajeswaran2018learning} & RSS 2018 & MuJoCo & - & 4 & 100 & RGB, D, T & ADROIT Hand & Dex, Uni \\
TriFinger~\cite{wuthrich2021trifinger} & CoRL 2021 & PyBullet & - & 2 & - & RGB & 3-DoF Hand & Dex, Uni \\
\midrule
PlasticineLab~\cite{huang2021plasticinelab} & ICLR 2021 & Taichi + DiffTaichi & 1 & 10 & - & RGB, D, P & Rigid End-Effector & De \\
SoftGym~\cite{lin2021softgym} & CoRL 2021 & NVIDIA FleX & 4 & 10 & - & RGB, D, P & Sawyer, Franka & De, Uni \\
DaXBench~\cite{chen2023daxbench} & ICLR 2023 & DaX & 4 & 9 & - & RGB, D, P & - & De \\
DLO-Lab~\cite{cao2026dlo} & ICML 2026 & Genesis/Taichi & - & 10 & 800 & RGB, V & Robot Arm(s) + Parallel Gripper & De, Uni, Bi \\
\midrule
ManipulaTHOR~\cite{ehsani2021manipulathor} & CVPR 2021 & Unity/AI2-THOR & 2.6K+ & - & - & RGB, D, N & Kinova Gen3 on Mobile Base & Mo, Uni \\
HomeRobot~\cite{yenamandra2023homerobot} & CoRL 2023 & AI Habitat & 7892 & - & - & RGB, D & Hello Robot Stretch & Mo, Uni \\
BEHAVIOR-1K~\cite{li2023behavior} & CoRL 2023 & OmniGibson & 5215 & 1K & - & RGB, D & Mobile Manipulator & Mo, Uni \\
\midrule
ODYSSEY~\cite{wang2026odyssey} & AAAI 2026 & Isaac Sim & 100+ & 12 & - & RGB, D, LiDAR & Unitree Go2 + ARX5 & Mo, Quad, Uni \\
\midrule
BiGym~\cite{chernyadev2025bigym} & CoRL 2024 & MuJoCo & 10+ & 40 & 2K & RGB, D & Unitree H1 & Hu, Bi \\
HumanoidBench~\cite{sferrazza2024humanoidbench} & RSS 2024 & MuJoCo & ~15 & 27 & - & RGB, LiDAR & Unitree H1 + Shadow-Hand & Hu, Bi, Dex \\
HumanoidGen~\cite{jing2025humanoidgen} & NeurIPS 2025 & SAPIEN & - & 20 & - & RGB, D & Unitree & Hu, Bi, Dex \\
SIMPLE~\cite{wei2026simple} & 2026 & MuJoCo + Isaac Sim & 1.5K+ & 60 & 6K & RGB & Unitree G1 & Hu, Bi, Dex \\
\bottomrule
\end{tabular}
}
\\[1ex]
\hfill {\footnotesize D = depth, S = segmentation, T = tactile sensing, P = particle-based state representations, N = normals}
\\[0ex]
\hfill {\footnotesize Ba = basic, De = deformable object, Dex = dexterous, Mo = mobile, Quad = quadrupedal, Hu = humanoid manipulation}
\\[0ex]
\hfill {\footnotesize Uni = single arm, Bi = bimanual arms}
\end{table*}

Single-embodiment manipulation benchmarks focus on a specific type of robotic platform, typically represented by single-arm manipulators or humanoids equipped with manipulators. For consistency in categorization, we group single-arm and dual-arm systems under the same embodiment type.
Building on the task taxonomy introduced in Section~\ref{sec:manipulation_tasks}, we provide a comprehensive overview of existing single-embodiment manipulation simulators and benchmarks in Table~\ref{tab: manipulation_simulator_benchmark}.

\noindent \textbf{Basic Manipulation Benchmarks.}
These benchmarks primarily target relatively constrained tabletop tasks performed by single- or dual-arm manipulators, including pick-and-place, sorting, pushing, insertion, opening, closing, and pouring. Early benchmarks largely focused on learning task-specific trajectories through reinforcement learning or imitation learning~\cite{yu2020meta,gupta2020relay}. More recent efforts have progressively expanded the evaluation scope toward more demanding settings, including long-horizon tasks that require sequential execution of multiple manipulation stages~\cite{mees2022calvin,zhang2025vlabench,lei2026robomemarena}, language-conditioned trajectory generation~\cite{jiang2022vima,zheng2022vlmbench,liu2023libero}, generalization under visual distractions~\cite{pumacay2024colosseum,garcia2025towards,fei2025libero} or unseen tasks~\cite{fang2025intention}, and robustness to environmental or execution perturbations~\cite{balaji2026oopsieverse,cui2026libero}. Other benchmarks seek to establish fairer and more standardized evaluation protocols for VLA models~\cite{li2025evaluating,zhang2025vlabench}, while recent platforms increasingly extend evaluation from single-arm settings to more capable bimanual manipulation~\cite{grotz2024peract2,mu2025robotwin,chen2025robotwin}. In parallel, growing attention has been devoted to tactile sensing for policy learning and evaluation in contact-rich manipulation tasks~\cite{xu2023efficient,akinola2025tacsl,luu2025manifeel}.

\noindent \textbf{Dexterous Manipulation Benchmarks.}
Dexterous manipulation benchmarks have progressed from evaluating individual algorithms and hardware platforms toward broader assessment across tasks and embodiments. Early work combined deep reinforcement learning with demonstrations to improve high-dimensional skill acquisition~\cite{rajeswaran2018learning}, while TriFinger provided a reproducible hardware testbed~\cite{wuthrich2021trifinger}. More recent benchmarks expand this scope: DexJoCo~\cite{wang2026dexjoco} evaluates task-oriented dexterous manipulation under diverse settings, whereas DexVerse~\cite{yao2026dexverse} emphasizes multi-task and multi-embodiment generalization. Together, they support more systematic evaluation of dexterous manipulation policies.

\noindent \textbf{Deformable Object Manipulation Benchmarks.}
Benchmarks for deformable object manipulation provide environments for evaluating robotic systems on non-rigid materials such as cloth, ropes, fluids, and elastic objects~\cite{huang2021plasticinelab,lin2021softgym}. DaXBench~\cite{chen2023daxbench} broadens task and object coverage through differentiable physics, while DLO-Lab~\cite{cao2026dlo} focuses on deformable linear objects with material properties, contact interactions, and topological complexity. RGBench~\cite{hu2026real} emphasizes physically accurate garment simulation and systematic measurement of the sim-to-real gap, whereas MoDeSuite~\cite{zhang2026modesuite} extends deformable-object evaluation to mobile manipulation requiring coordinated base and arm motion. Together, these benchmarks support more systematic comparison of planning and learning methods across physical properties, object categories, embodiments, and sim-to-real settings.

\noindent \textbf{Mobile Manipulation Benchmarks.}
Mobile manipulation benchmarks evaluate systems that integrate locomotion and manipulation~\cite{ehsani2021manipulathor, yenamandra2023homerobot, li2023behavior}. Typical tasks involve coordinating a mobile base (wheeled or legged) with an onboard manipulator to transport objects, navigate to target locations, and interact within cluttered or spatially extended environments. These benchmarks are critical for studying perception, planning, and control challenges faced by embodied agents operating in dynamic and unstructured settings.

\noindent \textbf{Quadrupedal Manipulation Benchmarks.}
Wang et al. introduced ODYSSEY~\cite{wang2026odyssey}, a benchmark and framework for open-world quadruped robots that unifies exploration and manipulation in long-horizon tasks. By integrating vision-language planning with whole-body control, ODYSSEY addresses key challenges in instruction decomposition, locomotion–manipulation coordination, and generalization across diverse open-world scenarios, with validation in both simulation and the real world.

\noindent \textbf{Humanoid Manipulation Benchmarks.}
Humanoid manipulation benchmarks evaluate whole-body coordination, dexterity, and stability in tasks involving upper-body manipulation and legged locomotion~\cite{chernyadev2025bigym,sferrazza2024humanoidbench,jing2025humanoidgen}. Recent benchmarks further expand this scope: SIMPLE~\cite{wei2026simple} provides a scalable simulation testbed for diverse loco-manipulation tasks and policy architectures, while HumanoidArena~\cite{wang2026humanoidarena} emphasizes egocentric hierarchical learning and transfer across motion-tracking controllers. Complementary frameworks such as OASIS~\cite{yu2026oasis} and GRAIL~\cite{xie2026grail} improve scalable sim-to-real learning through simulated teleoperation, reconstructed assets, domain randomization, and video-generated interaction priors. Together, these efforts support more comprehensive evaluation and data generation for generalizable humanoid loco-manipulation.

\subsection{Cross-Embodiment Manipulation Simulators and Benchmarks}

Cross-embodiment manipulation benchmarks support a diverse range of robotic platforms, including single-arm, dual-arm, mobile, quadrupedal, and humanoid robots. These settings are designed to evaluate whether a single model can consistently perform similar tasks across robots with varying morphologies, degrees of freedom, and control constraints.
Some benchmarks integrate multiple embodiment-specific benchmarks, each with different simulator backends and without a unified interface~\cite{majumdar2023we}. Others consolidate various simulator backends and provide a unified API for consistent interaction~\cite{kumar2023robohive, geng2025roboverse}. Additionally, there are benchmarks that rely on a single simulator backend, within which different embodiments are supported through carefully designed environments~\cite{zhu2020robosuite, mittal2023orbit, nasiriany2024robocasa, genesis2024genesis, tao2025maniskill3, zhang2025agentworld}.
We present a comprehensive overview of existing cross-embodiment simulators and benchmarks in Table~\ref{tab: cross_embodiment_simulator_benchmark}. 

\begin{table*}[t]
\centering
\caption{Summary of cross-embodiment robotic manipulation benchmarks. S denotes segmentation, T tactile sensing, A audio, L LiDAR, and F force/contact information.}
\label{tab: cross_embodiment_simulator_benchmark}
\vskip -0.05in
\resizebox{\textwidth}{!}{
\begin{tabular}{lcccccccc}
\toprule
\textbf{Name} & \textbf{Venue/Year} & \textbf{Simulator} & \textbf{\#Objects} & \textbf{\#Tasks} & \textbf{\#Demos} & \textbf{Observation} & \textbf{\#Robots} & \textbf{Manip. Type} \\
\midrule
RoboSuite~\cite{zhu2020robosuite}     & 2020         & MuJoCo             & 20               & 9 tasks        & -              & RGB, D                & 10                        & Ba, Mo, Hu, Quad, Uni, Bi, Dex \\
CortexBench~\cite{majumdar2023we}   & NeurIPS 2023 & Multiple           & -                & 17 tasks & 850+      & RGB, D                & 6                         & Ba, Mo, Uni \\
RoboHive~\cite{kumar2023robohive}   & NeurIPS 2023 & Multiple           & -                & 17 tasks & 850+      & RGB, D                & 10+                         & Ba, Mo, Quad, Hu, Uni, Bi, Dex \\
ORBIT/Isaac Lab~\cite{mittal2023orbit} & RA-L 2023  & Isaac Sim          & -                & 5 types & -     & RGB, D, S             & 16                        & Ba, Mo, Hu, Uni, Bi, Dex \\
RoboCasa~\cite{nasiriany2024robocasa}      & RSS 2024     & MuJoCo/\cite{zhu2020robosuite}   & 2509             & 100 tasks      & 100K+           & RGB, D                & 4+                        & Ba, Mo, Hu, Quad, Uni, Bi, Dex \\
Genesis~\cite{genesis2024genesis}       & 2024         & Genesis Engine     & -                & -              & -              & RGB, D, T, A          & 6                         & Ba, Mo, Hu, Uni, Bi, Dex \\
ManiSkill3~\cite{tao2025maniskill3}    & RSS 2025     & SAPIEN             & 10K+             & 12 types  & 1M frames       & RGB, D, S             & 20+                       & Ba, Mo, Hu, Uni, Bi, Dex \\
RoboVerse~\cite{geng2025roboverse}     & RSS 2025         & MetaSim            & 5.5K             & 276 tasks      & 500K            & RGB, D                & 5                         & Ba, Mo, Hu, Uni, Bi, Dex \\
AgentWorld~\cite{zhang2025agentworld}    & CoRL 2025     & Isaac Sim             & 9K             & -  & 1000+       & RGB, D             & 4                       & Mo, Hu, Uni, Bi, Dex \\
VIKI-Bench~\cite{kang2025viki}    & NeurIPS 2025         & \cite{nasiriany2024robocasa, tao2025maniskill3} & -             & 23,737 tasks   & -              & RGB, D                & 6                         & Ba, Mo, Hu, Uni, Bi, Dex \\
GS-Playground~\cite{jia2026gs} & RSS 2026 & GS-Playground & - & 3 types & - & RGB, D, L, F & 5+ & Ba, Mo, Hu, Quad, Uni \\
\bottomrule
\end{tabular}
}
\end{table*}

\begin{table*}[t]
\centering
\caption{Summary of trajectory datasets.}
\label{tab: trajectory_datasets}
\vskip -0.05in
\resizebox{\textwidth}{!}{
\begin{tabular}{lccccccc}
\toprule
\textbf{Dataset} & \textbf{Venue/Year} & \textbf{Domain} & \textbf{\#Demos} & \textbf{\#Verbs} & \textbf{Robot Type} & \textbf{Observation} \\
\midrule
MIME~\cite{sharma2018multiple} & CoRL 2018 & real & 8.3k & 20 & Baxter Robot & RGB, D \\
BridgeData~\cite{ebert2022bridge} & 2021 & real & 7.2k & 4 & WidowX250 & RGB, D \\
BC-Z~\cite{jang2022bc} & CoRL 2021 & real & 26k & 3 & Google Robot & RGB \\
RT-1~\cite{brohan2023rt} & RSS 2023 & real & 130k & 2 & Google Robot & RGB, D \\
RH20T~\cite{fang2023rh20t} & RSSW 2023 & real & 110k & 33 & Flexiv, UR5, Franka & RGB, D, T \\
BridgeData V2~\cite{walke2023bridgedata} & CoRL 2023 & real & 60.1k & 24 & WidowX 250 & RGB, D \\
RoboSet~\cite{bharadhwaj2024roboagent} & ICRA 2024 & real & 98.5k & 11 & Franka Panda & RGB, D \\
Open X-Embodiment~\cite{o2024open} & ICRA 2024 & real & 1.4M & 217 & 22 Embodiments & RGB, D \\
DROID~\cite{khazatsky2024droid} & RSS 2024 & real & 76k & 86 & Franka Panda & RGB, D \\
AgiBot World Dataset~\cite{bu2025agibot} & IROS 2025 & real & 1M+ & 87 &	Agibot & RGB, D, T \\
ARIO~\cite{wang2024all} & 2024 & real + sim & 3M & 20 & AgileX, UR5, Cloud Ginger XR-1 & RGB, D, T, A \\
RoboMind~\cite{wu2025robomind} & RSS 2025 & real & 107k & 38 & Franka Panda, Tien Kung, AgileX, UR5 & RGB, D \\
RoboFAC~\cite{lu2025robofac} & 2025 & sim (SAPIEN/ManiSkill) & 9.44k & 12 & Franka Panda & RGB, D \\
RoboCOIN~\cite{wu2025robocoin} & 2025 & real & 180K+ & 39 & 15 Dual-Arm/Humanoid Embodiments & RGB, D \\
Humanoid Everyday~\cite{zhao2026humanoid} & ICRA 2026 & real & 10.3k & 221 & Unitree G1, H1 & RGB, D, L, T \\
\bottomrule
\end{tabular}
}
\end{table*}

\subsection{Trajectory Datasets}

Trajectory datasets are structured collections of time-ordered data that capture the sequential states, actions, and sensory observations of an agent interacting with an environment. In the context of robotics and embodied AI, these datasets typically include robot joint states, end-effector poses, control inputs, and multimodal observations (e.g., RGB images, depth maps, force-torque readings) collected during the execution of specific tasks.
In addition to trajectory datasets included in benchmarks, some works focus on building dedicated datasets that vary in scale, quality, and diversity. These datasets range from small collections to large-scale repositories with millions of samples, and from low-fidelity teleoperated data to high-quality expert demonstrations. They also differ in embodiment types, control modes, and data sources such as human teleoperation, scripted agents, or learned policies. Many high-quality datasets include semantic labels, task definitions, and multimodal observations, making them valuable for learning general manipulation policies across tasks and robot types.
We provide a comprehensive summary of existing trajectory datasets in Table~\ref{tab: trajectory_datasets}.

\subsection{Embodied QA and Affordance Datasets}

\begin{table*}[t]
\centering
\caption{Summary of embodied QA and affordance datasets.}
\label{tab: embodied_qa_and_affordance_datasets}
\vskip -0.05in
\renewcommand{\arraystretch}{1.2}
\resizebox{\textwidth}{!}{
\begin{tabular}{lccc
>{\centering\arraybackslash}p{4cm}
>{\centering\arraybackslash}p{4cm}
>{\centering\arraybackslash}p{4cm}
}
\toprule
\textbf{Dataset} & \textbf{Venue} & \textbf{Domain} & \textbf{Size} & \textbf{Visual Perception Tasks} & \textbf{Spatial Reasoning Tasks} & \textbf{Functional and Commonsense Reasoning Tasks} \\
\midrule
OpenEQA~\cite{majumdar2024openeqa} & CVPR 2024 & real & 1.6K & Object, Attribute and Object State Recognition & Object Localization, Spatial Reasoning & Functional Reasoning, World Knowledge \\
ManipVQA~\cite{huang2024manipvqa} & IROS 2024 & real + sim & 84K & Physically Grounded Understanding & Object Detection & -- \\
ManipBench~\cite{zhao2025manipbench} & CoRL 2025 & real + sim & 12K+ & Keypoint Selection, Trajectory Understanding & Fabric Manipulation, Tool \& Drawer Contact & -- \\
RefSpatial~\cite{zhou2025roborefer} & NeurIPS 2025 & real & 20M & - & Object Location, Orientation and Topological Reasoning & -- \\
Robo2VLM~\cite{chen2025robo2vlm} & NeurIPS 2025 & real & 684K+ & Scene Understanding, Multiple View & Object State, Spatial Relationship & Goal-conditioned Reasoning, Interaction Reasoning \\
PAC Bench~\cite{gundawar2025pac} & NeurIPS 2025 & real + sim & 30K+ & Properties & Constraints & Affordance \\
PointArena~\cite{cheng2025pointarena} & 2025 & real & 982 & Pointing & Pointing & -- \\
\bottomrule
\end{tabular}
}
\end{table*}

While EQA and affordance understanding both require visual-semantic and spatial reasoning, EQA emphasizes high-level question answering based on environmental context, whereas affordance understanding targets low-level functional interaction with objects, such as grasping or tool use.
These datasets empower robotic models with the ability to perceive and understand the physical world, and we categorize their capabilities into three core task types. 
First, Visual Perception Tasks focus on the static recognition of visual information, enabling robots to identify what an object is, what color or material it has, and what state it is in (e.g., open or closed). This includes object recognition, attribute recognition, object state recognition~\cite{majumdar2024openeqa}, and keypoint selection to determine actionable locations for manipulation~\cite{zhao2025manipbench}. 
Second, Spatial Reasoning Tasks involve understanding object positions, spatial relationships, and reachability. They encompass 2D/3D object detection~\cite{huang2024manipvqa}, object localization~\cite{majumdar2024openeqa}, and reasoning about spatial relations between the robot and its environment~\cite{chen2025robo2vlm, zhou2025roborefer}, such as determining the relative direction between a gripper and a target object. 
Finally, Functional and Commonsense Reasoning Tasks address the robot’s understanding of affordances~\cite{gundawar2025pac} and functional uses~\cite{majumdar2024openeqa} of objects—for instance, recognizing that a dish wand is used for cleaning utensils or that a knife should be grasped by its handle. These tasks bridge perception with actionable, context-aware behavior grounded in physical interaction knowledge.
Table~\ref{tab: embodied_qa_and_affordance_datasets} offers detailed descriptions of representative datasets.

\subsection{Human Video Datasets and Video-World-Model Benchmarks}
\label{subsec:video_datasets_benchmarks}

Given the scarcity of robot trajectories, increasing attention has turned to human egocentric videos as a scalable complementary data source that captures diverse hand-object interactions and offers valuable priors for learning robot manipulation.
Ego4D~\cite{grauman2022ego4d} established a large-scale foundation for first-person activity understanding, while Ego-Exo4D~\cite{grauman2024ego} extended this setting with synchronized egocentric and exocentric views and rich multimodal annotations. More manipulation-oriented datasets have increasingly incorporated supervision that is directly relevant to robot learning. EgoDex~\cite{hoque2026egodex} provides large-scale dexterous manipulation videos with detailed 3D hand trajectories, and EgoVerse~\cite{punamiya2026egoverse} emphasizes diverse human demonstrations and systematic human-to-robot transfer across tasks, scenes, and robot embodiments. Hoi!~\cite{engelbracht2025hoi} further connects human and robotic interaction data through cross-view visual observations, force, torque, and tactile sensing. Collectively, these datasets shift human video from generic activity understanding toward scalable supervision for manipulation, although embodiment mismatch, incomplete action labels, and uncertain recovery of contact dynamics remain important limitations.

In parallel, recent benchmarks have begun evaluating whether video world models capture interaction dynamics that can support robot execution, rather than assessing generated videos solely through perceptual quality. Dream.exe~\cite{zhao2026dream} converts generated manipulation videos into robot trajectories and evaluates their execution in simulation, while RoboWM-Bench~\cite{jiang2026robowm} extends embodiment-grounded evaluation to both human-hand and robotic manipulation videos. WoW-World-Eval~\cite{fan2026wow} further introduces an embodied Turing test that evaluates perception, planning, prediction, generalization, and execution through complementary automatic, human-preference, and action-execution metrics. Together, these benchmarks shift world-model evaluation from visual realism toward physical consistency, long-horizon reasoning, and embodied executability.
\section{Manipulation Tasks}
\label{sec:manipulation_tasks}

Early grasping methods primarily focused on identifying stable grasp configurations through geometric analysis, force closure conditions, or task-specific heuristics. Candidate grasp poses were analytically generated from object geometry, contact normals, or predefined grasp templates, and then executed using inverse kinematics (IK) and motion planning. Building on such established grasps, early approaches to dexterous, deformable, mobile, quadrupedal, and humanoid manipulation typically assumed a known target pose or a task-specific goal. The emphasis was on optimizing motion trajectories or control commands to achieve these goals under physical and kinematic constraints, rather than learning end-to-end action generation from raw sensory input as in modern RL or IL methods. The comparison is summarized in Table~\ref{tab: method_comparison}.

Among these, basic manipulation is by far the most extensively studied, supported by a rich body of literature that enables fine-grained categorization of methods. We therefore dedicate Sections~\ref{sec:high_level_planning} and \ref{sec:low_level_action_modeling} to a detailed discussion of high-level planning and low-level learning-based action modeling in the context of basic manipulation, while for other task categories we summarize representative methods within their respective subsections.

\begin{table*}[t]
\centering
\caption{Comparison between non-learning and learning-based methods for grasping and manipulation. Here, grasping specifically refers to grasp detection and generation.}
\label{tab: method_comparison}
\vskip -0.05in
\resizebox{\textwidth}{!}{
\begin{tabular}{l
>{\centering\arraybackslash}p{3cm}
>{\centering\arraybackslash}p{4.5cm}
>{\centering\arraybackslash}p{4.5cm}
>{\centering\arraybackslash}p{2.5cm}
>{\centering\arraybackslash}p{2.5cm}
}
\toprule
\textbf{Mani. Type} & \textbf{Control Type} & \textbf{Pose Generation} & \textbf{Trajectory Generation} & \textbf{Generalization} & \textbf{Interpretability \& Stability} \\
\midrule
\multirow{2}{*}{Grasping} 
  & Non-learning & Analytical: geometric rules, force-closure analysis & IK + motion planning & Low  & High \\
  & Learning      & Learned from data                                & IK + motion planning & High & Low  \\
\midrule
\multirow{2}{*}{Manipulation} 
  & Non-learning & Task-specific or predefined goal pose             & IK + motion planning & Low  & High \\
  & Learning      & Learned implicitly via RL or IL & Learned policies via RL or IL & High & Low \\
\bottomrule
\end{tabular}
}
\end{table*}

\subsection{Grasping}

In the narrow sense considered in this work, grasping specifically refers to the tasks of grasp detection and grasp generation. These tasks involve identifying feasible grasp configurations from sensor inputs such as images or point clouds, allowing robotic end-effectors to securely pick up objects. The primary focus is on predicting the position and orientation of the gripper to ensure stable and reliable grasps, even in the presence of diverse object shapes, varying poses, and cluttered environments.

\begin{figure}[t]
\centering
\includegraphics[width=0.75\linewidth]{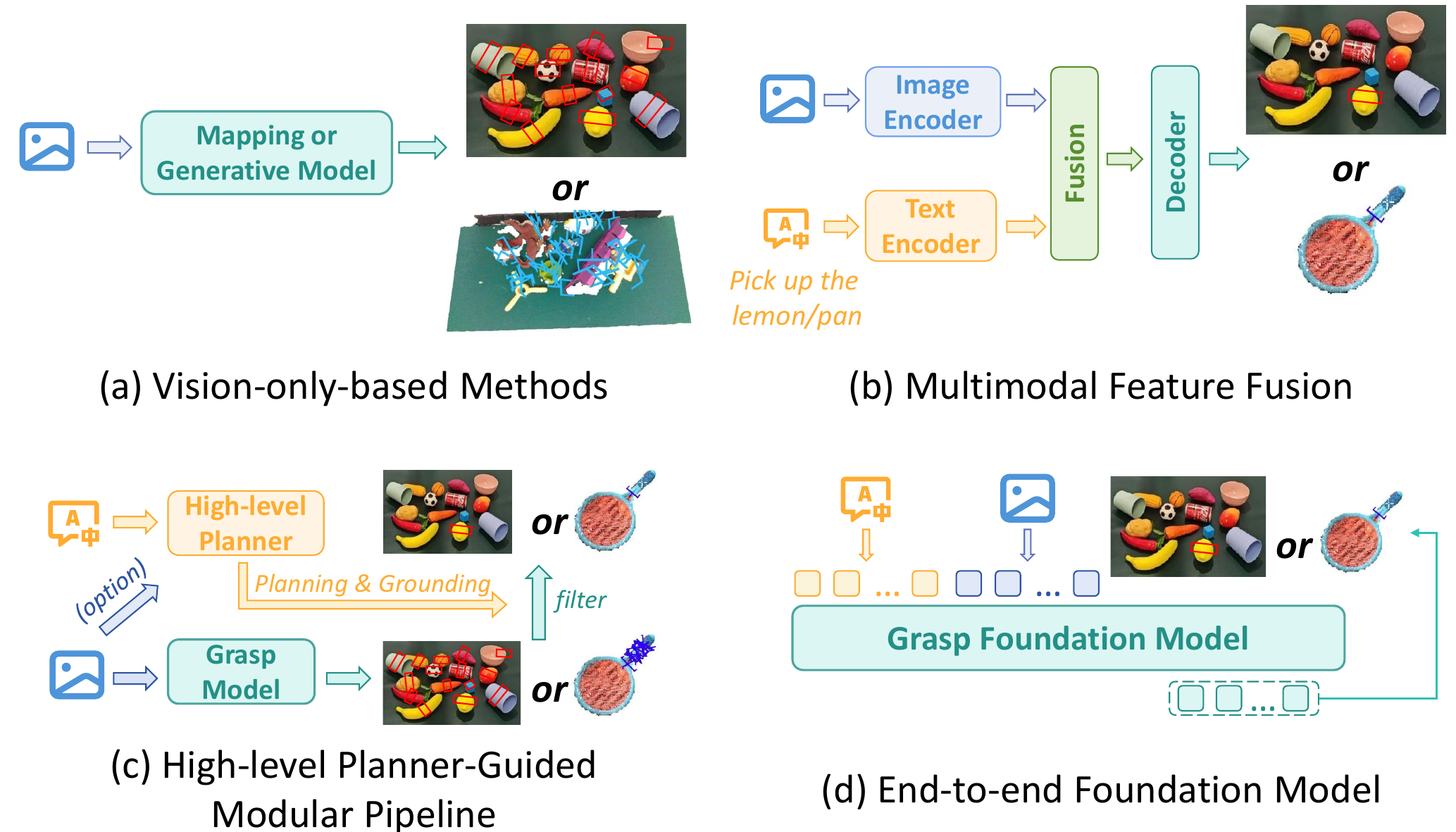}  
\caption{Comparison of different grasping methods. (a) Vision-only approaches, which map visual inputs to grasp poses using CNNs or spatial transformers, or generate poses via VAEs or diffusion models. With the introduction of language, three main categories emerge: (b) fusion of language and visual features through mechanisms such as cross-attention; (c) generation of multiple grasp candidates using pretrained grasp models, followed by selection with high-level planners (e.g., LLMs, VLMs, or 3D representations); and (d) end-to-end fine-tuning of grasp foundation models on large-scale grasping datasets. Figure adapted from~\cite{zhang2026vcot}.
}
\label{fig: grasp_methods}
\end{figure}

\noindent \textbf{Non-Learning-Based Grasp.} 
Early methods generate grasp poses by explicitly analyzing object geometry~\cite{jones1990planning}, performing contact-driven force analysis~\cite{miller2004graspit}, or applying task-specific rules~\cite{stoytchev2005behavior}, in combination with the gripper model. 
However, due to their limited generalization ability in handling complex shapes, occlusions, and unseen objects, these approaches have gradually been replaced by data-driven, learning-based methods, as discussed below. The taxonomy of learning-based grasping approaches is illustrated in Figure~\ref{fig: grasp_methods}.

\noindent \textbf{Rectangle-based Grasp.}
Grasping rectangles were first introduced by Jiang et al.~\cite{jiang2011efficient}. While they are visually similar to bounding boxes, their representational semantics are fundamentally different. A grasping rectangle is defined as a 5-dimensional representation, as previously described in Section~\ref{subsec: grasping_datasets}. 
Subsequent methods have progressively incorporated deep learning techniques, ranging from basic convolutional neural networks (CNNs)~\cite{redmon2015real, kumra2017robotic, morrison2018closing} to more advanced architectures such as ResNet~\cite{carion2020end}, GR-ConvNet~\cite{kumra2020grconvnet} and Transformer~\cite{wang2022transformer}, and further to CLIP-based models that integrate textual information through feature fusion methods~\cite{nguyen2024lightweight}. More recently, diffusion models have also been employed~\cite{vuong2024language}. Over time, the models have evolved from simple to complex, and the modalities have shifted from unimodal to increasingly multimodal.

\noindent \textbf{6-DoF Grasp.}
Two-dimensional rectangle-based representations, which typically assume parallel-jaw grippers executing top-down grasps, are limited to 2 or 3 degrees of freedom (DoF). This restricts the gripper’s orientation and reduces applicability in unstructured or complex environments. To address these limitations, researchers have proposed 6-DoF grasp representations~\cite{yan2018learning} that define a grasp as a 6-dimensional pose in 3D space. This formulation enables the robot to grasp objects from arbitrary orientations, improving flexibility and robustness. The 6-DoF representation is especially beneficial in cluttered scenes, non-planar object poses, and scenarios involving complex geometries.

Apart from a few early 2D-based approaches~\cite{zhai2023monograspnet}, the majority of recent methods adopt 3D representations. In particular, methods based on point cloud inputs have explored a wide range of architectures, ranging from mapping convolutional networks~\cite{fang2023anygrasp} and transformer-based models~\cite{fan2025miscgrasp}, to generative autoencoders~\cite{yan2018learning}, variational autoencoders~\cite{mousavian20196dof, iwase2025zerograsp}, and UNet-style frameworks~\cite{xie2025rethinking}. 
To improve efficiency, various methods have been proposed, including applying instance segmentation or heatmap to focus on task-relevant manipulation regions~\cite{murali20206, chen2023efficient}, reducing the grasp search space by representing grasp poses with contact points on the object surface~\cite{sundermeyer2021contact}, incorporating graph neural networks for geometric reasoning on point cloud data~\cite{moghadam2023grasp}, and adopting an economic supervision paradigm that selects key annotations and leverages focal representation to reduce training cost and improve performance~\cite{wu2024economic}.
Some works further address structural distortions commonly found in real-world point cloud data by introducing completion or denoising modules to convert the input into a clean and consistent style~\cite{cheng2025pcf, cai2025real}. 
Several methods also address the SE(3)-equivariance problem~\cite{lim2025equigraspflow, hu2025orbitgrasp} by modeling grasp generation as a continuous normalizing flow over SE(3) with equivariant vector fields, or by predicting per-point grasp quality over the orientation sphere using spherical harmonic basis functions.
Meanwhile, several methods also leverage 3D representations such as SDF~\cite{breyer2021volumetric, song2025implicit, jauhri2024learning}, NeRF~\cite{rashid2023language, dai2023graspnerf}, and 3DGS~\cite{zheng2024gaussiangrasper, ji2025graspsplats, yu2024sparsegrasp} to extract geometric features or to sample grasp points for downstream prediction.

\noindent \textbf{Language-driven Grasp.}
In recent years, language-driven grasping has gained increasing attention, aiming to achieve object-specific and instruction-guided manipulation. Existing approaches can be broadly categorized into three groups. The first adopts multimodal feature fusion~\cite{xu2023joint, nguyen2024language}, where textual and visual modalities are jointly encoded, often via cross-attention mechanisms. The second leverages existing grasp models to generate large numbers of grasp candidates, followed by ranking or scoring with LLMs or VLMs to select the most confident grasps. For instance, LLMs can generate task-specific descriptions~\cite{tang2023graspgpt}, while VLMs are used for visual grounding to compute grasp confidence scores~\cite{qian2025thinkgrasp, tang2025affordgrasp}. Representative methods include VL-Grasp, which employs VLMs to attend to the target object and generate grasps~\cite{lu2023vl}; OWG, which leverages VLM-based semantic priors for planning under occlusion~\cite{tziafas2025towards}; and Reasoning-Grasping, which integrates visual-linguistic inference for improved object-level understanding~\cite{jin2025reasoning}. Other efforts adapt MLLMs for environment-aware error correction~\cite{luo2025roboreflect}, or learn object-centric attributes to enable rapid grasp adaptation across tasks~\cite{yang2024attribute}. Affordance-driven approaches also ground grasp generation in language and vision cues~\cite{dong2025rtagrasp, song2025learning, deshpande2025graspmolmo}. The third line of work directly fine-tunes MLLMs on grasp foundation models with large-scale grasp datasets~\cite{xu2024rt, zhang2026vcot}.

\noindent \textbf{Challenges.}
Firstly, the aforementioned grasp detection and generation methods were originally designed for 2-DoF parallel-jaw grippers. However, with the increasing use of dexterous hands, research has shifted toward dexterous grasping, which requires complex annotations such as hand joint angles and contact force maps. To handle the high dimensionality of this task, various generative approaches, including diffusion-based models, have been proposed~\cite{xu2024dexterous, weng2024dexdiffuser, wei2024grasp}. Some approaches represent grasp configurations via contact points or maps~\cite{li2023gendexgrasp}, or leverage interaction-based representations such as D(R, O)~\cite{wei2024d} to infer grasps from geometric relationships.
Second, traditional grasping strategies typically rely on a single end-effector, such as suction or parallel grippers, which limits adaptability to diverse object geometries. To address this, several works have explored bimanual grasping using dual-arm or dual-gripper configurations~\cite{mahler2019learning, yamada2025combo}.
Lastly, grasping transparent objects remains a significant challenge, as depth sensors often fail to detect or localize such materials accurately. Recent efforts address this limitation by incorporating alternative modalities, such as LiDAR~\cite{deng2025fusegrasp} or 3D reconstruction~\cite{duisterhof2024residual, shi2024asgrasp, kim2025t}, to infer the geometry of transparent or reflective objects.

\subsection{Basic Manipulation}
Basic manipulation refers to relatively simple tabletop tasks performed by single- or dual-arm manipulators, such as pick-and-place, sorting, pushing, inserting, opening, closing, and pouring. 
Most current research remains focused on this category, as illustrated in Figure~\ref{fig: cat_basic} and further discussed in Sections~\ref{sec:high_level_planning}--\ref{sec:low_level_action_modeling}, where the majority of methods and benchmarks are developed around object-centric interactions in structured environments.
While these sections classify approaches within the scope of basic manipulation, the proposed taxonomy is general in nature and can be readily extended to other categories of manipulation tasks.

\begin{table*}[t]
\centering
\caption{Representative methods across manipulation types and learning paradigms.}
\label{tab: manipulation_methods}
\vskip -0.05in
\resizebox{\textwidth}{!}{
\begin{tabular}{l
>{\centering\arraybackslash}p{3cm}
>{\centering\arraybackslash}p{5cm}
>{\centering\arraybackslash}p{5cm}
>{\centering\arraybackslash}p{5cm}
>{\centering\arraybackslash}p{5cm}
}
\toprule
Mani. Type & RL & IL & RL+IL & VLA \\
\midrule
Basic & See Table~\ref{tab: rl} & See Figure~\ref{fig: il} & See Table~\ref{tab: rl-il} & See Figure~\ref{fig: vla-taxonomy} \\
Dexterous & PDDM~\cite{nagabandi2020deep}, \cite{rajeswaran2018learning}, \cite{zhu2019dexterous} & DexHandDiff~\cite{liang2025dexhanddiff}, CordViP~\cite{fu2025cordvip} & REBOOT~\cite{hu2023reboot}, ViViDex~\cite{chen2025vividex} & OFA~\cite{li2025object}, LBM~\cite{barreiros2025careful} \\
Soft Robotics & \cite{thuruthel2018model}, \cite{li2022towards} & Soft DAgger~\cite{nazeer2023soft}, KineSoft~\cite{yoo2025kinesoft} & SS‑ILKC~\cite{meng2025sensor} & -- \\
Deformable Object & \cite{matas2018sim} & DeformerNet~\cite{hu20193}, DexDeform~\cite{li2023dexdeform} & DMfD~\cite{salhotra2022learning} &
$\chi_0$~\cite{yu2026chi0} \\
Mobile & \cite{wu2020spatial}, MoMa~\cite{hu2023causal} & MOMA-Force~\cite{yang2023moma}, Skill Transformer~\cite{huang2023skill} & \cite{xiong2024adaptive} & MoManipVLA~\cite{wu2025momanipvla} \\
Quadrupedal & VBC~\cite{liu2025visual}, GAMMA~\cite{zhang2024gamma} &
Human2LocoMan~\cite{niu2025human2locoman} & \cite{he2024learning}, WildLMa~\cite{qiu2025wildlma} & QUAR-VLA~\cite{ding2024quar}, GeRM~\cite{song2024germ} \\
Humanoid & \cite{xie2023hierarchical}, FLAM~\cite{zhang2025flam} & OmniH2O~\cite{he2025omnih2o}, iDP3~\cite{ze2025generalizable} & \cite{schakkal2025hierarchical} & GR00T N1~\cite{bjorck2025gr00t}, Humanoid-VLA~\cite{ding2025humanoid} \\
\bottomrule
\end{tabular}
}
\end{table*}

\subsection{Dexterous Manipulation}

\begin{wrapfigure}{r}{0.48\textwidth}
  \centering
  \vspace{-12pt} 
  \includegraphics[width=0.48\textwidth]{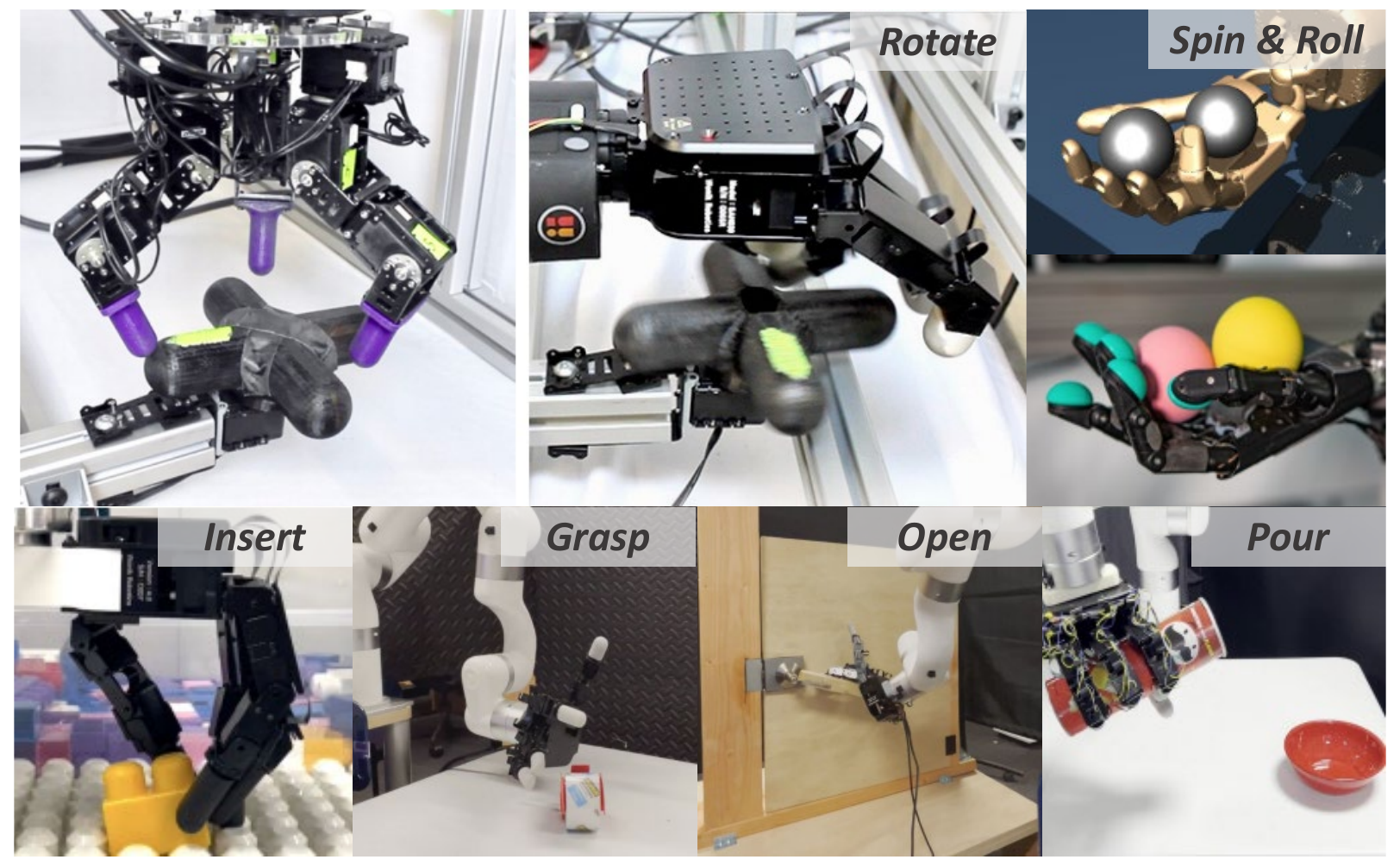}
  \caption{Tasks in dexterous manipulation, figure adapted from~\cite{zhu2019dexterous, nagabandi2020deep, qin2023dexpoint, chen2023sequential, wang2024cyberdemo}.
  }
  \label{fig: case_dex}
  \vspace{-5pt}
\end{wrapfigure}

Dexterous manipulation refers to the capability of robotic systems equipped with multi-fingered or anthropomorphic hands to achieve precise and coordinated object control through complex contact interactions. It involves in-hand reorientation, fine force modulation, and multi-point contact, enabling actions such as twisting, grasping, and rotating, as illustrated in Figure~\ref{fig: case_dex}. This capability is critical for tasks requiring high precision and adaptability, such as tool use, assembly, and manipulation of small or irregular objects. Human hand models typically include 20–25 DoF, with each finger modeled by 4 DoF, the thumb by 4–5 DoF, and additional DoF from the palm and wrist for enhanced realism~\cite{welte2025interactive}.


\noindent \textbf{Non-Learning-Based Methods.}
Early work on dexterous manipulation predominantly employed non-learning approaches, including optimization- and control-based techniques such as heuristic search and constrained optimization~\cite{bai2014dexterous}. These methods typically assume access to known dynamics and object models, and focus on trajectory planning under physical constraints.

\noindent \textbf{Learning-Based Methods.}
With the advent of RL, both model-based and model-free paradigms have been applied to dexterous manipulation. Model-based methods such as PDDM~\cite{nagabandi2020deep} exploit learned dynamics for efficient planning and control, while model-free approaches directly optimize policies through interaction~\cite{rajeswaran2018learning}, sometimes combined with few-shot imitation to accelerate adaptation to real hands~\cite{zhu2019dexterous}. More recently, human-in-the-loop frameworks such as HIL-SERL~\cite{luo2025precise} integrate demonstrations, online corrections, and reinforcement learning for sample-efficient acquisition of precise skills on physical robots. DexScrew~\cite{hsieh2026learning} further addresses imperfect simulation by first learning transferable finger-motion primitives with RL in simplified environments, then using these primitives to collect real multisensory demonstrations for tactile-conditioned behavior cloning.

IL has gained attention for its data efficiency, exemplified by DexMV~\cite{qin2022dexmv}, which employs kinematic retargeting from human videos to robot hands. Extensions incorporate auxiliary supervision such as contact prediction~\cite{liang2025dexhanddiff,fu2025cordvip} and inverse dynamics modeling~\cite{radosavovic2021state}, improving generalization and action consistency. SAT~\cite{lei2026structural} instead represents an action chunk as a variable-length sequence of joint-wise trajectories and incorporates joint-level functional and kinematic priors, enabling 3D dexterous policy learning across heterogeneous hand embodiments. Hybrid IL--RL approaches combine these strengths, typically using IL for policy initialization followed by RL refinement~\cite{kumar2016learning,arunachalam2023dexterous}. DexMV [294] instead translates human videos into robot demonstrations for imitation learning, whereas ViViDex~\cite{chen2025vividex} first performs RL in a privileged state space and subsequently distills the resulting policy through IL.

Recent VLA models expand task and embodiment generalization. DexGraspVLA~\cite{zhong2026dexgraspvla} integrates semantic reasoning with diffusion-based policies for grasping in cluttered scenes, while OFA~\cite{li2025object}, LBM~\cite{barreiros2025careful}, and Being-H0~\cite{luo2026being} leverage multimodal prompts or human videos to generate dexterous motions and follow language instructions. UniDex~\cite{zhang2026unidex} transforms egocentric human videos into robot-centric trajectories and introduces a function-aligned action space to support unified control across heterogeneous dexterous hands. Dexora~\cite{zhang2026dexora} extends VLA modeling to high-dimensional dual-arm and dual-hand manipulation, combining embodiment-matched simulated and real demonstrations with quality-aware training to reduce the influence of noisy teleoperation data.

Human guidance provides another supervision source: Chen et al.~\cite{chen2025object} use object and wrist trajectories from videos to guide RL, and Mandi et al.~\cite{mandi2025dexmachina} introduce functional retargeting to transition from demonstrations to autonomous control. Affordance reasoning further supports grasp-specific strategies~\cite{kannan2023deft, agarwal2023dexterous}, guiding object selection under semantic constraints.
Finally, recent works emphasize robustness and autonomy. Task decomposition reduces complexity by structuring subtasks~\cite{dasari2023learning}, while recovery mechanisms such as REBOOT~\cite{hu2023reboot} introduce IL-trained reset policies to handle failures in long-horizon settings.

\noindent \textbf{Challenges.} 
Beyond high-dimensional control and contact dynamics, dexterous manipulation faces broader challenges.
First, many real-world tasks are long-horizon and compositional, requiring sequential execution of fine-grained skills. Chen et al.~\cite{chen2023sequential} address this by decomposing tasks into discrete skills trained with RL and linking them via a high-level policy.
Second, sim-to-real transfer remains a bottleneck due to discrepancies in perception, dynamics, and actuation. CyberDemo~\cite{wang2024cyberdemo} mitigates this through data augmentation, improving robustness under domain shift.
Third, accurate perception is difficult in cluttered or partially observed scenes, where occlusion hampers object tracking. DexPoint~\cite{qin2023dexpoint} leverages point cloud completion to recover missing geometry and enhance spatial awareness.

\subsection{Soft Robotic Manipulation}

\begin{wrapfigure}[14]{r}{0.48\textwidth}
  \centering
  \vspace{-12pt} 
  \includegraphics[width=0.48\textwidth]{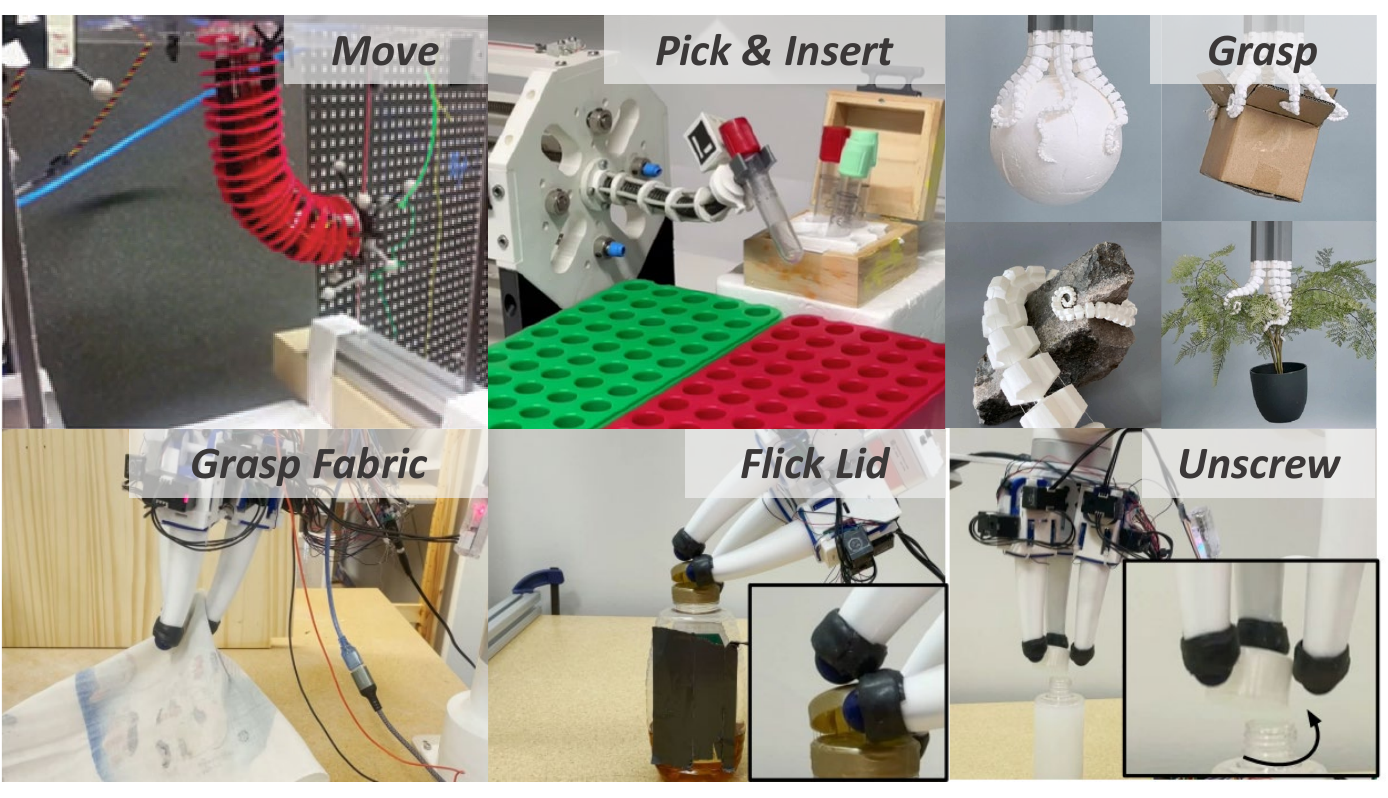}
  \caption{Tasks in soft robotic manipulation, figure adapted from~\cite{you2017model, yoo2025kinesoft, wang2025spirobs}.
  }
  \label{fig: case_soft}
  \vspace{-5pt}
\end{wrapfigure}

Soft manipulators, built with compliant materials or structures, are well-suited for human–robot collaboration, operation in uncertain environments, and safe, adaptive grasping, as illustrated in Figure~\ref{fig: case_soft}. They overcome the limitations of rigid manipulators when handling delicate or deformable objects~\cite{chen2022review}. To this end, a wide range of designs have been developed to support diverse applications~\cite{ansari2017towards, thuruthel2018stable, puhlmann2022rbo, liu2022touchless, wang2025spirobs}.

\noindent \textbf{Non-Learning-Based Methods.}
For soft manipulator control, analytical kinematic models based on the constant-curvature assumption map shape parameters to end-effector poses via virtual rigid-link chains and Denavit–Hartenberg transformations~\cite{jones2006kinematics}. Similarly, three-dimensional continuum manipulators can be approximated as rigid-jointed robots, with computed-torque control achieving closed-loop dynamics~\cite{wang2020dynamic}. Other approaches employ model predictive control with Koopman operators~\cite{bruder2019modeling, lv2025multi}, or rely on accurate Lagrangian models combined with adaptive dynamic sliding mode control to enhance robustness~\cite{kazemipour2022adaptive}. Hybrid schemes also integrate learning into non-learning frameworks: forward dynamics can be approximated with machine learning and combined with trajectory optimization for open-loop control~\cite{thuruthel2017learning}, while feedback-driven strategies exploit learned models to stabilize and restore system states~\cite{thuruthel2018stable}.


\textbf{Learning-based Methods.} 
RL and IL have been widely applied to soft manipulator control. 
Model-free RL avoids explicit physical modeling by training policies in simulation for closed-loop endpoint control~\cite{you2017model}. Forward dynamics learned with recurrent networks can be integrated into trajectory optimization to generate samples and train predictive controllers~\cite{thuruthel2018model}, while LSTM-based dynamics models enable feedback policy learning~\cite{centurelli2022closed}. Domain randomization combined with incremental offline training has improved task-space accuracy and adaptability~\cite{li2022towards}, and wavelet-based dynamics approximations paired with LSTM and TD3 controllers further enhance generalization under variability~\cite{zhou2024cable}.
IL approaches include Soft DAgger~\cite{nazeer2023soft}, which employs dynamic behavior mapping for online expert-like action generation, and KineSoft~\cite{yoo2025kinesoft}, which integrates kinesthetic teaching with diffusion-based policy learning. Hybrid methods also emerge, such as SS-ILKC~\cite{meng2025sensor}, which combines multi-objective RL with adversarial IL to learn goal-directed control strategies in sensor space, supported by sim-to-real pre-calibration for zero-shot transfer.

\noindent \textbf{Challenges.} 
Soft-hand-based manipulation faces several challenges, including modeling highly nonlinear and underactuated dynamics, achieving precise force and pose control with deformable or fragile objects, ensuring robustness under environmental uncertainty and sensor noise, and lowering the cost and complexity of data collection and teleoperation for policy training. To improve sample efficiency, Soft DAgger~\cite{nazeer2023soft} enables online imitation learning from limited demonstrations through dynamic behavior mapping. To reduce hardware complexity in teleoperation, Liu et al.~\cite{liu2022touchless} propose a flexible bimodal sensory interface that combines vision-based perception with wearable sensors, enabling intuitive and low-cost control without bulky equipment.

\subsection{Deformable Object Manipulation}

\begin{wrapfigure}{r}{0.48\textwidth}
  \centering
  \vspace{-15pt} 
  \includegraphics[width=0.48\textwidth]{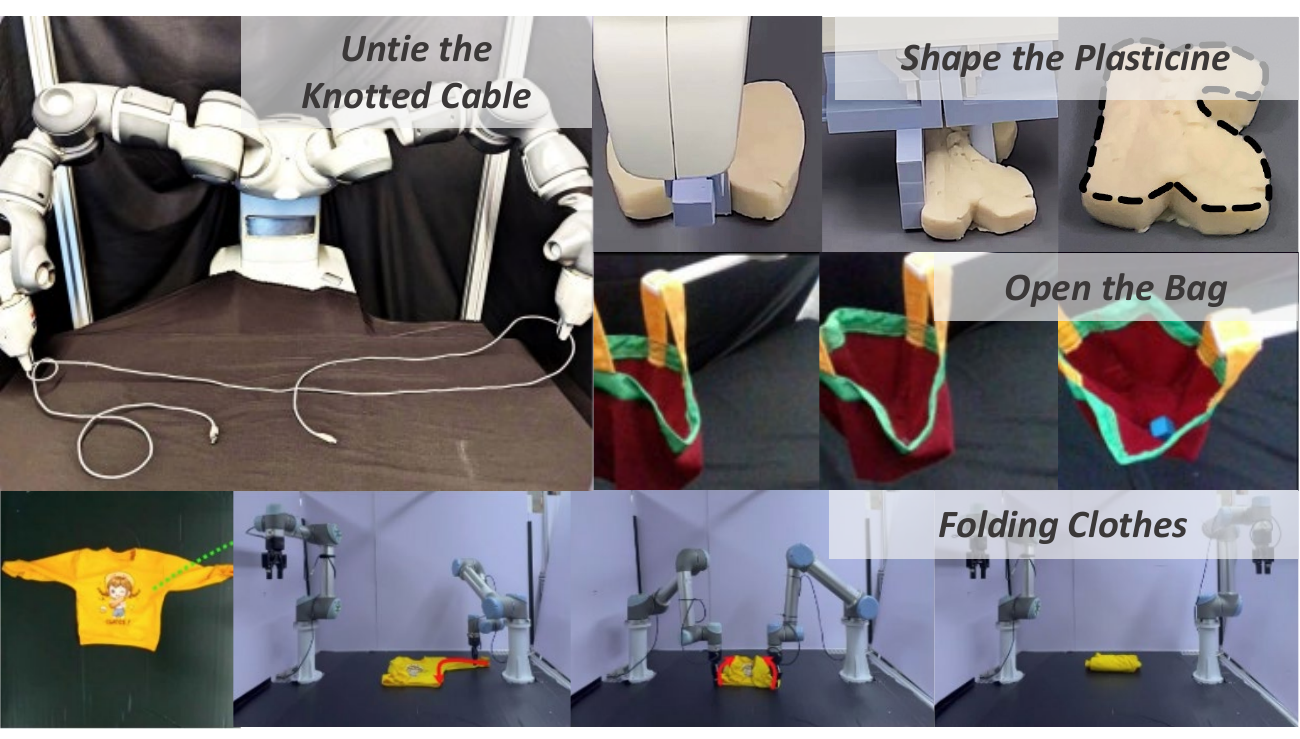}
  \caption{Tasks in deformable object manipulation, figure adapted from~\cite{shi2023robocook, viswanath2023handloom, weng2024interactive, luan2025learning}.
  }
  \label{fig: case_deformable}
  \vspace{-5pt}
\end{wrapfigure}

Deformable object manipulation (DOM) requires robots to perceive and control non-rigid objects whose shapes vary under applied forces. Unlike rigid-body manipulation, it demands reasoning over high-dimensional, continuous state spaces, coping with uncertain deformation dynamics, and interpreting subtle visual or tactile cues. As illustrated in Figure~\ref{fig: case_deformable}, tasks such as cloth folding, rope and cable tying, and food handling involve diverse deformations—including tension, compression, and bending—making DOM a complex yet crucial challenge in robotic manipulation~\cite{sanchez2018robotic, blanco2024t, gu2023survey}.

\noindent \textbf{Non-Learning-Based Methods.}
Traditional approaches to DOM, such as path planning~\cite{saha2007manipulation} and model-based control~\cite{luque2024model}, often rely on simplified physical models or analytical solutions. However, these methods typically struggle with generalization and real-time adaptability in complex environments, leading to a shift toward data-driven techniques such as RL, IL, and hybrid learning-control frameworks.


\noindent \textbf{Learning-based Methods.}
Jan et al.~\cite{matas2018sim} propose a deep RL method based on a modified DDPG algorithm that learns from visual and robot state inputs and achieves zero-shot sim-to-real transfer via domain randomization. In contrast, IL methods exploit demonstrations: DexDeform~\cite{li2023dexdeform} extracts latent skills from human demonstrations and fine-tunes them with limited robot data, DefGoalNet~\cite{thach2024defgoalnet} predicts goal configurations from few-shot demonstrations conditioned on state and context, and MPD~\cite{scheikl2024movement} generates movement primitives with diffusion models. To combine both paradigms, DMfD~\cite{salhotra2022learning} incorporates expert data into RL through advantage-weighted BC loss, expert-initialized replay buffers, and reset-to-state initialization, achieving efficient policy optimization.

Beyond learning paradigms, DOM has explored diverse strategies spanning geometric modeling, affordance prediction, and structure-aware perception. DeformGS~\cite{duisterhof2024deformgs} represents deformable objects with canonical Gaussians and learns a time-conditioned deformation field for temporally consistent 3D pose tracking. Affordance-based approaches include DeformerNet~\cite{hu20193}, which predicts end-effector displacements from partial point clouds, Foresightful Affordance~\cite{wu2023learning}, which integrates dense per-pixel affordances with long-horizon value estimation, and APS-Net~\cite{luan2025learning}, which ranks standardized folding and flattening trajectories guided by affordance heatmaps. Language-conditioned affordance prediction has also been explored~\cite{deng2024learning}.
Finally, estimating object-specific physical properties has emerged as a complementary direction~\cite{caporali2024deformable, kuroki2024gendom}. GenDOM~\cite{kuroki2024gendom} learns a parameter-conditioned policy and leverages a single human demonstration with differentiable simulation to infer unseen object properties, enabling generalization to novel deformable instances.

\noindent \textbf{Challenges.}
Deformable objects lack a fixed pose, and key deformation regions are often occluded during manipulation. To improve visibility and perception, recent work jointly optimizes camera and manipulator motion, guided by structure-of-interest cues~\cite{weng2024interactive}. Deformation dynamics also exhibit delayed responses, complicating real-time control; this is addressed by encoding states into latent spaces and predicting their dynamics. For example, DeformNet~\cite{li2024deformnet} encodes object geometry with PointNet and a conditional NeRF, and models temporal evolution with a recurrent state-space model for latent-level MPC. An even greater challenge is modeling topological changes. DoughNet~\cite{bauer2024doughnet} extends latent-space modeling to jointly capture geometric and topological variations from point clouds, enabling long-horizon planning with CEM for tool selection and manipulation.

\subsection{Mobile Manipulation}

\begin{wrapfigure}{r}{0.48\textwidth}
  \centering
  \vspace{-12pt} 
  \includegraphics[width=0.48\textwidth]{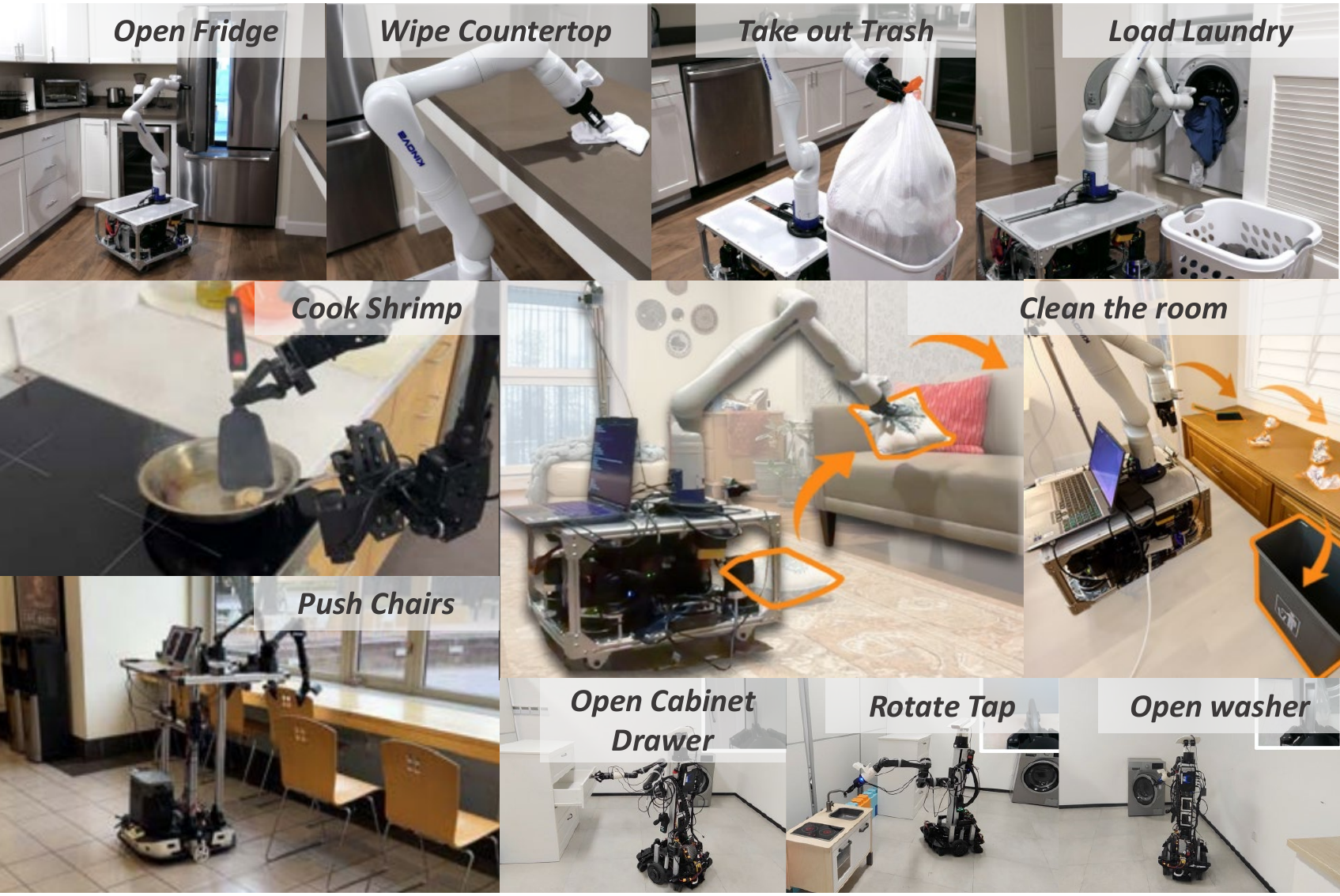}
  \caption{Tasks in mobile manipulation, figure adapted from~\cite{yang2023moma, fu2025mobile, wu2025tidybot++, sundaresan2025homer}.
  }
  \label{fig: case_mobile}
  \vspace{-5pt}
\end{wrapfigure}

Mobile manipulation is a robotic paradigm that combines navigation and manipulation capabilities within a single system. It allows robots to physically interact with objects beyond a fixed workspace by actively navigating the environment. This integration poses significant challenges in perception, planning, and control, as it requires coordinating whole-body motion, handling long-horizon tasks, and dealing with dynamic, partially observable scenes. 
Several studies have focused on designing robots specifically for mobile manipulation~\cite{holmberg2000development, hebert2015mobile, fu2025mobile}.
As illustrated in Figure~\ref{fig: case_mobile}, mobile manipulation is critical for real-world applications such as household assistance, warehouse automation, and service robotics~\cite{thakar2023survey}.

\noindent \textbf{Non-Learning-Based Methods.}
Traditionally, mobile manipulation is decomposed into two separate problems: navigation and manipulation. Navigation is typically handled by classic path planners, such as grid-based methods (e.g., Dijkstra) and sampling-based planners (e.g., RRT). Manipulation is often addressed using grasp models, motion primitives, or trajectory planning based on point clouds.
Some control methods perform joint optimization of navigation and manipulation. For example, Berenson et al.~\cite{berenson2008optimization} optimize full-body configurations and grasp poses simultaneously, then plan motions via sampling. Chitta et al.~\cite{chitta2012mobile} coordinate base-arm planning with ROS-based navigation and point cloud grasping. Others embed manipulation goals directly into MPC cost functions~\cite{pankert2020perceptive}.


\noindent \textbf{Learning-based Methods.}
These methods have emerged to learn policies for whole-body control. 
For RL, Wang et al.~\cite{wang2020learning} use RGB-D inputs to estimate object pose and train a policy that jointly predicts base velocity, arm trajectories, and gripper actions. HarmonicMM~\cite{yang2024harmonic} extends this by integrating visual and pose information into a unified RL framework, while Wu et al.~\cite{wu2020spatial} propose spatial Q-value learning at the map level for navigation guidance. To improve reward signals, Honerkamp et al.~\cite{honerkamp2021learning} design dense rewards based on end-effector reachability, and Causal MoMa~\cite{hu2023causal} introduces causality-aware modeling of control–reward relations to stabilize policy optimization.
For IL, MOMA-Force~\cite{yang2023moma} learns motion and force policies from visual-force demonstrations, while HoMeR~\cite{sundaresan2025homer} decomposes tasks into global keypose prediction and local refinement to map end-effector targets to whole-body actions. Wang et al.~\cite{zhicheng2025object} leverage SAM2-based perception for object-centric IL, and Skill Transformer~\cite{huang2023skill} treats manipulation as a skill prediction problem, learning both skill categories and corresponding low-level actions.
Hybrid methods combine IL with RL to enhance generalization. For example, Xiong et al.~\cite{xiong2024adaptive} initialize policies via behavior cloning and refine them through online RL to handle unseen articulated objects.

Beyond learning paradigms, recent work integrates high-level reasoning and multimodal representations. VLA-based methods such as MoManipVLA~\cite{wu2025momanipvla} process multimodal instructions to predict end-effector waypoints while delegating base motion to a trajectory optimizer. LLM-driven frameworks, including MoMa-LLM~\cite{honerkamp2024language} and SayPlan~\cite{rana2023sayplan}, combine open-vocabulary language, scene graphs, and reasoning for object search and high-level task planning. Complementary efforts in \textit{3D scene modeling} guide manipulation through active perception: ActPerMoMa~\cite{jauhri2024active} optimizes viewpoint selection and grasp reachability via incremental TSDF mapping, while TaMMa~\cite{hou2025tamma} employs sparse Gaussian localization and depth completion to generate accurate target poses.

\noindent \textbf{Challenges.}
In addition to perception–action coordination, dynamic obstacles, and long-horizon decision-making, mobile manipulation faces several additional challenges.
A first challenge is enabling effective human–robot collaboration. Ciocarlie et al.~\cite{ciocarlie2012mobile} developed an assistive system that combines user interfaces, shared control, and autonomy to transform a PR2 robot into an in-home assistant. Extending this line of work, Robi Butler~\cite{xiao2025robi} introduces a closed-loop household control framework that supports multimodal interaction, where high-level planners such as LLMs and VLMs enable natural language and gesture-based commands for intuitive and remote collaboration.
A second challenge is real-time manipulation during navigation, where robots must coordinate mobility and manipulation under environmental constraints. Whole-body motion control frameworks~\cite{haviland2022holistic, burgess2023architecture} address this problem by enabling reactive planning and execution for on-the-move manipulation.

\subsection{Quadrupedal Manipulation}

\begin{wrapfigure}{r}{0.48\textwidth}
  \centering
  \vspace{-12pt} 
  \includegraphics[width=0.48\textwidth]{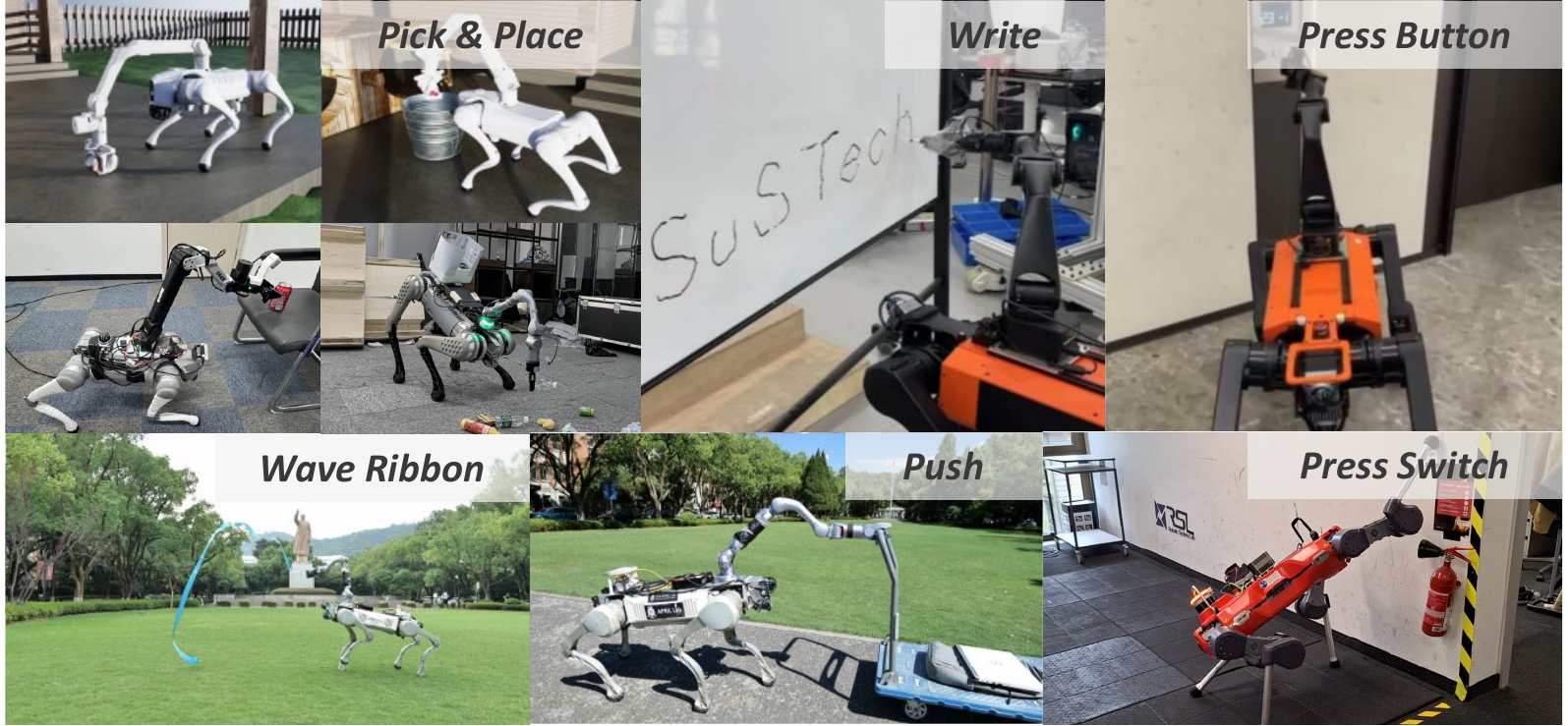}
  \caption{Tasks in quadrupedal manipulation, figure adapted from~\cite{jiang2024learning, arm2024pedipulate, hou2025efficient, wang2025quadwbg, wang2026odyssey}.
  }
  \label{fig: case_quad}
  \vspace{-5pt}
\end{wrapfigure}

Quadrupedal manipulation is an emerging paradigm that combines the agile mobility of quadruped robots with the ability to physically interact with objects. Unlike traditional manipulators or mobile bases, quadrupeds can traverse complex and unstructured terrains while maintaining dynamic stability, making them particularly suitable for applications such as search and rescue, field robotics, and autonomous exploration. Representative tasks are illustrated in Figure~\ref{fig: case_quad}.
Manipulation can be realized through several embodiments: whole-body loco-manipulation~\cite{jeon2023learning, huang2024hilma, ouyang2025long, ding2024quar, song2024germ, zhao2025more}; leg-as-manipulator designs that repurpose one or more legs for interaction~\cite{cheng2023legs, arm2024pedipulate, he2024learning}; back-mounted arms~\cite{wolfslag2020optimisation, fu2023deep, arcari2023bayesian, ferrolho2023roloma, yokoyama2023asc, qiu2025learning, mendonca2025continuously, zhang2024gamma, liu2025visual, jiang2024learning, wang2025quadwbg, pan2025roboduet, zhi2025learning, hou2025efficient}; and grippers integrated into front legs for simultaneous locomotion and manipulation~\cite{lin2024locoman}. Each embodiment introduces unique challenges in whole-body coordination, dynamic control, and perception-driven planning.


\noindent \textbf{Non-Learning-Based Methods.}
Recent advances in quadrupedal manipulation have explored diverse control strategies, ranging from model-based optimization to learning-based approaches. Optimization-based methods provide interpretable and physically grounded control with strong task generalization. For example, Arcari et al.~\cite{arcari2023bayesian} combine MPC with Bayesian multi-task error learning for real-time dynamics adaptation. Wolfslag et al.~\cite{wolfslag2020optimisation} incorporate SUF stability metrics and contact constraints into a quadratic programming framework for robust, support-leg-aware planning, a concept further extended by RoLoMa~\cite{ferrolho2023roloma} to improve trajectory robustness. LocoMan~\cite{lin2024locoman} demonstrates a hardware–control co-design approach, equipping quadrupeds with lightweight front-leg manipulators and employing unified WBC to achieve agile locomotion and precise manipulation.

\noindent \textbf{Learning-based Methods.}
In the realm of RL, recent research has focused on developing structured and informative control representations. Jeon et al.~\cite{jeon2023learning} introduce a hierarchical RL framework that encodes interaction experience, robot morphology, and action history into latent representations to facilitate effective policy learning. Fu et al.~\cite{fu2023deep} decouple leg and arm rewards in the policy-gradient formulation to enhance coordination between locomotion and manipulation, while Zhi et al.~\cite{zhi2025learning} present a unified force-position controller that estimates forces from perception instead of sensors. Hou et al.~\cite{hou2025efficient} incorporate explicit arm kinematics and feasibility-based rewards to promote physically plausible behaviors. Two-stage training schemes have also been explored: RoboDuet~\cite{pan2025roboduet} sequentially trains locomotion and arm policies with reward adaptation for coordination. GAMMA~\cite{zhang2024gamma} improves grasp precision by conditioning policies on grasp poses, while Wang et al.~\cite{wang2025quadwbg} propose GORM, a metric for grasp reachability under varying base poses, to guide base movement for effective grasping. Teacher-student frameworks further advance base-arm coordination, as in VBC~\cite{liu2025visual} and Jiang et al.~\cite{jiang2024learning}, where visual or pose-tracking guidance is distilled into student policies. SLIM~\cite{zhang2025learning} extends this paradigm to long-horizon pick-and-place by progressively expanding a privileged teacher across task stages and distilling it into an egocentric visuomotor policy through RL. Other works combine hierarchical and hybrid designs, such as HiLMa-Res~\cite{huang2024hilma}, which uses high-level RL to control Bézier parameters and base motion while relying on hybrid CPG-Bézier controllers for leg movements, and Pedipulate~\cite{arm2024pedipulate}, which demonstrates end-to-end RL for single-foot manipulation. HeLoM~\cite{yang2026helom} further develops hierarchical whole-body RL for hexapod robots, using a high-level planner to specify pushing commands and foreleg targets and a low-level controller to coordinate locomotion, balance, and forceful object interaction.

In the IL domain, Human2LocoMan~\cite{niu2025human2locoman} enables cross-embodiment transfer from XR-driven human demonstrations using a modular Transformer policy, effectively transferring human manipulation skills to quadrupeds. Hybrid IL--RL frameworks further improve efficiency and generalization. He et al.~\cite{he2024learning} employ BC to train a high-level planner for grasp trajectories, while a low-level RL controller coordinates locomotion and single-leg manipulation. WildLMa~\cite{qiu2025wildlma} builds an IL-based skill library from VR demonstrations and sequences these skills under LLM guidance.

VLA-based frameworks have also emerged for high-level semantic reasoning. QUAR-VLA~\cite{ding2024quar} pioneers the application of VLA models to quadrupeds, while GeRM~\cite{song2024germ} and MoRE~\cite{zhao2025more} integrate mixture-of-experts architectures with offline RL to learn generalist visuomotor policies and Q-functions with enhanced generalization and decision quality. QUART-Online~\cite{tong2025quart} advances this line by introducing action-chunk discretization and semantic alignment training, enabling latency-free inference for quadrupedal VLA tasks.

Finally, advances in 3D semantic perception and high-level planning extend quadrupedal capabilities. GeFF~\cite{qiu2025learning} performs real-time NeRF-based reconstruction with semantic relevance fields to guide locomotion, while manipulation is executed via learned grasp models. At the task level, LLMs have been combined with policy libraries: Ouyang et al.~\cite{ouyang2025long} propose a hierarchical framework where LLMs parse long-horizon, multi-skill instructions into structured subgoals executed by RL-based skills, bridging symbolic reasoning and continuous control.

\noindent \textbf{Challenges.}
In addition to common challenges such as balance–manipulation coupling, terrain adaptability, and perception delays, real-world deployment remains difficult due to the sim-to-real gap caused by modeling inaccuracies and sensor noise. To mitigate this, fully autonomous RL pipelines have been developed. Mendonca et al.~\cite{mendonca2025continuously} propose a framework that integrates on-policy data collection with continuous training, while ASC~\cite{yokoyama2023asc} addresses long-horizon tasks by decomposing them into modular skills trained via RL, coordinating them through a skill-switching policy jointly trained with IL and RL, and introducing a corrective policy to improve robustness during deployment.
Long-horizon task execution itself poses another significant challenge. Cheng et al.~\cite{cheng2023legs} propose a stage-wise RL framework that independently learns locomotion and single-leg manipulation skills, which are later composed via a behavior tree to accomplish temporally extended tasks.

\subsection{Humanoid Manipulation}


\begin{wrapfigure}{r}{0.48\textwidth}
  \centering
  \vspace{-13pt} 
  \includegraphics[width=0.48\textwidth]{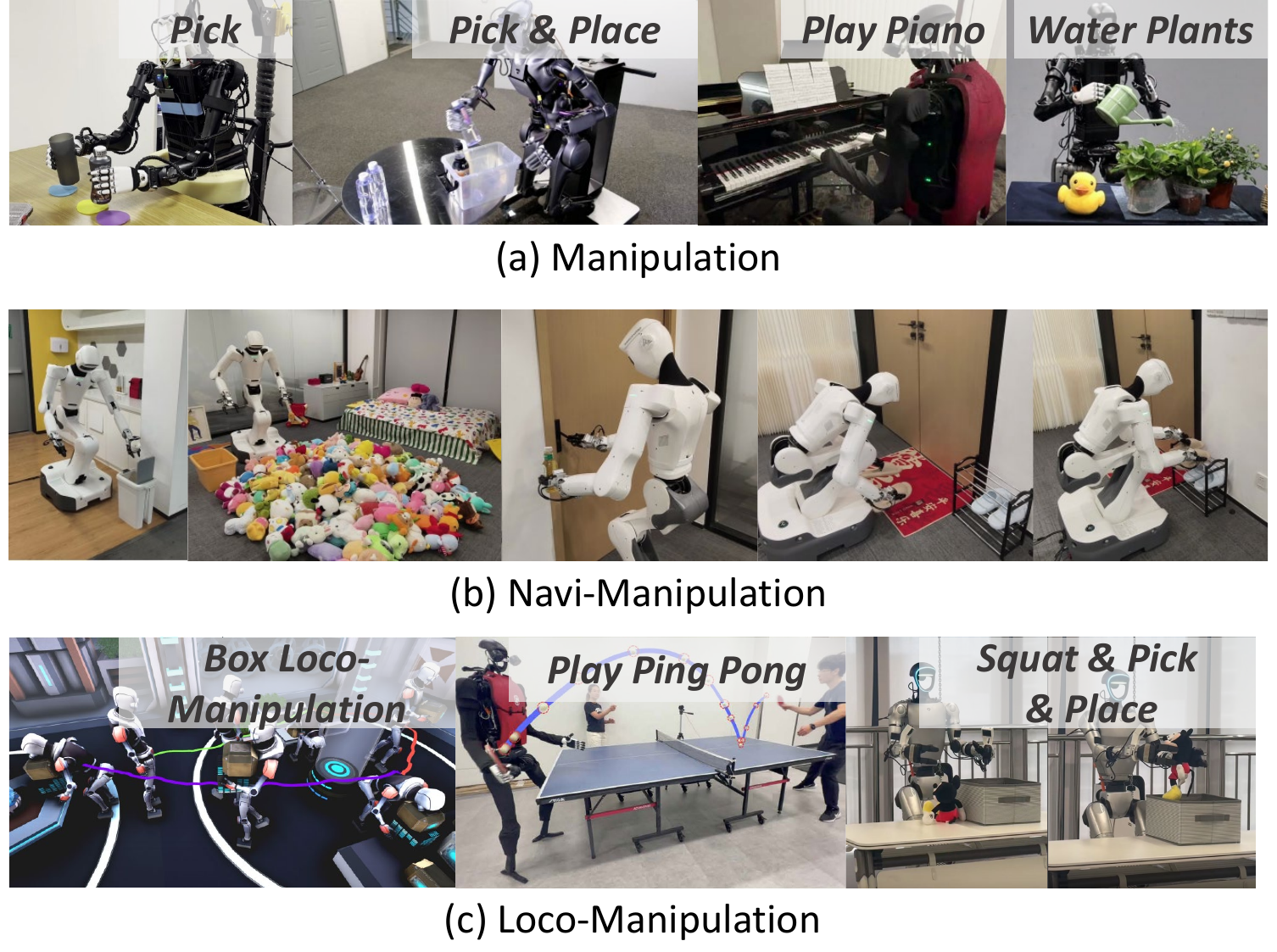}
  \caption{Tasks in humanoid manipulation, categorized into manipulation~\cite{xie2023hierarchical, fu2025humanplus, ze2025generalizable}, navi-manipulation~\cite{gao2025towards}, and loco-manipulation~\cite{qiu2025humanoid, fu2025humanplus, liu2026trajbooster}.
  }
  \label{fig: case_human}
  \vspace{-5pt}
\end{wrapfigure}

Humanoid manipulation involves robotic platforms with human-like morphology, typically comprising a torso, two arms, and either simplified 2-DoF grippers or fully dexterous hands, with locomotion provided by a mobile base or bipedal legs. Designed for object interaction in human-centered environments, these systems aim to replicate or extend human manipulation capabilities through tasks such as grasping, lifting, tool use, and coordinated bimanual operation, as illustrated in Figure~\ref{fig: case_human}. Although humanoid morphology offers natural compatibility with tools, workspaces, and infrastructures designed for humans, it also introduces substantial challenges in balance maintenance, whole-body coordination, and fine motor control, making humanoid manipulation a central topic in robotics and embodied intelligence.

\noindent \textbf{Non-Learning-Based Methods.}
Early humanoid control primarily relied on traditional approaches such as rule-based and analytical controllers~\cite{harada2006dynamics, bouyarmane2012humanoid}. Some works also explored the integration of learning-based models into control frameworks. For example, OKAMI~\cite{li2025okami} leverages 3D human pose estimation from a single human video to infer joint positions and derive executable actions for humanoid robots via inverse kinematics.

\noindent \textbf{Learning-Based Methods.}
Recent humanoid manipulation research has shifted from hand-engineered control stacks toward learning-based systems spanning RL, IL, hybrid RL--IL, VLA, and world-action-model paradigms~\cite{gu2025humanoid}. In RL, FLAM~\cite{zhang2025flam} uses a pretrained human-motion model to score pose stability as an auxiliary reward, while Xie et al.~\cite{xie2023hierarchical} combine diffusion-based motion generation with RL for whole-body manipulation. OmniRetarget~\cite{yang2026omniretarget} further preserves robot--object and robot--environment interactions when retargeting human demonstrations into kinematic references for RL. In IL, TRILL~\cite{seo2023deep} maps RGB observations to pose commands, OmniH2O~\cite{he2025omnih2o} unifies goals from language, vision, motion capture, and VR, and iDP3~\cite{ze2025generalizable} adapts diffusion policies to egocentric whole-body control. Other methods leverage large-scale egocentric demonstrations~\cite{qiu2025humanoid}, tactile feedback~\cite{murooka2025tact}, or data augmentation from a single demonstration~\cite{fu2026demohlm}. MimicDroid~\cite{shah2026mimicdroid} meta-trains an in-context policy from human play videos for few-shot adaptation. Hybrid RL--IL methods combine efficient supervision with feasible execution, as in André et al.~\cite{schakkal2025hierarchical}, who train a mid-level trajectory generator with IL and a low-level whole-body tracker with RL.

VLA methods jointly ground language and visual observations in coordinated locomotion and manipulation. HumanVLA~\cite{xu2024humanvla} separately encodes images and language, while GR00T N1~\cite{bjorck2025gr00t} fuses visual and tokenized language features through a VLM and predicts actions with a diffusion head. Humanoid-VLA~\cite{ding2025humanoid} aligns language and action representations via cross-attention. TrajBooster~\cite{liu2026trajbooster} uses dual-arm end-effector trajectories to transfer wheeled-humanoid demonstrations to bipedal whole-body control. WholeBodyVLA~\cite{jiang2026wholebodyvla} learns latent representations from action-free egocentric videos and decodes them into arm and locomotion commands. $\Psi_0$~\cite{wei2026psi0} adopts staged training with egocentric videos and humanoid trajectories, while HEX~\cite{bai2026hex} leverages humanoid-aligned states and expert-routed proprioceptive prediction for cross-embodiment learning. OpenHLM~\cite{hu2026openhlm} further studies teleoperation interfaces, action-space design, and heterogeneous co-training for whole-body-native VLAs.
Beyond conventional VLAs, MotionWAM~\cite{zheng2026motionwam} incorporates intermediate denoising features from a video world model and predicts unified whole-body motion tokens spanning locomotion, torso motion, height regulation, foot interaction, and hand manipulation. This direction seeks to combine visual dynamics priors with real-time whole-body action generation.

\noindent \textbf{Challenges.}
In addition to challenges such as multimodal perception, balance–manipulation coupling, and high degrees of freedom, humanoid manipulation also struggles with enabling multi-agent collaboration. For instance, CooHOI~\cite{gao2024coohoi} exploits object dynamics as an implicit communication channel to achieve coordinated manipulation across multiple agents. To further address the sim-to-real gap, Lin et al.~\cite{lin2025sim} introduce a generalizable reward formulation and a decoupled RL architecture, which improve sample efficiency through staged training.

\subsection{Aerial Manipulation}
\label{subsec:aerial_manipulation}

Aerial manipulation combines the mobility of unmanned aerial vehicles with robotic end-effectors, such as grippers or articulated arms, to enable physical interaction beyond the reach of ground-based platforms. Representative tasks include aerial grasping, object transportation, pick-and-place, tool use, surface interaction, and infrastructure maintenance. Unlike fixed-base manipulation, aerial manipulation must coordinate flight and manipulation while maintaining stability under arm motion, contact forces, and payload changes. Its floating-base dynamics, limited onboard sensing and computation, and strict payload constraints therefore create distinctive challenges for perception, planning, and action modeling.

\begin{wrapfigure}{r}{0.48\textwidth}
  \centering
  \vspace{-12pt} 
  \includegraphics[width=0.48\textwidth]{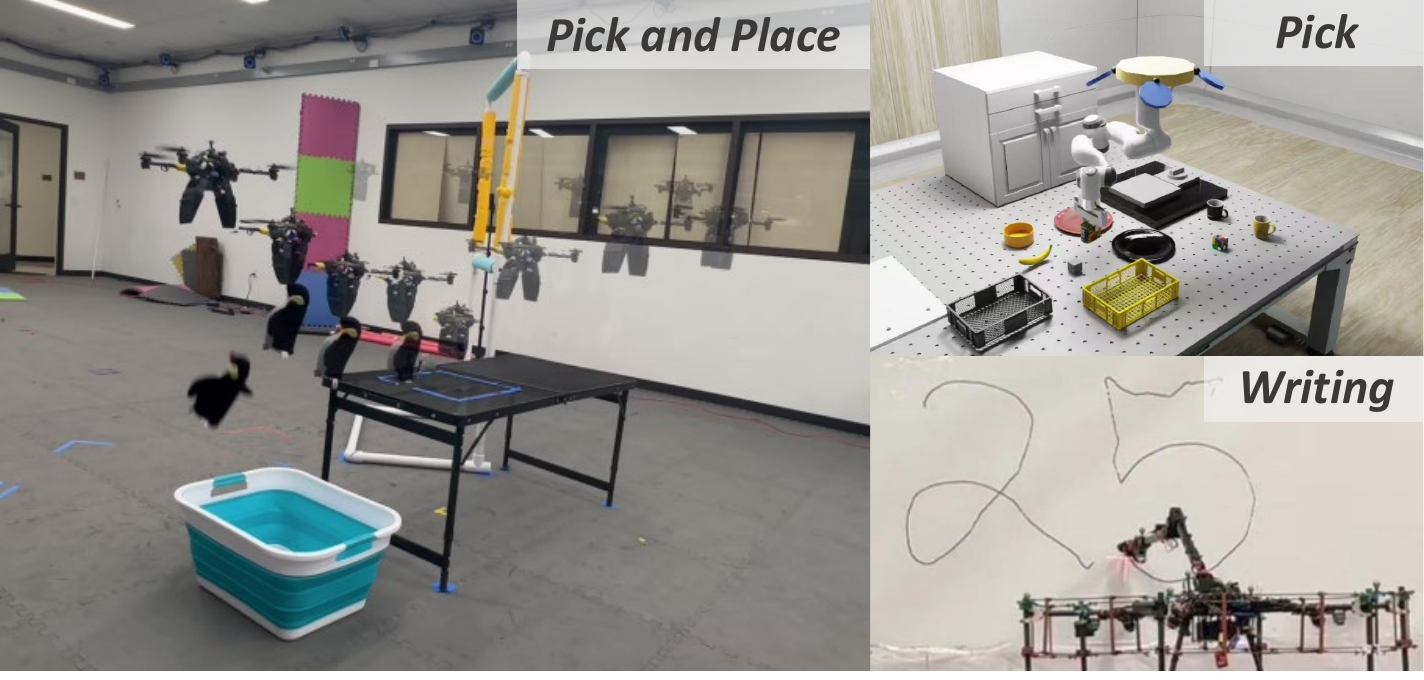}
  \caption{Tasks in aerial manipulation, figure adapted from~\cite{tucker2026pi, sun2026air, he2025flying}.
  }
  \label{fig: case_aerial}
  \vspace{-5pt}
\end{wrapfigure}

\noindent \textbf{Non-Learning-Based Methods.}
Early aerial manipulation systems primarily relied on analytical modeling, motion planning, and model-based control. Pounds et al.~\cite{pounds2011grasping} studied hovering capture and load stability during aerial grasping, while Kim et al.~\cite{kim2013aerial} modeled a quadrotor and a 2-DoF arm as a coupled system and designed an adaptive sliding-mode controller for autonomous object transportation. Orsag et al.~\cite{orsag2017dexterous} further organized aerial manipulation according to momentary, loose, and strong environmental coupling, demonstrating tasks such as pick-and-place, insertion, and valve operation. More recent systems extend these foundations toward greater workspace and robustness. Lee et al.~\cite{lee2026autonomous} combine geometric robust control with optimization-based whole-body planning, allowing an omnidirectional aerial manipulator to operate at arbitrary poses in $\mathrm{SE}(3)$. FlyAware~\cite{ye2026flyaware} instead addresses payload-dependent dynamics through vision-based pre-grasp inertia estimation and post-grasp adaptive control.

\noindent \textbf{Learning-Based Methods.}
Learning-based aerial manipulation has gradually expanded from task-specific RL and demonstration-based learning toward general-purpose VLA systems. In RL, Dimmig and Kobilarov~\cite{dimmig2024non} jointly learn a world model and interaction policy for non-prehensile aerial manipulation under unknown object dynamics. Swooper~\cite{huang2026swooper} adopts a two-stage deep RL strategy that first learns flight control and subsequently coordinates high-speed flight with active gripper control using a unified lightweight policy.
Demonstration-based approaches reduce the cost and risk of direct interaction during policy acquisition. Zito and Ferrante~\cite{zito2022one} learn transferable contact distributions from a single demonstration to identify attachment points on unseen payloads. Flying Hand~\cite{he2025flying} introduces an end-effector-centric interface that combines whole-body model predictive control with intuitive teleoperation, enabling imitation policies to learn diverse tasks including writing, insertion, pick-and-place, and tool interaction.

VLA models seek to unify semantic understanding, long-horizon reasoning, flight, and manipulation. AIR-VLA~\cite{sun2026air} establishes an aerial-manipulation benchmark and multimodal dataset covering spatial understanding, base motion, manipulator control, and multi-stage task execution. AirVLA~\cite{tucker2026pi} investigates the transfer of a manipulation-pretrained $\pi_0$ model to aerial pick-and-place, using synthetic navigation data and payload-aware guidance to address data scarcity and the mismatch between fixed-base and aerial dynamics. AIR-VLA+~\cite{sun2026airvlaplus} separates UAV movement from arm manipulation through cascaded action decoders and an asymmetric mixture-of-experts architecture, reducing interference between coarse platform motion and precise end-effector control.

\noindent \textbf{Challenges.}
Aerial manipulation remains constrained by strong coupling between flight and arm motion, time-varying payload dynamics, limited sensing and onboard computation, and the destabilizing effects of physical contact. Learning-based methods additionally face scarce demonstrations, costly and potentially unsafe exploration, and substantial sim-to-real discrepancies in aerodynamics and contact behavior. For VLA systems, a central challenge is coordinating semantically driven platform movement with precise manipulation despite their different action scales and dynamics. Future progress will require safer data collection, more accurate payload and contact estimation, embodiment-aware action representations, and tighter integration of high-level reasoning with real-time stability guarantees.

\subsection{Underwater Manipulation}
\label{subsec:underwater_manipulation}

Underwater manipulation involves robotic platforms equipped with one or more arms or grippers to physically interact with submerged objects, structures, and environments. Representative tasks include object retrieval, valve operation, sampling, inspection, maintenance, and archaeological intervention. Compared with terrestrial manipulation, underwater systems must account for coupled vehicle--manipulator dynamics, buoyancy, hydrodynamic drag, external currents, degraded visibility, and limited communication bandwidth. These factors complicate perception, contact regulation, and coordinated motion while placing stringent requirements on operational reliability and safety.

\noindent \textbf{Non-Learning-Based Methods.}
Traditional underwater manipulation primarily relies on model-based control, task-priority optimization, and human teleoperation. The TRIDENT framework~\cite{simetti2014floating} coordinates a floating vehicle and redundant manipulator through an extended task-priority controller, enabling dexterous grasping while satisfying operational constraints such as joint limits and target visibility. Ocean One~\cite{khatib2016ocean,brantner2021controlling} combines a whole-body controller for manipulation, posture, and constraint regulation with visual and haptic interfaces, allowing a human operator to supervise deep-sea dexterous manipulation without directly controlling every joint. DexROV~\cite{gancet2015dexrov} addresses long-distance underwater teleoperation under communication latency through predictive simulation, force-feedback interfaces, and shared autonomy. More recently, MR-UBi~\cite{nishi2026mr} integrates bilateral control with a mixed-reality reaction-torque indicator, providing visual and haptic feedback to improve contact-force regulation during underwater grasping and pick-and-place operations.

\noindent \textbf{Learning-Based Methods.}
Learning-based underwater manipulation remains less explored than its terrestrial counterpart, but recent work has begun to address model uncertainty, demonstration scarcity, and challenging visual conditions. In RL, Carlucho et al.~\cite{carlucho2020reinforcement} introduce an actor--critic controller that directly predicts joint torques for an underwater manipulator while respecting position and torque constraints, reducing dependence on an accurate hydrodynamic model. In IL, AquaBot~\cite{liu2025self} first learns from human teleoperation through behavior cloning and subsequently improves the policy using autonomous real-world trials, supporting tasks such as grasping, sorting, and object retrieval. Bi-AQUA~\cite{tsunoori2025bi} learns from bilateral-control demonstrations and introduces lighting-aware feature modulation and transformer conditioning to maintain manipulation performance under varying underwater illumination. UMI-Underwater~\cite{li2026umi} further reduces reliance on underwater teleoperation by autonomously collecting underwater grasp demonstrations and transferring knowledge from on-land human demonstrations through a depth-based affordance representation. An affordance-conditioned diffusion policy then generates robot actions that remain robust to changes in lighting, color, and background appearance. Together, these methods indicate a shift from manually operated systems toward autonomous and data-scalable underwater manipulation.

\noindent \textbf{Challenges.}
Underwater manipulation remains constrained by uncertain hydrodynamics, strong coupling between vehicle and manipulator motion, limited visibility, communication latency, and the difficulty of regulating contact forces in moving fluids. Learning-based approaches additionally face scarce real-world demonstrations, costly and risky data collection, limited simulation fidelity, and substantial distribution shifts across water conditions, lighting, payloads, and deployment sites. Future progress will require tighter integration of model-based control with learned policies, scalable self-supervised data collection, robust multimodal perception using vision, sonar, proprioception, and force sensing, and safety-aware adaptation under changing environmental dynamics.


\begin{figure}[t]
\centering
\includegraphics[width=0.55\linewidth]{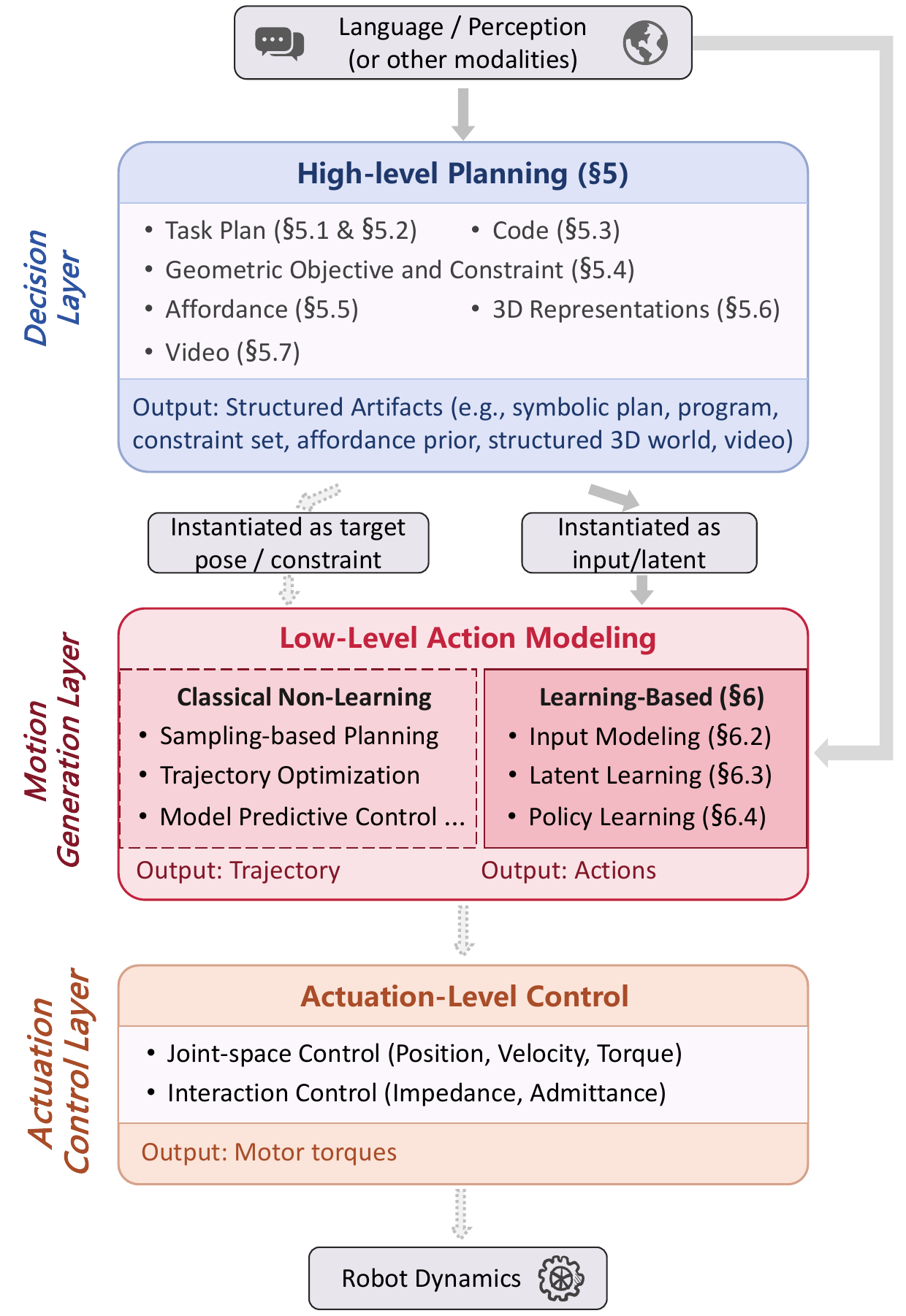}  
\caption{Method taxonomy for basic manipulation, extensible to other manipulation tasks. It comprises High-Level Planning, Low-Level Action Modeling, and Actuation-Level Control. High-level modules produce planning artifacts that guide downstream action generation, while low-level methods operate on these artifacts or raw multimodal observations. Solid arrows indicate the survey’s primary focus: High-Level Planning and Learning-Based Low-Level Action Modeling.
}
\label{fig: cat_basic}
\end{figure}

\section{High-level Planning}
\label{sec:high_level_planning}

\definecolor{catBlue}{HTML}{87CEFA}
\definecolor{catGreen}{HTML}{8FBC8F}
\definecolor{catOrange}{HTML}{FF7F0E}
\definecolor{catPurple}{HTML}{9467BD}

\newcommand{\taxmain}[1]{%
  \parbox[c]{25mm}{\raggedright #1}%
}

\newcommand{\taxmid}[1]{%
  \parbox[c]{36mm}{\centering #1}%
}

\newcommand{\taxitem}[1]{%
  \parbox[c]{76mm}{\raggedright #1}%
}

\begin{figure*}[!t]
\centering
\scriptsize

\forestset{
  taxo/.style={
    for tree={
      grow'=0,
      parent anchor=east,
      child anchor=west,
      anchor=west,
      edge path'={
        (!u.parent anchor) -- +(3pt,0) |- (.child anchor)
      },
      edge={draw, line width=0.6pt},
      rounded corners,
      draw,
      minimum height=4mm,
      inner xsep=3pt,
      inner ysep=1.5pt,
      s sep=4pt,
      l sep=9pt,
      minimum height=4.8mm,
      inner ysep=2.2pt,
      font=\scriptsize
    },
    sub/.style={
      draw,
      rounded corners=2pt,
      fill opacity=.5,
      inner xsep=3pt,
      inner ysep=2pt
    },
    sub_1/.style={
      draw,
      rounded corners=2pt,
      fill=hidden-blue!32,
      inner xsep=3pt,
      inner ysep=2pt
    },
    item/.style={
      draw,
      rounded corners=2pt,
      fill=yellow!32,
      inner xsep=3pt,
      inner ysep=2pt
    }
  }
}

\begin{forest} taxo
[{\rotatebox{90}{\textbf{High-level Planner (\S~\ref{sec:high_level_planning})}}},
 draw, inner sep=2pt, fill opacity=.5,
  [{\taxmain{LLM-based Task Planning~\ref{subsec: llm-based_planners}}}, sub
    [{\taxmid{Grounded \& Closed-Loop Planning}}, sub_1
      [{\taxitem{
        SayCan~\cite{brohan2023can},
        Grounded Decoding~\cite{huang2023grounded},
        Inner Monologue~\cite{huang2022inner},
        LLM-Planner~\cite{song2023llm},
        Polaris~\cite{wang2024polaris}
      }}, item]
    ]
    [{\taxmid{Feasibility-Aware Planning}}, sub_1
      [{\taxitem{
        LLM+P~\cite{liu2023llm+},
        LLM$^3$~\cite{wang2024llm},
        REFLECT~\cite{liu2023reflect}
      }}, item]
    ]
    [{\taxmid{Preference \& Collaborative Planning}}, sub_1
      [{\taxitem{
        APRICOT~\cite{wang2024apricot},
        MALMM~\cite{singh2024malmm},
        RoCo~\cite{mandi2024roco}
      }}, item]
    ]
  ]
  [{\taxmain{MLLM-based Task Planning~\ref{subsec: mllm-based_planners}}}, sub
    [{\taxmid{General-Purpose MLLM Reasoning}}, sub_1
      [{\taxitem{
        PaLM-E~\cite{driess2023palm},
        VILA~\cite{du2024video},
        PG-InstructBLIP~\cite{gao2024physically},
        EVLP~\cite{guan2026evlp},
        EmbodiedGPT~\cite{mu2023embodiedgpt}
      }}, item]
    ]
    [{\taxmid{Spatial Reasoning \& Grounding}}, sub_1
      [{\taxitem{
        SpatialVLM~\cite{chen2024spatialvlm},
        SpatialBot~\cite{cai2025spatialbot},
        RoboSpatial~\cite{song2025robospatial},
        EmbodiedVSR~\cite{zhang2025embodiedvsr},
        SoFar~\cite{qi2025sofar},
        RoboRefer~\cite{zhou2025roborefer}
      }}, item]
    ]
    [{\taxmid{Reward \& Progress Reasoning}}, sub_1
      [{\taxitem{
        Robo-Dopamine~\cite{tan2026robo},
        RoboReward~\cite{lee2026roboreward},
        Robometer~\cite{liang2026robometer},
        ProcVLM~\cite{feng2026procvlm}
      }}, item]
    ]
    [{\taxmid{Failure Reasoning \& Recovery}}, sub_1
      [{\taxitem{
        AHA~\cite{duan2025aha},
        Code-as-Monitor~\cite{zhou2025code},
        I-FailSense~\cite{grislain2026failsense},
        Rewind-IL~\cite{zheng2026rewind},
        AgentChord~\cite{xu2026reaction}
      }}, item]
    ]
    [{\taxmid{Robot-Specialized Embodied Reasoning}}, sub_1
      [{\taxitem{
        RoboBrain~\cite{ji2025robobrain},
        RoboBrain 2.0~\cite{team2025robobrain},
        RynnBrain~\cite{dang2026rynnbrain},
        FSD~\cite{yuan2026seeing},
        Embodied-R1~\cite{yuan2026embodied}, Gemini Robotics~\cite{team2025gemini}
      }}, item]
    ]
  ]
  [{\taxmain{Programmatic Planning~\ref{subsec: programmatic_planning}}}, sub
    [{\taxmid{Language-to-Program}}, sub_1
      [{\taxitem{
        Code as Policies~\cite{liang2023code},
        ProgPrompt~\cite{singh2023progprompt},
        ChatGPT for Robotics~\cite{vemprala2024chatgpt}
      }}, item]
    ]
    [{\taxmid{Demonstration-to-Program}}, sub_1
      [{\taxitem{
        Instruct2Act~\cite{huang2023instruct2act},
        Demo2Code~\cite{wang2023demo2code},
        SHOWTELL~\cite{murray2024teaching},
        Statler~\cite{yoneda2024statler}
      }}, item]
    ]
    [{\taxmid{Grounded \& Verifiable Programs}}, sub_1
      [{\taxitem{
        InterPreT~\cite{han2024interpret},
        Reliable CaP~\cite{ahn2026towards},
        HyCodePolicy~\cite{liu2025hycodepolicy}
      }}, item]
    ]
  ]
  [{\taxmain{Geometric Constraint-Based Planning~\ref{subsec: geometric_constraint_planning}}}, sub
    [{\taxmid{Spatial Objectives}}, sub_1
      [{\taxitem{
        VoxPoser~\cite{huang2023voxposer},
        IKER~\cite{patel2025real}
      }}, item]
    ]
    [{\taxmid{Relational \& Keypoint Constraints}}, sub_1
      [{\taxitem{
        CoPa~\cite{huang2024copa},
        ReKep~\cite{huang2025rekep},
        GeoManip~\cite{tang2025geomanip}
      }}, item]
    ]
    [{\taxmid{Functional Constraints}}, sub_1
      [{\taxitem{
        CoDex~\cite{jiang2026codex}
      }}, item]
    ]
  ]
  [{\taxmain{Affordance-Based Planning~\ref{subsec: affordance_based_Planning}}}, sub
    [{\taxmid{Geometric Affordance}}, sub_1
      [{\taxitem{
        Ditto~\cite{jiang2022ditto},
        GAPartNet~\cite{geng2023gapartnet},
        CPM~\cite{liu2023composable},
        RoboPCA~\cite{xiao2026robopca}
      }}, item]
    ]
    [{\taxmid{Visual Affordance}}, sub_1
      [{\taxitem{
        Transporter Networks~\cite{zeng2021transporter},
        VAPO~\cite{borja2022affordance},
        RAAP~\cite{zhuang2026raap}
      }}, item]
    ]
    [{\taxmid{Semantic Affordance}}, sub_1
      [{\taxitem{
        Affordance-Based Imitation Learning~\cite{lopes2007affordance},
        SAGE~\cite{geng2024sage}
      }}, item]
    ]
    [{\taxmid{Multimodal Affordance}}, sub_1
      [{\taxitem{
        CLIPort~\cite{shridhar2022cliport},
        ManipLLM~\cite{li2024manipllm},
        MOKA~\cite{liu2024moka}, A3VLM~\cite{huang2024a3vlm}, RoboPoint~\cite{yuan2025robopoint},
        UniAff~\cite{yu2025uniaff}, BiPreManip~\cite{shen2026bipremanip}
      }}, item]
    ]
  ]
  [{\taxmain{3D Representation-Based Planning~\ref{subsec: 3d_representation_based_planning}}}, sub
    [{\taxmid{Point-Cloud Representations}}, sub_1
      [{\taxitem{
        Point-Cloud Planning~\cite{saha2025planning},
        PA3FF~\cite{chen2026pa3ff}
      }}, item]
    ]
    [{\taxmid{Implicit \& Descriptor Fields}}, sub_1
      [{\taxitem{
        NDF~\cite{simeonov2022neural},
        R-NDF~\cite{simeonov2023se},
        F3RM~\cite{shen2023distilled},
        $D^{3}$Fields~\cite{wang2025d},
        PA3FF~\cite{chen2026pa3ff}
      }}, item]
    ]
    [{\taxmid{Gaussian-Splatting Representations}}, sub_1
      [{\taxitem{
        Splat-MOVER~\cite{shorinwa2025splat},
        Physically Embodied GS~\cite{abou2025physically},
        MSGField~\cite{sheng2024msgfield},
        RoboSplat~\cite{yang2025novel}
      }}, item]
    ]
    [{\taxmid{Generative \& Structured 3D Representations}}, sub_1
      [{\taxitem{
        Imagination Policy~\cite{huang2025imagination},
        RoboEXP~\cite{jiang2025roboexp}
      }}, item]
    ]
  ]
]
\end{forest}

\caption{Taxonomy of high-level planning approaches, organized into major planning paradigms, including LLM- and MLLM-based task planning, programmatic planning, and geometric constraint-based planning, together with supporting capabilities in affordance learning and 3D representations.}
\label{fig: high-level_planner_taxonomy}
\end{figure*}

\begin{figure}[!t]
\centering
\includegraphics[width=1.0\textwidth]{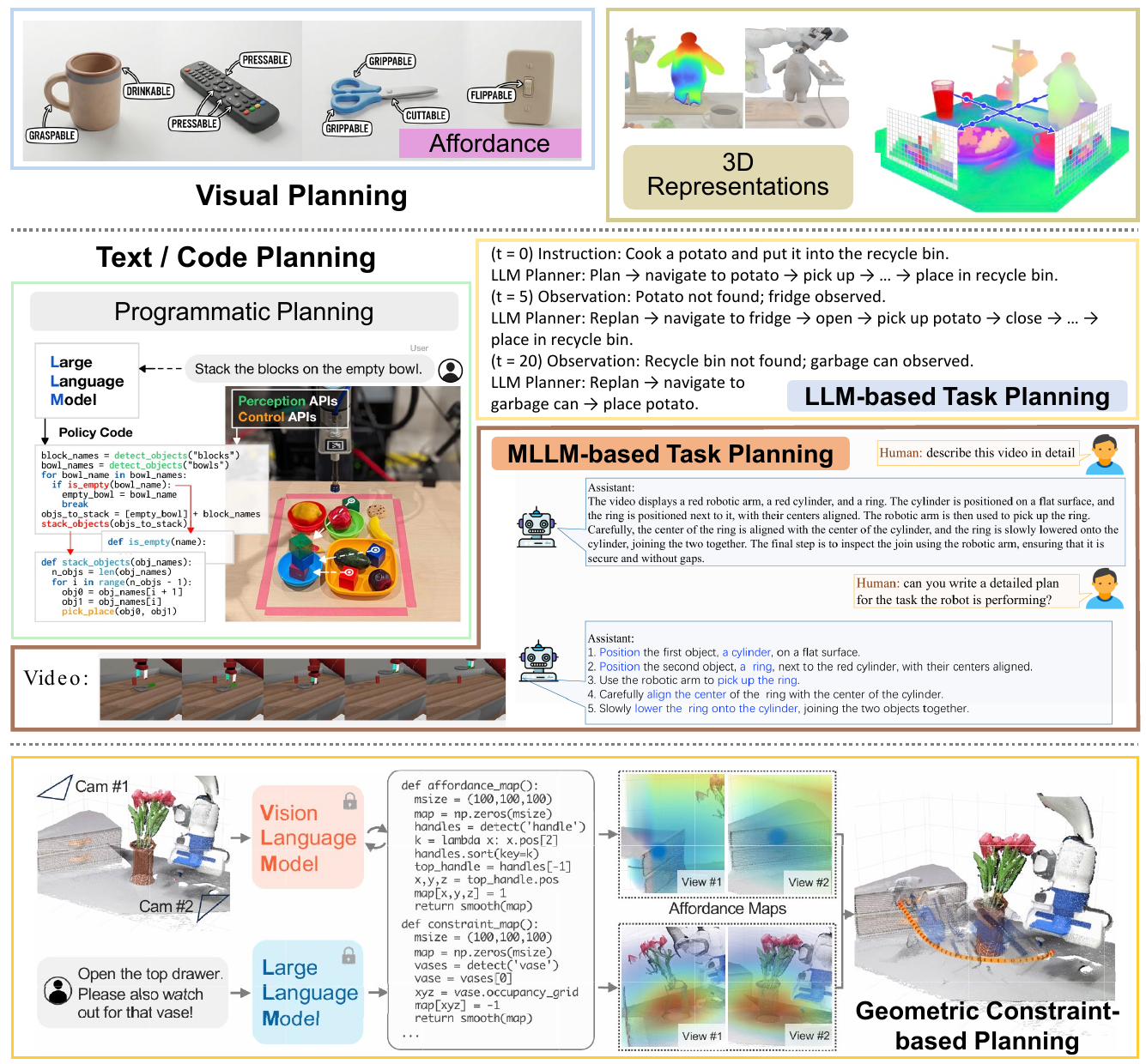}  
\caption{Overview of the taxonomy of high-level planners, highlighting six core components: LLM-based task planning, MLLM-based task planning, code generation, motion planning, affordance learning, and 3D scene representations. Figure adapted from~\cite{song2023llm, mu2023embodiedgpt, liang2023code, huang2023voxposer, jiang2022ditto, shen2023distilled}.
}
\label{fig: high-level_planner}
\end{figure}

High-level planning in robot manipulation provides structured guidance for low-level execution by determining what actions to perform, in which order, and which aspects of the environment are relevant. LLMs and MLLMs are increasingly used for \textbf{task planning}, \textbf{programmatic planning}, and \textbf{geometric constraint-based planning}, supporting task decomposition, skill sequencing, executable program generation, and spatially grounded reasoning. In parallel, \textbf{affordance-based planning} and \textbf{3D representation-based planning} provide actionable intermediate representations, such as functional regions, contact cues, geometric relations, and structured scene states. More recently, \textbf{video-based planning} extends this paradigm by representing future task evolution through generated visual trajectories or predictive world models, enabling prospective reasoning over action outcomes and plan refinement. Together, these approaches establish high-level planning as a flexible guidance layer that integrates semantic reasoning, spatial grounding, scene understanding, and future prediction to support reliable execution across diverse manipulation tasks. We summarize this taxonomy in Figure~\ref{fig: high-level_planner_taxonomy} and provide a visualization in Figure~\ref{fig: high-level_planner}.

\subsection{LLM-Based Task Planning}
\label{subsec: llm-based_planners}

Early work followed a symbolic planning and grounding paradigm, in which neural networks mapped demonstrations and observations to symbolic states and goals represented by predicate truth values~\cite{huang2019continuous}. LLMs have subsequently extended this paradigm by providing semantic knowledge, instruction decomposition, and open-vocabulary reasoning, while frequently retaining symbolic representations or predefined robot skills for grounding and execution. SayCan~\cite{brohan2023can} is an early influential example that combines language-model-based task relevance with learned affordance scores estimating skill executability, thereby selecting actions that are both semantically appropriate and physically feasible. Grounded Decoding~\cite{huang2023grounded} further integrates language and grounded model likelihoods during sequence decoding. However, these methods primarily perform planning without incorporating execution feedback. Inner Monologue~\cite{huang2022inner} addresses this limitation by feeding task-success signals, scene descriptions, and human feedback back into the LLM reasoning process, enabling closed-loop plan revision in unstructured environments. Related feedback-driven replanning is also explored in LLM-Planner~\cite{song2023llm}.

Another line of work improves the logical and physical feasibility of LLM-generated plans. LLM+P \cite{liu2023llm+} translates natural-language problems into PDDL representations and delegates plan search to a classical planner, combining the semantic flexibility of LLMs with the correctness guarantees of symbolic planning. LLM$^3$~\cite{wang2024llm} more directly connects task and motion planning by using motion-planning failures to revise action sequences and continuous parameters, reducing repeated exploration of geometrically infeasible plans. REFLECT~\cite{liu2023reflect} similarly diagnoses unsuccessful interactions and uses the resulting failure information to generate corrective plans. These methods shift LLM-based planning from unconstrained language generation toward iterative planning grounded in symbolic validity, motion feasibility, and execution outcomes.

Recent studies further incorporate user preferences, safety requirements, and broader interaction structures. APRICOT~\cite{wang2024apricot} combines LLM-based active preference learning with constraint-aware task planning, querying users to resolve ambiguous preferences while adapting plans to geometric limitations. MALMM~\cite{singh2024malmm} employs multiple LLM agents to improve planning and decision-making, while Polaris~\cite{wang2024polaris} combines LLM reasoning with Syn2Real visual grounding for open-ended tabletop manipulation. RoCo~\cite{mandi2024roco} extends LLM-based planning to multi-robot collaboration and introduces RoCoBench for evaluating collaborative manipulation tasks.

\subsection{MLLM-based Task Planning}
\label{subsec: mllm-based_planners}

Text-only LLMs require visual observations and other sensory inputs to be processed by separate perception modules before reasoning. In contrast, MLLMs~\cite{liu2023visual,karamcheti2024prismatic,li2023blip} jointly process visual and linguistic information, enabling tighter integration of perception, reasoning, planning, and execution monitoring. We therefore view MLLM-based task planning as a closed-loop process that encompasses plan generation, spatial grounding, progress evaluation, failure diagnosis, and corrective reasoning.

\noindent \textbf{General-Purpose MLLM Reasoning.}
Early studies adapt general-purpose MLLMs to robotic decision-making. PaLM-E~\cite{driess2023palm} incorporates continuous embodied observations into a pretrained language model, enabling visually grounded sequential decision-making, while VILA~\cite{du2024video} directly leverages GPT-4V for visual grounding and manipulation planning without task-specific fine-tuning. PG-InstructBLIP~\cite{gao2024physically} instead fine-tunes InstructBLIP~\cite{dai2023instructblip} on physical concepts to strengthen physical reasoning. EmbodiedGPT~\cite{mu2023embodiedgpt} introduces embodied chain-of-thought reasoning~\cite{wei2022chain}, and Zhang et al.~\cite{zhang2025learning2} further incorporate fine-grained reward guidance. Matcha~\cite{zhao2023chat} enables interactive multimodal perception, whereas EVLP~\cite{guan2026evlp} jointly generates linguistic subplans and visual subgoals. Socratic Models~\cite{zeng2023socratic} and Socratic Planner~\cite{shin2024socratic} further coordinate specialized perception, reasoning, and control modules through language.

\noindent \textbf{Spatial Reasoning and Grounding.}
Spatial reasoning supplies the geometric and relational information required to ground high-level instructions into manipulation plans. SpatialVLM~\cite{chen2024spatialvlm}, SpatialBot~\cite{cai2025spatialbot}, and RoboSpatial~\cite{song2025robospatial} improve metric, directional, and 3D scene understanding, while EmbodiedVSR~\cite{zhang2025embodiedvsr} organizes evolving relations through dynamic scene graphs. For manipulation-specific grounding, SoFar~\cite{qi2025sofar} represents object poses and action directions using language-grounded functional orientations, and RoboRefer~\cite{zhou2025roborefer} combines depth-aware perception with multi-step spatial referring. These methods provide intermediate representations, including orientations, referring regions, scene graphs, affordances, and geometric relations, that constrain downstream action generation.

\noindent \textbf{Reward and Progress Reasoning.}
MLLMs can also serve as critics that evaluate execution progress and provide feedback for replanning or policy improvement. Robo-Dopamine~\cite{tan2026robo} models fine-grained manipulation progress, RoboReward~\cite{lee2026roboreward} learns general-purpose vision--language rewards from diverse robot trajectories, and Robometer~\cite{liang2026robometer} combines within-trajectory progress estimation with cross-trajectory comparison. ProcVLM~\cite{feng2026procvlm} further grounds dense rewards in procedural stages and visual state transitions. Together, these methods extend multimodal reasoning from plan generation to execution evaluation.

\noindent \textbf{Failure Reasoning and Recovery.}
Failure reasoning closes the planning loop by detecting deviations, diagnosing their causes, and selecting corrective actions. AHA~\cite{duan2025aha} detects manipulation failures and produces natural-language explanations for plan revision, while Code-as-Monitor~\cite{zhou2025code} translates task constraints into executable visual programs for reactive and proactive monitoring. I-FailSense~\cite{grislain2026failsense} targets failure detection across robots and tasks, and self-refining VLMs~\cite{qi2026self} improve failure recognition and reasoning through iterative feedback. Rewind-IL~\cite{zheng2026rewind} leverages detected failures for recovery learning, whereas AgentChord~\cite{xu2026reaction} incorporates anticipated failure branches and corrective transitions into task graphs. These approaches extend one-shot planning toward closed-loop monitoring, diagnosis, recovery, and replanning.

\noindent \textbf{Robot-Specialized Embodied Reasoning.}
Recent work increasingly trains MLLMs specifically for robotic perception and reasoning. RoboBrain~\cite{ji2025robobrain} learns from robot-centric annotations spanning task plans, affordances, and end-effector trajectories, while RoboBrain 2.0~\cite{team2025robobrain} extends these capabilities to spatial--temporal reasoning and closed-loop interaction. RynnEC~\cite{dang2025rynnec} emphasizes region-centric visual representations, and RynnBrain~\cite{dang2026rynnbrain} further integrates embodied reasoning within a unified foundation model. FSD~\cite{yuan2026seeing} connects spatial reasoning to manipulation through affordance regions and visual traces, whereas Embodied-R1~\cite{yuan2026embodied} uses pointing as an intermediate interface between multimodal reasoning and action generation. Other embodied models, including ERA~\cite{chen2025era}, GenieReasoner~\cite{liu2025unified}, PhysBrain~\cite{lin2025physbrain}, HY-Embodied~\cite{team2026hy}, and ACE-Brain~\cite{gong2026ace}, further integrate task reasoning, spatial grounding, affordance prediction, and action-related representations. Overall, this trend shifts MLLMs from external planners toward embodied foundation models that support planning, grounding, evaluation, and closed-loop adaptation within a unified framework.

\subsection{Programmatic Planning}
\label{subsec: programmatic_planning}

Programmatic planning represents robot plans as executable programs that connect high-level semantic reasoning with perception, motion primitives, and control APIs. Early work maps natural-language instructions to robot programs~\cite{venkatesh2021translating}, while Code as Policies~\cite{liang2023code} prompts LLMs with perception and control APIs to synthesize executable policies. ProgPrompt~\cite{singh2023progprompt} similarly generates situated robot programs with explicit action and state structures, and Vemprala et al.~\cite{vemprala2024chatgpt} establish a reusable framework for integrating ChatGPT with robotic APIs. Subsequent methods broaden the available supervision and program representations. Instruct2Act~\cite{huang2023instruct2act} integrates visual foundation models for multimodal instruction following, Demo2Code~\cite{wang2023demo2code} abstracts demonstrations into reusable programmatic skills, and SHOWTELL~\cite{murray2024teaching} generates policy code directly from visual demonstrations. Statler~\cite{yoneda2024statler} further maintains an explicit world state to support longer-horizon reasoning.

Recent work increasingly focuses on grounding, verification, and adaptation of generated programs. InterPreT~\cite{han2024interpret} learns executable predicates from language feedback and compiles them with learned symbolic operators into PDDL domains for generalizable planning. Towards Reliable Code-as-Policies~\cite{ahn2026towards} combines symbolic verification with interactive environment exploration to validate task-relevant conditions before execution. HyCodePolicy~\cite{liu2025hycodepolicy} closes the programming loop by integrating geometric grounding, visual monitoring, failure diagnosis, and iterative code repair. 
More recently, Functional Cache Grafting~\cite{chun2026functional} retrieves and composes validated function-level code structures, reducing redundant generation while improving policy robustness. Overall, programmatic planning is evolving from one-shot language-to-code generation toward grounded, verifiable, self-correcting, and reusable executable representations.

\subsection{Geometric Constraint-based Planning}
\label{subsec: geometric_constraint_planning}

Geometric objectives and constraints provide an intermediate representation between semantic task reasoning and motion execution. Rather than directly mapping observations to robot actions, these approaches translate language and perception into spatial costs, keypoint relations, target poses, or geometric constraints that can be optimized by downstream planners. VoxPoser~\cite{huang2023voxposer} composes language-conditioned 3D value maps as spatial objectives for model-based trajectory generation. CoPa~\cite{huang2024copa} identifies task-relevant object parts and their spatial relations to derive target end-effector poses, while ReKep~\cite{huang2025rekep} represents manipulation tasks as sequences of optimizable relational constraints over semantic 3D keypoints. GeoManip~\cite{tang2025geomanip} further treats stage-specific geometric constraints as a general interface between semantic reasoning and trajectory optimization.

Recent work extends this paradigm toward adaptive objectives and more complex physical interactions. IKER~\cite{patel2025real} uses VLMs to generate and iteratively refine keypoint-based reward functions from language and RGB-D observations, enabling geometric task specification for downstream policy learning. CoDex~\cite{jiang2026codex} similarly infers semantic constraints for dexterous functional manipulation and uses constrained optimization to generate physically feasible grasp candidates before policy refinement. Collectively, these methods separate semantic task interpretation from numerical motion generation through explicit, optimizable geometric representations, improving the modularity and physical grounding of manipulation planning.

\subsection{Affordance-Based Planning}
\label{subsec: affordance_based_Planning}

Affordance-based planning represents manipulation tasks through action-oriented properties that specify where and how an agent can interact with objects or environments. Originating from Gibson's notion of affordance~\cite{gibson2014theory}, this perspective provides an intermediate abstraction between semantic understanding and motion generation. Rather than directly predicting complete trajectories, affordance-based methods identify functional parts, interaction regions, keypoints, contact poses, action directions, or kinematic constraints that guide downstream execution. We organize these methods by their dominant information source and representation into geometric, visual, semantic, and multimodal affordances, while noting that recent approaches increasingly combine these cues.

\noindent \textbf{Geometric Affordance.}
Geometric affordances derive interaction possibilities primarily from object shape, articulation, and kinematic structure. Early methods infer parts, joints, and motion constraints directly from 3D observations~\cite{jiang2022ditto,geng2023gapartnet,geng2023partmanip}. Ditto~\cite{jiang2022ditto}, for example, reconstructs articulation models from physical interaction, while GAPartNet~\cite{geng2023gapartnet} introduces cross-category generalizable actionable parts that support skill transfer across object instances. CPM~\cite{liu2023composable} further represents manipulation through compositional geometric relations between functional parts. For articulated objects, Kinematic-aware Prompting~\cite{xia2024kinematic} converts joints and contact locations into a unified kinematic description and prompts an LLM to generate 3D manipulation waypoints, explicitly connecting object kinematics with low-level action generation. ScrewSplat~\cite{kim2025screwsplat} instead reconstructs object geometry, movable parts, and screw axes directly from RGB observations, providing an explicit kinematic model that can subsequently support text-guided manipulation. More recently, RoboPCA~\cite{xiao2026robopca} jointly predicts task-conditioned contact regions and end-effector poses from RGB-D observations, coupling \textit{where} to interact with \textit{how} the interaction should be geometrically realized.

\noindent \textbf{Visual Affordance.}
Visual affordances infer interaction opportunities directly from images, commonly representing them as pixels, regions, keypoints, or action-conditioned heatmaps. Transporter Networks~\cite{zeng2021transporter} establish a spatially equivariant formulation for pick-and-place by predicting image-space action locations, while VAPO~\cite{borja2022affordance} learns visual affordances from unstructured interaction data. KITE~\cite{sundaresan2023kite} grounds language instructions into 2D image keypoints and conditions downstream manipulation skills on these keypoints, enabling fine-grained object- and part-level interaction across scene variations. UAD~\cite{tang2025uad} reduces the dependence on manually annotated affordances by distilling complementary knowledge from pretrained vision and vision--language models into a lightweight task-conditioned affordance predictor. Recent work further improves transfer beyond fixed training categories. RAAP~\cite{zhuang2026raap} combines retrieval with cross-image alignment, decoupling contact localization from post-contact action direction to transfer affordances across unseen objects. Together, these methods turn visual grounding and correspondence into actionable intermediate representations for manipulation.

\noindent \textbf{Semantic Affordance.}
Semantic affordances associate object identity, functional parts, and task concepts with possible interactions. Early affordance-based imitation learning~\cite{lopes2007affordance} showed that semantic properties can provide transferable priors for mapping observations to manipulation behaviors. More recent methods shift from category-level semantics toward functional and part-level representations. SAGE~\cite{geng2024sage} bridges semantic parts with Generalizable Actionable Parts, combining instruction interpretation with part-level motion knowledge to construct executable policies. PartInstruct~\cite{yin2025partinstruct} further highlights the importance of such fine-grained grounding through a benchmark in which instructions explicitly refer to object parts and their task-dependent roles, exposing the difficulty of jointly grounding part semantics and predicting manipulation actions in 3D space. This distinction between semantic identity and physical functionality is especially important for articulated and multi-part objects, where similarly named parts may exhibit different kinematics and afford different interactions.

\noindent \textbf{Multimodal Affordance.}
Foundation models increasingly combine language, vision, geometry, demonstrations, and interaction feedback to infer richer affordance representations. CLIPort~\cite{shridhar2022cliport} couples a semantic ``what'' pathway with a spatial ``where'' pathway for language-conditioned manipulation. ManipLLM~\cite{li2024manipllm} further reasons about object categories, affordance priors, contact points, and end-effector directions within an embodied MLLM. MOKA~\cite{liu2024moka} uses marked visual prompts to elicit point-based affordances from pretrained VLMs, providing a compact interface between semantic reasoning and robot motion, while RoboPoint~\cite{yuan2025robopoint} predicts spatial affordance points from language and images. PIVOT~\cite{nasiriany2024pivot} similarly turns continuous spatial decisions into iterative visual selection by overlaying candidate actions, locations, or trajectories onto images and repeatedly querying a VLM to refine them, enabling actionable spatial reasoning without task-specific fine-tuning.

Retrieval provides another route to generalizable affordance reasoning. RAM~\cite{kuang2025ram} constructs an affordance memory from heterogeneous robot, human--object interaction, and custom demonstrations, retrieves task-relevant examples from language instructions, and transfers the retrieved 2D affordance into an executable 3D affordance in the target scene. For articulated manipulation, A3VLM~\cite{huang2024a3vlm} learns robot-agnostic articulation and action affordances that can be instantiated through downstream motion primitives, whereas UniAff~\cite{yu2025uniaff} jointly models affordance understanding and object-centric 3D motion constraints for both articulated objects and tools. AIC MLLM~\cite{xiong2025autonomous} further closes the interaction loop by using visual and textual feedback from failed interactions to correct predicted SE(3) contact poses, showing how affordance-related geometric decisions can be refined through physical experience. BiPreManip~\cite{shen2026bipremanip} uses affordance prediction prospectively, reasoning about the desired final interaction to plan preparatory actions for coordinated bimanual manipulation. These developments shift affordances from passive interaction descriptors toward explicit planning interfaces that connect multimodal task reasoning, geometric grounding, interaction feedback, and executable robot behavior.

\subsection{3D Representation-Based Planning}
\label{subsec: 3d_representation_based_planning}

3D representation-based planning uses structured scene representations to bridge perception and action by exposing planning-relevant geometry, semantics, relations, or goal states. Rather than directly mapping observations to control signals, these methods support manipulation through intermediate representations such as point-cloud transformations, continuous descriptor fields, editable Gaussian scenes, imagined goal geometries, or action-conditioned scene graphs. We focus on representations that explicitly support goal specification, correspondence, search, or downstream motion generation.

\noindent \textbf{Point-Cloud Representations.}
Point clouds provide an explicit geometric state space for manipulation planning. Saha et al.~\cite{saha2025planning} formulate multi-object rearrangement as A* search directly over point-cloud transformations, avoiding discrete symbolic state and action representations. Related methods use point clouds to encode functional parts or geometric targets for manipulation. For example, PA3FF~\cite{chen2026pa3ff} learns a continuous part-aware 3D feature field from point-cloud inputs, providing functional-part representations that support correspondence and downstream policy learning.

\noindent \textbf{Implicit and Descriptor Fields.}
Implicit fields provide continuous, geometry-aware representations for pose correspondence and language-conditioned manipulation. NDF~\cite{simeonov2022neural} learns SE(3)-equivariant descriptors for pose transfer, while R-NDF~\cite{simeonov2023se} extends them to relational rearrangement. F3RM~\cite{shen2023distilled} distills 2D foundation-model features into 3D fields for few-shot language-guided grasping and placing. $D^3$Fields~\cite{wang2025d} further models dynamic semantic and instance-aware descriptors for zero-shot rearrangement, while PA3FF~\cite{chen2026pa3ff} incorporates functional part structure to improve robust articulated-object generalization.

\noindent \textbf{Gaussian-Splatting Scene Representations.}
Gaussian Splatting provides an explicit and editable 3D scene representation combining geometry, appearance, and semantics. Splat-MOVER~\cite{shorinwa2025splat} embeds semantic and affordance features into 3DGS and generates grasp candidates for multi-stage open-vocabulary manipulation. Physically Embodied Gaussian Splatting~\cite{abou2025physically} couples visual Gaussians with particle-based physical states for predictive and correctable scene modeling, while MSGField~\cite{sheng2024msgfield} augments Gaussian primitives with semantic and motion attributes for dynamic manipulation. RoboSplat~\cite{yang2025novel} instead exploits scene editability to synthesize diverse demonstrations and improve downstream policy generalization.

\noindent \textbf{Generative and Structured 3D Representations.}
Beyond reconstructing the current scene, 3D representations can encode desired future states and action-relevant structure. Imagination Policy~\cite{huang2025imagination} generates target point clouds and recovers keyframe actions through rigid registration, formulating action inference as goal-state generation. RoboEXP~\cite{jiang2025roboexp} incrementally builds action-conditioned scene graphs through interaction, combining geometry, semantics, and relational structure for downstream manipulation. These approaches extend 3D representations from static perception toward explicit goal and interaction representations for planning.

\subsection{Video-Based Planning}
\label{subsec:video_based_planning}

Video-based planning represents task evolution through predicted visual futures, providing a temporal interface between task understanding and robot execution. Unlike static goals or geometric constraints, videos encode object motion, interaction outcomes, and intermediate task states. Recent world models further condition future prediction on language or robot actions, enabling visual rollouts to support action extraction, model-based planning, and planner optimization. We focus on methods where predicted futures directly contribute to plan representation, action selection, or planner learning.

\noindent \textbf{Generative Video Planning.}
One line of work treats generated future observations as visual plans. Large Video Planner~\cite{chen2025large} learns from large-scale human and task videos to produce zero-shot video plans that are converted into robot actions. RIGVid~\cite{patel2026robotic} generates manipulation videos from language and initial observations, filters them with a VLM, and recovers object trajectories through 6-DoF pose tracking. Geometry-aware 4D Video Generation~\cite{liu2026geometry} improves cross-view geometric consistency in generated RGB-D videos, while GEM-4D~\cite{zhou2026gem} adds dense 4D correspondence supervision and inverse dynamics for trajectory recovery. These methods use video as an explicit spatiotemporal plan that can be translated into robot motion.

\noindent \textbf{Predictive World-Model Planning.}
A second direction predicts the consequences of candidate actions for action selection. SAMPO~\cite{wang2025sampo} performs action-conditioned video prediction for model-based control. Grounded World Model~\cite{li2026grounded} predicts outcomes in a vision--language-aligned latent space and scores them against task instructions, enabling language-conditioned model-predictive control. Here, the world model functions as a predictive dynamics model within the planning loop.

\noindent \textbf{World-Model-Guided Planner Learning.}
World models can also provide simulated interaction for improving planners with fewer physical rollouts. DreamPlan~\cite{jia2026dreamplan} uses imagined rollouts from an action-conditioned video world model to construct preferences for reinforcement fine-tuning of a VLM planner. RoboEvolve~\cite{chen2026roboevolve} co-evolves a VLM planner and video simulator using simulated successes and near-miss failures. GE-Sim 2.0~\cite{qiu2026ge} further supports closed-loop simulation by predicting visual and proprioceptive evolution and providing rollout-level supervision for policy learning. These approaches position world models as learned environments for planner refinement.

\section{Low-level Learning-based Action Modeling}
\label{sec:low_level_action_modeling}

Low-level action modeling focuses on how robots transform perception and task context into executable actions, trajectories, or action sequences, providing the interface between high-level planning and physical execution. While high-level planning determines what to do and in what order, such as task decomposition, skill sequencing, or goal reasoning, low-level action modeling determines how planned intent is instantiated into concrete manipulation behaviors through learned visuomotor mappings. The two are complementary: high-level planners provide structured intent and semantic constraints, whereas low-level action models translate them, or directly map multimodal observations, into action representations that can be executed by controllers. Within this framework, learning paradigms such as imitation learning and reinforcement learning define how action models are optimized. We further decompose learning-based low-level action modeling into three interdependent components: input modeling, which determines what sensory and contextual information is used and how it is encoded; latent learning, which studies compact and transferable internal representations; and policy learning, which determines how these representations are decoded into executable actions. Together, this taxonomy provides a unified view of learning-based action generation by connecting perception, representation, and policy optimization between high-level reasoning and actuation-level control.

\subsection{Learning Strategy}
\label{sec:learning_strategy}

\subsubsection{Reinforcement Learning}
\label{subsubsec: rl}

In robotic manipulation, reinforcement learning (RL) has emerged as a central paradigm for acquiring complex skills. By leveraging high-dimensional perceptual inputs (e.g., vision or proprioception) and reward signals as feedback, RL enables agents to learn control policies through trial-and-error interaction with the environment.
This section reviews RL methods for robotic manipulation from both theoretical and application perspectives. We categorize existing approaches into two main classes: model-free and model-based algorithms, depending on whether the agent exploits an explicit or learned dynamics model to guide the learning process. Representative methods across these categories are summarized in Table~\ref{tab: rl}.

\textbf{i) Model-Free Methods}

In robotic manipulation, model-free RL learns policies or value functions directly from interaction data without explicitly modeling environment dynamics. Its flexibility makes it suitable for high-dimensional and contact-rich manipulation, but typically at the cost of substantial interaction data and limited sample efficiency. Recent work therefore increasingly combines RL with offline data, pre-trained policies, learned rewards, and imitation priors to improve scalability and real-world applicability. We categorize these methods into three groups.

\begin{table*}[t]
\centering
\caption{Representative RL methods for manipulation tasks.}
\label{tab: rl}
\resizebox{\textwidth}{!}{
\begin{tabular}{lll}
\toprule
\textbf{Category} & \textbf{Subcategory} & \textbf{Representative Methods} \\
\midrule
\multirow{3}{*}{\textbf{Model-Free RL}}
    & Pre-Training & QT-Opt~\cite{dmitry2018qt}, PTR~\cite{kumar2022pre}, V-PTR~\cite{bhateja2023robotic}\\
\multirow{3}{*}{}
    & Fine-Tuning & Residual RL~\cite{johannink2019residual}, RLDG~\cite{xu2025rldg}, V-GPS~\cite{nakamoto2025steering}, PA-RL~\cite{mark2024policy}\\
\multirow{3}{*}{}
    & VLA-RL & iRe-VLA~\cite{guo2025improving}, RIPT~\cite{tan2025interactive}, VLA-RL~\cite{lu2025vla}, ConRFT~\cite{chen2025conrft}\\
\midrule
\multirow{3}{*}{\textbf{Model-Based RL}}
    & Imagination Trajectory Generation & Dreamer~\cite{hafner2020dream}, MWM~\cite{seo2023masked}\\
\multirow{3}{*}{}
    & Planning & GPS~\cite{levine2013guided}, TD-MPC~\cite{hansen2022temporal}\\
\multirow{3}{*}{}
    & Differentiable RL & SAPO~\cite{xing2025stabilizing}, SAM-RL~\cite{lv2023sam}, DiffTORI~\cite{wan2024difftori}\\
\bottomrule
\end{tabular}
}
\end{table*}

\noindent \textbf{RL in Pre-Training.}
RL pre-training aims to acquire reusable policies, value functions, rewards, or exploration behaviors before downstream adaptation. QT-Opt~\cite{dmitry2018qt} demonstrates scalable self-supervised RL for vision-based manipulation, while PTR~\cite{kumar2022pre} and Q-Transformer~\cite{chebotar2023q} show how offline RL and value learning can support rapid adaptation from large robot datasets. V-PTR~\cite{bhateja2023robotic} further incorporates human videos into value pre-training. Beyond offline supervision, PEAC~\cite{ying2024peac} performs reward-free cross-embodiment pre-training to learn task-agnostic and embodiment-aware knowledge, while TaskExp~\cite{zhu2025taskexp} uses multi-task pre-training with decision and perceptual objectives to improve generalization. ROBOFUME~\cite{yang2024robot} and ReWiND~\cite{zhang2025rewind} extend pre-training to learned rewards and multi-task policies, and DEAS~\cite{kim2026deas} improves long-horizon offline value learning over action sequences. Together, these methods broaden RL pre-training from policy and value initialization toward reward learning, unsupervised skill acquisition, and generalizable multi-task control.

\noindent \textbf{RL in Fine-Tuning.}
RL fine-tuning improves pre-trained policies through task-specific interaction while preserving prior capabilities and training stability. Residual methods~\cite{johannink2019residual,ankile2025imitation} and Policy Decorator~\cite{yuan2025policy} refine frozen base policies through corrective actions, while RLDG~\cite{xu2025rldg} distills RL-improved trajectories back into a generalist policy. Value-guided approaches such as V-GPS~\cite{nakamoto2025steering} and PA-RL~\cite{mark2024policy} improve action selection without directly modifying the base policy, and batch online RL~\cite{dong2026matters} alternates autonomous data collection with offline updates. For expressive generative policies, EXPO~\cite{dong2026expo} stabilizes online RL by learning a lightweight value-guided edit policy around an imitation-trained base policy, while Behavioral Mode Discovery~\cite{longhini2026behavioral} regularizes fine-tuning to prevent mode collapse. These approaches highlight a shift toward policy-agnostic, value-guided, and distribution-preserving RL post-training.

\noindent \textbf{RL for VLA Models.}
VLA, as a special class of generalist models, has attracted significant research attention in the field, with numerous studies focusing on designing efficient reinforcement learning post-training schemes for VLA models~\cite{liu2025can, guo2025improving, tan2025interactive, chen2025conrft, lu2025vla, shu2025rftf, chen2025tgrpo, song2025hume, wagenmaker2025steering}.
iRe-VLA~\cite{guo2025improving} proposed a robust post-training pipeline that iteratively executes online RL for efficient exploration and IL on both expert and collected online data.
RIPT~\cite{tan2025interactive} introduces a simple and critic-free VLA-RL framework that extends the Leave-One-Out PPO (LOOP~\cite{chen2025reinforcement}) algorithm to estimate advantage functions for each sampled trajectory, allowing significant performance improvements over supervised fine-tuning with minimal demonstrations through online RL.
ConRFT~\cite{chen2025conrft} proposes a comprehensive post-training pipeline that employs a unified consistency-based training objective that combines RL and IL in both the offline and online phases.

\textbf{ii) Model-Based Methods}

In robotic manipulation, model-based methods exploit explicit or learned environment dynamics to facilitate policy learning and action optimization. By predicting future states and rewards, they enable planning, imagination-based policy learning, and improved sample efficiency. However, their effectiveness depends on model fidelity, which remains challenging in high-dimensional and contact-rich manipulation. We categorize these methods into three groups.

\noindent \textbf{Imagination Trajectory Generation.}
The classical Dyna framework~\cite{sutton1991dyna} uses learned dynamics to generate additional transitions for policy learning. Modern approaches increasingly perform such imagination in compact latent spaces. The Dreamer series~\cite{hafner2020dream,hafner2021mastering,hafner2025mastering} learns latent dynamics from observations and optimizes policies through imagined rollouts, forming the basis of many subsequent visual MBRL methods~\cite{seo2022reinforcement,wu2023daydreamer,seo2023masked,rafailov2021offline}. DayDreamer~\cite{wu2023daydreamer} extends latent imagination to real physical robots, while MWM~\cite{seo2023masked} improves visual world modeling through masked representation learning.

Recent work further improves the data sources and representations used for imagination. VeoRL~\cite{pan2025video} learns an interactive world model from unlabeled videos and uses predicted long-term behaviors to guide offline RL. GWM~\cite{lu2025gwm} predicts action-conditioned future scenes using 3D Gaussian representations, enabling scalable world-model learning for robotic manipulation. MIST-WM~\cite{feng2026learning} instead couples active exploration with structured world-model learning to discover compact task-sufficient latent states, improving sample efficiency and generalization across tasks. These approaches extend imagination-based RL from generic latent prediction toward richer data sources, explicit 3D structure, and task-relevant dynamics.

\noindent \textbf{Planning.}
Model-based RL can also exploit learned dynamics for online action optimization. The GPS series~\cite{levine2013guided,levine2013variational,levine2014learning,levine2014learning2,levine2016end} uses locally fitted dynamics and trajectory optimization to generate supervision for policy learning. TD-MPC~\cite{hansen2022temporal} and TD-MPC2~\cite{hansen2024td} jointly learn latent dynamics, value functions, and policies, and optimize action sequences through model-predictive control. VLAPS~\cite{neary2025improving} combines VLA-derived action priors with model-based MCTS, enabling test-time search over candidate action sequences. More recently, MBDPO~\cite{cheng2026scaling} integrates search and policy improvement by formulating optimization over imagined world-model trajectories as diffusion policy optimization. Together, these methods shift model-based planning from local trajectory optimization toward scalable latent search and policy refinement.

\noindent \textbf{Differentiable RL.}
Differentiable RL exploits differentiable dynamics and rewards to optimize actions or policy parameters directly through model gradients. Although real-world dynamics are generally not differentiable, this capability can be provided by differentiable simulators or learned dynamics models. SAPO~\cite{xing2025stabilizing} leverages analytic simulation gradients for efficient policy optimization in dexterous manipulation. SAM-RL~\cite{lv2023sam} combines differentiable physics and rendering to jointly adapt simulation parameters and learn manipulation policies. DiffTORI~\cite{wan2024difftori} further integrates differentiable trajectory optimization with actor-critic learning, using model gradients to refine policy-generated trajectories.

\subsubsection{Imitation Learning}
\label{subsubsec: il}

In 1999, Schaal et al.~\cite{schaal1999imitation} proposed imitation learning (IL) as a key pathway for enabling robots to acquire efficient motor skills, visuomotor coordination, and modular control. Since then, IL has evolved into a central paradigm for learning complex manipulation behaviors, with its algorithms, recent developments, and challenges comprehensively reviewed in~\cite{zare2024survey}. Compared with RL, IL avoids costly reward design and extensive environment interaction. Early studies were largely grounded in control-theoretic formulations with limited perception--action integration, while subsequent work progressively adopted deep architectures such as ResNets and Transformers, large-scale visual and multimodal pretraining, and more recently large language and vision models. Current advances emphasize multimodal fusion, robust and efficient policy learning, scalable data utilization, and generative modeling.

Looking forward, promising directions include foundation-model-driven IL that integrates VLA models and LLMs with robotics, causal IL based on counterfactual reasoning, generalization and transfer across tasks and embodiments, safe and reliable deployment, and more efficient human--robot interaction with reduced dependence on continuous human feedback. In this subsection, we primarily focus on state-based and visual imitation learning, as summarized in Figure~\ref{fig: il}. Language-related approaches are discussed separately in Section~\ref{subsec: vla}.


\newcommand{\ilorient}[1]{%
  \rotatebox{90}{#1}%
}

\newcommand{\ilcat}[1]{%
  \parbox[c]{40mm}{\raggedright #1}%
}

\newcommand{\ilsubcat}[1]{%
  \parbox[c]{28mm}{\raggedright #1}%
}

\newcommand{\ilitem}[1]{%
  \parbox[c]{72mm}{\raggedright #1}%
}

\begin{figure*}[!t]
\centering
\scriptsize

\forestset{
  iltaxo/.style={
    for tree={
      grow'=0,
      parent anchor=east,
      child anchor=west,
      anchor=west,
      edge path'={
        (!u.parent anchor) -- +(3pt,0) |- (.child anchor)
      },
      edge={draw=darkgray, line width=0.6pt},
      rounded corners,
      draw,
      minimum height=4.8mm,
      inner xsep=2.5pt,
      inner ysep=2.0pt,
      s sep=3pt,
      l sep=6pt,
      font=\scriptsize
    },
    orient/.style={
      draw,
      rounded corners=2pt,
      fill opacity=.5,
      inner xsep=2.5pt,
      inner ysep=2pt
    },
    cat-lfa/.style={
      draw,
      rounded corners=2pt,
      fill=yellow!18,
      inner xsep=2.5pt,
      inner ysep=2pt
    },
    item-lfa/.style={
      draw,
      rounded corners=2pt,
      fill=yellow!32,
      inner xsep=2.5pt,
      inner ysep=2pt
    },
    cat-lfo/.style={
      draw,
      rounded corners=2pt,
      fill=hidden-blue!18,
      inner xsep=2.5pt,
      inner ysep=2pt
    },
    item-lfo/.style={
      draw,
      rounded corners=2pt,
      fill=hidden-blue!32,
      inner xsep=2.5pt,
      inner ysep=2pt
    }
  }
}

\begin{forest} iltaxo
[{\rotatebox{90}{\textbf{Imitation Learning Methods (\S~\ref{subsubsec: il})}}},
 draw, inner sep=2pt, fill opacity=.5,
  [{\ilorient{\textbf{Imitation from Action}}}, orient
    [{\ilcat{Action-level IL}}, cat-lfa
      [{\ilsubcat{BC from Action}}, cat-lfa
        [{\ilitem{
          \cite{xie2020deep},
          Data Scaling Laws~\cite{lin2025data},
          BC-Z~\cite{jang2022bc},
          Most of VA/VLA/TVLA,
          Diffusion Policies,
          \textit{etc}
        }}, item-lfa]
      ]
      [{\ilsubcat{BC from Pose}}, cat-lfa
        [{\ilitem{
          PoseInsert~\cite{sun2025exploring},
          ZeroMimic~\cite{shi2025zeromimic},
          RiEMann~\cite{gao2025riemann}
        }}, item-lfa]
      ]
      [{\ilsubcat{MP-based IL}}, cat-lfa
        [{\ilitem{
          DMP~\cite{schaal2003computational,ijspeert2013dynamical},
          ProMP~\cite{paraschos2013probabilistic},
          HSMM~\cite{tanwani2018generalizing},
          Language Movement Primitives~\cite{dai2026language},
          FODMP~\cite{shi2026fodmp}
        }}, item-lfa]
      ]
      [{\ilsubcat{Search-based IL}}, cat-lfa
        [{\ilitem{
          \cite{billard2004discovering},
          \cite{ratliff2009learning},
          \cite{duan2024structured},
          Guided TAMP Imitation~\cite{mcdonald2022guided},
          SAILOR~\cite{jain2025smooth}
        }}, item-lfa]
      ]
      [{\ilsubcat{Optimization-based IL}}, cat-lfa
        [{\ilitem{
          \cite{mronga2021learning},
          \cite{dhonthi2022optimizing},
          LeTO~\cite{xu2024leto},
          NODE-IL~\cite{zhao2024neural},
          SCDS~\cite{abyaneh2025contractive}
        }}, item-lfa]
      ]
    ]
    [{\ilcat{Reward-based IL}}, cat-lfa
      [{\ilsubcat{Inverse Reinforcement Learning}}, cat-lfa
        [{\ilitem{
          Inverse KKT~\cite{englert2017inverse},
          SPRINQL~\cite{hoang2024sprinql},
          Masked IRL~\cite{hwang2025masked},
          Constrained Demonstrators~\cite{li2026robot}
        }}, item-lfa]
      ]
      [{\ilsubcat{Adversarial Imitation Learning}}, cat-lfa
        [{\ilitem{
          GAIL~\cite{ho2016generative},
          Option-GAIL~\cite{jing2021adversarial},
          LAPAL~\cite{wang2022latent},
          DRAIL~\cite{lai2024diffusion},
          OLLIE~\cite{yue2024ollie},
          C-LAIfO~\cite{giammarino2025visually}
        }}, item-lfa]
      ]
    ]
    [{\ilcat{Representation Learning for IL}}, cat-lfa
      [{\ilsubcat{Latent Learning for IL}}, cat-lfa
        [{\ilitem{
          \cite{tamar2018imitation}, \cite{ren2021generalization}, \cite{stepputtis2022system}, Riemannian IL~\cite{zeestraten2017approach},
          TRAIL~\cite{zolna2021task},
          MCNN~\cite{sridhar2024memory},
          BC-IB~\cite{bai2025rethinking},
          Latent Policy Barrier~\cite{sun2025latent}
        }}, item-lfa]
      ]
      [{\ilsubcat{In-Context Imitation Learning}}, cat-lfa
        [{\ilitem{
          \cite{dasari2021transformers},
          ICRT~\cite{fu2024context},
          Instant Policy~\cite{vosylius2025instant},
          Robust Instant Policy~\cite{oh2025robust},
          HiST-AT~\cite{fateh2026hierarchical},
          ICLR~\cite{nguyen2026iclr}
        }}, item-lfa]
      ]
    ]
    [{\ilcat{Interactive IL}}, cat-lfa
      [{\ilitem{
        TIPS~\cite{jauhri2021interactive},
        IWR~\cite{mandlekar2020human},
        Sirius~\cite{liu2022robot},
        LILAC~\cite{cui2023no},
        Lazy~\cite{hoque2021lazydagger}, Thrift~\cite{hoque2022thriftydagger},
        CHG-DAgger~\cite{takahashi2024chg},
        FlowDAgger~\cite{murray2026flowdagger},
        Set-Supervised DP~\cite{li2026set},
        ARMADA~\cite{yu2026armada}
      }}, item-lfa]
    ]
    [{\ilcat{Robustness and Efficiency}}, cat-lfa
      [{\ilsubcat{Robustness}}, cat-lfa
        [{\ilitem{
          DART~\cite{laskey2017dart},
          \cite{valletta2021imitation},
          \cite{oh2023bayesian},
          Counter-BC~\cite{sagheb2025counterfactual},
          ADC~\cite{huang2025adversarial},
          CCIL~\cite{ke2024ccil},
          CREST~\cite{lee2021causal},
          RAIL~\cite{jung2025rail},
          FAIL-Detect~\cite{xu2025can}
        }}, item-lfa]
      ]
      [{\ilsubcat{Efficiency}}, cat-lfa
        [{\ilitem{
          \cite{lin2022efficient},
          TEDA~\cite{ge2024bridging},
          SRIL~\cite{xie2024subconscious},
          DemoSpeedup~\cite{guo2025demospeedup},
          SAIL~\cite{arachchige2025sail}
        }}, item-lfa]
      ]
    ]
  ]
    [{\ilorient{\textbf{Imitation from Observation}}}, orient
      [{\ilcat{Reward or Occupancy Recovery from Observation}}, cat-lfo
        [{\ilitem{
          \cite{alakuijala2023learning},
          \cite{liu2024imitation},
          DIFO~\cite{huang2024diffusion},
          DILO~\cite{sikchi2025dual}
        }}, item-lfo]
      ]
      [{\ilcat{State or Representation Alignment and Matching}}, cat-lfo
        [{\ilitem{
          \cite{chella2006cognitive},
          \cite{sieb2020graph},
          SILO~\cite{lee2020follow},
          VGS-IL~\cite{jin2020geometric},
          WHIRL~\cite{bahl2022human},
          ZeST~\cite{cui2022can},
          NIFT~\cite{huang2023nift},
          VIP~\cite{ma2023vip},
          LIV~\cite{ma2023liv}
        }}, item-lfo]
      ]
      [{\ilcat{Model- and Physics-Grounded LfO}}, cat-lfo
        [{\ilitem{
          \cite{chen2023imitation},
          Diff-LfD~\cite{zhu2023diff}
        }}, item-lfo]
      ]
      [{\ilcat{Behavior Cloning from Observation}}, cat-lfo
        [{\ilitem{
          Sensorimotor Primitives~\cite{liang2022learning},
          Point Policy~\cite{haldar2025point}
        }}, item-lfo]
      ]
      [{\ilcat{Observation-to-Action Reconstruction and Retargeting}}, cat-lfo
        [{\ilitem{
          \cite{bharadhwaj2023zero},
          MimicFunc~\cite{tang2025mimicfunc},
          Do as I Do~\cite{paliwal2026dexterous},
          EgoInfinity~\cite{wang2026egoinfinity}
        }}, item-lfo]
      ]
      [{\ilcat{Command-, Goal-, and Planning-Based LfO}}, cat-lfo
        [{\ilitem{
          Learning Actions from Human Demonstration Video~\cite{yang2019learning},
          Motion Reasoning~\cite{huang2020motion},
          Cago~\cite{duan2025learning}
        }}, item-lfo]
      ]
    ]
]
\end{forest}
\caption{A structured taxonomy of imitation learning methods organized by the source of supervision, including imitation from action and imitation from observation.}
\label{fig: il}
\end{figure*}

i) \textbf{Imitation from Action}

Imitation Learning from Action assumes access to expert demonstrations in the form of state–action pairs, where both the sensory states (e.g., robot configurations and environment observations) and the corresponding expert control commands (e.g., end-effector actions and poses) are available.

\noindent \textbf{Action-level IL.}
Action-level IL focuses on directly learning control policies from expert demonstrations at the action or trajectory level. Rather than inferring high-level goals or task abstractions, these methods map observed states to executable actions, poses, or motion primitives, thereby emphasizing precise motor control, trajectory generation, and stability in manipulation.

\textbullet\hspace{0.2em} \textbf{Behavior Cloning (BC).} 
\textit{BC from Action} learns a direct mapping from states to low-level control commands based on expert state–action pairs. Early studies relied on non-parametric or regression-based methods~\cite{huang2019non}, later expanding into hierarchical architectures~\cite{katz2016imitation, sharma2019third, xie2020deep}, multimodal generalization~\cite{hao2023masked}, stability-constrained training~\cite{totsila2023end, jung2025rail}, cross-embodiment transfer, human–robot collaboration~\cite{wang2024genh2r}, and the integration of structural priors~\cite{kawaharazuka2021imitation, wagenmaker2025behavioral, dai2025civil, tayal2025genosil}. More recently, efficient Transformer-based multi-task policies have improved scalability and reusability under limited data conditions~\cite{xu2025speci}.
Additional studies investigate fine-tuning strategies for model components~\cite{zhang2025effective}, architectural comparisons~\cite{drolet2024comparison}, and scaling laws~\cite{lin2025data}, while others propose end-to-end frameworks encompassing data collection, training, and deployment~\cite{mandi2022cacti, dalal2023imitating}. Overall, BC from action has evolved beyond supervised imitation into a paradigm emphasizing efficiency, robustness, structural priors, and adaptability, with growing emphasis on language-integrated interaction that advances action-level supervision toward general-purpose manipulation.

\textit{BC from Pose} abstracts away low-level control by predicting end-effector poses in SE(3), leaving tracking to IK or force controllers. This geometry-aware formulation improves precision, transferability, and robustness. Representative works include PoseInsert~\cite{sun2025exploring}, which leverages relative SE(3) as a core representation to learn pose-guided policies for sub-millimeter insertion, optionally fusing RGB-D cues; ZeroMimic~\cite{shi2025zeromimic}, which distills skills from in-the-wild human videos into deployable, image-goal–conditioned policies via pose and geometry alignment; and RiEMann~\cite{gao2025riemann}, which develops an SE(3)-equivariant pipeline to predict 6-DoF target poses in real time without explicit segmentation. Collectively, these approaches highlight the strengths of pose-level cloning for high-precision tasks and generalization across scenes and embodiments.

\textbullet\hspace{0.2em} \textbf{MP-based IL.} 
Movement primitives (MPs) encode demonstrations as parameterized trajectories or dynamical systems, providing compact and adaptable skill representations. Dynamic Movement Primitives (DMPs) introduced stable attractor systems for reproducing and adapting demonstrated motions~\cite{schaal2003computational,ijspeert2013dynamical}, while Probabilistic Movement Primitives (ProMPs) model trajectory distributions for uncertainty handling and skill blending~\cite{paraschos2013probabilistic}. Hidden Semi-Markov Models (HSMMs) further capture temporal variability through skill segmentation and sequencing~\cite{tanwani2018generalizing}. MPs have also been integrated with task constraints for safe and flexible execution~\cite{perico2019combining}, hybrid force/motion primitives for compliant manipulation~\cite{wang2020framework}, and dynamical systems for visual servoing~\cite{paolillo2023dynamical}. Recent work connects MPs with foundation and generative models. Language Movement Primitives~\cite{dai2026language} grounds VLM reasoning into DMP parameters for language-conditioned manipulation, while FODMP~\cite{shi2026fodmp} combines ProDMP trajectory representations with one-step diffusion to generate temporally structured motions efficiently. Overall, MP-based IL has evolved from trajectory reproduction toward probabilistic, constraint-aware, multimodal, and generative skill representations.

\textbullet\hspace{0.2em} \textbf{Search-based IL.} 
Search-based IL treats imitation as planning or search over trajectories, policies, or outcomes rather than direct action regression. Early work formulated imitation as a search for optimal strategies~\cite{billard2004discovering}, followed by Learning to Search approaches using functional-gradient optimization~\cite{ratliff2009learning}. Planner-in-the-loop methods further combine imitation with structured search. Guided Imitation of Task and Motion Planning~\cite{mcdonald2022guided} distills TAMP-generated subgoals and trajectories into hierarchical policies, while partially learned policies in turn accelerate subsequent planning. More recently, SAILOR~\cite{jain2025smooth} learns a world model and reward model from demonstrations and performs test-time search toward expert outcomes, enabling recovery from states outside the demonstration distribution. These methods shift imitation from reproducing expert actions toward learning how to search for expert-like outcomes, improving robustness to covariate shift and long-horizon errors.

\textbullet\hspace{0.2em} \textbf{Optimization-based IL.} 
Optimization-based IL integrates imitation objectives with trajectory optimization, constraint satisfaction, or differentiable planning. Early methods incorporate task constraints into policy learning~\cite{mronga2021learning} and optimize demonstrated skills under temporal-logic specifications~\cite{dhonthi2022optimizing}. More recent approaches integrate optimization more tightly with policy learning. LeTO~\cite{xu2024leto} embeds differentiable trajectory optimization into end-to-end visuomotor learning, while NODE-IL~\cite{zhao2024neural} uses continuous-time Neural ODE dynamics for long-horizon multi-skill manipulation. Contractive dynamical imitation policies~\cite{abyaneh2025contractive} further impose contractive dynamics to improve stability and recovery under out-of-distribution perturbations. Overall, this line has evolved from constraint-aware trajectory optimization toward differentiable and stability-aware policy learning.

\noindent \textbf{Reward-based IL.}
Reward-based IL infers reward or cost functions from demonstrations and optimizes policies under the learned objectives. Unlike behavior cloning, which directly fits expert actions, reward-based methods model the underlying objectives that explain demonstrated behavior, potentially improving generalization and adaptation beyond the demonstration distribution.

\textbullet\hspace{0.2em} \textbf{Inverse Reinforcement Learning (IRL).}
IRL seeks a reward or cost under which expert trajectories are approximately optimal and derives a policy through planning or RL. Inverse KKT~\cite{englert2017inverse} formulates manipulation IRL as inverse optimal control, recovering task costs and active constraints from demonstrations through KKT conditions. SPRINQL~\cite{hoang2024sprinql} instead learns an inverse soft Q-function from mixed-quality offline demonstrations, emphasizing expert data without adversarial training or online interaction. A divergence-minimization perspective~\cite{ghasemipour2020divergence} further unifies BC, GAIL, AIRL, and related methods through occupancy-distribution matching.
Recent work addresses ambiguity and limitations in demonstrations. Masked IRL~\cite{hwang2025masked} combines demonstrations with language and uses LLMs to identify task-relevant state factors, reducing reward ambiguity and spurious correlations. Learning from constrained demonstrators~\cite{li2026robot} infers state-based progress rewards from restricted demonstrations, allowing subsequent optimization to surpass the demonstrated behavior. These developments extend IRL toward language-guided reward disambiguation, imperfect demonstrations, and interactive model-based learning.

\textbullet\hspace{0.2em} \textbf{Adversarial Imitation Learning (AIL).}
AIL learns imitation objectives by discriminating expert behavior from agent-generated trajectories, with GAIL~\cite{ho2016generative} establishing the occupancy-matching paradigm. Subsequent work improves visual representation and robustness through latent world models and contrastive learning~\cite{rafailov2021visual,giammarino2025visually}. Reward modeling has also been strengthened through diffusion-based discriminators and preference supervision~\cite{lai2024diffusion,taranovic2023adversarial}. Efficiency is improved through offline-to-online learning~\cite{yue2024ollie}, discriminator-guided model-based imitation~\cite{zhang2023discriminator}, and trajectory augmentation and correction~\cite{antotsiou2021adversarial}. Structured policies further address long-horizon and high-dimensional control through hierarchical options and latent action spaces~\cite{jing2021adversarial,wang2022latent}, while auxiliary-task exploration, actor-critic formulations, and empirical studies improve exploration and training stability~\cite{ablett2021learning,deka2023arc,orsini2021matters}.

\noindent \textbf{Representation Learning for IL.}
Representation learning for IL extracts task-relevant, structured, and transferable representations from demonstrations to improve policy efficiency, robustness, and adaptation. Rather than directly fitting observation--action mappings, these methods shape latent spaces, temporal representations, or demonstration-conditioned contexts that capture the factors underlying expert behavior.

\textbullet\hspace{0.2em} \textbf{Latent Learning for IL.}
Latent representation learning seeks compact task-relevant embeddings while suppressing nuisance information that can impair imitation. Task-relevant adversarial IL~\cite{zolna2021task} restricts representations to behaviorally relevant information, while information-bottleneck behavior cloning~\cite{bai2025rethinking} explicitly compresses perceptual inputs into task-sufficient latents to reduce representational redundancy. Geometry-aware formulations further model demonstrations on Riemannian manifolds~\cite{zeestraten2017approach}, while temporal and memory-based approaches capture dependencies beyond individual observations~\cite{jia2020vision,liu2021robotic,sridhar2024memory}. More recently, Latent Policy Barrier~\cite{sun2025latent} treats the latent distribution of expert demonstrations as an implicit in-distribution region and optimizes predicted future latents to remain within this region, improving robustness to covariate shift. Together, these methods extend latent learning from compact representation toward geometry-, memory-, and distribution-aware policy representations.

\textbullet\hspace{0.2em} \textbf{In-Context Imitation Learning (ICIL).}
ICIL conditions a pre-trained policy on demonstrations at inference time, enabling adaptation to new tasks without parameter updates. Early approaches represent demonstrations through visual sequences, relational graphs, or sensorimotor tokens~\cite{dasari2021transformers,vosylius2023few,fu2024context,vosylius2025instant}. Robust Instant Policy~\cite{oh2025robust} further aggregates candidate trajectories with a Student's-$t$ formulation to reduce sensitivity to outliers. Recent work increasingly focuses on richer context representations and scalable training. HiST-AT~\cite{fateh2026hierarchical} learns hierarchical spatiotemporal action tokens that capture both action structure and temporal progression, while ICLR~\cite{nguyen2026iclr} augments demonstrations with visual reasoning traces of anticipated robot motion. These developments shift ICIL from direct demonstration conditioning toward structured action tokenization, intermediate visual reasoning, and scalable context learning.

\noindent \textbf{Interactive Imitation Learning (IIL).}
IIL improves imitation policies through corrective feedback collected during policy execution, mitigating distribution shift while reducing the need for complete expert demonstrations. Early approaches primarily rely on human teleoperation, shared control, or language feedback to correct policy failures~\cite{muhlig2012interactive,mandlekar2020human,chisari2022correct,mandlekar2023human,cui2023no,kasaei2024vital,wu2025robocopilot}. Language-based methods further enable intuitive online correction, while LLM-based teachers provide corrective or evaluative feedback with reduced human supervision~\cite{werner2025llm}.

Recent work increasingly adapts generative policies from sparse interventions. FlowDAgger~\cite{murray2026flowdagger} maps human corrective actions into the latent noise space of frozen flow- or diffusion-based policies and learns a latent policy for adaptation. Set-Supervised Diffusion Policy~\cite{li2026set} exploits paired erroneous and corrected action chunks, using positive and negative supervision to improve diffusion-policy learning. For dexterous manipulation, Hand-in-the-Loop~\cite{li2026hand} smoothly blends human corrections with autonomous execution to avoid discontinuities during hand--arm takeover. LazyDAgger and ThriftyDAgger~\cite{hoque2021lazydagger,hoque2022thriftydagger} reduce unnecessary queries through switching- and risk-aware intervention strategies, while Sirius~\cite{liu2022robot} integrates human intervention with continual learning during deployment. Model-based runtime monitoring~\cite{liu2024model} anticipates failures and triggers corrective interaction, while ARMADA~\cite{yu2026armada} combines failure detection with shared control for scalable real-world adaptation. Overall, IIL is evolving from continuous human correction toward selective intervention and lightweight adaptation of generative and foundation policies.

\noindent \textbf{Robustness and Efficiency.}
Robustness and efficiency in IL address two critical requirements for deploying manipulation policies in real-world environments: reliability under uncertainty and scalability under resource constraints. Robustness focuses on ensuring safe and stable execution in the presence of imperfect data, sensor noise, and distribution shifts, while efficiency emphasizes accelerating learning and inference to achieve real-time performance and practical deployment.

\textbullet\hspace{0.2em} \textbf{Robustness.}
Recent work on robust and safe imitation learning for manipulation consolidates three threads. 
First, robustness to imperfect data is advanced along three lines. Methods either clean and reweight demonstrations via counterfactual consistency and uncertainty-aware handling of label conflicts~\cite{valletta2021imitation, sagheb2025counterfactual}, augment and stress-test policies through adversarial human-in-the-loop collection~\cite{huang2025adversarial}, continuity-based corrective augmentation~\cite{ke2024ccil}, and Bayesian or classical noise injection~\cite{oh2023bayesian, laskey2017dart}, or exploit causal interventions~\cite{lee2021causal} to isolate task-relevant state variables. 
Second, safety and constraints are enforced via reachability-based safety filters that wrap IL policies and provably reduce collisions without sacrificing task success~\cite{jung2025rail}. 
Third, failure handling under distribution shift combines object-centric inverse-policy recovery to drive states back toward the training manifold and failure detection without failure data via sequential OOD tests with uncertainty quantification and conformal guarantees~\cite{xu2025can, gao2025out}. Collectively, these advances move IL toward data-efficient, safety-aware, and deployment-ready manipulation.

\textbullet\hspace{0.2em} \textbf{Efficiency.}
A recent line of work pushes imitation learning for manipulation toward faster execution and practical deployment.
Current techniques chiefly accelerate training and inference through accelerated or resampled demonstrations, model slimming via parameter reduction, quantization, pruning and distillation, action scheduling, and asynchronous parallelism, thereby improving both throughput and latency~\cite{lin2022efficient, ge2024bridging, xie2024subconscious, guo2025demospeedup, arachchige2025sail}.

\textbf{ii) Imitation from Observation}

Imitation Learning from Observation (LfO) assumes access to expert demonstrations as state or visual trajectories without corresponding robot action labels. Compared with imitation from action, the central challenge is therefore to bridge the missing action supervision between observed behavior and executable robot control. Existing approaches address this gap in different ways, including recovering learning objectives, aligning observation spaces, exploiting physical models, learning policies from reconstructed supervision, retargeting observed motions into robot trajectories, or inferring task goals and commands.

\noindent \textbf{Reward or Occupancy Recovery from Observation.}
This line infers rewards, utilities, or occupancy-matching objectives directly from observation-only demonstrations and subsequently optimizes policies through RL or planning. Representative approaches include observation-only IRL with automatic discount scheduling~\cite{alakuijala2023learning}, human-video reward learning for manipulation~\cite{liu2024imitation}, and diffusion-based adversarial LfO~\cite{huang2024diffusion}. DILO~\cite{sikchi2025dual} instead derives a dual occupancy-matching objective that learns a multi-step utility from offline interaction data, avoiding explicit inverse dynamics or adversarial discriminators. Rather than reconstructing expert actions, these methods transform observed behavior into an optimization objective that guides subsequent policy learning.

\noindent \textbf{State or Representation Alignment and Matching.}
A second strand reduces the discrepancy between demonstrations and robot execution by aligning states or latent representations across viewpoints, domains, and embodiments~\cite{chella2006cognitive,sieb2020graph,lee2020follow,jin2020geometric,kalinowska2021ergodic,bahl2022human}. VIP and LIV~\cite{ma2023vip,ma2023liv} learn transferable visual representations and progress-aware objectives from videos, while ZeST~\cite{cui2022can} leverages foundation-model representations for zero-shot task specification. NIFT~\cite{huang2023nift} represents human--object interactions through transferable interaction fields, while selective and ergodic matching methods determine which aspects of demonstrated behavior should be reproduced~\cite{lee2020follow,kalinowska2021ergodic}. These methods do not necessarily reconstruct explicit expert actions, but instead establish shared task-relevant representations in which observed human and robot behaviors can be compared or matched.

\noindent \textbf{Model- and Physics-Grounded LfO.}
Model- and physics-grounded methods exploit environment dynamics, contact models, or differentiable simulation to constrain the mapping from observed state evolution to executable behavior. Imitation Learning as State Matching~\cite{chen2023imitation} optimizes control by differentiating through physical dynamics to match demonstrated state trajectories, while Diff-LfD~\cite{zhu2023diff} combines differentiable physics and rendering to recover physically consistent manipulation behavior from visual demonstrations. Such physical grounding is particularly useful when similar visual trajectories may arise from substantially different actions, contacts, or underlying dynamics.

\noindent \textbf{Behavior Cloning from Observation.}
A more direct family of LfO methods ultimately learns executable policies from observation-only demonstrations, despite the absence of original robot action labels. Instead of treating the observed trajectory only as a reward or goal, these approaches infer task-relevant action supervision, intermediate control targets, or robot-aligned representations and use them to learn observation-to-action policies. Liang et al.~\cite{liang2022learning} learn both low-level sensorimotor primitives and high-level manipulation policies directly from raw human demonstration videos, jointly recovering intermediate subgoals and sequential task structure without manually annotated actions or subtask boundaries. Point Policy~\cite{haldar2025point} similarly learns robot policies exclusively from offline human videos by representing both human observations and robot actions through semantically meaningful 3D keypoints, thereby translating human hand motion into morphology-agnostic robot-space supervision. These methods extend the behavior-cloning paradigm from directly provided expert actions to action supervision inferred or structured from observation-only demonstrations.

\noindent \textbf{Observation-to-Action Reconstruction and Retargeting.}
Another family focuses primarily on reconstructing executable robot trajectories from observed human behavior rather than directly learning a policy within the reconstruction procedure. Bharadhwaj et al.~\cite{bharadhwaj2023zero} extract agent-agnostic action representations from passive human videos and transform predicted human hand trajectories into robot-executable motions for zero-shot manipulation. MimicFunc~\cite{tang2025mimicfunc} introduces function-centric correspondences for tool manipulation, constructing a local function frame from keypoints in a single RGB-D human demonstration and transferring its motion to geometrically different but functionally equivalent tools. More recent pipelines explicitly reconstruct 3D or 4D hand--object interactions before retargeting them to robotic embodiments. Do as I Do~\cite{paliwal2026dexterous} recovers hand--object interactions from monocular human videos and retargets them into executable trajectories for dexterous robotic hands, while EgoInfinity~\cite{wang2026egoinfinity} lifts in-the-wild videos into metric 4D hand--object representations and compiles recovered human motions into robot trajectories across diverse embodiments. The resulting robot-space trajectories can either be executed directly or serve as automatically generated demonstrations for subsequent behavior cloning, making reconstruction and retargeting an important bridge between passive human video and policy learning.

\noindent \textbf{Command-, Goal-, and Planning-Based LfO.}
A complementary direction abstracts away the demonstrator's exact motion and instead recovers semantic commands, symbolic goals, or intermediate states that specify what the robot should accomplish. Yang et al.~\cite{yang2019learning} formulate human-video imitation as video-to-command learning, combining manipulation-oriented visual understanding with video captioning to infer semantic commands that are subsequently grounded into robot execution. Motion Reasoning~\cite{huang2020motion} instead infers symbolic task goals from third-person demonstrations and combines task and motion reasoning to distinguish intended task effects from incidental state changes, allowing the recovered goal to be replanned in a different environment. Cago~\cite{duan2025learning} uses demonstrated states as goal supervision and selects intermediate goals according to the learner's current capabilities, constructing an adaptive curriculum for long-horizon policy learning. By preserving task intent rather than reproducing embodiment-specific actions, these approaches are particularly suitable when human and robot motions differ substantially but the desired task outcome remains shared.

\subsubsection{Bridging Reinforcement and Imitation Learning}

Reinforcement learning (RL) and imitation learning (IL) provide complementary supervision for robotic manipulation: IL offers efficient learning from demonstrations, while RL enables autonomous exploration and improvement beyond demonstrated behavior. Their integration has evolved from sequential pre-training and fine-tuning toward shared formulations, learned rewards, interleaved optimization, and human-guided adaptation. We organize these approaches into seven categories, as summarized in Table~\ref{tab: rl-il}.

\noindent \textbf{Methods Supporting RL and IL.}
Some methods provide shared formulations or components applicable to both RL and IL. Dual RL~\cite{liu2023frame} establishes a unified optimization perspective that connects several reinforcement and imitation learning objectives through common dual formulations. Other techniques, such as frame mining, provide reusable data-processing mechanisms that benefit both paradigms. These approaches expose methodological commonalities between RL and IL without prescribing a specific direction of knowledge transfer.

\noindent \textbf{IL Pre-training + RL Fine-tuning.}
A dominant paradigm initializes policies from demonstrations and subsequently uses RL to improve performance beyond the demonstrated distribution. Earlier approaches combine imitation-derived skills, action priors, or value objectives with RL refinement~\cite{pastor2011skill,pertsch2021guided,zhan2021framework,james2022q,watson2023coherent,bhaskar2024planrl}. Recent work increasingly targets generative and large robot policies. TMRL modulates diffusion timesteps during pre-training to preserve exploration capacity for subsequent RL, while Q2RL extracts Q-values from behavior-cloned policies to bootstrap on-robot RL. RL-100 further combines imitation initialization with offline and online RL for scalable real-world manipulation, and latent-steering approaches constrain RL refinement to compact policy representations. Overall, this direction shifts IL from a final policy-learning objective toward an initialization for autonomous post-training.

\begin{table*}[t]
\centering
\caption{Representative methods combining RL and IL for manipulation tasks.}
\label{tab: rl-il}
\resizebox{\textwidth}{!}{
\begin{tabular}{lp{0.75\textwidth}}
\toprule
\textbf{Category} & \textbf{Representative Methods} \\
\midrule

Methods Supporting RL and IL
& Frame Mining~\cite{liu2023frame}, Dual RL~\cite{sikchi2024dualrl} \\

IL Pre-training + RL Fine-tuning
& \cite{pastor2011skill}, SkiLD~\cite{pertsch2021guided}, FERM~\cite{zhan2021framework},
Q-attention~\cite{james2022q}, CSIL~\cite{watson2023coherent},
PLANRL~\cite{bhaskar2024planrl}, Sketch-to-Skill~\cite{yu2025sketch},
TMRL~\cite{hong2026tmrl},
RL-100~\cite{lei2025rl}, ZPRL~\cite{yu2026beyond}, \cite{dodeja2026life} \\

RL Pre-training + IL Fine-tuning
& MoPA-PD~\cite{liu2022distilling}, \cite{lee2022spend} \\

Reward Shaping from IL
& MaxEnt IOC~\cite{finn2016guided}, \cite{brown2020better},
\cite{liu2018imitation}, LbW~\cite{xiong2021learning},
\cite{lee2021generalizable}, \cite{haldar2023watch},
RoboCLIP~\cite{sontakke2023roboclip}, IRIS~\cite{mandlekar2020iris},
ORIL~\cite{zolna2020offline}, \cite{zhu2018reinforcement},
ResiP~\cite{ankile2025imitation}, ORCA~\cite{huey2025imitation},
Concept2Robot~\cite{shao2021concept2robot} \\

Joint RL--IL Optimization
& \cite{jena2021augmenting}, AW-Opt~\cite{lu2022aw},
\cite{mosbach2022accelerating}, IN-RIL~\cite{gao2025ril},
TSIL~\cite{jia2026temporal} \\

Human-in-the-Loop RL
& HIL-SERL~\cite{luo2025precise}, SiLRI~\cite{zhao2025real} \\

In-Context RL
& DCRL~\cite{dance2021conditioned} \\

\bottomrule
\end{tabular}
}
\end{table*}

\noindent \textbf{RL Pre-training + IL Fine-tuning.}
The reverse direction first obtains strong behaviors through RL, planning, or privileged supervision and subsequently transfers them into deployable policies through imitation. Motion-planner-augmented policies can be distilled into visual control policies~\cite{liu2022distilling}, while kickstarting and offline RL can generate informative experience that is later transferred to downstream policies~\cite{lee2022spend}. This paradigm is particularly useful when powerful training-time teachers rely on privileged states, planners, or expensive interaction that cannot be retained during deployment.

\noindent \textbf{Reward and Goal Shaping from IL.}
Demonstrations can also guide RL without directly supervising actions by defining rewards, values, or goal-progress signals. Early approaches infer objectives through inverse optimal control or demonstration ranking~\cite{finn2016guided,brown2020better}. Other methods derive perceptual or semantic rewards through representation similarity, temporal progress, or goal proximity~\cite{sermanet2017unsupervised,liu2018imitation,xiong2021learning,lee2021generalizable}. More recent approaches use optimal transport and foundation-model representations to convert demonstrations or videos into dense objectives~\cite{haldar2023watch,sontakke2023roboclip}, while temporally misaligned demonstrations can still provide useful reward signals for RL refinement~\cite{huey2025imitation}. These methods allow imitation information to supervise reinforcement learning without requiring direct action matching.

\noindent \textbf{Joint RL--IL Optimization.}
Rather than separating imitation and reinforcement into different stages, joint approaches combine their objectives throughout training. Demonstration losses can regularize RL updates or be alternated with reinforcement objectives to improve sample efficiency and stability~\cite{jena2021augmenting,lu2022aw,mosbach2022accelerating,gao2025ril}. Recent interleaved methods explicitly alternate RL and IL updates to reduce conflicts between their optimization signals, while temporal self-imitation converts successful trajectories discovered by RL into new imitation targets. This bidirectional interaction allows autonomous experience and demonstration supervision to improve each other during training.

\noindent \textbf{Human-in-the-Loop RL.}
Human interventions bridge imitation and reinforcement learning by providing corrections during autonomous exploration. HIL-SERL~\cite{luo2025precise} combines intervention data with real-world RL for precise manipulation while reducing exploration risk. Recent methods further handle imperfect interventions by weighting suboptimal corrections. Human feedback thus provides both demonstrations and safer exploration guidance.

\noindent \textbf{In-Context RL.}
In-context RL uses demonstrations as contextual information rather than explicit imitation targets. Demonstration-conditioned RL~\cite{dance2021conditioned} conditions policies on example trajectories while optimizing task return, enabling adaptation to new behaviors from only a few demonstrations. This formulation separates demonstration conditioning from imitation loss and provides an alternative route for combining few-shot imitation with reinforcement-based optimization.

Overall, RL--IL integration has moved beyond sequential training. Demonstrations can initialize policies, provide rewards or goals, regularize optimization, guide human-supervised exploration, or serve as contextual inputs. Conversely, RL experience can refine demonstrated behaviors and provide improved supervision for imitation, making RL and IL increasingly complementary within a unified policy-learning pipeline.

\subsubsection{Learning with Auxiliary Tasks}

\begin{figure}[!t]
\centering
\includegraphics[width=1.0\textwidth]{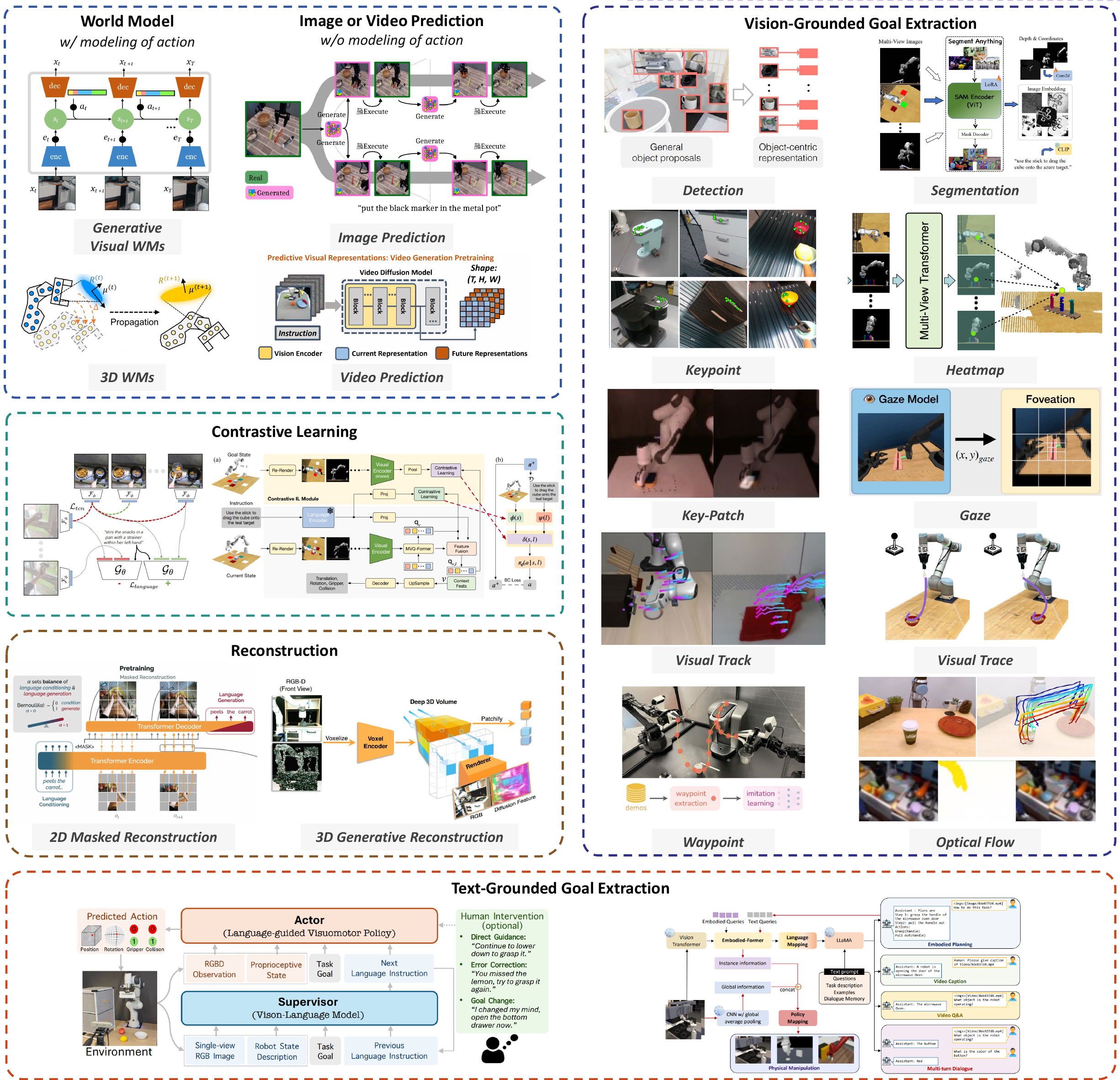}  
\caption{Overview of the taxonomy of methods for learning with auxiliary tasks, highlighting six core components: world models~\cite{mendonca2023structured, lu2024manigaussian}, image or video prediction~\cite{black2024zero, hu2025video}, contrastive learning~\cite{nair2023r3m, ma2025contrastive}, masking reconstruction~\cite{ze2023gnfactor, karamcheti2023language}, text-grounded goal extraction~\cite{mu2023embodiedgpt, dai2025racer}, and vision-grounded goal extraction~\cite{wang2025discovering, chuang2025look, fang2025keypoint, wi2023calamari, yu2024bikc, shi2023waypoint, mehta2025l2d2, wen2024any, xu2025flow}.
}
\label{fig: learning_with_auxiliary_tasks}
\end{figure}

Learning with auxiliary tasks refers to training paradigms that enhance policy learning by introducing additional self-supervised or weakly supervised objectives beyond the primary manipulation goal. These auxiliary objectives encourage the model to capture structured representations of the environment, actions, and goals, thereby improving sample efficiency, generalization, and robustness. As illustrated in Figure~\ref{fig: learning_with_auxiliary_tasks}, we summarize the most commonly used auxiliary tasks in current robotic learning frameworks.

\textbf{i) World Model}

World models (WMs) have become a central paradigm in robotic manipulation, supporting predictive planning, policy generalization, and data-efficient control. Formally, a WM learns an internal representation of environment dynamics, typically parameterizing the conditional distribution $p_{\theta}(s_{t+1} \mid s_{t}, a_{t})$, where $s_t \in S$ denotes the state (or observation) at time $t$ and $a_t \in A$ the agent’s action. By modeling such transitions, WMs enable agents to anticipate the outcomes of actions and perform rollouts in latent space, thereby reducing reliance on costly real-world interaction. This capability is especially critical in robotic manipulation, where physical contact, safety constraints, and limited data render direct trial-and-error learning impractical. Beyond the following research directions, WM-related approaches also overlap with the model-based RL methods discussed in Section~\ref{subsubsec: rl}.

\noindent \textbf{Generative Visual WMs.}
Generative visual world models learn to represent environment dynamics by synthesizing future visual observations conditioned on past states and actions. By treating video or image generators as interactive environments, such models enable rollouts in visual or latent space that can be queried for planning and policy learning. 
Recent advances span video–action diffusion pretraining on large robotic datasets~\cite{zhu2025unified, song2025physical}, transforming generic video diffusion into interactive predictive models~\cite{huang2026vid2world}, leveraging world models for trajectory generation in one-shot imitation~\cite{goswami2025osvi}, and compositional imagination for skill generalization~\cite{barcellona2025dream}. 
VLMPC further demonstrates how action-conditioned video prediction can be integrated with vision–language constraints and model-predictive control to enable closed-loop manipulation~\cite{zhao2024vlmpc}. 
Structured WMs from human videos also highlight how large-scale visual modeling can support manipulation via transfer and representation learning~\cite{mendonca2023structured}. 
Collectively, these works establish generative visual WMs as a promising route toward scalable, data-driven predictive control.

\noindent \textbf{3D-Structured or Physics-Grounded WMs.}
Beyond pixel-based prediction, these methods incorporate explicit spatial structures or physical priors, such as 3D Gaussians, occupancy fields, point clouds, and particle-based representations, to improve geometric and physical consistency. Gaussian-based approaches include Physically Embodied Gaussian Splatting~\cite{abou2025physically}, ManiGaussian~\cite{lu2024manigaussian}, ManiGaussian++~\cite{yu2025manigaussian++}, and GWM~\cite{lu2025gwm}. ORV~\cite{yang2026orv} instead uses 4D semantic occupancy to provide geometric and semantic guidance for action-conditioned video generation, improving temporal and multi-view consistency as well as sim-to-real transfer. Other approaches model dynamics through 3D point clouds, continuous fields, or flow-based representations, including ParticleFormer~\cite{huang2025particleformer}, GAF~\cite{chai2025gaf}, and 3DFlowAction~\cite{zhi20253dflowaction}. Collectively, these methods introduce explicit spatial or physical structure into predictive modeling to produce more reliable and control-relevant future states.

\noindent \textbf{Policy Learning with WMs.}
World models not only serve as predictive environments but also provide structured interfaces for policy learning and long-horizon control. A key direction lies in offline-to-online adaptation, where methods such as MOTO and FOWM pretrain on large offline datasets and then finetune policies in the real world~\cite{rafailov2023moto,feng2023finetuning}. Diffusion-policy adaptation has also been integrated into world models, as in DiWA, enabling efficient policy transfer across tasks~\cite{chandra2025diwa}. Robustness to language variation and viewpoint shifts is pursued in works such as LUMOS and ReViWo~\cite{nematollahi2025lumos,pang2025learning}. Another line of research focuses on generalist pretraining or multi-stage learning pipelines that couple reward, policy, and world model optimization~\cite{ren2023surfer,seo2023multi,wang2025modeling,zhao2025generalist,escoriza2025multi}. Beyond architecture, several methods exploit the latent dynamics of world models: Coupled distillation approaches design stable online imitation rewards directly in the latent space~\cite{li2025coupled}, while WoMAP explicitly models $p(z_{t+1} \mid z_t, a_t)$ and a reward predictor, using rollouts with MPC to guide latent-space action selection~\cite{yin2025womap}. Residual Plan further refines this by searching for corrective residuals within latent rollouts~\cite{jain2025smooth}. Together, these approaches demonstrate how structured WM interfaces can transform policy learning from raw trial-and-error into efficient, robust, and generalizable control.

\noindent \textbf{Systemization and Deployment.}
Beyond single-robot settings, recent efforts focus on scaling world models to fleet- and platform-level deployments, making them more practical and accessible in real-world robotics. Sirius-Fleet demonstrates multi-task fleet learning with shared visual world models across distributed robots~\cite{liu2025multi}, highlighting collaborative scalability. Cosmos proposes a foundation platform for physical AI, integrating modular world models to support diverse robotic systems~\cite{agarwal2025cosmos}. Similarly, Genie Envisioner introduces a unified world foundation platform for robotic manipulation, aiming at standardized pretraining and deployment pipelines~\cite{liao2026genie}. Together, these system-level initiatives emphasize robustness, interoperability, and scalability, pushing WMs beyond isolated benchmarks toward large-scale embodied AI platforms.

\textbf{ii) Image or Video Prediction}

Pure image or video prediction methods for robotics synthesize future visual signals or goal imagery without learning explicit action-conditioned forward dynamics. At deployment, generators may serve as control surrogates by providing visual plans, subgoals, or trajectories that controllers or inverse dynamics can follow, or as data and representation engines for pretraining, augmentation, and evaluation. 
Recent work includes video-as-policy and latent video planning~\cite{du2024video, xu2025vilp, liang2025video}, foundation-model pretraining with generative video~\cite{wu2024unleashing, li2025gr, cheang2024gr}, human-to-robot transfer through generated demonstrations~\cite{sharma2019third, bharadhwaj2024towards, bharadhwaj2025gen2act, heng2025rwor}, policy or inverse-dynamics learning $p(a_{t} \mid s_t,s_t+1)$ from predictive visual representations~\cite{liu2026foam, hu2025video, tian2025predictive}, controllable editing and goal imagery for data and goal synthesis~\cite{black2024zero, nie2025ermv}, visual planning for robust control with goal-expressive plans and subgoal filtering~\cite{bu2024closed, zhang2025gevrm, hatch2024ghil, chen2025world4omni}, self-improving or interpolation-based video supervision~\cite{oba2023future, soni2024videoagent}, and mining implicit dynamics in diffusion generators to support manipulation~\cite{wen2024vidman}. Collectively, these approaches use generative models as visual surrogates or data engines rather than explicit world models, yet they improve generalization and sample efficiency by providing structured goals, broad visual coverage, and reusable predictive representations.

\textbf{iii) Vision-Grounded Goal Extraction}

Vision-grounded goal extraction refers to deriving task-relevant information directly from visual inputs to support manipulation tasks. Examples include object detection and segmentation, visual tracking, and optical flow, all of which provide structured cues that guide the robot’s actions.

\noindent \textbf{Detection and Segmentation.}
These methods transform raw pixels into goal-bearing symbols, such as sparse object proposals that indicate what or where to act, and dense masks that capture precise geometry. Proposal-centric policies~\cite{zhu2023viola, li2025llara} leverage pre-trained object proposals and transformer reasoning to focus control on task-relevant entities. Segmentation-centric approaches ground actions with language-conditioned masks or incorporate fine-grained semantic cues through diffusion, enabling precise localization of task-critical regions even in cluttered scenes~\cite{dong2025imit, chen2025gondola}. For open-world generalization, open-vocabulary detectors (e.g., Grounding-DINO~\cite{liu2024grounding}) and promptable segmenters (e.g., SAM~\cite{kirillov2023segment}) provide zero- or few-shot category grounding, which can be seamlessly integrated into policy learning pipelines.

\noindent \textbf{Gaze.}
Human gaze, either measured or predicted, serves as a compact attention prior with little supervision that tells policies what to look at and when. 
First, foveated perception crops or reweights features around fixations to enable high precision control and better sample efficiency in manipulation~\cite{kim2021gaze, chuang2025look}. 
Second, gaze acts as a robustness and few-shot prior, suppressing distractors and improving generalization under limited demonstrations~\cite{kim2020using, hamano2022using}. 
When eye trackers are unavailable, video-based gaze prediction can supply similar priors during learning or inference~\cite{kim2022memory}. 
Beyond attention shaping, gaze also structures behavior, revealing subtask boundaries and reusable skill bottlenecks that facilitate modular policies~\cite{takizawa2025gaze, takizawa2025enhancing}. 
Related hand and eye action spaces offer spatial invariance that complements gaze-driven attention~\cite{wang2021generalization}. 
Recent systems further integrate gaze with foveated and vision transformers, scaling these benefits to cluttered scenes with improved data and compute efficiency~\cite{chuang2025look, li2024virt}.

\noindent \textbf{Keypoint.}
Keypoint-based goal extraction encodes task-relevant landmarks in image or 3D space into compact, object-relative representations that policies can consume directly. Compared with raw pixels, keypoints provide geometric structure, interpretability, and improved generalization across viewpoints and scenes. 
Recent work falls into several directions: supervised or semantic keypoints that align with task semantics and affordances, enabling data-efficient manipulation and even bimanual settings~\cite{gao2023k, gao2024bi, wang2025skil}; automatic discovery or task-driven selection that chooses a small set of informative landmarks for robust policy learning~\cite{zhang2025atk}; large model-guided abstraction that infers object-relative frames and stable anchors from language and vision~\cite{fang2025keypoint}; tokenizing actions through keypoint sequences to enable in-context imitation and rapid adaptation~\cite{di2024keypoint}; and hierarchical or open-world frameworks that use keypoints as an interface for composing reusable skills~\cite{li2025hamster}. Knowledge-guided pipelines further stabilize keypoint definitions under distribution shifts by injecting priors about parts, constraints, and contact patterns~\cite{miao2025knowledge}. In practice, keypoints serve as inputs, conditioning variables, or attention anchors for predicting poses, waypoints, and action parameters, offering a clean bridge from perception to control.

\noindent \textbf{Key-Patch.}
CALAMARI~\cite{wi2023calamari} formulates action as contact itself by predicting language-conditioned contact formation maps in the image plane, so the policy decides where on a surface to make contact rather than selecting a single point. The architecture factorizes perception and control at the natural boundary of contact. A multimodal spatial action module aligns per-pixel action predictions with observations, and a low-level controller then optimizes motion to maintain contact while avoiding penetration. By leveraging visual language pretraining for spatial features, CALAMARI improves grounding and sim-to-real transfer and demonstrates strong performance on contact-rich tasks such as sweeping, wiping, and erasing, providing a clean key-patch interface from instruction to control.

\noindent \textbf{Keypose.}
Keypose-based goal extraction selects a small set of SE(3) anchor states that summarize task progress and provide precise geometric targets for control. Compared with pixel features or sparse keypoints, keyposes carry subgoal semantics, mark completion events, and reduce compounding error in multi-stage manipulation. Two representative directions illustrate this interface. KOI accelerates online imitation by deriving hybrid key states from demonstrations, where semantic key states from a vision language pipeline describe what to do and motion key states from optical flow describe how to do it, yielding task-aware rewards that improve exploration efficiency in simulation and real-world settings~\cite{lu2025koi}. BiKC targets bimanual manipulation with a hierarchical policy in which a high-level keypose predictor segments stages and conditions a low-level consistency model that generates trajectories with fast inference and improved success and efficiency~\cite{yu2024bikc}.

\noindent \textbf{Waypoint.}
Waypoint-based goal extraction summarizes manipulation into a short sequence of executable subgoals in task space, typically a handful of SE(3) poses that bridge perception and control. By decoupling “what to achieve” from “how to move,” waypoints reduce covariate shift, ease long-horizon planning, and provide a stable interface to motion controllers and consistency policies. Two representative lines illustrate this interface. AWE learns to predict object-centric pick and place waypoints directly from demonstrations, then delegates time parameterization and smoothing to a low-level controller, yielding strong sample efficiency and robustness in cluttered scenes~\cite{shi2023waypoint}. VIEW further systematizes the perception to waypoint pipeline with object relative frames and consistency regularization so that the predicted subgoals remain geometrically coherent across views and instances, improving generalization while retaining simple downstream control~\cite{jonnavittula2025view}.

\noindent \textbf{Visual Trace.}
Visual trace methods encode human intent as drawable primitives, such as sketches, contours, and reticles, aligned with the scene and used as lightweight goal representations. Compared with keypoints or waypoints, traces provide richer semantic guidance at low annotation cost and can support both training and inference. Sketch-guided methods map freehand drawings to object-relative motion plans for controllable trajectory following~\cite{mehta2025l2d2}, while minimal visual cues use simple reticles or marks to guide attention and spatial priors in clutter without modifying the policy backbone~\cite{dai2026aimbot}.

\noindent \textbf{Visual Track.}
Visual track methods encode identity-consistent trajectories of scene points across time and use them as an action or guidance space that bridges perception and control with low supervision cost. 
Tracks can directly serve as actions for cross-embodiment transfer from human videos, with 2D motion tracks predicted from multiple views and then lifted to 6-DoF robot trajectories, yielding strong few-shot performance in the real world~\cite{ren2025motion}. 
Web videos can supervise track prediction to produce an open-loop plan that is subsequently refined by a residual closed-loop policy, improving generalization to novel objects and scenes~\cite{bharadhwaj2024track2act}. 
Pre-training a trajectory model to predict future paths of arbitrary points further supplies dense correspondence priors that boost visuomotor learning with minimal action labels and enable transfer across morphologies~\cite{wen2024any}. 
Diffusion-generated trajectories can guide policy optimization at the sequence level, reducing compounding error in long-horizon tasks~\cite{fan2025diffusion}. 
Complementary work prescribes a small set of semantically meaningful points and propagates them through data with off-the-shelf vision models, creating stable anchors that improve out-of-distribution generalization and can be tracked to condition policies~\cite{levy2025p3}.

\noindent \textbf{Flow.}
Flow-based goal extraction treats motion fields as the interface between perception and control, specifying how objects should move rather than only where they are. Under this lens, it is useful to separate four flavors.
Optical flow uses dense 2D correspondences from videos without action labels to supervise policies or provide priors for planning and control, exemplified by the action-from-video pipeline~\cite{ko2024learning, zhong2025flowvla}.
Action flow encodes policy- or task-conditioned motion fields in time and space, which can be fused with memory or generated by diffusion to coordinate multi-arm behaviors and improve precision, as in ActionSink~\cite{guo2025actionsink} and VLM-SFD~\cite{chen2025vlm}. 
Object-centric flow defines motion on manipulated objects or parts, enabling cross-embodiment transfer and even reward shaping~\cite{xu2025flow, chen2025ec, yu2025genflowrl}, for example, Im2Flow2Act’s object-flow interface~\cite{xu2025flow} and GenFlowRL’s generative object-centric flow~\cite{yu2025genflowrl}. 
3D flow lifts signals to scene flow, voxels, or point clouds and often augments them with semantics, serving as pose-aware priors or world-model latents; representative systems include G3Flow~\cite{chen2025g3flow}, 3DFlowAction~\cite{zhi20253dflowaction}, and ManiTrend~\cite{he2025manitrend}, while VIP~\cite{li2025vip} uses sparse point flows during pre-training and ToolFlowNet~\cite{seita2023toolflownet} predicts per-point tool flow to ground contact-rich skills.

\textbf{iii) Text-Grounded Goal Extraction}

Recent advances in text-grounded goal extraction highlight the shift from simple language conditioning toward explicitly parsing instructions into actionable goals, recovery strategies, or reasoning steps. RACER introduces rich language-guided recovery policies that allow imitation learning agents to adapt when failures occur, demonstrating how natural language can serve as a corrective signal rather than only a task specifier \cite{dai2025racer}. EmbodiedGPT extends this idea to large-scale pretraining by incorporating embodied chain-of-thought, enabling policies to follow multi-step reasoning trajectories derived from textual instructions \cite{mu2023embodiedgpt}. Along similar lines, CoTPC formulates chain-of-thought predictive control, translating instructions into structured predictive sub-goals that can be optimized in control loops~\cite{jia2024chain}. Complementing these reasoning-oriented approaches, OCI leverages object-centric instruction augmentation to link linguistic references more directly to specific entities in the environment, thereby reducing ambiguity and improving manipulation precision~\cite{wen2024object}. Together, these works demonstrate that grounding language into structured goals—whether through recovery cues, reasoning chains, or object-level references—substantially improves robustness, generalization, and interpretability in robotic manipulation.

\textbf{iv) Contrastive Learning}

The integration of contrastive learning into robotic manipulation can be organized into five categories.
First, universal representation learning employs large-scale contrastive pretraining on internet or egocentric videos to obtain transferable embeddings that accelerate downstream robot learning. R3M demonstrates how time-contrastive objectives yield general-purpose features for control and language grounding~\cite{nair2023r3m}.
Second, language-conditioned alignment applies contrastive objectives to couple visual states, natural language instructions, and robot actions. This line includes BC-Z \cite{jang2022bc} and HULC~\cite{mees2022matters}, which align multimodal representations for zero-shot task generalization and robust imitation over unstructured datasets.
Third, video-to-action alignment maps human demonstration videos to robot policies via cross-modal contrastive learning. Vid2Robot~\cite{jain2024vid2robot} leverages cross-attention transformers with video-conditioned contrastive losses to bridge video-to-robot execution.
Fourth, contrastive imitation learning directly integrates contrastive losses into imitation pipelines, supervising alignment of embeddings with action trajectories or policy heads. $\Sigma$-agent~\cite{ma2025contrastive} exemplifies this by combining contrastive imitation with multi-task, language-guided manipulation.
Finally, action-sequence contrastive supervision contrasts correct and incorrect action trajectories to shape representations tailored for policy optimization, as in CLASS~\cite{lee2026class}.
This five-category taxonomy disentangles overlapping use cases—representation pretraining, multimodal alignment, video-conditioned imitation, contrastive IL, and trajectory-level supervision—providing a clear structure for surveying contrastive methods in manipulation.

\textbf{v) Reconstruction}

Reconstruction has emerged as a powerful self-supervised pretraining strategy for robot manipulation, encompassing both masked modeling and generative reconstruction paradigms. The central idea is to recover or predict missing or novel observations such as masked pixels, tokens, or 3D representations from partial inputs. By compelling the model to infer the underlying structure of the environment and action space, reconstruction-based pretraining promotes the learning of rich and structured representations. These representations can then be transferred to downstream imitation or reinforcement learning tasks, improving sample efficiency, robustness, and cross-task generalization. Existing works can be broadly categorized into two groups: (1) 2D masked reconstruction, which focuses on reconstructing intentionally occluded or masked inputs, and (2) 3D reconstruction, which encompasses both 3D masked reconstruction and generative reconstruction approaches that predict unseen views, 3D feature fields, or future observations.

\noindent \textbf{2D Masked Reconstruction.}
Early methods focused on reconstructing masked pixels or spatiotemporal tokens from egocentric robot data. 
MVP introduced masked visual pretraining for real-world robotic control, showing strong transfer across manipulation tasks~\cite{xiao2022masked}. 
STP extended this to spatiotemporal predictive pretraining for motor control~\cite{yang2024spatiotemporal}, while sensorimotor pretraining methods coupled visual masking with proprioceptive prediction to learn multimodal embeddings~\cite{radosavovic2023robot}. 
MUTEX~\cite{shah2023mutex} and Voltron~\cite{karamcheti2023language} further demonstrated how masked reconstruction can be combined with multimodal task specifications and language grounding.

\noindent \textbf{3D Reconstruction.}
Beyond pixel masking, 3D reconstruction approaches leverage multiview images, point clouds, or implicit feature fields to learn spatially consistent representations, often in SE(3)-equivariant forms~\cite{chen2023polarnet, ze2023visual, ze2023gnfactor, zhang2025leveraging, lu2024manigaussian, zhu2025spa, qian20253d, jia2024lift3d, zhu2025equact, cui2025cl3r, li2025train}. 
3D-MVP performed masked multiview pretraining for robust manipulation policies~\cite{qian20253d}, while SPA introduced spatial-awareness masking to enhance embodied representations~\cite{zhu2025spa}. 
Lift3D demonstrated how to “lift” 2D pretrained models into 3D spaces~\cite{jia2024lift3d}, and CL3R combined 3D reconstruction with contrastive learning for enhanced manipulation features~\cite{cui2025cl3r}. 
Perceiver-actor models such as PerAct~\cite{shridhar2023perceiver}, M2T2~\cite{yuan2023m2t2}, and RVT-2~\cite{goyal2024rvt} integrate masked pretraining into transformer-based architectures for multi-task pick-and-place. 
Recent generative 3D methods extend this trend: GNFactor~\cite{ze2023gnfactor} employs neural feature fields for volumetric reconstruction, while NeRF-style formulations explore corrective augmentation and novel-view synthesis for manipulation~\cite{zhou2023nerf}.

Reconstruction thus serves as a unifying self-supervised scaffold across modalities. 
2D masked methods emphasize scalable pretraining from large-scale egocentric video and multimodal signals, while 3D reconstruction approaches—whether masked or generative—focus on spatial consistency, equivariance, and viewpoint robustness. 
Together, these methods establish reconstruction-based pretraining as a key enabler of sample-efficient and generalizable manipulation policies.

\subsection{Input Modeling}
\label{sec:input_learning}

Input modeling defines how robots perceive and represent the world through various sensory modalities, determining what inputs are used and how they are processed before being fed into control or policy models. It encompasses the selection and encoding of multimodal observations—such as vision, language, touch, force, or audio—and the transformation of these raw signals into structured representations suitable for learning and decision-making. Effective input modeling ensures that sensory data are aligned, fused, and abstracted in a way that preserves essential spatial, temporal, and semantic information, thereby enabling robust perception, reasoning, and control across diverse manipulation tasks.

\subsubsection{Vision-Action Models}

\noindent \textbf{2D Vision as Input.}
The development of Vision Action Models largely relies on 2D visual observations~\cite{jing2023exploring,ma2024hierarchical,wang2024scaling,chi2023diffusion,gong2025carp,li2024crossway,prasad2024consistency,su2025freqpolicy,kim2025uniskill}, often captured by multi-view RGB cameras. Representative architectures include convolutional, Transformer-based, and diffusion-based policies.
Vi-PRoM~\cite{jing2023exploring} employs ResNet-50~\cite{he2016deep} to extract visual features for manipulation, while HDP~\cite{ma2024hierarchical} introduces hierarchical policy learning for efficient visuomotor control. HPT~\cite{wang2024scaling} uses a Transformer backbone to align vision and proprioception across embodiments. Diffusion Policy~\cite{chi2023diffusion} represents a major shift by modeling visuomotor control as conditional denoising, while Consistency Policy~\cite{prasad2024consistency} substantially accelerates diffusion-based action generation through consistency distillation. Beyond policy architectures, recent work highlights the importance of visual sensing itself: Rethinking Camera Choice~\cite{xue2026rethinking} systematically studies fisheye cameras and shows how field of view and visual diversity affect spatial localization and generalization. Despite their effectiveness, 2D visual-action models lack explicit 3D grounding, motivating 3D-aware policies that directly align actions with scene geometry.

\noindent \textbf{3D Vision as Input.}
An increasing number of vision–action models incorporate explicit 3D observations of the physical environment~\cite{goyal2023rvt, goyal2024rvt, wang2024gendp, ze20243d, noh20253d, gervet2023act3d, ke2024diffuser, cao2025mamba, lu2024manicm, gkanatsios20253d, wilcox2025adapt3r}. 
By leveraging multi-view transformations, RVT~\cite{goyal2023rvt} and RVT-2~\cite{goyal2024rvt} explicitly enhance task execution efficiency and generalization, thereby improving performance in real-world applications. GenDP~\cite{wang2024gendp} significantly improves success rates on unseen instances and enables category-level generalization by addressing the generalization limitations of diffusion policies through 3D semantic fields derived from multi-view RGBD observations. DP3~\cite{ze20243d} conditions the diffusion model on 3D point-cloud representations and robot states, enabling efficient execution of complex manipulation tasks with the Allegro Hand. 3D-FDP~\cite{noh20253d} leverages scene-level 3D flow as a structured intermediate
representation to capture fine-grained local motion cues. However, while these models provide 3D-aware capabilities lacking in 2D Vision-Action models, their absence of textual semantic information ultimately limits their generalization and robustness in the real world.

\subsubsection{Vision-Language-Action Models}
\label{subsec: vla}

\newcommand{\vlaorient}[1]{%
  \rotatebox{90}{#1}%
}

\newcommand{\vlacat}[1]{%
  \parbox[c]{30mm}{\raggedright #1}%
}

\newcommand{\vlaitem}[1]{%
  \parbox[c]{105mm}{\raggedright #1}%
}

\begin{figure*}[!t]
\centering
\scriptsize

\forestset{
  vlataxo/.style={
    for tree={
      grow'=0,
      parent anchor=east,
      child anchor=west,
      anchor=west,
      edge path'={
        (!u.parent anchor) -- +(3pt,0) |- (.child anchor)
      },
      edge={draw=darkgray, line width=0.6pt},
      rounded corners,
      draw,
      minimum height=4.8mm,
      inner xsep=2.5pt,
      inner ysep=2.0pt,
      s sep=3pt,
      l sep=6pt,
      font=\scriptsize
    },
    orient/.style={
      draw,
      rounded corners=2pt,
      fill opacity=.5,
      inner xsep=2.5pt,
      inner ysep=2pt
    },
    cat2d/.style={
      draw,
      rounded corners=2pt,
      fill=yellow!18,
      inner xsep=2.5pt,
      inner ysep=2pt
    },
    item2d/.style={
      draw,
      rounded corners=2pt,
      fill=yellow!32,
      inner xsep=2.5pt,
      inner ysep=2pt
    },
    cat3d/.style={
      draw,
      rounded corners=2pt,
      fill=hidden-blue!18,
      inner xsep=2.5pt,
      inner ysep=2pt
    },
    item3d/.style={
      draw,
      rounded corners=2pt,
      fill=hidden-blue!32,
      inner xsep=2.5pt,
      inner ysep=2pt
    }
  }
}

\begin{forest} vlataxo
[{\rotatebox{90}{\textbf{Vision-Language-Action Models (\S~\ref{subsec: vla})}}},
 draw, inner sep=2pt, fill opacity=.5,
  [{\rotatebox{90}{\textbf{2D Vision as Input}}}, orient
    [{\vlaorient{Model-oriented}}, orient
      [{\vlacat{Non-LLM-based VLA}}, cat2d
        [{\vlaitem{
          RT-1~\cite{brohan2023rt},
          VIMA~\cite{jiang2022vima},
          HULC~\cite{mees2022matters},
          HULC++~\cite{mees2023grounding},
          Octo~\cite{ghosh2024octo},
          MDT~\cite{reuss2024multimodal},
          BAKU~\cite{haldar2024baku},
          MResT~\cite{saxena2023mrest},
          LISA~\cite{garg2022lisa},
          Dita~\cite{hou2025dita},
          OTTER~\cite{huang2025otter},
          RoboGround~\cite{huang2025roboground},
          Bi-LAT~\cite{kobayashi2025bi},
          GR-1~\cite{wu2024unleashing},
          TwinVLA~\cite{im2026twinvla}, STEER~\cite{smith2025steer}
        }}, item2d]
      ]
      [{\vlacat{LLM/VLM-based VLA}}, cat2d
        [{\vlaitem{
          RT-2~\cite{zitkovich2023rt},
          RoboFlamingo~\cite{li2024vision},
          OpenVLA~\cite{kim2025openvla},
          OpenVLA-OFT~\cite{kim2025fine},
          $\pi_0$~\cite{black2025pi_0},
          $\pi_{0.5}$~\cite{intelligence2025pi_},
          CogACT~\cite{li2024cogact},
          RoboVLMs~\cite{li2024towards},
          RoboMamba~\cite{liu2024robomamba},
          Magma~\cite{yang2025magma},
          Diffusion-VLA~\cite{wen2025diffusion},
          DexVLA~\cite{wen2025dexvla},
          ChatVLA~\cite{zhou2025chatvla},
          HybridVLA~\cite{liu2026hybridvla},
          InstructVLA~\cite{yang2026vision},
          X-VLA~\cite{zheng2026x},
          UD-VLA~\cite{chen2026unified}, GR-3~\cite{cheang2025gr}, Interleave-VLA~\cite{fan2025interleave}
        }}, item2d]
      ]
      [{\vlacat{Latent Learning for VLA}}, cat2d
        [{\vlaitem{
          QueST~\cite{mete2024quest},
          UniVLA~\cite{bu2025univla},
          VQ-VLA~\cite{wang2025vq},
          villa-X~\cite{chen2026villa},
          MemoryVLA~\cite{shi2026memoryvla},
          ViPRA~\cite{routray2026vipra},
          LARA~\cite{liu2026lara},
          LDA-1B~\cite{lyu2026lda},
          From Pixels to Tokens~\cite{lin2026from},
          LaRA-VLA~\cite{bai2026latent},
          Fast-ThinkAct~\cite{huang2026fast},
          Chain of World~\cite{yang2026chain},
          JALA~\cite{luo2026jointaligned},
          VLA-JEPA~\cite{sun2026vla}
        }}, item2d]
      ]
      [{\vlacat{Hierarchical VLA}}, cat2d
        [{\vlaitem{
          RT-H~\cite{belkhale2024rt},
          PIVOT-R~\cite{zhang2024pivot},
          Hi Robot~\cite{shi2025hi},
          HiBerNAC~\cite{wu2025hibernac},
          H-GAR~\cite{zhu2026h},
          MemER~\cite{sridhar2026scaling}
        }}, item2d]
      ]
      [{\vlacat{Dual-/Multi-system VLA}}, cat2d
        [{\vlaitem{
          LCB~\cite{shentu2024llms},
          DP-VLA~\cite{han2024dual},
          HiRT~\cite{zhang2025hirt},
          RoboDual~\cite{bu2024towards},
          Fast-in-Slow~\cite{chen2025fast},
          Hume~\cite{song2025hume},
          OpenHelix~\cite{cui2025openhelix},
          RationalVLA~\cite{song2025rationalvla},
          TriVLA~\cite{liu2025trivla},
          G0~\cite{jiang2025galaxea},
          Libra-VLA~\cite{wei2026libra}
        }}, item2d]
      ]
    ]
    [{\vlaorient{Model-agnostic}}, orient
      [{\vlacat{Training-time Optimization}}, cat2d
        [{\vlaitem{
          Knowledge Insulation~\cite{driess2025knowledge},
          RETAIN~\cite{yadav2026robust},
          FAN~\cite{niu2026boosting},
          T-MEE~\cite{bai2026reshaping},
          MAPS~\cite{huang2026maps},
          RICL~\cite{sridhar2025ricl},
          CronusVLA~\cite{li2026cronusvla}, \cite{zhang2026grounding}, MoIRA~\cite{kuzmenko2026moira}
        }}, item2d]
      ]
      [{\vlacat{Inference-time Optimization}}, cat2d
        [{\vlaitem{
          VOTE~\cite{lin2025vote},
          RoboMonkey~\cite{kwok2025robomonkey},
          AR-VRM~\cite{yang2025ar},
          Mechanistic Steering~\cite{haon2025mechanistic},
          VLA-Reasoner~\cite{guo2026vlareasoner},
          TapSampling~\cite{zhao2026tapsampling}, CO-RFT~\cite{huang2025co},
          \cite{jang2026verifier},
          \cite{liang2026adaptive}
        }}, item2d]
      ]
      [{\vlacat{Reinforcement Learning and Post-training}}, cat2d
        [{\vlaitem{
          V-GPS~\cite{nakamoto2025steering},
          iRe-VLA~\cite{guo2025improving},
          ConRFT~\cite{chen2025conrft},
          RLDG~\cite{xu2025rldg},
          ThinkAct~\cite{huang2025thinkact},
          VLA-RL~\cite{lu2025vla},
          RIPT-VLA~\cite{tan2025interactive},
          SimpleVLA-RL~\cite{li2025simplevla},
          RL4VLA~\cite{liu2025can},
          ManipLVM-R1~\cite{song2026maniplvm},
          \cite{xiao2026self},
          \cite{zhang2026balancing},
          DyGRO-VLA~\cite{lin2026dygrovla}, VLA-RFT~\cite{li2025vla}
        }}, item2d]
      ]
      [{\vlacat{Learning with Auxiliary Tasks}}, cat2d
        [{\vlaitem{
          ECoT~\cite{zawalski2025robotic},
          CoT-VLA~\cite{zhao2025cot},
          TraceVLA~\cite{zheng2025tracevla},
          Seer~\cite{tian2025predictive},
          UP-VLA~\cite{zhang2025up},
          MOO~\cite{stone2023open},
          DreamVLA~\cite{zhang2025dreamvla},
          ReconVLA~\cite{song2026reconvla},
          Emma-X~\cite{sun2025emma},
          RAD~\cite{clark2025action},
          VLA-OS~\cite{gao2025vla},
          OneTwoVLA~\cite{lin2026onetwovla},
          ACoT-VLA~\cite{zhong2026acot},
          RoboInter~\cite{li2026robointer},
          HiF-VLA~\cite{lin2026hif}, A0~\cite{xu2025a0},
          Action-Sketcher~\cite{tan2026action},
          CrayonRobo~\cite{li2025object2},
          ControlVLA~\cite{li2025controlvla}, LLARVA~\cite{niu2024llarva}
        }}, item2d]
      ]
      [{\vlacat{Efficiency}}, cat2d
        [{\vlaitem{
          TinyVLA~\cite{wen2025tinyvla}, NoRA~\cite{hung2025nora},
          SmolVLA~\cite{shukor2025smolvla},
          Flower~\cite{reuss2025flower},
          VLA-Adapter~\cite{wang2026vla}, \cite{liu2024tail},
          Evo-1~\cite{lin2026evo},
          DeeR-VLA~\cite{yue2024deer}, EfficientVLA~\cite{yang2025efficientvla},
          MoLe-VLA~\cite{zhang2026mole},
          CogVLA~\cite{li2025cogvla},
          SP-VLA~\cite{li2026sp},
          VLA-Cache~\cite{xu2025vla},
          QVLA~\cite{xu2026qvla},
          QuantVLA~\cite{zhang2026quantvla}, \cite{park2025saliency},
          Spec-VLA~\cite{wang2025spec},
          BEAST~\cite{zhou2025beast},
          PD-VLA~\cite{song2025pd},
          Fast-dVLA~\cite{song2026fast},
          BlockVLA~\cite{wang2026blockvla}
          RTC~\cite{black2025real},
          REMAC~\cite{wang2026real},
          TTF-VLA~\cite{liu2026ttf}, CEED-VLA~\cite{song2026ceed}
        }}, item2d]
      ]
      [{\vlacat{Robustness}}, cat2d
        [{\vlaitem{
          SAFE~\cite{gu2025safe},
          FAIL-Detect~\cite{xu2025can},
          RACER~\cite{dai2025racer},
          BYOVLA~\cite{hancock2025run},
          RobustVLA~\cite{guo2026robustness},
          StableVLA~\cite{fu2026stablevla},
          BadVLA~\cite{zhou2025badvla},
          Phantom Menace~\cite{lu2026phantom},
          \cite{wang2025exploring}, \cite{lu2026robots}
        }}, item2d]
      ]
      [{\vlacat{Generalization}}, cat2d
        [{\vlaitem{
          LongVLA~\cite{fan2025long}, LoHoVLA~\cite{yang2025lohovla},
          BehaviorVLA~\cite{hu2026from},
          \cite{jiang2026crosshand},
          \cite{hui2026seeing},
          FOCA~\cite{nguyen2026foca},
          AtomicVLA~\cite{zhang2026atomicvla},
          \cite{zhu2026beyond},
          BagelVLA~\cite{hu2026bagelvla},
          PALM~\cite{liu2026palm},
          Spatial Memory~\cite{li2026spatial}
        }}, item2d]
      ]
    ]
  ]
    [{\rotatebox{90}{\textbf{3D Vision as Input}}}, orient
    [{\vlaorient{Model-oriented}}, orient
      [{\vlacat{3D Embedding and Fusion}}, cat3d
        [{\vlaitem{
          SpatialVLA~\cite{qu2025spatialvla},
          3D-CAVLA~\cite{bhat20253d},
          GeoVLA~\cite{sun2025geovla},
          FP3~\cite{yang2025fp3},
          VoxAct-B~\cite{liu2025voxact},
          VIHE~\cite{wang2024vihe},
          OG-VLA~\cite{singh2025og},
          RoboMM~\cite{yan2025robomm},
          PointACT~\cite{chen2026pointact},
          PointVLA~\cite{li2026pointvla}
        }}, item3d]
      ]
      [{\vlacat{Spatial Alignment and Guidance}}, cat3d
        [{\vlaitem{
          BridgeVLA~\cite{li2025bridgevla},
          FALCON~\cite{zhang2026spatial}
        }}, item3d]
      ]
      [{\vlacat{Multi-view and Active Perception}}, cat3d
        [{\vlaitem{
          Learning to See and Act~\cite{bai2025learning},
          Cortical Policy~\cite{zhang2026cortical}
        }}, item3d]
      ]
      [{\vlacat{3D/4D Reasoning and Prediction}}, cat3d
        [{\vlaitem{
          evo-0~\cite{lin2025evo},
          3D-VLA~\cite{zhen20243d},
          ConsisVLA-4D~\cite{li2026consisvla4d},
          PhysMani~\cite{yun2026physmani},
          GraphCoT-VLA~\cite{huang2026graphcot}, ChainedDiffuser~\cite{xian2023chaineddiffuser},
        }}, item3d]
      ]
    ]
    [{\vlaorient{Model-agnostic}}, orient
      [{\vlacat{Training-time Optimization}}, cat3d
        [{\vlaitem{
          Spatial Forcing~\cite{li2026spatialforcing},
          GeoPredict~\cite{qian2026geopredict}
        }}, item3d]
      ]
      [{\vlacat{Inference-time Optimization}}, cat3d
        [{\vlaitem{
          Affordance Field Intervention~\cite{xu2026affordance},
          DepthCache~\cite{li2026depthcache}
        }}, item3d]
      ]
    ]
  ]
]
\end{forest}

\caption{A structured taxonomy of VLA models organized by input modality (2D vs.~3D) and methodological orientation (model-oriented architectures vs.~model-agnostic strategies).}
\label{fig: vla-taxonomy}
\end{figure*}

Recent progress in Vision--Language--Action (VLA) models has created a unified paradigm for mapping multimodal perception into executable robotic behaviors. 
Compared with earlier vision-only or language-conditioned controllers, VLAs integrate semantic grounding, spatial reasoning, and sequential action generation within a single architecture. 
To provide a systematic overview, we summarize the landscape in a taxonomy (Figure~\ref{fig: vla-taxonomy}) that organizes existing work by input modality (2D and 3D) and by methodological orientation, distinguishing \textbf{model-oriented} approaches (architectural innovations) from \textbf{model-agnostic} ones (inference, training, or efficiency enhancements). 
This taxonomy clarifies technical differences, shared design trade-offs, and the evolution of VLAs toward scalable, robust, and general-purpose robotic intelligence.

\textbf{i) 2D Vision as Input}

Most VLAs rely on 2D images from RGB cameras, sometimes with multi-view setups, as the primary perceptual stream.
Within this setting, a diverse set of methods has emerged, which can be broadly divided into \textbf{model-oriented} approaches that redesign the policy architecture itself, and \textbf{model-agnostic strategies} that improve inference or training without altering the core model.

\noindent \textbf{Model-oriented Approaches.}
Model-oriented approaches focus on advancing the architecture and internal structure of VLA models to improve their representation learning, reasoning ability, and task adaptability. Instead of modifying training schemes or inference strategies, these methods emphasize redesigning the policy backbone, integrating multimodal modules, and structuring control hierarchies to better align perception, reasoning, and action.

\textbullet\hspace{0.2em} \textbf{Non-LLM-based VLA.}
Early approaches directly mapped visual observations and language instructions to robot actions without relying on large language-model backbones. RT-1~\cite{brohan2023rt} pioneered scalable Transformer-based action prediction from large robot datasets, while VIMA~\cite{jiang2022vima} and HULC~\cite{mees2022matters} extended language-conditioned control toward multimodal prompts and unstructured demonstrations. More recent methods improve scalability and action modeling. Octo~\cite{ghosh2024octo} trains a generalist Transformer policy on large-scale cross-embodiment robot data and supports efficient adaptation to new robots and action spaces, while MDT~\cite{reuss2024multimodal} combines multimodal goal conditioning with diffusion-based action generation under sparse language annotations. Dita~\cite{hou2025dita} further scales Diffusion Transformers to directly model continuous action sequences across heterogeneous embodiments. Collectively, these methods demonstrate that scalable multimodal policy learning can achieve broad manipulation capabilities without large language backbones, while motivating stronger semantic priors for open-world generalization.

\textbullet\hspace{0.2em} \textbf{LLM/VLM-based VLA.}
With the rise of LLMs and VLMs, robotic policies increasingly leverage web-scale semantic priors for action generation. RT-2~\cite{zitkovich2023rt} co-trains vision-language and robot data with tokenized actions, enabling improved generalization to novel objects and instructions and emergent semantic reasoning. RoboFlamingo~\cite{li2024vision} adapts pretrained VLMs to manipulation through lightweight fine-tuning and an explicit policy head, while OpenVLA~\cite{kim2025openvla} scales open-source VLA training across diverse robot demonstrations. $\pi_0$~\cite{black2025pi_0} and $\pi_{0.5}$~\cite{intelligence2025pi_} further introduce flow-matching action generation and large-scale cross-embodiment training.
Recent work increasingly focuses on coupling semantic reasoning with expressive action models. DiffusionVLA~\cite{wen2025diffusion} combines autoregressive VLM reasoning with diffusion-based action generation, while DexVLA~\cite{wen2025dexvla} introduces a large diffusion action expert for cross-embodiment and dexterous control. Together, these developments shift VLA research from simply transferring semantic priors toward tighter integration of reasoning, generalization, and continuous motor generation.

\textbullet\hspace{0.2em} \textbf{Latent Learning for VLA.}
Latent learning introduces compact intermediate representations between multimodal perception and action generation, supporting action abstraction, internal reasoning, and predictive modeling. Early approaches primarily focus on action representation: QueST~\cite{mete2024quest} learns quantized temporal skill abstractions for continuous control, while VQ-VLA~\cite{wang2025vq} scales vector-quantized action tokenization over large trajectory datasets. UniVLA~\cite{bu2025univla} further learns task-centric latent actions from videos, enabling pre-training across embodiments and viewpoints. Recent work extends latent learning beyond action compression toward internal reasoning. Latent Reasoning VLA~\cite{bai2026latent} internalizes multimodal reasoning and future prediction into continuous latent states, while Fast-ThinkAct~\cite{huang2026fast} distills explicit reasoning into compact latent plans for efficient action generation. Chain of World~\cite{yang2026chain} further combines latent motion representations with world-model prediction, allowing temporal dynamics to guide downstream actions. Overall, latent learning is evolving from compact action representations toward unified latent spaces for action abstraction, reasoning, and future prediction, with the key challenge of preserving task relevance, physical grounding, and alignment with executable actions.

\textbullet\hspace{0.2em} \textbf{Hierarchical VLA.}  
Long-horizon and compositional tasks motivate hierarchical designs that decompose control into intermediate instructions, primitives, waypoints, or skills. Hi Robot~\cite{shi2025hi} separates high-level VLM reasoning from low-level VLA execution, enabling open-ended instruction following and interactive task refinement. RT-H~\cite{belkhale2024rt} introduces hierarchical action abstractions, while PIVOT-R~\cite{zhang2024pivot} combines primitive parsing, waypoint prediction, and low-level action decoding through an asynchronous hierarchical executor. Such structures improve interpretability and long-horizon execution, but introduce challenges in learning reliable intermediate abstractions and preventing high-level errors from propagating downstream.

\textbullet\hspace{0.2em} \textbf{Dual- and Multi-System VLA.} 
Inspired by dual-process cognition, these architectures decouple slow semantic reasoning from fast visuomotor execution. LCB~\cite{shentu2024llms} introduces a learnable latent action code to communicate high-level LLM intent to low-level policies beyond explicit language interfaces. DP-VLA~\cite{han2024dual} and HiRT~\cite{zhang2025hirt} further decouple execution frequencies, using slowly updated VLM representations to guide lightweight policies that react to real-time observations. OpenHelix~\cite{cui2025openhelix} systematically studies this interface and shows the importance of policy pre-training, projector pre-alignment, and multimodal supervision, while revealing that high-level latent tokens may otherwise encode instruction semantics more strongly than dynamic visual changes. Fast-in-Slow~\cite{chen2025fast} instead embeds the fast action module within the slow VLM through partial parameter sharing and joint co-training. Libra-VLA~\cite{wei2026libra} explicitly communicates discrete macro-intents to a continuous action refiner through asynchronous coarse-to-fine execution. Beyond dual systems, TriVLA~\cite{liu2025trivla} adds a video-based dynamics module to couple semantic perception, future prediction, and real-time control. These developments shift structured VLAs from simple frequency separation toward tighter reasoning--action communication and predictive multi-system coordination.

\noindent \textbf{Model-agnostic Strategies.}
Model-agnostic strategies aim to enhance the performance, reliability, and efficiency of VLA models without modifying their underlying architecture. Rather than redesigning model structures, these approaches operate at the inference, training, or auxiliary supervision level to improve decision quality, generalization, and deployment practicality.

\textbullet\hspace{0.2em} \textbf{Training-Time Optimization.}
Training-time strategies improve VLA adaptation through optimization objectives or parameter-update schemes without redesigning the policy architecture. Knowledge Insulation~\cite{driess2025knowledge} mitigates interference between newly initialized action modules and pretrained VLM representations through gradient insulation and multimodal co-training. RETAIN~\cite{yadav2026robust} merges pretrained and task-finetuned parameters to preserve generalist capabilities during adaptation, while FAN-guided finetuning~\cite{niu2026boosting} exploits feasible action neighborhoods to improve sample efficiency and generalization. T-MEE~\cite{bai2026reshaping} instead reshapes trajectory-level action error distributions through minimum error entropy, improving robustness to few-shot, noisy, and imbalanced training data. Together, these methods improve VLA training through knowledge preservation, structured action supervision, and robust optimization.

\textbullet\hspace{0.2em} \textbf{Inference-Time Optimization.}
Inference-time methods improve VLA execution without modifying the underlying policy architecture. VOTE~\cite{lin2025vote} and RoboMonkey~\cite{kwok2025robomonkey} generate multiple action candidates and improve selection through ensemble voting or learned verification. More recent methods explore stronger test-time scaling mechanisms. TapSampling~\cite{zhao2026tapsampling} samples actions in a learned latent space and selects them using a task-progress verifier, while verifier-free test-time sampling~\cite{jang2026verifier} exploits the model's internal distributional confidence to rank candidates without an external verifier. Adaptive Action Chunking~\cite{liang2026adaptive} instead dynamically adjusts the execution horizon according to action uncertainty, balancing temporal consistency and responsiveness. Together, these approaches shift VLA inference from single-shot prediction toward adaptive sampling, verification, and execution.

\textbullet\hspace{0.2em} \textbf{Reinforcement Learning and Post-training.}
RL-based post-training improves pretrained VLAs through reward-driven interaction, enabling policies to move beyond imitation and adapt to deployment distributions. SimpleVLA-RL~\cite{li2025simplevla} demonstrates scalable online RL with sparse outcome rewards, while ConRFT~\cite{chen2025conrft} combines offline and online reinforced fine-tuning for sample-efficient real-world adaptation. iRe-VLA~\cite{guo2025improving} alternates RL and supervised learning to stabilize online VLA optimization, whereas adaptive offline RL~\cite{zhang2026balancing} balances advantage signals and gradient variance for flow-based policies. Recent studies further examine broader roles of RL: empirical analysis~\cite{liu2025can} shows that RL primarily improves semantic generalization and execution robustness, while residual-RL-based self-improvement~\cite{xiao2026self} uses RL to discover failure-recovery behaviors and distills the resulting trajectories back into a generalist policy. Together, these developments shift VLA post-training from simple task-specific fine-tuning toward scalable online adaptation, robust offline optimization, and self-improving data generation.

\textbullet\hspace{0.2em} \textbf{Learning with Auxiliary Tasks.}
Auxiliary objectives enrich VLA learning with intermediate reasoning, structured representations, and predictive supervision beyond action imitation. ECoT~\cite{zawalski2025robotic} and CoT-VLA~\cite{zhao2025cot} introduce embodied and visual chain-of-thought supervision, while ACoT-VLA grounds intermediate reasoning in action generation. Visual intermediate representations are explored by TraceVLA~\cite{zheng2025tracevla}, whereas RoboInter learns broader robotic intermediate representations to improve perception--action grounding. Predictive objectives provide temporal supervision: Seer~\cite{tian2025predictive} learns future-aware inverse dynamics, and HiF-VLA incorporates hindsight and foresight through motion representations. Reconstruction- and world-aware objectives such as DreamVLA~\cite{zhang2025dreamvla} and ReconVLA~\cite{song2026reconvla} further improve representation learning. Overall, auxiliary tasks are evolving from explicit reasoning traces toward multimodal intermediate representations and predictive supervision that better connect perception, reasoning, and action.

\textbullet\hspace{0.2em} \textbf{Efficiency.}  
Recent work improves VLA efficiency across model size, computation, representation, and execution. Compact policies such as TinyVLA~\cite{wen2025tinyvla}, FLOWER~\cite{reuss2025flower}, and Evo-1~\cite{lin2026evo} reduce model complexity while retaining competitive manipulation performance. At the computation level, DeeR-VLA~\cite{yue2024deer} dynamically allocates inference depth, EfficientVLA~\cite{yang2025efficientvla} performs training-free acceleration and compression, and VLA-Cache~\cite{xu2025vla} exploits temporal redundancy through adaptive token caching. Quantization methods such as QVLA~\cite{xu2026qvla} and QuantVLA~\cite{zhang2026quantvla} further reduce memory and computation through VLA-specific low-precision inference. Action-side compression provides another route: BEAST~\cite{zhou2025beast} encodes trajectories with compact B-spline representations to reduce autoregressive decoding steps. Finally, RTC~\cite{black2025real} enables asynchronous execution of action-chunking flow policies to mitigate inference-induced control latency. Together, these methods extend VLA efficiency from model compression toward joint optimization of computation, memory, action representation, and real-time execution.

\textbullet\hspace{0.2em} \textbf{Robustness.}  
Robustness has become a central requirement for reliable VLA deployment under distribution shifts, execution failures, and adversarial disturbances. RobustVLA~\cite{guo2026robustness} systematically studies perturbations across actions, observations, instructions, and environments, and jointly improves input and output robustness through consistency and robust optimization. For visual disturbances, BYOVLA~\cite{hancock2025run} performs test-time observation intervention, while StableVLA~\cite{fu2026stablevla} suppresses unseen visual corruptions through information-bottleneck-based feature filtering. Failure awareness provides another line of defense: SAFE~\cite{gu2025safe} learns task-general failure signals from VLA representations, whereas FAIL-Detect~\cite{xu2025can} performs uncertainty-aware sequential OOD detection without requiring failure demonstrations; RACER~\cite{dai2025racer} further enables language-guided failure recovery. Security-oriented studies expose complementary vulnerabilities, with BadVLA~\cite{zhou2025badvla} revealing persistent backdoor attacks and Phantom Menace~\cite{lu2026phantom} analyzing and mitigating physical sensor attacks. Together, these works broaden VLA robustness from visual resilience toward multimodal perturbation tolerance, failure awareness and recovery, and adversarial security.

\textbullet\hspace{0.2em} \textbf{Generalization.}
Generalization remains a central challenge for VLAs across task horizon, embodiment, environment, and data regimes. Long-VLA~\cite{fan2025long} improves long-horizon execution through phase-aware modeling, while BehaviorVLA~\cite{hu2026from} learns temporally coherent behavior representations robust to distribution shifts. Cross-embodiment approaches such as XL-VLA~\cite{jiang2026crosshand} introduce shared latent action spaces across heterogeneous robot hands. Beyond embodiment transfer, sim-to-real methods~\cite{hui2026seeing} increase environmental diversity by translating simulated trajectories into realistic training videos, whereas FOCA~\cite{nguyen2026foca} targets few-shot adaptation through future-oriented conditioning. AtomicVLA~\cite{zhang2026atomicvla} supports continual skill acquisition through extensible atomic skill experts, while long-tail learning~\cite{zhu2026beyond} addresses data-scarce tasks. Together, these studies broaden VLA generalization from unseen objects and scenes toward new embodiments, longer task horizons, limited-data adaptation, and expanding skill distributions.

Together, the 2D VLA landscape illustrates a clear trajectory: starting from direct end-to-end policies, evolving into architectures that leverage semantic priors from LLMs/VLMs, and further into modular, hierarchical, or multi-system designs. Complemented by model-agnostic techniques for optimization, reinforcement, and robustness, these developments collectively point toward a new generation of VLAs that balance efficiency, interpretability, and scalability.

\textbf{ii) 3D Vision as Input}

Compared with 2D inputs, 3D representations provide richer spatial grounding for contact-rich manipulation and long-horizon planning. However, because most VLM backbones are pre-trained on 2D image–text data, they lack intrinsic 3D understanding. This has motivated research into how VLAs can incorporate explicit 3D perception, which can be broadly divided into \textbf{Model-oriented} approaches that redesign architectures to integrate 3D information, and \textbf{Model-agnostic} strategies that introduce auxiliary mechanisms without altering the backbone.

\noindent \textbf{Model-oriented Approaches.}
Model-oriented approaches explicitly augment the representation or policy architecture with geometry-aware components.
Rather than relying solely on planar image features, these methods introduce point-cloud, depth, voxel, multi-view, or geometry-foundation representations to improve spatial grounding and action prediction.

\textbullet\hspace{0.2em} \textbf{3D Embedding and Fusion.}
A primary direction is to integrate explicit 3D observations with pretrained 2D vision--language representations.
SpatialVLA~\cite{qu2025spatialvla} introduces geometry-aware position encodings and adaptive action grids to inject spatial structure into VLA prediction, while VoxAct-B~\cite{liu2025voxact} and VIHE~\cite{wang2024vihe} exploit voxelized or virtual-view representations for fine-grained 3D manipulation.
Other methods explore complementary geometric representations: 3D-CAVLA~\cite{bhat20253d} incorporates depth and 3D context, GeoVLA~\cite{sun2025geovla} introduces point-based geometric features, OG-VLA~\cite{singh2025og} converts 3D observations into orthographic views, and FP3~\cite{yang2025fp3} investigates large-scale 3D policy learning.
More recently, PointACT~\cite{chen2026pointact} directly couples hierarchical point-cloud representations with action decoding through multi-scale point--action interaction, allowing action tokens to exploit both local geometric details and global scene structure.
Together, these approaches demonstrate different ways of coupling explicit 3D geometry with pretrained semantic representations for spatially grounded control.

\textbullet\hspace{0.2em} \textbf{Spatial Alignment and Guidance.}
A complementary direction focuses on aligning geometric and semantic representations rather than directly replacing image features with 3D inputs.
BridgeVLA~\cite{li2025bridgevla} projects 3D observations into multiple 2D views and formulates action prediction as spatial heatmaps, thereby aligning both perception and action prediction with representations readily processed by pretrained vision models.
FALCON~\cite{zhang2026spatial} instead extracts spatial priors from spatial foundation models and injects the resulting spatial tokens into an enhanced action pathway, preserving the pretrained vision--language representation while strengthening geometric reasoning.
These approaches reflect a shift from directly encoding geometric observations toward explicitly bridging pretrained semantic features and transferable spatial priors.

\textbullet\hspace{0.2em} \textbf{Multi-view and Active Perception.}
Multiple viewpoints provide another mechanism for reducing occlusion and depth ambiguity.
Learning to See and Act~\cite{bai2025learning} reconstructs scenes and selects task-relevant virtual viewpoints, allowing the policy to acquire informative observations rather than relying on fixed cameras.
Cortical Policy~\cite{zhang2026cortical} models complementary static and dynamic visual streams and exploits cross-view geometric correspondence to strengthen spatial reasoning while preserving motion-sensitive information for control.
Such methods extend 3D perception from passive multi-view fusion toward task-conditioned viewpoint selection and cross-view reasoning, particularly useful under occlusion, viewpoint variation, and dynamic interaction.

\textbullet\hspace{0.2em} \textbf{3D/4D Reasoning and Prediction.}
Beyond static geometric representations, another line models spatial structures and their evolution over time.
3D-VLA~\cite{zhen20243d} is an early representative that combines language-conditioned 3D generation with action prediction, casting VLA learning as a generative 3D world-modeling problem.
ConsisVLA-4D~\cite{li2026consisvla4d} extends spatial modeling toward spatiotemporal consistency by reasoning across views, objects, and temporally evolving scene states.
For dynamic-object manipulation, PhysMani~\cite{yun2026physmani} couples a physics-principled 3D Gaussian world model with a future-aware policy to predict physically grounded scene dynamics before action generation.
GraphCoT-VLA~\cite{huang2026graphcot} complements predictive world modeling with structured spatial reasoning, representing 3D relations as graphs and connecting intermediate reasoning states to downstream action generation.
Together, these methods extend 3D VLAs from static geometric perception toward structured reasoning and temporally coherent prediction of geometry, motion, and interaction outcomes.

\noindent \textbf{Model-agnostic Strategies.}
Model-agnostic strategies preserve the underlying VLA architecture to a greater extent and instead use 3D information as auxiliary supervision or external guidance.
Compared with model-oriented 3D VLAs, this direction remains less extensively explored, but recent work has begun to reveal how geometric information can improve existing policies without making explicit 3D processing an indispensable component of the deployed backbone.

\textbullet\hspace{0.2em} \textbf{Training-time Optimization.}
Training-time strategies transfer geometric knowledge into a VLA primarily through additional supervision.
Spatial Forcing~\cite{li2026spatialforcing} aligns intermediate VLA visual representations with geometric features produced by pretrained 3D foundation models, encouraging spatially informative representations without requiring explicit depth or point-cloud inputs during inference.
GeoPredict~\cite{qian2026geopredict} instead introduces predictive supervision over multi-step 3D robot keypoints and future 3D Gaussian geometry; these geometric predictors are used during training, whereas deployment avoids explicit 3D decoding.
The two approaches therefore inject 3D information at different levels---representation alignment and predictive supervision, respectively---while retaining relatively lightweight inference.

\textbullet\hspace{0.2em} \textbf{Inference-time Optimization.} A smaller but emerging line exploits geometric information at inference time without retraining the underlying VLA policy. Affordance Field Intervention (AFI)~\cite{xu2026affordance} introduces an external 3D spatial affordance field as an on-demand plug-in to detect unfavorable behaviors, guide rollback and waypoint search, and score candidate trajectories during execution. DepthCache~\cite{li2026depthcache} instead uses depth as a structural prior for training-free visual-token merging, preserving geometrically important near-field regions while compressing redundant background tokens. These studies suggest that 3D information can serve not only as a perception modality but also as an external signal for runtime correction and inference efficiency. Nevertheless, compared with the mature inference-time optimization literature for 2D VLAs, systematic studies of sampling, verification, caching, and lightweight spatial guidance remain relatively limited in 3D settings.

Across both 2D and 3D modalities, VLA research shows converging trends. Model-oriented approaches aim to strengthen representational capacity by incorporating LLM and VLM priors, latent structures, and hierarchical reasoning in 2D, and by leveraging embeddings, alignment mechanisms, and world models in 3D. Complementing these, model-agnostic strategies emphasize improving inference robustness and efficiency with minimal retraining. Looking ahead, key directions include the standardization of 3D representations in VLAs, the development of hybrid frameworks that integrate reactive control with deliberative planning through safety-aware dual systems, and co-training across 2D and 3D modalities to couple large-scale priors with spatial grounding. Together, these advances move beyond scaling backbone models toward architectures that explicitly integrate perception, memory, planning, and control, evaluated under long-horizon, cross-embodiment, and real-world deployment challenges.

\subsubsection{Tactile-based Action Models}
\label{subsubsec: tactile}
Tactile sensing provides direct information about contact, force, slip, compliance, and other physical properties that are difficult to infer reliably from vision alone. Such feedback is particularly important for contact-rich and partially observable manipulation, where small geometric or force errors can lead to task failure. Recent tactile-based action models have therefore evolved from task-specific tactile feedback integration toward pretrained tactile representations, visuo-tactile policy learning, language-grounded touch understanding, and tactile-augmented generalist policies.

\noindent \textbf{Tactile Representation and Latent Learning.}
Representation learning provides reusable tactile features that can be transferred across tasks and policies. T-DEX~\cite{guzey2023dexterity} learns self-supervised tactile representations from robotic play to support dexterous manipulation, while Sparsh~\cite{higuera2025sparsh} scales self-supervised pretraining to learn generalizable representations from vision-based tactile sensing. CLTP~\cite{ma2026cltp} introduces contrastive language--tactile pretraining to align tactile geometry with natural language for 3D contact understanding. Moving toward interaction-aware representations, exUMI~\cite{xu2025exumi} proposes action-aware Tactile Predictive Pretraining that learns contact dynamics by predicting future tactile observations conditioned on robot actions and visual context. Tactile Beyond Pixels~\cite{higuera2025tactile} extends tactile representation learning beyond image-based signals by jointly modeling sensory cues such as force, motion, and sound. Together, these works show a progression from generic self-supervised tactile encoding toward action-conditioned and multisensory representations that directly capture physical interaction dynamics.

\tikzset{
    tactile-box/.style={
        rectangle,
        draw=black,
        rounded corners,
        text opacity=1,
        minimum height=1.5em,
        minimum width=5em,
        inner sep=2pt,
        align=center,
        fill opacity=.5,
    },
    tactile-leaf/.style={
        tactile-box,
        fill=yellow!32,
        text=black,
        font=\normalsize,
        inner xsep=5pt,
        inner ysep=4pt,
        align=left,
        text width=32.5em,
    }
}

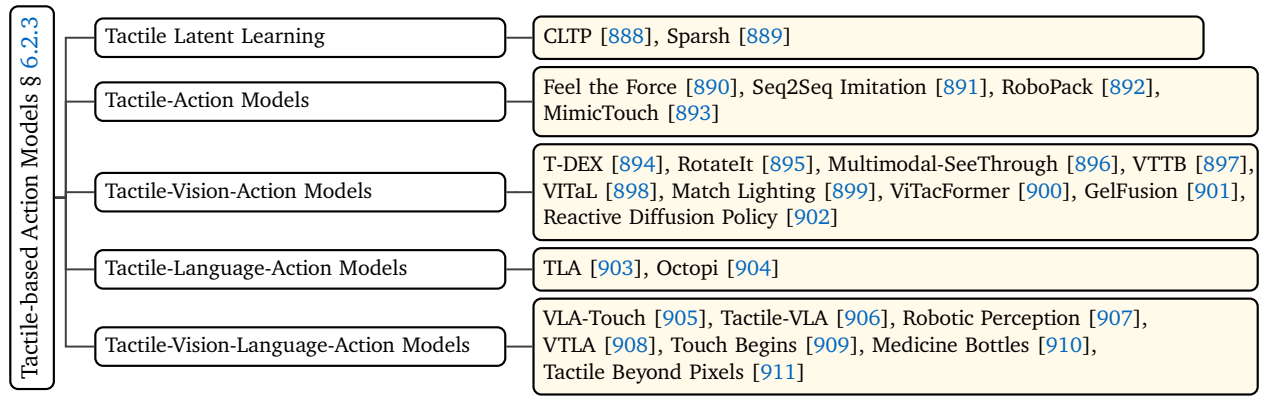
\begin{figure}[t]
\centering
\resizebox{\textwidth}{!}{
\begin{forest}
forked edges,
for tree={
    grow=east,
    reversed=true,
    anchor=base west,
    parent anchor=east,
    child anchor=west,
    base=left,
    font=\large,
    rectangle,
    draw=black,
    rounded corners,
    align=center,
    minimum width=4em,
    edge+={darkgray, line width=1pt},
    s sep=3pt,
    inner xsep=2pt,
    inner ysep=4pt,
    line width=1.1pt,
    ver/.style={
        rotate=90,
        child anchor=north,
        parent anchor=south,
        anchor=center,
        inner xsep=5pt
    },
},
where level=1{
    text width=18em,
    font=\normalsize,
    align=center,
    inner xsep=5pt,
    inner ysep=3pt
}{},
where level=2{
    text width=7em,
    font=\normalsize,
    align=left,
    inner xsep=3pt
}{},
where level=3{
    text width=15em,
    font=\normalsize,
    align=left,
    inner xsep=5pt
}{},
where level=4{
    font=\normalsize,
    align=left
}{},
[{Tactile-based Action Models \S~\ref{subsubsec: tactile}}, ver
    [{Tactile Representation and\\ Latent Learning}
        [{T\mbox{-}DEX~\cite{guzey2023dexterity}, Sparsh~\cite{higuera2025sparsh}, CLTP~\cite{ma2026cltp}, exUMI~\cite{xu2025exumi},\\ Tactile Beyond Pixels~\cite{higuera2025tactile}}, tactile-leaf]
    ]
    [{Tactile-Action Models}
        [{Seq2Seq Imitation~\cite{yang2023seq2seq}, RoboPack~\cite{ai2024robopack}, MimicTouch~\cite{yu2025mimictouch}, \\ Feel the Force~\cite{adeniji2025feel}, SimShear~\cite{freud2025simshear}}, tactile-leaf]
    ]
    [{Tactile-Vision-Action Models}
        [{RotateIt~\cite{qi2023general}, \cite{ablett2024multimodal}, VTTB~\cite{gu2024vttb}, VITaL~\cite{george2025vital}, VT-Refine~\cite{huang2025vt}, \\ Reactive Diffusion Policy~\cite{xue2025reactive},  ViTaS~\cite{tian2026vitas}, ViTacGen~\cite{wu2025vitacgen}, \\ ViTacFormer~\cite{heng2025vitacformer}, GelFusion~\cite{jiang2025gelfusion}}, tactile-leaf]
    ]
    [{Tactile-Language Grounding}
        [{Octopi~\cite{yu2024octopi}, TLA~\cite{hao2025tla}}, tactile-leaf]
    ]
    [{Tactile-Vision-Language-Action Models}
        [{FuSe~\cite{jones2025beyond}, VLA\mbox{-}Touch~\cite{bi2026vla}, Tactile\mbox{-}VLA~\cite{huang2025tactile}, OmniVTLA~\cite{cheng2026omnivtla}, \\ VTLA~\cite{zhang2025vtla}, T-Rex~\cite{niu2026t}, FTP-1~\cite{yuan2026ftp}}, tactile-leaf]
    ]
]
\end{forest}
}
\caption{A structured overview of tactile-based action models.}
\label{fig:tactile}
\end{figure}

\noindent \textbf{Tactile-Action Models.}
Tactile-action models directly condition manipulation policies or control models on tactile feedback without necessarily relying on vision--language reasoning. Seq2Seq Imitation~\cite{yang2023seq2seq} formulates tactile feedback as a sequential input for imitation learning under partial observability. RoboPack~\cite{ai2024robopack} instead incorporates tactile observations into learned dynamics models for model-predictive control, enabling precise dense-packing behaviors. MimicTouch~\cite{yu2025mimictouch} transfers multimodal human tactile demonstrations to contact-rich robotic manipulation, while Feel the Force~\cite{adeniji2025feel} captures human contact forces with tactile gloves to train transferable force-aware manipulation policies. SimShear~\cite{freud2025simshear} further addresses sim-to-real tactile control by synthesizing shear-aware tactile observations and transferring tactile servoing policies to real contact-sensitive tasks. These methods illustrate how tactile feedback can support not only reactive control but also dynamics prediction, human-to-robot transfer, and sim-to-real policy learning.

\noindent \textbf{Tactile-Vision-Action Models.}
Combining vision and touch enables policies to exploit complementary global and local information, with vision providing scene-level context and tactile sensing resolving fine-grained contact states. RotateIt~\cite{qi2023general} combines visual and tactile observations to generalize in-hand object rotation, while Multimodal-SeeThrough~\cite{ablett2024multimodal} uses a transparent visuotactile sensor together with force-matched demonstrations for imitation learning. VTTB~\cite{gu2024vttb} integrates visual and tactile feedback for adaptive robot-assisted bed bathing, and VITaL~\cite{george2025vital} introduces visuo-tactile pretraining to improve both tactile-based and vision-only manipulation policies. Reactive Diffusion Policy~\cite{xue2025reactive} adopts a slow--fast visual-tactile diffusion architecture to improve responsiveness during contact-rich manipulation. More recent work explores stronger multimodal training and transfer: VT-Refine~\cite{huang2025vt} combines real visuo-tactile demonstrations with tactile simulation and reinforcement-learning fine-tuning for precise bimanual assembly; ViTaS~\cite{tian2026vitas} uses soft-fusion contrastive learning to model both the alignment and complementarity between vision and touch; and ViTacGen~\cite{wu2025vitacgen} generates standardized tactile representations from visual observations, allowing tactile-informed policies to operate even when high-resolution tactile sensors are unavailable during deployment. Overall, visuo-tactile policies are shifting from direct modality concatenation toward structured fusion, predictive tactile generation, and simulation-assisted policy improvement.

\noindent \textbf{Tactile-Language Grounding.}
Language provides a shared semantic space for interpreting tactile observations and connecting physical properties with task intent. Octopi~\cite{yu2024octopi} develops a large tactile-language model for reasoning about object properties such as hardness and material characteristics from touch. TLA~\cite{hao2025tla} extends tactile-language alignment toward action generation by learning mappings among tactile sequences, language instructions, and manipulation actions. These approaches establish tactile-language grounding as an interface between low-level contact signals and higher-level semantic reasoning.

\noindent \textbf{Tactile-Vision-Action Models.}
Combining vision and touch enables policies to exploit complementary global scene context and local contact information. RotateIt~\cite{qi2023general} combines visual and tactile observations for generalized in-hand rotation, while Multimodal-SeeThrough~\cite{ablett2024multimodal} uses visuotactile sensing with force-matched demonstrations for imitation learning. Funk et al.~\cite{funk2025match} further demonstrate the importance of tactile feedback for fast contact-rich manipulation through robotic match lighting. VTTB~\cite{gu2024vttb} integrates vision and touch for adaptive bed bathing, whereas VITaL~\cite{george2025vital} introduces visuo-tactile pretraining for manipulation. Touch Begins Where Vision Ends~\cite{zhao2025touch} combines VLM-based object localization with a reusable local visuo-tactile policy to generalize contact-rich skills across scenes. Reactive Diffusion Policy~\cite{xue2025reactive} adopts a slow--fast visual-tactile diffusion architecture for responsive control. More recent methods improve multimodal fusion and transfer: VT-Refine~\cite{huang2025vt} combines real demonstrations, tactile simulation, and RL fine-tuning; ViTaS~\cite{tian2026vitas} models vision--touch alignment and complementarity through contrastive learning; and ViTacGen~\cite{wu2025vitacgen} generates standardized tactile representations from vision. Overall, visuo-tactile policies are progressing from direct sensory fusion toward structured multimodal learning, reusable local control, and simulation-assisted transfer.

\subsubsection{Extra Modalities as Input}
\label{subsubsec: extra_modalities}

Beyond vision, language, and tactile sensing, additional physical modalities such as force, audio, infrared, and radar provide complementary information that is difficult to infer reliably from images alone. These signals are particularly useful for contact-rich manipulation, partially observable interactions, and environments in which visual observations are ambiguous or degraded. Existing approaches have progressed from incorporating individual sensing channels into task-specific control policies toward integrating heterogeneous physical feedback into generalist VLA models.

\noindent \textbf{Force.}
Force and torque provide direct measurements of physical interaction and have long been used to improve manipulation in contact-sensitive tasks. Early work focused on imitation learning from demonstrations that capture motion and force. AR-Haptic~\cite{kormushev2011imitation} learns positional and force profiles from kinesthetic and haptic demonstrations, while Bilateral~\cite{adachi2018imitation} employs bilateral control to record position and force simultaneously for precise manipulation imitation. Feeling the Force~\cite{edmonds2017feeling} jointly models pose and contact force from human demonstrations to discover latent interaction stages in multi-step bottle-opening tasks, highlighting the value of force for revealing physical state changes that may be visually ambiguous. Wang et al.~\cite{wang2021robotic} further combine trajectory and force demonstrations for contact-sensitive assembly, and ImmersiveDemo~\cite{li2023immersive} shows that immersive demonstrations with force feedback can provide informative supervision for imitation learning.

More recent work shifts from demonstration-level force supervision toward closed-loop force-aware policy learning. FoAR~\cite{he2025foar} introduces a force-aware reactive policy that combines visual observations with real-time force feedback to adapt manipulation behaviors during contact. TA-VLA~\cite{zhang2025ta} systematically studies how torque signals should be incorporated into VLA architectures, highlighting the importance of where and how physical feedback is integrated into action generation. ForceVLA~\cite{yu2026forcevla} further introduces a force-aware mixture-of-experts mechanism that combines vision--language representations with real-time force sensing for physically grounded contact-rich manipulation. In contrast to methods that require force sensing during deployment, FD-VLA~\cite{zhao2026fd} distills force information from force-supervised training into latent force representations inferred from vision and robot states, enabling force-aware action generation without relying on physical force sensors at inference time. These developments illustrate a progression from explicit force imitation toward force-conditioned VLA architectures and the transfer of physical interaction knowledge into deployable latent representations.

\noindent \textbf{Audio.}
Audio provides an additional source of information about interaction events that may be visually occluded or difficult to detect from images alone. Play It By Ear~\cite{du2022play} attaches a microphone to the robot gripper and uses contact sounds during imitation learning, allowing policies to recognize interaction events that are hidden from visual observations. See, Hear, and Feel~\cite{li2023see} jointly models visual, tactile, and auditory signals through multimodal attention, where sound provides immediate evidence of events such as contact or liquid pouring. MUTEX~\cite{shah2023mutex} extends the role of audio from environmental feedback to task specification by learning a unified manipulation policy conditioned on heterogeneous inputs, including speech instructions and spoken goal descriptions.

Recent approaches increasingly exploit audio at larger scales and integrate it more tightly with manipulation policies. ManiWAV~\cite{liu2025maniwav} learns from in-the-wild audio--visual demonstrations and uses contact sounds together with vision to improve contact-rich manipulation. MS-Bot~\cite{feng2025play} introduces stage-guided multisensory fusion that dynamically adjusts the contribution of visual, tactile, and auditory signals according to the current manipulation stage. VLAS~\cite{zhao2025vlas} directly incorporates speech instructions into a VLA model through speech--text alignment, enabling spoken commands to condition downstream robot actions. More recently, ELLSA~\cite{wang2026end} unifies visual perception, speech understanding, language generation, and robot action within an end-to-end framework, supporting concurrent listening, seeing, speaking, and acting during embodied interaction. Together, these works show that audio is evolving from a supplementary contact cue toward a general interface for both physical-event perception and natural human--robot interaction.

\noindent \textbf{Other Physical Sensors.}
Beyond force and audio, emerging work has begun to integrate heterogeneous physical sensing into generalist robot policies. OmniVLA~\cite{guo2026omnivla} develops a unified multi-sensor VLA framework that incorporates signals from infrared cameras, mmWave radar, and microphone arrays. Rather than designing an independent policy for each sensor, it maps heterogeneous observations into spatially grounded representations that can be incorporated into an RGB-pretrained VLA through lightweight modality-specific components. Such approaches broaden VLA perception beyond conventional visual inputs and suggest a path toward policies that can selectively exploit sensing modalities according to environmental and task requirements. Related directions include thermal perception~\cite{guo2026omnivla}, event cameras~\cite{zhai2026vla}, and physiological sensing~\cite{he2026forceband}, although these modalities remain substantially less explored than force and audio for general-purpose manipulation.

\subsection{Latent Learning}
\label{subsec:latent_learning}

\begin{figure}[!t]
\centering
\includegraphics[width=\linewidth]{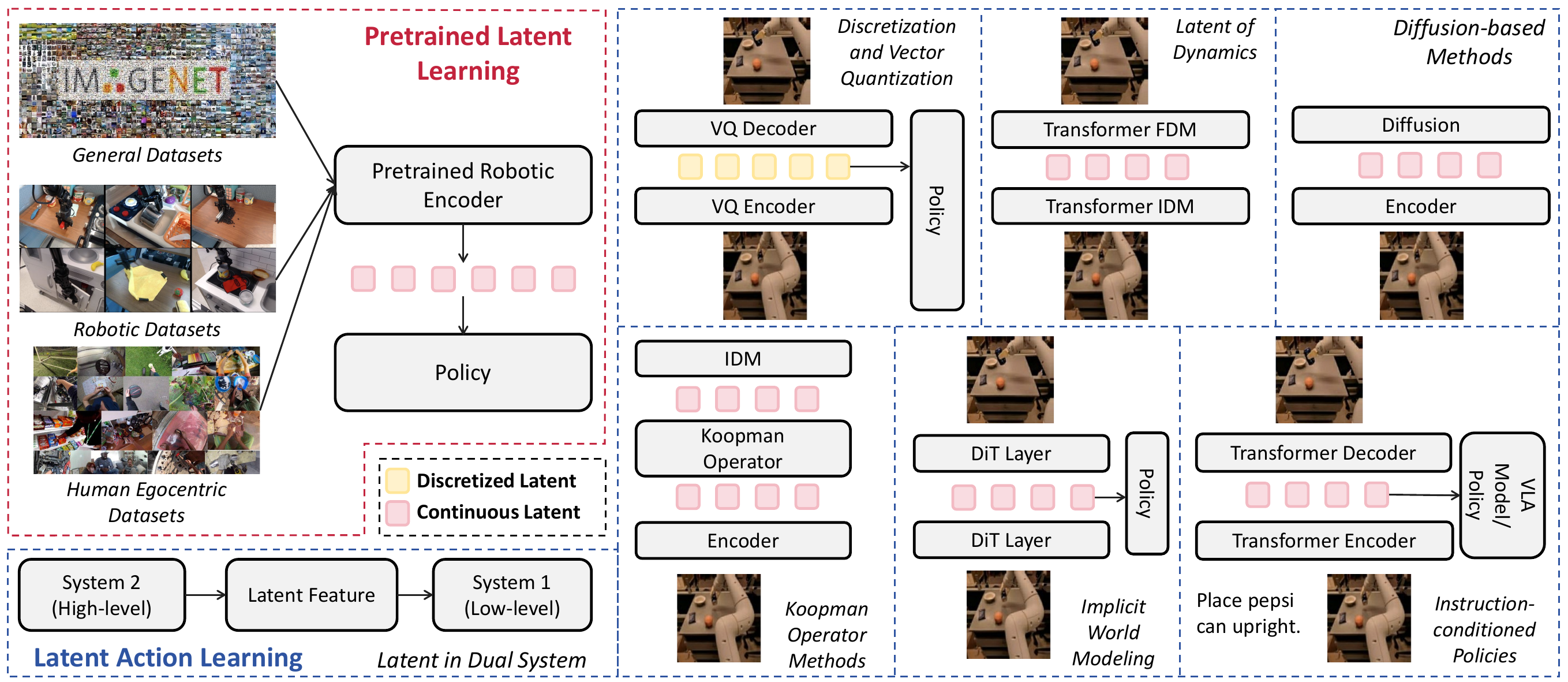}  
\caption{Overview of Latent Learning.
Left Top: Robotic encoders are pretrained on general datasets, human egocentric datasets, and robotic datasets to produce latents for policies.
Left Bottom: In the dual system, system 2 outputs latent to guide system 1 to generate action.
Right: Latent action learning is conducted through discretized (yellow) or continuous latents (pink). 
}
\label{fig: latent_learning}
\end{figure}

Latent learning investigates how robots acquire and leverage compact, structured, and transferable representations that bridge perception and control. It focuses on discovering intermediate representations that capture task-relevant semantics, dynamics, and affordances, thereby improving generalization and sample efficiency. Existing approaches can be broadly categorized into two complementary directions. \textbf{Pretrained latent learning} aims to learn general-purpose visual or multimodal representations through large-scale pre-training, typically using self-supervised or multimodal objectives to distill task-relevant structure from human egocentric videos or robotic demonstrations. These pretrained encoders provide robust perceptual embeddings that serve as transferable inputs for downstream control across diverse tasks and embodiments. \textbf{Latent action learning}, in contrast, goes beyond representation acquisition to explore how latent spaces can be effectively utilized for control. It jointly models latent representations and their temporal or causal mappings to actions, often through quantized tokens, continuous latent dynamics, or implicit world models. In this paradigm, the latent space not only encodes environmental and task information but also serves as an internal interface for reasoning, planning, and action generation, offering a unified perspective on representation and policy learning. We summarize these two paradigms in Figure~\ref{fig: latent_learning}.

\subsubsection{Pretrained Latent Learning}

Learning encoder-grounded and generalizable visual representations, often referred to as robotic representations, is essential for real-world visuomotor control. Pre-training such representations on large domain-relevant data has emerged as an effective strategy for robot learning, inspired by the success of representation pre-training in computer vision~\cite{he2022masked} and natural language processing~\cite{devlin2019bert}. Depending on the source of pre-training data, existing robotic representations can be broadly categorized into three groups: those learned from general-purpose visual datasets (\textit{e.g.}, ImageNet~\cite{deng2009imagenet}), human egocentric datasets (\textit{e.g.}, Ego4D~\cite{grauman2022ego4d}), and robot interaction datasets (\textit{e.g.}, BridgeV2~\cite{walke2023bridgedata}). Across these regimes, representation learning has progressively moved from static semantic features toward task-adaptive, interaction-aware, and dynamics-aware latent representations.

\noindent \textbf{Training on General Datasets.}
General-purpose image and video datasets provide broad visual priors that can be adapted to robotic control. Parisi et al.~\cite{parisi2022unsurprising} aggregate features from multiple layers of a MoCo-v2 encoder to construct a pre-trained visual representation for control, showing that appropriately adapted visual features can rival hand-crafted state inputs. Theia~\cite{shang2025theia} further distills complementary knowledge from multiple vision foundation models into a compact representation specialized for robot learning, while VER~\cite{wang2026ver} extends this idea toward task-adaptive representation selection by distilling multiple visual experts and learning a lightweight router to activate task-relevant features. Beyond static semantics, Token Bottleneck (ToBo)~\cite{kim2025token} compresses the current scene into a compact latent token and predicts subsequent observations, encouraging the representation to preserve both scene content and temporal dynamics. Generalizable Imitation Learning Through Pre-Trained Representations~\cite{chang2025generalizable} instead exploits patch-level features from self-supervised vision transformers to extract stable semantic keypoints, improving policy generalization across unseen object appearances, geometries, and categories.

\noindent \textbf{Training on Human Egocentric Datasets.}
Human egocentric videos provide large-scale observations of object interaction without the cost of robot data collection. MVP~\cite{xiao2022masked} applies masked visual pre-training to egocentric imagery and transfers the frozen encoder to downstream control, while VC-1~\cite{majumdar2023we} scales masked pre-training to more than 4,000 hours of egocentric video and ImageNet data to establish a general visual backbone for embodied intelligence. R3M~\cite{nair2023r3m} combines temporal contrastive learning and video--language alignment on Ego4D, and Voltron~\cite{karamcheti2023language} jointly learns language-conditioned visual reconstruction and visually grounded language generation to capture both perceptual and semantic information. Vi-PRoM~\cite{jing2023exploring} further combines contrastive self-supervision and supervised human--object interaction learning. More interaction-oriented approaches explicitly model manipulation structure: MPI~\cite{zeng2024learning} predicts intermediate interaction states and active objects, HRP~\cite{srirama2024hrp} transfers affordance priors such as hand locations and contact regions from human videos, and VIP~\cite{ma2023vip} formulates representation learning as goal-conditioned value pre-training. Ag2Manip~\cite{li2024ag2manip} further addresses the human--robot embodiment gap by learning agent-agnostic visual and action representations that suppress embodiment-specific cues. Together, these methods move egocentric pre-training from generic visual reconstruction toward interaction-aware, affordance-aware, and embodiment-invariant representations.

\noindent \textbf{Training on Robotic Datasets.}
Robot datasets provide direct supervision over observations, actions, proprioception, and interaction dynamics, allowing representations to be aligned more closely with downstream control. RPT~\cite{radosavovic2023real} pre-trains transformers by masking and reconstructing visual observations, proprioceptive states, and actions from large-scale robot trajectories, while Premier-TACO~\cite{zheng2024premier} improves temporal action contrastive learning through more effective negative sampling for few-shot policy adaptation. MCR~\cite{jiang2025robots} introduces \emph{manipulation centricity} to quantify the alignment between visual representations and manipulation performance and learns manipulation-centric features from large-scale robot data. Recent methods increasingly model how actions transform the environment: LaVA-Man~\cite{zhu2025lavaman} learns language-guided visual-action representations by predicting manipulation-induced visual changes, whereas DynaRend~\cite{tian2025dynarend} combines masked reconstruction, future prediction, and differentiable volumetric rendering to jointly encode 3D geometry, semantics, and temporal dynamics. Complementing these methods, CVESC~\cite{dong2026capturing} shows that the ability of a visual representation to recover latent environment structure, including object configuration, geometry, and physical state, strongly correlates with downstream control performance. Overall, these results suggest that pretrained robotic representations are evolving from broadly transferable visual features toward latent representations that explicitly preserve task-relevant physical state, interaction structure, and action-conditioned dynamics.

\subsubsection{Latent Action Learning}

Recent advances in latent action learning have introduced diverse paradigms that connect video representation, world modeling, and policy generation, allowing robots to acquire reusable action abstractions beyond explicit action supervision. In addition to the latent representations discussed in Section~\ref{subsubsec: il} and intermediate latent structures in dual-/multi-system VLAs in Section~\ref{subsec: vla}, latent actions provide an explicit interface between observations and executable robot controls. Existing methods instantiate latent actions primarily as discrete tokens, continuous vectors, or hierarchical structures that combine reusable skill abstractions with continuous parameterization. The following sections review these major formulations.

\textbf{i) Discretization and Vector Quantization}

Discrete latent action methods map continuous behaviors into a finite vocabulary of reusable action or motion tokens, simplifying action modeling and enabling large-scale pre-training from data without explicit robot controls. ILPO~\cite{edwards2019imitating} pioneered learning latent action variables from observation-only demonstrations, while LAPO~\cite{schmidt2023learning} reconstructs latent action structure from unlabeled videos through a latent inverse dynamics model. Behavior Generation with Latent Actions~\cite{lee2024behavior} applies vector quantization to continuous control for efficient generative behavior modeling, and Discrete Policy~\cite{wu2025discrete} learns disentangled discrete action codes for multi-task manipulation. STAR~\cite{li2025star} further introduces rotation-augmented vector quantization to learn diverse and orientation-robust skill abstractions, whereas MOTO~\cite{chen2025moto} treats latent motion tokens as a bridging language between large-scale video pre-training and downstream robot manipulation. LAPA~\cite{ye2025latent} learns latent actions by modeling transitions between current and future visual observations and subsequently trains a vision-language model to predict these latent actions before grounding them to executable robot controls using action-labeled data. DreamGen~\cite{jang2025dreamgen} further scales this video-driven learning paradigm by generating diverse synthetic robot trajectories with video world models and recovering corresponding robot actions through inverse dynamics, thereby expanding the supervision available for downstream policy learning.

\textbf{ii) Continuous Latent Action Representations}

Continuous latent action representations encode behaviors as vectors in a continuous space, preserving fine-grained variations in motion and supporting interpolation across related behaviors. Compared with discrete action tokens, continuous latent spaces are particularly suitable for modeling smooth dynamics, cross-embodiment correspondence, and predictive structures that condition downstream action generation.

\noindent \textbf{Latent Dynamics Representations.}
One line of work learns latent variables that capture changes in observations or motion dynamics and subsequently uses them to guide control. MimicPlay~\cite{wang2023mimicplay} learns long-horizon manipulation from human play by constructing latent plans conditioned on visual goals. CLAM~\cite{liang2025clam} learns continuous latent actions from unlabeled demonstrations and grounds them to executable motor commands using limited action supervision. CoMo~\cite{yang2026learning} further scales continuous latent motion learning to internet videos, extracting motion-centric representations that can serve as pseudo-actions across substantial domain and embodiment differences. These approaches exploit continuous representations to separate transferable motion structure from embodiment-specific low-level controls.

\noindent \textbf{Implicit World Modeling.}
Another direction derives action-relevant latent representations from predictive world models. VPP~\cite{hu2025video} adapts a video foundation model for language-conditioned future prediction and aggregates its intermediate representations through a Video-Former to condition a diffusion policy. FLARE~\cite{zheng2025flare} learns action-relevant representations through implicit future prediction, allowing predictive latent features to improve policy generalization without explicitly generating future observations during deployment. Genie Envisioner~\cite{liao2026genie} further aligns intermediate latent features of a video diffusion model with representations inside an action diffusion policy, transferring predictive structures from video generation to robot control. These methods indicate that useful action representations can emerge from predicting how the physical scene evolves, even when the world model itself is not explicitly decoded at policy inference time.

\noindent \textbf{Diffusion-based Methods.}
Continuous latent representations can also provide structured spaces in which diffusion policies generate robot behaviors. LAD~\cite{bauer2025latent} learns a shared latent action space and trains diffusion policies within this space to facilitate cross-embodiment manipulation transfer. KOAP~\cite{bi2025imitation} combines diffusion-based planning with Koopman controllers, allowing a small set of learned latent actions to be composed into stable long-horizon behaviors. LaDi-WM~\cite{huang2025ladi} instead trains a latent diffusion world model to predict semantic and geometric dynamics, which are subsequently used as conditioning information for diffusion-based action generation.

\noindent \textbf{Koopman Operator Methods.}
Koopman-based approaches seek latent spaces in which nonlinear manipulation dynamics become easier to model through approximately linear evolution. KoDex~\cite{han2023utility} investigates Koopman operator representations for dexterous manipulation and demonstrates their utility for learning structured manipulation dynamics. KOROL~\cite{chen2024korol} further learns interpretable object-centric latent features whose evolution can be rolled out using a Koopman operator, providing visualizable predictions that support downstream manipulation.

\noindent \textbf{Goal- or Instruction-conditioned Representations.}
Latent spaces can additionally provide interfaces between high-level task specifications and low-level behavior. Procedure Cloning~\cite{yang2022chain} learns intermediate procedural representations that capture the sequence of decisions underlying long-horizon demonstrations. GRIF~\cite{myers2023goal} constructs a shared goal representation that aligns language instructions with visual goal observations using partially labeled data, while IGOR~\cite{chen2024igor} represents image goals as atomic control units that bridge high-level foundation models and downstream robot controllers. UniVLA~\cite{bu2025univla} learns task-centric latent actions without action supervision from videos spanning different embodiments and viewpoints and subsequently trains a VLA to predict these latent action tokens for downstream control. Oishi et al.~\cite{oishi2025imitation} further learn disentangled latent behavioral characteristics, allowing qualitative task modifiers to selectively adjust motion attributes during execution.

\textbf{iii) Hierarchical and Parameterized Latent Skills}

Recent work moves beyond flat discrete or continuous representations toward structured latent spaces that explicitly capture the hierarchy and compositionality of manipulation skills. HiMaCon~\cite{liu2025himacon} discovers hierarchical manipulation concepts from unlabeled multimodal trajectories through cross-modal correlation learning and multi-horizon prediction, organizing interaction patterns across temporal scales from short-term behaviors to longer-horizon subgoals. DEPS~\cite{gupta2025learning} instead learns parameterized skills through a hierarchical latent policy consisting of a discrete skill selector, a continuous skill parameter, and a low-level action policy. Such hybrid representations preserve reusable discrete skill identities while capturing task-specific variations through continuous parameters. Together, these methods suggest that latent action learning is evolving from flat action compression toward structured skill representations that model temporal hierarchy, behavioral variation, and compositionality, thereby providing more scalable interfaces between large-scale observational data and executable robot control.

\subsection{Policy Learning}
\label{subsec:policy_learning}

Policy learning defines how a robot transforms internal representations, such as latent features or encoded observations, into executable actions that interact with the physical world. It focuses on decoding perceptual and latent information into control outputs (for example, end-effector poses) through learned mappings rather than manually designed rules. In essence, policy learning establishes the computational mechanism that connects perception and decision-making with motor execution. We summarize these methods in Figure~\ref{fig: policy_learning}.

\subsubsection{MLP-based Policy}
\label{subsubsec: mlp}

MLP-based policies~\cite{nair2023r3m,majumdar2023we,radosavovic2023real,lyu2025dywa} provide a simple and efficient baseline for visuomotor control by directly mapping encoded observations or latent representations to robot actions. Such architectures are computationally lightweight and particularly effective when coupled with strong pretrained encoders, but their limited temporal modeling capacity makes performance strongly dependent on the quality of the input representation.

\subsubsection{Transformer-based Policy}
\label{subsubsec: transformer}

Transformer-based policies exploit attention to model dependencies among observations, actions, goals, and language over time. Self-attention captures long-range temporal structure, while cross-attention allows action generation to selectively retrieve task-relevant perceptual or linguistic information. These properties make Transformers particularly suitable for action chunking, autoregressive control, multimodal conditioning, and long-horizon manipulation.

\tikzset{
    my-box/.style={
        rectangle,
        draw=black,
        rounded corners,
        text opacity=1,
        minimum height=1.5em,
        minimum width=5em,
        inner sep=2pt,
        align=center,
        fill opacity=.5,
    },
    leaf-2d/.style={
        my-box,
        fill=yellow!32,
        text=black,
        font=\normalsize,
        inner xsep=5pt,
        inner ysep=4pt,
        align=left,
    },
    category/.style={
        align=left,
        inner xsep=3pt,
        inner ysep=4pt,
    }
}
\newcommand{\policybranch}[1]{\parbox[c]{38mm}{\raggedright #1}}
\newcommand{\policycat}[1]{\parbox[c]{33mm}{\raggedright #1}}
\newcommand{\policyitem}[1]{\parbox[c]{96mm}{\raggedright #1}}

\begin{figure}[!t]
\centering
\resizebox{\textwidth}{!}{
\begin{forest}
forked edges,
for tree={
    grow=east,
    reversed=true,
    anchor=base west,
    parent anchor=east,
    child anchor=west,
    base=left,
    font=\normalsize,
    rectangle,
    draw=black,
    rounded corners,
    align=left,
    minimum width=4em,
    edge+={darkgray, line width=1pt},
    s sep=3pt,
    l sep=5pt,
    inner xsep=3pt,
    inner ysep=4pt,
    line width=1.1pt,
    ver/.style={
        rotate=90,
        child anchor=north,
        parent anchor=south,
        anchor=center,
        inner xsep=5pt
    },
},
[{Policy Learning (\S~\ref{subsec:policy_learning})}, ver
[{\policybranch{MLP-based Policy \S~\ref{subsubsec: mlp}}}
[{\policyitem{R3M~\cite{nair2023r3m}, VC-1~\cite{majumdar2023we}, RPT~\cite{radosavovic2023real}, DyWA~\cite{lyu2025dywa}}}, leaf-2d]
]
[{\policybranch{Transformer-based Policy \S~\ref{subsubsec: transformer}}}
[{\policycat{ACT-based Policy}}, category
[{\policyitem{ACT~\cite{zhao2023learning}, RoboAgent~\cite{bharadhwaj2024roboagent}, Bi-ACT~\cite{buamanee2024bi}, BAKU~\cite{haldar2024baku}, InterACT~\cite{lee2024interact}, Haptic-ACT~\cite{li2025haptic}, ALOHA Unleashed~\cite{zhao2024aloha}, Bi-LAT~\cite{kobayashi2025bi}, Chain-of-Action~\cite{pan2025chain}, Q-chunking~\cite{li2025reinforcement}}}, leaf-2d]
]
[{\policycat{Autoregressive Policy}}, category
[{\policyitem{Act3D~\cite{gervet2023act3d}, ICRT~\cite{fu2024context}, CARP~\cite{gong2025carp}, Dense Policy~\cite{su2025dense}, HAT~\cite{qiu2025humanoid}, RVT~\cite{goyal2023rvt}}}, leaf-2d]
]
]
[{\policybranch{Diffusion Policy \S~\ref{subsubsec: dp}}}
[{\policycat{3D DP}}, category
[{\policyitem{DP3~\cite{ze20243d}, 3D Diffuser Actor~\cite{ke2024diffuser}, iDP3~\cite{ze2025generalizable}, G3Flow~\cite{chen2025g3flow}, DP4~\cite{liu2025spatial}, H\textsuperscript{3}DP~\cite{lu2025h}}}, leaf-2d]
]
[{\policycat{Scaling and Foundation DP}}, category
[{\policyitem{ScaleDP~\cite{zhu2025scaling}, RDT-1B~\cite{liu2025rdt}}}, leaf-2d]
]
[{\policycat{Accelerated DP}}, category
[{\policyitem{OneDP~\cite{wang2025one}, Consistency Policy~\cite{prasad2024consistency}, ManiCM~\cite{lu2024manicm}}}, leaf-2d]
]
[{\policycat{MoE DP}}, category
[{\policyitem{SDP~\cite{wang2025sparse}, MoDE~\cite{raposo2024mixture}}}, leaf-2d]
]
[{\policycat{Geometry- and Equivariance-aware DP}}, category
[{\policyitem{EquiBot~\cite{yang2025equibot}, EDP~\cite{wang2025equivariant}}}, leaf-2d]
]
[{\policycat{DP with Video Prediction}}, category
[{\policyitem{UniPi~\cite{du2023learning}, UVA~\cite{li2025unified}, UWM~\cite{zhu2025unified}}}, leaf-2d]
]
[{\policycat{DP with Other Techniques}}, category
[{\policyitem{HDP~\cite{ma2024hierarchical}, DPPO~\cite{ren2024diffusion}, Latent Diffusion Planning~\cite{xie2025latent}, MBA~\cite{su2025motion}, RDP~\cite{xue2025reactive}, Past-Token Prediction~\cite{villasevil2025learning}, DP-Attacker~\cite{chen2024diffusion}}}, leaf-2d]
]
]
[{\policybranch{Flow Matching Policy \S~\ref{subsubsec: fmp}}}
[{\policycat{General FM Policy}}, category
[{\policyitem{AdaFlow~\cite{hu2024adaflow}, X-IL~\cite{jia2025xil}, FLOWER~\cite{reuss2025flower}, Streaming Flow Policy~\cite{jiang2025streaming}, RTC~\cite{black2025real}, H-RDT~\cite{bi2026hrdt}, VFP~\cite{zhai2026vfp}, GenFlowRL~\cite{yu2025genflowrl}, ManiFlow~\cite{yan2025maniflow}}}, leaf-2d]
]
[{\policycat{Efficient and One-step FM}}, category
[{\policyitem{MP1~\cite{sheng2026mp1}, Action-to-Action FM~\cite{jia2026action}}}, leaf-2d]
]
[{\policycat{3D FM Policy}}, category
[{\policyitem{PointFlowMatch~\cite{chisari2024learning}, FlowPolicy~\cite{zhang2025flowpolicy}, Affordance FM~\cite{zhang2024affordance}, FlowRAM~\cite{wang2025flowram}, 3D FlowMatch Actor~\cite{gkanatsios20253d}}}, leaf-2d]
]
]
[{\policybranch{SSM-based Policy \S~\ref{subsubsec: ssm}}}
[{\policyitem{MaIL~\cite{jia2024mail}, RoboMamba~\cite{liu2024robomamba}, FlowRAM~\cite{wang2025flowram}, MTIL~\cite{zhou2025mtil}, Mamba as Motion Encoder~\cite{tsuji2025mamba}}}, leaf-2d]
]
[{\policybranch{SNN-based Policy \S~\ref{subsubsec: snn}}}
[{\policyitem{SDP~\cite{hou2025sdp}, STMDP~\cite{wang2024brain}, Multimodal SNN~\cite{zhang2025multimodal}, Fully Spiking A2C~\cite{zhang2025fully}}}, leaf-2d]
]
[{\policybranch{Frequency-based Policy \S~\ref{subsubsec: freq}}}
[{\policyitem{Fourier Transporter~\cite{huang2024fourier}, FreqPolicy~\cite{zhong2025freqpolicy}, Wavelet Policy~\cite{huang2025wavelet}, ManipForce~\cite{lee2026manipforce}}}, leaf-2d]
]
[{\policybranch{Action Tokenization and Action Representation \S~\ref{subsubsec: action_tokenization}}}
[{\policycat{Action Tokenization and Compression}}, category
[{\policyitem{FAST~\cite{pertsch2025fast}, \cite{wang2026learning}, \cite{chuang2026the}}}, leaf-2d]
]
[{\policycat{Trajectory Parameterization}}, category
[{\policyitem{ABPolicy~\cite{yang2026abpolicy}}}, leaf-2d]
]
]
[{\policybranch{Drift-based Policy \S~\ref{subsubsec: drift}}}
[{\policyitem{DBPO~\cite{gao2026drift}, Implicit Drifting Policy~\cite{yang2026implicit}}}, leaf-2d]
]
]
\end{forest}
}
\caption{A structured overview of policy learning for visuomotor control, organized by modeling paradigm.}
\label{fig: policy_learning}
\end{figure}

\noindent \textbf{ACT-based Policy.}
The Action Chunking Transformer (ACT)~\cite{zhao2023learning} predicts short action sequences using a CVAE-based Transformer, reducing compounding errors and enabling fine-grained bimanual manipulation with low-cost hardware. RoboAgent~\cite{bharadhwaj2024roboagent} extends action chunking to multi-task manipulation through semantic augmentation, while Bi-ACT~\cite{buamanee2024bi} incorporates bilateral position--force information for contact-sensitive imitation. BAKU~\cite{haldar2024baku} develops an efficient multi-task Transformer capable of fusing heterogeneous observations and decoding chunked actions, and InterACT~\cite{lee2024interact} explicitly models inter-arm dependencies through hierarchical attention for coordinated bimanual manipulation. Haptic-ACT~\cite{li2025haptic} combines action chunking with immersive haptic demonstrations, whereas Bi-LAT~\cite{kobayashi2025bi} further introduces language conditioning to control both motion and contact force. Chain-of-Action~\cite{pan2025chain} formulates trajectory generation as autoregressive action reasoning, while Q-chunking~\cite{li2025reinforcement} incorporates action chunks into reinforcement learning to improve exploration in sparse-reward settings. These developments extend action chunking from short-horizon imitation toward multimodal, interactive, and decision-aware sequence control.

\noindent \textbf{Autoregressive Policy.}
Autoregressive policies explicitly factorize robot trajectories into sequential predictions. Act3D~\cite{gervet2023act3d} constructs an adaptive-resolution 3D feature field and uses coarse-to-fine attention for end-effector prediction, while ICRT~\cite{fu2024context} treats demonstrations as in-context prompts and autoregressively predicts actions for few-shot imitation. CARP~\cite{gong2025carp} learns multi-scale action embeddings and predicts trajectories through coarse-to-fine autoregressive refinement, whereas Dense Policy~\cite{su2025dense} introduces bidirectional autoregressive action generation using a BERT-style encoder. Robotic View Transformer~\cite{goyal2023rvt} predicts 3D manipulation locations through multi-view reprojection, and 3D-MVP~\cite{qian20253d} pretrains a multi-view 3D Transformer through masked reconstruction before adapting its features to action generation. Together, these approaches demonstrate that autoregressive policies can model action dependencies at multiple temporal and spatial resolutions rather than predicting individual motor commands independently.

\subsubsection{Diffusion Policy}
\label{subsubsec: dp}

Diffusion Policy (DP)~\cite{chi2023diffusion} formulates visuomotor control as conditional denoising in action space, enabling expressive modeling of multimodal demonstrations while supporting receding-horizon execution. Its success has motivated extensions along several directions, including 3D perception, model scaling, acceleration, expert specialization, geometric structure, predictive modeling, and reinforcement-learning adaptation.

\noindent \textbf{3D DP.}
Three-dimensional diffusion policies incorporate explicit geometric structure to improve spatial grounding and manipulation generalization. DP3~\cite{ze20243d} conditions diffusion policies on compact point-cloud representations and demonstrates strong data efficiency across complex manipulation tasks. 3D Diffuser Actor~\cite{ke2024diffuser} lifts multi-view visual features into 3D scene representations and performs diffusion-based prediction of end-effector keyposes and trajectories using 3D relative-position attention. iDP3~\cite{ze2025generalizable} extends 3D diffusion policies toward egocentric deployment, while G3Flow~\cite{chen2025g3flow} maintains dynamic object-centric 3D semantic features during manipulation. CordViP~\cite{fu2025cordvip} explicitly models hand--object spatial correspondences, DP4~\cite{liu2025spatial} incorporates 3D spatial and 4D temporal information through dynamic Gaussian representations, and H$^3$DP~\cite{lu2025h} hierarchically organizes depth, multi-scale visual features, and coarse-to-fine action generation.

\noindent \textbf{Scaling and Foundation DP.}
Diffusion policies have also progressed from task-specific models toward large-scale generalist policies. ScaleDP~\cite{zhu2025scaling} redesigns conditioning and attention mechanisms to scale diffusion Transformers from relatively small policies toward billion-parameter models, demonstrating systematic performance gains with increasing capacity. RDT-1B~\cite{liu2025rdt} develops a billion-scale diffusion foundation policy for bimanual manipulation and introduces a unified action representation to learn transferable behaviors from heterogeneous multi-robot datasets. These works suggest that scaling laws and heterogeneous data aggregation, previously central to foundation vision and language models, are increasingly relevant to low-level robot policies.

\noindent \textbf{Accelerated DP.}
Because iterative denoising introduces substantial control latency, several approaches seek to reduce the number of policy evaluations. OneDP~\cite{wang2025one} distills an iterative diffusion policy into one-step generation, while Consistency Policy~\cite{prasad2024consistency} learns self-consistent predictions across diffusion times to enable fast sampling with limited degradation. ManiCM~\cite{lu2024manicm} applies consistency modeling to 3D manipulation and achieves approximately one-step action generation while maintaining competitive manipulation performance. These approaches shift computation from iterative inference toward training-time distillation or consistency constraints.

\noindent \textbf{MoE DP.}
Mixture-of-Experts architectures improve the capacity and specialization of diffusion policies. SDP~\cite{wang2025sparse} sparsely activates task-specific experts to enable multi-task learning without proportionally increasing computation, while MoDE~\cite{raposo2024mixture} introduces noise-conditioned routing among expert denoisers. More recently, skill-oriented MoE diffusion policies explicitly interpret expert specialization as reusable manipulation abstractions, coupling temporally stable routing with selective expert activation to support compositional and few-shot skill learning.

\noindent \textbf{Geometry- and Equivariance-aware DP.}
Another direction incorporates the geometry of robot actions directly into policy design. EquiBot~\cite{yang2025equibot} introduces SIM(3)-equivariant diffusion for data-efficient generalization under scale, rotation, and translation changes, while EDP~\cite{wang2025equivariant} exploits SO(2)/SE(3) symmetry for 6-DoF manipulation. More recently, Lie-manifold diffusion formulations model pose generation directly on $\mathrm{SE}(3)$ rather than treating rotations and translations as unconstrained Euclidean vectors, predicting updates in the tangent space and mapping them back to valid poses through geometric operators. Such methods reduce representation mismatch between generative objectives and the underlying kinematic structure of robot actions.

\noindent \textbf{DP with Video Prediction.}
Predictive visual modeling provides another source of structure for diffusion policies. UniPi~\cite{du2023learning} formulates planning as language-conditioned video generation and subsequently extracts actions from imagined task execution. UVA~\cite{li2025unified} jointly models video prediction and action generation through shared latent representations, while UWM~\cite{zhu2025unified} combines video and action diffusion within a unified framework and can exploit both action-labeled demonstrations and action-free videos. These methods use future visual evolution as an intermediate representation that constrains downstream action generation.

\noindent \textbf{DP with Other Techniques.}
Diffusion policies have further been combined with hierarchical planning, reinforcement learning, latent prediction, tactile sensing, and task-specific structural priors. HDP~\cite{ma2024hierarchical} separates high-level keypose generation from low-level trajectory diffusion, while DPPO~\cite{ren2024diffusion} applies policy-gradient fine-tuning to pretrained diffusion policies. Latent Diffusion Planning~\cite{xie2025latent} predicts future states in a learned latent space and decodes them into executable controls with inverse dynamics, allowing action-free and suboptimal trajectories to contribute to policy learning. MBA~\cite{su2025motion} first predicts object motion and subsequently conditions robot action diffusion on the inferred motion, whereas RDP~\cite{xue2025reactive} combines a low-frequency diffusion planner with high-frequency tactile feedback for contact-rich manipulation. YOTO~\cite{zhou2025you} synthesizes demonstrations from human videos and trains a specialized bimanual diffusion policy for long-horizon coordination.

Recent analysis has also questioned which ingredients are responsible for the strong performance of diffusion policies. Much Ado About Noising~\cite{pan2026much} shows that their advantages cannot be explained solely by the ability to represent multimodal action distributions; noise injection and iterative supervised refinement can themselves induce useful regularization toward plausible action manifolds. This perspective suggests that future policies may retain the optimization benefits of diffusion-style training while simplifying the generative process used at inference.

\subsubsection{Flow Matching Policy}
\label{subsubsec: fmp}

Flow-matching policies model action generation as continuous transport between an initial distribution and demonstrated robot trajectories. Compared with conventional diffusion objectives, flow matching directly learns a velocity field and often permits straighter generation paths, fewer inference steps, and smoother trajectory synthesis. Recent work explores both general visuomotor flow policies and geometry-aware 3D variants.

\noindent \textbf{General Flow-Matching Policy.}
X-IL~\cite{jia2025xil} systematically studies architectural and representation choices across diffusion and flow-matching imitation policies. AdaFlow~\cite{hu2024adaflow} introduces variance-adaptive flow integration, dynamically adjusting the number of ODE steps according to uncertainty. FLOWER~\cite{reuss2025flower} develops a compact vision-language-action flow policy with strong multi-task manipulation performance, while Streaming Flow Policy~\cite{jiang2025streaming} directly interprets action trajectories as continuous flow trajectories and incrementally generates actions during execution. Real-Time Action Chunking~\cite{black2025real} overlaps action generation with execution through action inpainting, improving continuity between consecutive chunks. H-RDT~\cite{bi2026hrdt} combines human-video pre-training with flow-matching fine-tuning for bimanual manipulation, and VFP~\cite{zhai2026vfp} introduces variational flow matching to improve multimodal trajectory generation. GenFlowRL~\cite{yu2025genflowrl} uses generated object-centric motion flows as reward signals for reinforcement learning, whereas VITA~\cite{gao2026vita} learns a direct vision-to-action flow mapping with a lightweight architecture. ManiFlow~\cite{yan2025maniflow} combines flow matching with consistency-style training to straighten generative trajectories and reduce sampling cost. Action-to-Action Flow Matching~\cite{jia2026action} further replaces an uninformed noise source with representations of previously executed actions, exploiting temporal continuity to shorten the transport path toward future actions.

\noindent \textbf{3D Flow-Matching Policy.}
FlowPolicy~\cite{zhang2025flowpolicy} introduces consistency flow matching conditioned on 3D point clouds for efficient manipulation, while PointFlowMatch~\cite{chisari2024learning} directly conditions continuous flow trajectories on point-cloud and proprioceptive observations. FlowRAM~\cite{wang2025flowram} couples flow matching with region-aware RGB-D representations and Mamba-based temporal modeling, improving generalization under occlusion and spatial variation. MP1~\cite{sheng2026mp1} applies MeanFlow to point-cloud-conditioned manipulation and compresses trajectory generation to a single network evaluation. 3D FlowMatch Actor~\cite{gkanatsios20253d} combines pretrained 3D representations with flow matching to unify single- and dual-arm control, while Affordance-based Flow Matching~\cite{zhang2024affordance} conditions flow generation on task-relevant spatial affordances. Together, these approaches show that flow matching can combine efficient generative control with increasingly structured spatial representations.

\subsubsection{SSM-based Policy}
\label{subsubsec: ssm}

State-space-model policies use architectures such as Mamba to efficiently encode long observation and action histories while retaining approximately linear sequence complexity. MaIL~\cite{jia2024mail} demonstrates that Mamba can replace conventional sequence models in imitation learning and provide strong long-horizon control. RoboMamba~\cite{liu2024robomamba} combines vision, language, and robot states in a multimodal state-space architecture, while FlowRAM~\cite{wang2025flowram} integrates region-aware Mamba representations with flow-based action generation. MTIL~\cite{zhou2025mtil} explicitly encodes the complete interaction history to improve temporally dependent manipulation. More recently, SUREFlow~\cite{islam2026sureflow} combines Mamba-based sequence modeling with uncertainty-aware residual flow matching, selectively refining unreliable components of generated actions while retaining efficient long-context processing.

\subsubsection{SNN-based Policy}
\label{subsubsec: snn}

Spiking neural network (SNN) policies explore event-driven computation for energy-efficient and precise robot control. SDP~\cite{hou2025sdp} develops a spiking diffusion policy with learnable channel-wise membrane thresholds, allowing spiking activity to adapt across modalities. STMDP~\cite{wang2024brain} combines spiking computation with Transformer-based diffusion control to retain generative trajectory modeling while introducing event-driven dynamics. Multimodal Spiking Neural Network~\cite{zhang2025multimodal} incorporates multiple sensing channels for manipulation in high-dimensional action spaces, while Fully Spiking Actor-Critic~\cite{zhang2025fully} extends spiking architectures to continuous-control reinforcement learning. Although these methods demonstrate the feasibility of neuromorphic policy learning, SNN-based manipulation remains substantially less mature than Transformer- and generative-policy paradigms.

\subsubsection{Frequency-based Policy}
\label{subsubsec: freq}

Frequency-based policies exploit spectral representations of actions or observations to capture long-horizon structure and separate dynamics occurring at different temporal scales. Fourier Transporter~\cite{huang2024fourier} uses Fourier-domain representations to construct bi-equivariant manipulation policies in 3D. FreqPolicy~\cite{zhong2025freqpolicy} constrains flow-based action generation through frequency consistency, while a complementary frequency-autoregressive formulation represents continuous action trajectories through spectral tokens. Wavelet Policy~\cite{huang2025wavelet} uses hierarchical wavelet decomposition to capture both coarse long-horizon behavior and fine-grained local motion. ManipForce~\cite{lee2026manipforce} further extends frequency-aware modeling to contact-rich manipulation by jointly processing asynchronous visual observations and high-frequency force--torque feedback. These approaches suggest that explicitly separating temporal frequency components can provide useful inductive biases for long-horizon and contact-sensitive control.

\subsubsection{Action Tokenization and Structured Action Representation}
\label{subsubsec: action_tokenization}

In addition to the policy architecture itself, recent work increasingly studies how dense continuous robot trajectories should be represented before prediction. Compact action representations can reduce sequence length, remove temporal redundancy, impose trajectory smoothness, and provide more efficient interfaces for autoregressive or generative policies. FAST~\cite{pertsch2025fast} transforms high-frequency action chunks into the frequency domain and combines quantization with byte-pair encoding to obtain compact discrete action tokens, substantially reducing the sequence length required by autoregressive VLA policies. Learning High-Frequency Continuous Action Chunks in Latent Space~\cite{wang2026learning} instead learns a continuous latent representation of dense high-frequency action chunks and introduces reuse-and-refinement across consecutive predictions to maintain smooth execution. ABPolicy~\cite{yang2026abpolicy} represents trajectories using B-spline control points and performs flow-based generation in this compact parameter space, naturally enforcing intra-chunk smoothness while reducing redundancy. These methods highlight action representation as an independent design dimension: rather than only improving the network that predicts actions, policies can benefit from choosing structured coordinates in which action sequences are easier to model.

\subsubsection{Drift-based Policy}
\label{subsubsec: drift}

Drift-based policies have recently emerged as an alternative route to efficient generative control. Unlike diffusion and flow-matching policies that typically construct actions through iterative denoising or numerical integration, drifting methods attempt to internalize iterative trajectory refinement into the training objective and directly generate actions in a single policy evaluation. Drift-Based Policy~\cite{gao2026drift} learns a native one-step generative policy through fixed-point drifting, while its policy-optimization extension enables reinforcement-learning refinement without abandoning one-step deployment. Implicit Drifting Policy~\cite{yang2026implicit} further avoids explicitly estimating the drifting vector field and instead exploits the local geometry of expert demonstrations to constrain predictions toward plausible conditional action manifolds. Although still an emerging research direction, drifting provides a conceptually distinct path toward real-time multimodal control by replacing iterative inference with training-time geometric refinement.

\section{Approaches to the Key Bottlenecks}
\label{sec:bottlenecks}

Despite the remarkable progress of robot manipulation, achieving general-purpose deployment in unstructured real-world environments remains constrained by both fundamental learning bottlenecks and system-level challenges. At the learning level, \textbf{data} and \textbf{generalization} remain central: the quality, diversity, and utilization of data determine the scalability of imitation and reinforcement learning, while generalization governs whether learned skills can transfer across unseen environments, novel tasks, and diverse embodiments. Beyond these foundations, real-world deployment increasingly requires higher-level capabilities for coordinating complex behaviors and interacting with humans. \textbf{Agentic systems} provide mechanisms for planning, tool orchestration, memory, verification, and continual adaptation, whereas \textbf{human--robot interaction and collaboration} enable robots to infer human intent, incorporate feedback, share autonomy, and coordinate safely with users. Accordingly, this section discusses these four complementary directions---data, generalization, agents, and human--robot collaboration---that jointly shape the scalability, adaptability, and practical deployment of modern robot manipulation systems.

\subsection{Data Collection and Utilization}
\label{subsec:data}

Data serves as the cornerstone of learning-based and data-driven approaches in embodied intelligence, as its quality, diversity, and scale fundamentally determine the effectiveness and generalization of learned policies. Unlike conventional machine learning, robot learning depends on data that jointly capture perception, action, physical interaction, and embodiment-specific dynamics, making data simultaneously a critical resource and a principal bottleneck. This section focuses on two complementary dimensions. \textbf{Data collection} concerns how new demonstrations and interaction experiences are acquired or generated, encompassing human teleoperation, human-in-the-loop refinement, synthetic and automatic generation, and crowdsourced acquisition. \textbf{Data utilization} concerns how collected data are selected, retrieved, relabeled, augmented, expanded, and reweighted to improve learning efficiency and robustness. Together, these two dimensions form the data foundation for scaling robot learning toward increasingly generalizable real-world manipulation. We summarize representative data collection paradigms in Figure~\ref{fig: data-taxonomy-colored} and illustrate their specific forms in Figure~\ref{fig: data_collection}.

\subsubsection{Data Collection}
\label{subsubsec: data_collect}

Data collection for robot learning spans multiple paradigms that differ in cost, scalability, embodiment alignment, data fidelity, and reliance on humans. Broadly, existing approaches can be grouped into four categories. First, \textbf{human teleoperation} systems acquire direct supervision through interfaces ranging from replica arms and wearable devices to XR and robot-free systems. Second, \textbf{human-in-the-loop} methods reduce human effort by allowing operators to selectively intervene in or correct autonomous policy execution. Third, \textbf{synthetic and automatic data generation} amplifies limited human supervision through simulation, demonstration transformation, foundation models, optimization, or autonomous policy rollouts. Finally, \textbf{crowdsourcing frameworks} distribute data acquisition across large groups of contributors, lowering access barriers and broadening data coverage.

\textbf{i) Human Teleoperation}

Human teleoperation remains the most direct and widely used paradigm for collecting high-quality robot demonstrations. Existing systems differ primarily in interface cost, embodiment alignment, sensing fidelity, portability, and operator burden, ranging from direct robot teleoperation to robot-free systems that capture human manipulation trajectories for subsequent transfer to robotic embodiments.

\noindent \textbf{Cost--Effective and Scalable Teleoperation.}
Several systems~\cite{young2021visual,qin2023anyteleop,wu2024gello,chi2024universal,iyer2025open} reduce hardware and deployment requirements to support scalable demonstration collection. Young et al.~\cite{young2021visual} use inexpensive reacher--grabber tools to collect visual demonstrations that can be transferred to robot manipulation. AnyTeleop~\cite{qin2023anyteleop} develops a modular vision-based teleoperation framework applicable across robot arms, dexterous hands, and camera configurations, while GELLO~\cite{wu2024gello} introduces a low-cost replica-arm interface that provides intuitive embodiment-aligned control. UMI~\cite{chi2024universal} further decouples demonstration collection from access to the target robot by using a portable manipulation interface to acquire in-the-wild robot-compatible trajectories. OPEN TEACH~\cite{iyer2025open} instead uses consumer VR hardware and hand tracking to provide a broadly accessible teleoperation interface across different robot morphologies.

\noindent \textbf{Feedback--Rich Teleoperation for Contact and Dexterity.}
For contact-rich manipulation, data collection systems increasingly augment kinematic trajectories with force, tactile, or haptic information. M2R~\cite{kim2023training} introduces a master-to-robot interface with force or torque sensing to capture physical interaction information without expensive bilateral hardware. ALPHA-BiACT~\cite{kobayashi2025alpha} combines bilateral position and force control to collect richer demonstrations for uni- and bimanual manipulation, while Bunny-VisionPro~\cite{ding2025bunny} couples real-time VR hand tracking with low-cost haptic interfaces for dexterous teleoperation. FreeTacMan~\cite{wu2026freetacman} further develops a robot-free wearable interface with visuo-tactile grippers and optical pose tracking, jointly recording visual, tactile, and kinematic signals from human manipulation for contact-rich robot learning. These systems complement conventional motion demonstrations with direct interaction signals that are particularly important for insertion, compliant manipulation, and fine contact control.

\noindent \textbf{Egocentric and XR Interfaces Aligning Viewpoints.}
XR and egocentric interfaces provide intuitive human control while reducing discrepancies between human and robot observations. Zhang et al.~\cite{zhang2018deep} use consumer VR controllers to place operators directly in the robot observation--action loop, while EgoMimic~\cite{kareer2025egomimic} leverages egocentric wearable observations and human motion for viewpoint-aligned imitation learning. ARCap~\cite{chen2025arcap} provides augmented-reality feedback that guides users toward robot-feasible demonstrations, lowering the expertise required for high-quality data collection. Active perception can further improve demonstration quality when manipulation is affected by occlusion; AV-ALOHA~\cite{chuang2024active}, for example, incorporates an auxiliary controllable camera to maintain task-relevant visual observations during bimanual manipulation.


\definecolor{cjyBlue}{HTML}{87CEFA}
\definecolor{cjyOrange}{HTML}{FF7F0E}
\definecolor{cjyGreen}{HTML}{8FBC8F}
\definecolor{cjyPurple}{HTML}{9467BD}

\newcommand{\datamain}[1]{\parbox[c]{28mm}{\raggedright #1}}
\newcommand{\datasub}[1]{\parbox[c]{38mm}{\raggedright #1}}
\newcommand{\dataitem}[1]{\parbox[c]{70mm}{\raggedright #1}}

\begin{figure}[!t]
\centering
\scriptsize

\forestset{
  taxo/.style={
    for tree={
      grow'=0,
      parent anchor=east,
      child anchor=west,
      anchor=west,
      edge path'={(!u.parent anchor) -- +(3pt,0) |- (.child anchor)},
      edge={draw, line width=0.6pt},
      rounded corners,
      draw,
      minimum height=4mm,
      inner xsep=4pt,
      inner ysep=2pt,
      s sep=3pt,
      l sep=6pt,
      align=left,
      font=\scriptsize
    },
    root/.style={
      draw,
      rounded corners=2pt,
      inner xsep=2pt,
      inner ysep=2pt,
      align=center,
      font=\bfseries
    },
    group/.style={
      draw,
      rounded corners=2pt,
      inner xsep=3pt,
      inner ysep=3pt,
      align=center,
      font=\bfseries
    },
    sub_1/.style={
      draw,
      rounded corners=2pt,
      fill opacity=.5,
      inner xsep=4pt,
      inner ysep=3pt,
      align=left
    },
    sub/.style={
      draw,
      rounded corners=2pt,
      fill=hidden-blue!32,
      fill opacity=.5,
      inner xsep=4pt,
      inner ysep=3pt,
      align=left
    },
    item/.style={
      draw,
      rounded corners=2pt,
      fill=yellow!32,
      fill opacity=.5,
      inner xsep=4pt,
      inner ysep=3pt,
      align=left
    },
    sub_new/.style={
      draw,
      rounded corners=2pt,
      fill=yellow!32,
      fill opacity=.5,
      inner xsep=4pt,
      inner ysep=3pt,
      align=left
    }
  }
}

\begin{forest} taxo
[{\rotatebox{90}{\textbf{DATA (\S~\ref{subsec:data})}}}, root
[{\rotatebox{90}{\textbf{Collection (\S~\ref{subsubsec: data_collect})}}}, group
[{\datamain{Human Teleoperation}}, sub_1
[{\datasub{Cost-Effective and Scalable Teleoperation}}, sub
[{\dataitem{Young et al.~\cite{young2021visual}, AnyTeleop~\cite{qin2023anyteleop}, GELLO~\cite{wu2024gello}, UMI~\cite{chi2024universal}, OPEN TEACH~\cite{iyer2025open}}}, item]
]
[{\datasub{Feedback-Rich Teleoperation for Contact and Dexterity}}, sub
[{\dataitem{M2R~\cite{kim2023training}, ALPHA-BiACT~\cite{kobayashi2025alpha}, Bunny-VisionPro~\cite{ding2025bunny}, FreeTacMan~\cite{wu2026freetacman}}}, item]
]
[{\datasub{Egocentric and XR Interfaces Aligning Viewpoints}}, sub
[{\dataitem{Zhang et al.~\cite{zhang2018deep}, EgoMimic~\cite{kareer2025egomimic}, ARCap~\cite{chen2025arcap}, AV-ALOHA~\cite{chuang2024active}}}, item]
]
[{\datasub{Embodiment-Aligned and Dexterous Demonstration Interfaces}}, sub
[{\dataitem{AirExo~\cite{fang2024airexo}, DexCap~\cite{wang2024dexcap}, DexUMI~\cite{xu2025dexumi}, Tilde~\cite{si2024tilde}, TeleMoMa~\cite{dass2024telemoma}, TypeTele~\cite{lin2025typetele}}}, item]
]
]
[{\datamain{Human-in-the-Loop Enhancement}}, sub_1
[{\dataitem{IntervenGen~\cite{hoque2024intervengen}, CR-DAgger~\cite{xu2025compliant}}}, sub_new]
]
[{\datamain{Synthetic and Automatic Data Generation}}, sub_1
[{\datasub{Foundation-Model-Guided Generation}}, sub
[{\dataitem{Scaling Up and Distilling Down~\cite{ha2023scaling}, Manipulate-Anything~\cite{duan2025manipulate}, SOAR~\cite{zhou2025autonomous}}}, item]
]
[{\datasub{Demonstration Transformation and Synthesis}}, sub
[{\dataitem{MimicGen~\cite{mandlekar2023mimicgen}, DexMimicGen~\cite{jiangdexmimicgen}, SkillMimicGen~\cite{garrett2025skillmimicgen}, DemoGen~\cite{xue2025demogen}, Lucid-XR~\cite{ravan2025lucidxr}}}, item]
]
[{\datasub{Physics- and Constraint-Aware Generation}}, sub
[{\dataitem{Physics-Driven Data Generation~\cite{yang2025physics}, CP-Gen~\cite{lin2025constraint}, MoMaGen~\cite{li2026momagen}}}, item]
]
[{\datasub{Autonomous and Self-Improving Generation}}, sub
[{\dataitem{RoboCat~\cite{bousmalis2023robocat}, DexFlyWheel~\cite{zhu2025dexflywheel}}}, item]
]
]
[{\datamain{Crowdsourcing-based Data Collection}}, sub_1
[{\dataitem{Chung et al.~\cite{chung2014accelerating}, RoboTurk~\cite{mandlekar2018roboturk}, David et al.~\cite{david2024unpleasantness}, MART~\cite{tung2021learning}, AR2-D2~\cite{duan2023ar2}, EgoZero~\cite{liu2025egozero}, COBALT~\cite{agarwal2026cobalt}}}, sub_new]
]
]
[{\rotatebox{90}{\textbf{Utilization (\S~\ref{subsubsec: data_util})}}}, group
[{\datamain{Data Selection}}, sub_1
[{\dataitem{EAD~\cite{gandhi2023eliciting}, EIL~\cite{zheng2023extraneousness}, DC-IL~\cite{belkhale2023data}, UVP~\cite{dasari2023unbiased}, L2D~\cite{kuhar2023learning}, ILID~\cite{yue2024leverage}, Re-Mix~\cite{hejna2025remix}, MimicLabs~\cite{saxena2025matters}, CUPID~\cite{agia2025cupid}}}, sub_new]
]
[{\datamain{Data Retrieval}}, sub_1
[{\dataitem{VINN~\cite{pari2022surprising}, SAILOR~\cite{nasiriany2023learning}, Behavior Retrieval~\cite{du2023behavior}, Di Palo \& Johns~\cite{di2024effectiveness}, DINOBot~\cite{di2024dinobot}, RAEA~\cite{zhu2024retrieval}, GSR~\cite{yin2024offline}, FlowRetrieval~\cite{lin2025flowretrieval}, STRAP~\cite{memmel2025strap}, COLLAGE~\cite{kumar2025collage}}}, sub_new]
]
[{\datamain{Data Augmentation}}, sub_1
[{\datasub{Language, Goal, and Trajectory Relabeling}}, sub
[{\dataitem{DIAL~\cite{xiao2023robotic}, SPRINT~\cite{zhang2024sprint}, NILS~\cite{blank2025scaling}, DAAG~\cite{di2025diffusion}, S2I~\cite{chen2025towards}, GoalGAIL~\cite{ding2019goal}, OILCA~\cite{sun2023offline}}}, item]
]
[{\datasub{Physics- and Geometry-Consistent Augmentation}}, sub
[{\dataitem{Mitrano \& Berenson~\cite{mitrano2022data}, Wang et al.~\cite{wang2023identifying}, SEIL~\cite{jia2023seil}}}, item]
]
[{\datasub{Generative Visual, Viewpoint, and Embodiment Augmentation}}, sub
[{\dataitem{GenAug~\cite{chen2023genaug}, RoVi-Aug~\cite{chen2025rovi}, RoboPearls~\cite{tang2025robopearls}, ROSIE~\cite{yu2023scaling}, DMD~\cite{zhang2024diffusion}, VISTA~\cite{tian2025view}, DABI~\cite{kobayashi2025dabi}}}, item]
]
]
[{\datamain{Data Expansion}}, sub_1
[{\datasub{Synthesis and Imagined Rollouts}}, sub
[{\dataitem{Generative Predecessor Models~\cite{schroecker2019generative}, SAFARI~\cite{di2020safari}, TASTE-Rob~\cite{zhao2025taste}, Self-Imitation by Planning~\cite{luo2021self}, Category-Level Manipulation~\cite{shen2022learning}}}, item]
]
[{\datasub{Reuse and Corrective Expansion}}, sub
[{\dataitem{Scalable Multi-Task IL~\cite{singh2020scalable}, JUICER~\cite{ankile2024juicer}, Diff-DAgger~\cite{lee2025diff}}}, item]
]
]
[{\datamain{Data Reweighting}}, sub_1
[{\dataitem{Beliaev et al.~\cite{beliaev2022imitation}, FABCO~\cite{takahashi2025feasibility}, PLARE~\cite{luu2025policy}}}, sub_new]
]
]
]
\end{forest}

\caption{A structured taxonomy of data collection and utilization in robot learning.}
\label{fig: data-taxonomy-colored}
\end{figure}

\noindent \textbf{Embodiment-Aligned and Dexterous Demonstration Interfaces.}
A complementary line explicitly addresses kinematic and morphological mismatches between human operators and target robots. AirExo~\cite{fang2024airexo} introduces a low-cost dual-arm exoskeleton that provides embodiment-aligned demonstrations while remaining portable for in-the-wild collection. DexCap~\cite{wang2024dexcap} develops a portable motion-capture system that jointly records wrist and finger motion together with 3D environmental observations, supporting dexterous policy learning from human demonstrations. DexUMI~\cite{xu2025dexumi} further combines wearable human-hand interaction capture with a transfer pipeline for dexterous manipulation, while Tilde~\cite{si2024tilde} uses a paired TeleHand--DeltaHand design for precise in-hand teleoperation. TeleMoMa~\cite{dass2024telemoma} extends teleoperation to whole-body mobile manipulation through a modular multimodal interface. More recently, TypeTele~\cite{lin2025typetele} moves beyond direct human-to-robot pose retargeting by introducing reusable dexterous manipulation types and an MLLM-assisted retrieval mechanism, allowing operators to exploit robot-specific hand capabilities that are difficult to express through direct human-hand imitation. Overall, teleoperation systems are evolving from direct trajectory recording toward portable, multimodal, and embodiment-aware interfaces that jointly improve scalability and demonstration fidelity.

\textbf{ii) Human-in-the-Loop Enhancement}

Human-in-the-loop methods reduce the need for operators to demonstrate complete trajectories by allowing an autonomous policy to execute most of the task while humans selectively provide guidance or correction around difficult states and failure modes. IntervenGen~\cite{hoque2024intervengen} uses a small number of human interventions as seeds and automatically generates additional corrective trajectories around encountered failure states, expanding the state-space coverage of intervention data without requiring proportional human effort. CR-DAgger~\cite{xu2025compliant} further targets contact-rich manipulation through a compliant intervention interface that allows operators to provide continuous delta-action corrections without interrupting policy execution and uses the resulting correction signals to learn a force-aware residual policy. These approaches shift human supervision from exhaustive demonstration toward targeted intervention, concentrating human effort on states where autonomous policies are most prone to failure.

\textbf{iii) Synthetic and Automatic Data Generation}

Human demonstrations remain expensive and inherently limited in coverage. Synthetic and automatic approaches amplify a small amount of supervision into substantially larger and more diverse datasets through foundation-model guidance, demonstration transformation, simulation, constrained optimization, and autonomous policy improvement.

\noindent \textbf{Foundation-Model-Guided Generation.}
Foundation models can provide semantic task specifications and planning priors for generating robot experiences. Scaling Up and Distilling Down~\cite{ha2023scaling} uses language-guided task generation and planning to acquire diverse robot skills and subsequently distills the resulting experience into a multitask visuomotor policy. Manipulate-Anything~\cite{duan2025manipulate} leverages vision--language models to autonomously generate and correct real-world robot behaviors without relying on privileged task information or manually specified skill libraries. SOAR~\cite{zhou2025autonomous} similarly uses foundation models to support improvement of instruction-following skills, illustrating how semantic reasoning can increasingly replace manually designed task-generation pipelines.

\noindent \textbf{Demonstration Transformation and Synthesis.}
Another major direction transforms a small number of source demonstrations into new trajectories under varied task configurations. MimicGen~\cite{mandlekar2023mimicgen} decomposes source demonstrations and adapts their object-relative segments to novel scene configurations, enabling large-scale generation from limited human supervision. DexMimicGen~\cite{jiangdexmimicgen} extends this strategy to coordinated bimanual and dexterous manipulation, while SkillMimicGen~\cite{garrett2025skillmimicgen} organizes generation around reusable manipulation skills. DemoGen~\cite{xue2025demogen} further combines trajectory adaptation with 3D point-cloud editing to transform both actions and observations, enabling synthetic demonstration generation from extremely limited source data. Lucid-XR~\cite{ravan2025lucidxr} complements trajectory-level transformation with an XR-based synthetic data engine that combines on-device physics simulation, human-to-robot retargeting, and physics-guided video generation to amplify virtual interactions into diverse visuomotor training data.

\noindent \textbf{Physics- and Constraint-Aware Generation.}
Pure geometric transformation does not necessarily preserve robot kinematics, contact dynamics, collision constraints, or task feasibility. Physics-Driven Data Generation~\cite{yang2025physics} addresses this by combining embodiment-flexible human demonstrations with kinematic retargeting and trajectory optimization, producing physically consistent contact-rich trajectories across robot embodiments and physical parameters. CP-Gen~\cite{lin2025constraint} represents manipulation skills through keypoint-trajectory constraints and generates demonstrations under novel object poses and geometries while preserving task-relevant geometric relations. MoMaGen~\cite{li2026momagen} extends constraint-aware generation to multi-step bimanual mobile manipulation by satisfying hard constraints such as reachability and collision avoidance while balancing soft constraints associated with visibility during navigation and manipulation. These methods mark a shift from unconstrained demonstration replay toward data generation that preserves physical and task structure.

\noindent \textbf{Autonomous and Self-Improving Generation.}
Recent methods further close the loop between policy learning and data generation, allowing progressively improved policies to generate subsequent training experience. RoboCat~\cite{bousmalis2023robocat} alternates policy adaptation with autonomous experience generation, incorporating newly acquired successful trajectories into subsequent training rounds to progressively expand task coverage. DexFlyWheel~\cite{zhu2025dexflywheel} develops a more explicit data flywheel for dexterous manipulation by integrating imitation learning, residual reinforcement learning, policy rollouts, trajectory filtering, and data augmentation in an iterative cycle. Rather than treating synthetic generation as a one-shot preprocessing step, these systems progressively expand the data distribution as the underlying policy improves.

\begin{figure}[!t]
\centering
\includegraphics[width=\linewidth]{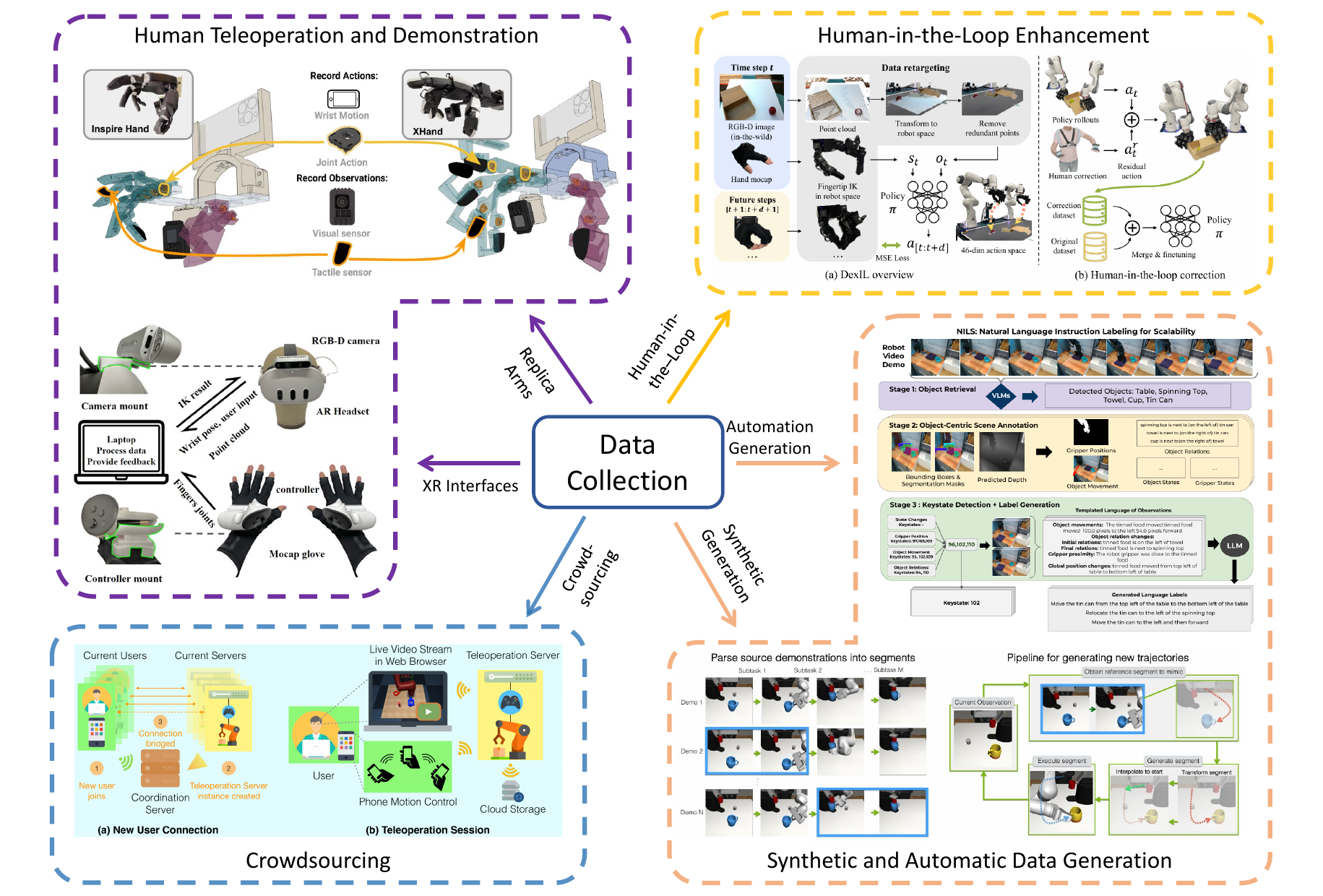}  
\caption{Overview of Data Collection.
DEXUMI~\cite{xu2025dexumi} employs Replica arms, and ARCAP~\cite{chen2025arcap} uses XR Interfaces to represent the data collection methods for Human Teleoperation and Demonstration. DexCap~\cite{wang2024dexcap} stands for Human-in-the-Loop Enhancement, NILS~\cite{blank2025scaling} represents Automatic Data Generation, MimicGen~\cite{mandlekar2023mimicgen} refers to Synthetic Data Generation, and RoboTurk represents Crowdsourcing-based Data Collection.
}
\label{fig: data_collection}
\end{figure}

\textbf{iv) Crowdsourcing-based Data Collection}

Crowdsourcing reduces the cost and expertise barriers of robot demonstration collection by distributing data acquisition across large numbers of contributors. Early work established the feasibility of platform-based crowdsourced imitation learning~\cite{chung2014accelerating}, while RoboTurk~\cite{mandlekar2018roboturk} introduced smartphone-based remote teleoperation with a cloud backend, allowing geographically distributed workers to collect large-scale manipulation demonstrations using commodity devices. Subsequent systems extended crowdsourcing toward real-robot and multi-arm settings~\cite{david2024unpleasantness,tung2021learning}. AR2-D2~\cite{duan2023ar2} reduces dependence on robot hardware during collection by using an iOS augmented-reality interface to capture human--object interactions and scene geometry for subsequent robot retargeting, while EgoZero~\cite{liu2025egozero} explores lightweight smart glasses for distributed in-the-wild human demonstration collection. More recently, COBALT~\cite{agarwal2026cobalt} combines cloud-based infrastructure with concurrent teleoperation and commodity interfaces including smartphones and VR devices, substantially broadening the geographic and hardware accessibility of crowdsourced robot-data acquisition.

Overall, robot data collection is evolving along several complementary axes. Teleoperation systems increasingly improve accessibility, multimodal feedback, and embodiment alignment; human-in-the-loop methods concentrate supervision around informative failures; synthetic generation increasingly preserves physical and task constraints; self-improving systems close the loop between data and policy learning; and crowdsourcing distributes collection across broader populations. These paradigms are complementary rather than mutually exclusive and can be combined within large-scale data engines that jointly balance human effort, data quality, diversity, embodiment fidelity, and scalability.

\subsubsection{Data Utilization}
\label{subsubsec: data_util}

Effectively utilizing existing data is essential for improving robot policy performance, especially under constraints of limited data collection budgets. Recent research has explored strategies for selecting, retrieving, relabeling, augmenting, expanding, and reweighting data so that existing experience contributes more effectively to downstream policy learning.

\textbf{i) Data Selection}

Raw robot datasets often contain noise, redundancy, inconsistent demonstrations, or imbalanced domains. Recent work~\cite{gandhi2023eliciting,zheng2023extraneousness,belkhale2023data,dasari2023unbiased,kuhar2023learning,yue2024leverage,hejna2025remix,saxena2025matters} therefore studies how to select, filter, and adjust training data through trajectory filtering, preference-based pruning, domain-mixture curation, and source selection. EIL~\cite{zheng2023extraneousness} filters extraneous demonstration segments by learning action-conditioned embeddings with temporal cycle consistency and applying an unsupervised voting-based alignment procedure. L2D~\cite{kuhar2023learning} evaluates heterogeneous human demonstrations through latent trajectory representations and preference learning, then selects higher-quality examples for offline imitation learning. Re-Mix~\cite{hejna2025remix} formulates dataset curation as minimax reweighting over domain mixtures using excess behavior-cloning loss, automatically emphasizing domains that improve generalist policy performance. CUPID~\cite{agia2025cupid} further introduces policy-aware data curation by estimating the influence of individual demonstrations on downstream closed-loop policy return, allowing harmful trajectories to be removed and newly collected trajectories to be prioritized according to their expected benefit. Other approaches include EAD~\cite{gandhi2023eliciting}, which elicits demonstrations through compatibility signals; DC-IL~\cite{belkhale2023data}, which characterizes data quality using action divergence and transition diversity; UVP~\cite{dasari2023unbiased}, which highlights the importance of pretraining-data distribution; ILID~\cite{yue2024leverage}, which selects state--action pairs using a state-only discriminator; and MimicLabs~\cite{saxena2025matters}, which emphasizes camera-pose and spatial diversity when constructing training subsets.

\textbf{ii) Data Retrieval}

Retrieval-based approaches address the data bottleneck by mining task-relevant demonstrations or sub-trajectories from large prior corpora, thereby reducing sample complexity and improving transferability~\cite{pari2022surprising,nasiriany2023learning,du2023behavior,di2024effectiveness,di2024dinobot,zhu2024retrieval,yin2024offline,lin2025flowretrieval,memmel2025strap,kumar2025collage}. A first line of work couples representation learning with non-parametric retrieval. VINN~\cite{pari2022surprising} learns a self-supervised visual encoder and performs nearest-neighbor search in latent space, yielding a strong policy without gradient-based fine-tuning. SAILOR~\cite{nasiriany2023learning} organizes prior experience into a latent space of short-horizon skills and retrieves similar sub-trajectories as reusable motion primitives, accelerating few-shot adaptation. Behavior Retrieval~\cite{du2023behavior} introduces a variational encoder to score state--action similarity, seeding retrieval with limited expert rollouts and jointly training on expert and retrieved data.

More recent methods emphasize adaptation and integration of retrieved experience. Di Palo and Johns~\cite{di2024effectiveness} retrieve and replay demonstrations through a decide--align--execute framework, while DINOBot~\cite{di2024dinobot} adapts demonstrations to novel objects through pixel-level alignment using DINO-ViT features. RAEA~\cite{zhu2024retrieval} integrates trajectories retrieved from multimodal memory directly into action prediction, and GSR~\cite{yin2024offline} organizes prior interactions as a graph over pretrained embeddings to identify high-value behaviors through search. FlowRetrieval~\cite{lin2025flowretrieval} leverages optical flow to mine cross-task motion segments, STRAP~\cite{memmel2025strap} retrieves sub-trajectories using visual foundation models and time-invariant alignment, and COLLAGE~\cite{kumar2025collage} combines multiple similarity signals with adaptive weighting to curate task-relevant subsets. Together, these methods show that retrieval can serve as both an alternative and a complement to parametric policy learning, enabling efficient reuse of large heterogeneous experience corpora.

\textbf{iii) Data Augmentation}

Data augmentation increases the effective diversity of existing datasets without requiring proportional collection effort. We organize existing approaches into three families: language and trajectory relabeling, physics- and geometry-consistent augmentation, and generative visual, viewpoint, and embodiment augmentation.

\noindent \textbf{Language, Goal, and Trajectory Relabeling.}
This family reinterprets existing demonstrations by modifying their semantic labels, goals, temporal decomposition, or trajectory assignments. DIAL~\cite{xiao2023robotic} automatically generates language instructions for unlabeled demonstrations using pretrained vision--language models, while SPRINT~\cite{zhang2024sprint} relabels and composes language instructions across existing trajectories to support long-horizon skill chaining. NILS~\cite{blank2025scaling} performs zero-shot natural-language annotation of long-horizon robot videos using foundation models. Beyond language labels, DAAG~\cite{di2025diffusion} generates temporally and geometrically consistent trajectory relabelings using diffusion models, and S2I~\cite{chen2025towards} segments demonstrations, selects useful segments, and optimizes trajectories to better exploit mixed-quality data. Related efforts include goal relabeling in GoalGAIL~\cite{ding2019goal} and counterfactual sample generation through variational reasoning in OILCA~\cite{sun2023offline}.

\noindent \textbf{Physics- and Geometry-Consistent Augmentation.}
This family transforms existing trajectories while preserving physical or geometric validity. Mitrano and Berenson~\cite{mitrano2022data} formulate augmentation as an optimization problem that applies rigid-body transformations while maintaining manipulation constraints. Wang et al.~\cite{wang2023identifying} identify high-quality rollouts and exploit environmental symmetries to construct principled augmentations. SEIL~\cite{jia2023seil} similarly exploits task symmetries to combine expert demonstrations with simulation-augmented equivariant transitions, improving data efficiency while preserving task structure.

\noindent \textbf{Generative Visual, Viewpoint, and Embodiment Augmentation.}
Generative models provide another mechanism for increasing perceptual and embodiment diversity while preserving the underlying behavior. GenAug~\cite{chen2023genaug} applies pretrained generative models to perform semantic scene edits that preserve demonstrated actions. RoVi-Aug~\cite{chen2025rovi} uses diffusion-based image translation to synthesize demonstrations across robot embodiments and camera viewpoints, enabling transfer to unseen configurations. RoboPearls~\cite{tang2025robopearls} leverages editable 3D Gaussian Splatting reconstructions for language-guided demonstration synthesis. Additional approaches include text-guided visual modification in ROSIE~\cite{yu2023scaling}, novel-view generation in DMD~\cite{zhang2024diffusion}, single-image 3D-aware viewpoint synthesis in VISTA~\cite{tian2025view}, and multi-rate sensory alignment in DABI~\cite{kobayashi2025dabi}.

\textbf{iv) Data Expansion}

Data expansion goes beyond modifying individual training samples and instead increases the amount of usable experience by synthesizing, recombining, or selectively collecting additional trajectories. Existing approaches can be broadly divided into synthesis and imagined rollouts, and reuse and corrective expansion.

\noindent \textbf{Synthesis and Imagined Rollouts.}
Generative Predecessor Models~\cite{schroecker2019generative} learn distributions over predecessor states and synthesize state--action samples that lead into expert trajectories, densifying the space of successful behaviors. Self-Imitation by Planning~\cite{luo2021self} uses planners to generate improved rollouts for existing tasks and iteratively appends them to the dataset, creating an improve--add--imitate loop. Category-Level Manipulation~\cite{shen2022learning} applies adversarial self-imitation to generate additional category-level trajectories, improving generalization across object instances. SAFARI~\cite{di2020safari} generates imagined rollouts under safety and corrective constraints, while TASTE-Rob~\cite{zhao2025taste} produces task-oriented hand--object interaction videos that provide additional visual and interaction supervision.

\noindent \textbf{Reuse and Corrective Expansion.}
Scalable Multi-Task Imitation Learning~\cite{singh2020scalable} relabels collected trajectories across tasks so that individual demonstrations provide supervision for multiple goals. JUICER~\cite{ankile2024juicer} decomposes demonstrations into reusable motion segments and recomposes them into new task sequences, enabling combinatorial expansion from a limited skill library. Diff-DAgger~\cite{lee2025diff} estimates policy uncertainty with diffusion models and selectively acquires corrective demonstrations in uncertain regions, concentrating additional supervision around likely failures.

\textbf{v) Data Reweighting}

Data reweighting assigns different importance to demonstrations according to quality, feasibility, expertise, or preference signals rather than treating all data uniformly. Beliaev et al.~\cite{beliaev2022imitation} estimate individual demonstrator expertise and use the resulting estimates to weight demonstrations during imitation learning. FABCO~\cite{takahashi2025feasibility} evaluates demonstration feasibility using robot dynamics and assigns higher training weight to trajectories that are more compatible with the target embodiment. PLARE~\cite{luu2025policy} instead queries a large vision--language model for preference labels over trajectory segments and optimizes the policy directly from these pairwise preferences using a contrastive objective. Together, these approaches show that weighting data according to expertise, feasibility, or preference can improve policy learning under heterogeneous and imperfect demonstrations.

\subsection{Generalization}
\label{subsec:generalization_tasks}

Robotic manipulation generalization can be broadly categorized into three dimensions: \textbf{environment generalization}, \textbf{task generalization}, and \textbf{cross-embodiment generalization}. Environment generalization concerns robustness to variations in conditions such as Sim2Real transfer, spatial transformations, lighting, and background or distractor changes. Task generalization emphasizes maintaining performance across different task configurations, including long-horizon tasks, few-shot or meta-learning, continual learning, and skill composition. Cross-embodiment generalization focuses on transferring skills across robots with diverse morphologies, kinematics, dynamics, or sensing modalities, which is crucial for building general-purpose embodied agents. In what follows, we structure our survey along these three perspectives.

\begin{figure}[!t]
\centering
\includegraphics[width=1.\linewidth]{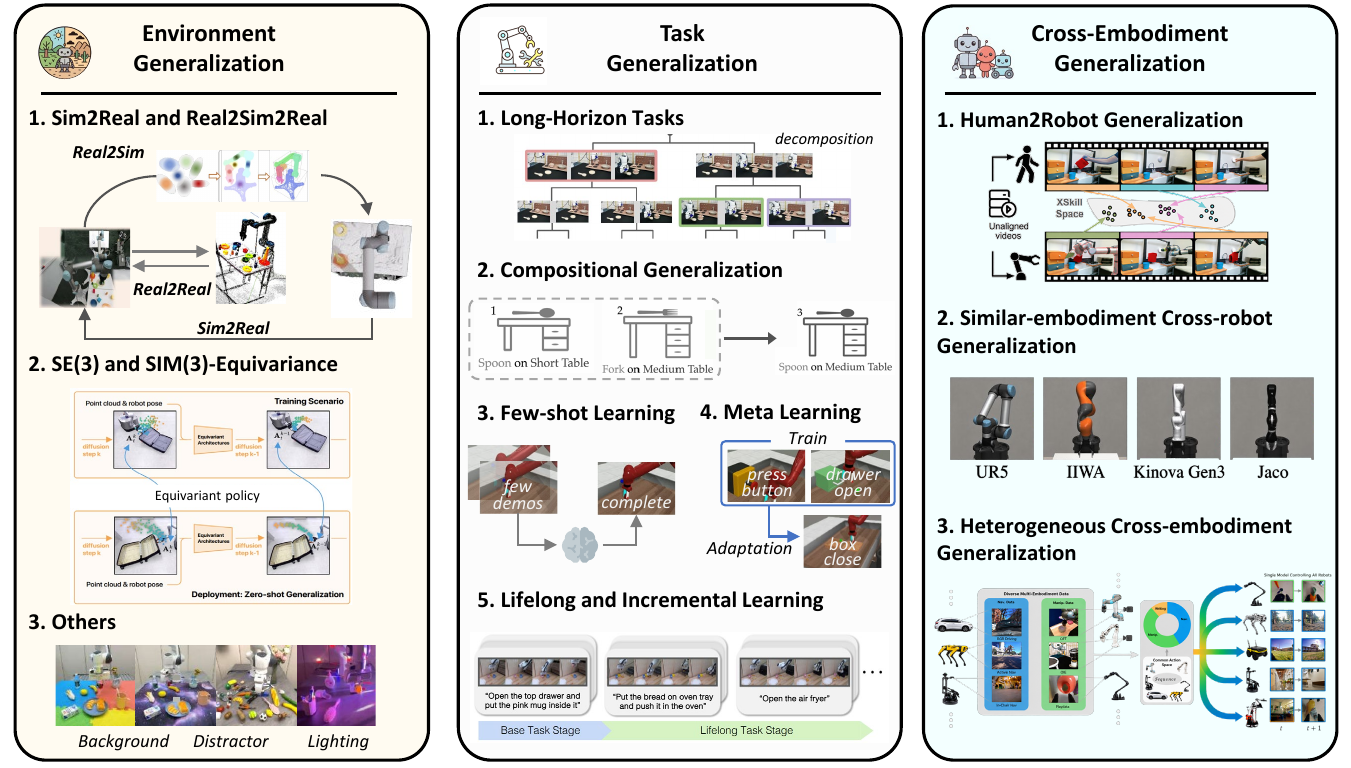}  
\caption{Overview of generalization. The figure is adapted from representative studies in environment generalization~\cite{li2024robogsim, yang2025equibot, deng2025graspvla}, task generalization~\cite{gao2024efficient, wan2024lotus}, and cross-embodiment generalization~\cite{xu2023xskill, yang2024pushing, chen2024mirage}.
}   
\label{fig: generalization}
\end{figure}

\subsubsection{Environment Generalization}

Environment generalization in robotic manipulation refers to the ability of policies trained under specific conditions to maintain performance across varying environments, scenes, viewpoints, and physical settings. Such generalization is challenged by discrepancies between simulation and reality, spatial transformations of objects and cameras, and visual appearance changes such as lighting, backgrounds, and distractors. Existing approaches address these challenges from several complementary perspectives, including sim-to-real and real-to-sim-to-real transfer, geometric equivariance and viewpoint robustness, and robustness to other visual environmental variations.

\textbf{i) Sim2Real and Real2Sim2Real Generalization}

Simulation offers a scalable alternative for training and evaluating manipulation policies, alleviating the cost, operational risk, and safety concerns associated with large-scale real-world data collection. However, discrepancies in dynamics, sensing, visual appearance, and environmental complexity often result in poor transferability, commonly referred to as the sim-to-real gap~\cite{lyu2024scissorbot}. Existing approaches can be broadly organized into three paradigms according to how simulated and real-world experience are combined: simulation-only training, sim--real adaptation and co-training, and real2sim2real reconstruction.

\noindent \textbf{Simulation-Only Training.}
This paradigm assumes that sufficient variability introduced during simulation training can enable policies to transfer directly to the real world without further adaptation. Domain randomization is a representative strategy, where physical parameters such as mass, friction, and damping~\cite{chen2021understanding}, as well as perceptual factors such as textures~\cite{tobin2017domain}, lighting~\cite{tremblay2018training}, and object poses~\cite{ren2019domain}, are systematically perturbed to prevent policies from overfitting to a specific simulator configuration. Peng et al.~\cite{peng2018sim} demonstrate that randomized dynamics can substantially improve real-world transfer, while more recent systems such as DexScale~\cite{liu2025dexscale} scale simulation diversity and augmentation to increasingly complex manipulation settings. Nevertheless, zero-shot transfer remains difficult when simulated variability fails to capture fine-grained visual, contact, or dynamic properties of real environments.

\noindent \textbf{Sim--Real Adaptation and Co-Training.}
Rather than relying exclusively on simulation, another line of work introduces a limited amount of real-world experience to anchor policy learning. Earlier approaches primarily adapt simulation-trained policies through real-world fine-tuning or continual learning; for example, Josifovski et al.~\cite{josifovski2025safe} study safe continual adaptation under deployment constraints. Natural Language Can Help Bridge the Sim2Real Gap~\cite{yu2024natural} instead uses language descriptions as a shared semantic signal across simulated and real observations, encouraging domain-invariant representations while training with abundant simulated and sparse real demonstrations. More recently, Sim-and-Real Co-Training~\cite{maddukuri2025sim} systematically demonstrates that directly training a shared policy on mixed simulated and real datasets can substantially improve real-world manipulation performance even when noticeable visual and behavioral discrepancies remain between domains. Generalizable Domain Adaptation~\cite{cheng2025generalizable} further aligns the joint distributions of observations and actions across simulation and reality through optimal-transport-based objectives, learning task-relevant domain-invariant representations and enabling simulated experience to improve real-world performance even in scenarios not represented by the limited real demonstrations. These developments shift sim-to-real learning from post-hoc policy correction toward joint exploitation of large-scale simulation and sparse real-world supervision.

\noindent \textbf{Real2Sim2Real Reconstruction.}
A complementary paradigm closes the loop by reconstructing simulatable or renderable environments from real-world observations and subsequently using these reconstructions for policy learning~\cite{torne2024reconciling,yu2025real2render2real}. RialTo~\cite{torne2024reconciling} constructs task-specific digital twins from limited real-world sensing and introduces inverse distillation to transfer real demonstrations into simulation, where reinforcement learning can safely improve policy robustness before deployment back in the real world. Real2Render2Real~\cite{yu2025real2render2real} reduces dependence on both dynamics simulation and physical robot collection by reconstructing object geometry and appearance from smartphone scans and tracking object motion from a single human demonstration, enabling large-scale rendering of robot-compatible training trajectories. Real2Edit2Real~\cite{zhao2026real2edit2real} further introduces a metric 3D control interface that edits reconstructed scenes and manipulation trajectories and uses depth-conditioned video generation to synthesize geometrically consistent demonstrations under new spatial configurations. These approaches illustrate an emerging transition from manually constructed simulation environments toward real-world-grounded digital reconstruction and editable synthetic environments for scalable policy generalization.

\textbf{ii) SE(3) and SIM(3)-Equivariance Generalization}

A complementary strategy for environment generalization is to explicitly encode geometric symmetry into policy architectures. SE(3)- and SIM(3)-equivariant models constrain representations or actions to transform predictably under changes in translation, rotation, and, for SIM(3), scale, thereby reducing the need to independently observe every spatial configuration during training. EquiBot~\cite{yang2025equibot} introduces a SIM(3)-equivariant diffusion policy that improves data efficiency and generalization across changes in object pose and scale. ET-SEED~\cite{tie2025seed} develops an efficient trajectory-level SE(3)-equivariant diffusion policy and relaxes the requirement that every transition in the diffusion process must independently employ expensive equivariant computation. More recently, E3Flow~\cite{zhang2026efficient} combines spherical-harmonic-based equivariant representations with rectified flow and multimodal visual features, extending geometric equivariance toward more efficient flow-based visuomotor policy learning. Together, these methods demonstrate how explicit geometric structure can provide strong inductive biases for spatial generalization.

\noindent \textbf{Viewpoint Robustness Beyond Explicit Equivariance.}
Related approaches address viewpoint changes without requiring the policy to satisfy explicit SE(3) or SIM(3) equivariance. Adapt3R~\cite{wilcox2025adapt3r} combines pretrained semantic visual features with end-effector-relative 3D localization, enabling imitation policies to transfer across unseen camera viewpoints and robot embodiments. Another direction actively changes the observation process itself rather than passively requiring the policy to tolerate arbitrary viewpoints. Vision in Action~\cite{xiong2025vision} learns task-dependent searching, tracking, and focusing behaviors from human demonstrations using an actuated robotic neck, allowing the robot to recover informative viewpoints when objects become occluded. SaPaVe~\cite{liu2026sapave} extends active perception to VLA-based manipulation by decoupling camera motion from manipulation actions and coordinating semantic camera control with geometry-aware execution. These methods complement explicit equivariance by either constructing viewpoint-robust spatial representations or actively controlling perception to maintain informative observations during manipulation.

\textbf{iii) Others}

Lighting, background, and distractor variations constitute another major source of environment shift, as visuomotor policies can overfit to visual cues that correlate with demonstrations but are irrelevant to task execution. Existing approaches address these variations through augmentation, invariant representation learning, and training across heterogeneous visual contexts. Physically based augmentation methods explicitly vary illumination during training to improve robustness under lighting changes~\cite{jin2025physically}. Decomposing the Generalization Gap~\cite{xie2024decomposing} systematically analyzes how individual visual factors contribute to imitation-learning generalization failures, highlighting the importance of separating task-relevant information from nuisance variations. Maniwhere~\cite{yuan2024learning} combines multi-view representation learning with curriculum-based randomization and augmentation to improve robustness across combinations of visual disturbances and support sim-to-real transfer. Other approaches reduce dependence on incidental scene context through position-invariant regularization~\cite{yin2023spatial} or by learning from demonstrations collected under changing contexts~\cite{yang2020cross,qian2023robot}. At a larger scale, Diffusion-VLA~\cite{wen2025diffusion} demonstrates that large multimodal foundation policies can retain manipulation performance under unseen objects, distractors, and new backgrounds, suggesting that broad pretraining can complement explicit augmentation and invariance mechanisms. Collectively, these methods aim to prevent policies from relying on spurious visual correlations and preserve task-relevant behavior under substantial changes in environmental appearance.

\subsubsection{Task Generalization}

Task generalization in robotic manipulation refers to the ability of learned policies to adapt to new, complex, or unseen tasks without extensive retraining. It encompasses the generalization of learned skills, structures, and adaptation mechanisms across variations in task composition, semantics, and temporal scope. Robust task generalization requires policies not only to execute individual skills but also to compose, transfer, and refine them under new configurations and objectives.

\textbf{i) Generalization for Long-horizon Tasks}

Long-horizon robotic manipulation represents a core challenge for embodied intelligence. Unlike short-horizon tasks that typically involve single-skill execution, long-horizon problems demand the coordination of multiple sub-skills, hierarchical reasoning, and temporally abstract decision-making. Robust generalization requires agents not only to plan and execute extended action sequences but also to adapt across variations in task structure, object arrangement, and dynamic environments. Recent advances approach this challenge along three interconnected axes: skill compositionality, semantic task decomposition, and structure-aware representation learning.

\noindent \textbf{Skill Compositionality.}
At the foundation lies the ability to compose and reuse modular skills~\cite{chenu2022divide,lee2021adversarial,zhu2022bottom,belkhale2023hydra,agia2022stap,zhang2023bootstrap,liu2024learning}. STAP~\cite{agia2022stap} addresses sequencing by optimizing the feasibility of action chains. Generative frameworks such as BOSS~\cite{zhang2023bootstrap} and BLADE~\cite{liu2024learning} leverage large language models to autonomously expand skill libraries and incorporate semantic grounding into action spaces, thereby enabling flexible skill reuse and extension.

\noindent \textbf{Semantic Task Decomposition.}
Beyond static skill composition, multimodal semantics have been introduced to parse and decompose tasks into modular components~\cite{mu2025look, myers2024policy, dalal2025local, lin2024hierarchical, sundaresan2025s, mao2024dexskills}. PALO~\cite{myers2024policy} employs vision-language models to translate high-level task descriptions into reusable sub-tasks, enabling rapid adaptation with minimal supervision. ManipGen~\cite{dalal2025local} integrates local policies that encode invariances in pose, skill order, and scene layout, combining them with foundational models in vision and motion planning to generalize across unseen long-horizon sequences.

\noindent \textbf{Structure-aware Representation Learning.}
A finer level of generalization is achieved through task-level abstraction and representation learning~\cite{triantafyllidis2023hybrid, tanwani2021sequential, wu2020squirl, chen2025backbone, chen2025robohorizon, yang2025bootstrapping}. TBBF~\cite{chen2025backbone} decomposes complex tasks into primitive therbligs, facilitating efficient action-object mappings and trajectory synthesis. RoboHorizon~\cite{chen2025robohorizon} formalizes a Recognize-Sense-Plan-Act pipeline by fusing LLMs with multi-view world models, addressing sparse reward supervision and perceptual complexity. HD-Space~\cite{yang2025bootstrapping} identifies atomic sub-task boundaries from demonstrations, improving sample efficiency and policy robustness with limited data.

\textbf{ii) Compositional Generalization}

Compositional generalization in robotic manipulation emphasizes the ability to solve novel tasks by systematically recombining known skills, objects, and instructions. Recent research explores diverse strategies to achieve this capability. One line of work focuses on data-driven efficiency, where targeted data collection strategies are designed to maximize coverage of compositional variations with minimal demonstrations, enabling scalable policy training~\cite{gao2024efficient}. Another approach grounds policies in programmatic or symbolic structures, allowing robots to leverage modular task representations for systematic recomposition and generalization to unseen task combinations~\cite{wang2023programmatically}. Complementary efforts investigate policy architectures that explicitly factorize control into entities or modules, facilitating the recombination of learned components to improve adaptability in complex environments~\cite{zhou2022policy}. Together, these methods illustrate a growing effort to move beyond rote memorization of tasks toward systematic generalization, enabling robots to flexibly adapt to combinatorial task spaces.

\textbf{iii) Few-shot Learning}

Few-shot learning enables robots to acquire skills from only a handful of demonstrations~\cite{chen2025vidbot, valassakis2022demonstrate, jain2024cobt, heppert2024ditto, li2025elastic}, alleviating the high cost of large-scale data collection. Research in this area can be broadly grouped into two directions: embedding structured priors and leveraging semantic transfer. 
One line of work embeds structured inductive biases into perception and control to maximize the utility of limited demonstrations. Domain-invariant constraints, including spatial equivariance, 3D geometry, or action continuity, are incorporated to reduce data requirements and improve generalization. For example, Ren et al.~\cite{ren2024learning} combine point cloud representations with a diffusion-based policy, achieving robust generalization from as few as ten demonstrations.
A complementary line of research emphasizes semantic transfer and compositional generalization. These methods extract transferable knowledge from alternative sources, such as human demonstrations or previously solved tasks, and adapt it to new objectives. For instance, YOTO~\cite{zhou2025you} maps human hand motions from video to dual-arm robotic skills, while TOPIC~\cite{song2025few} constructs task prompts and a dynamic relation graph to systematically reuse prior experience for new policy learning.

\textbf{iv) Meta Learning}

While closely related to few-shot learning, meta learning distinguishes itself by focusing on the ability to \emph{amortize} task adaptation. Instead of relying directly on structural priors or semantic transfer to generalize from a handful of demonstrations, meta learning trains over a distribution of tasks so that the model acquires a generic adaptation procedure. This paradigm equips robots with rapid adaptability to unseen tasks, even under extremely limited supervision~\cite{duan2017one,james2018task,di2023one,zhang2024one,bing2023meta,haldar2023teach,biza2023one}.

\noindent \textbf{Task-Conditioned Meta-Representations.}
One direction focuses on learning task-conditioned embeddings that serve as meta-representations. Early approaches such as Duan et al.~\cite{duan2017one} and TecNets~\cite{james2018task} encode demonstrations into low-dimensional vectors for policy conditioning. Later works extend this idea with relational graph networks~\cite{di2023one} or invariance-based objectives~\cite{zhang2024one}, enforcing structural consistency across tasks. These embedding-based approaches highlight the importance of compact task representations in accelerating policy adaptation.

\noindent \textbf{Cross-Modal and Instruction-Driven Adaptation.}
Another direction emphasizes the integration of additional modalities and task instructions into the adaptation process. MILLION~\cite{bing2023meta} incorporates natural language into the meta-learning loop, enabling semantically guided adaptation. FISH~\cite{haldar2023teach} demonstrates versatile skill acquisition from just one minute of demonstrations, while interaction-warping~\cite{biza2023one} aligns novel trajectories with past interaction structures to facilitate transfer. Collectively, these works establish meta learning as a principled framework for scalable and data-efficient generalization.

\textbf{v) Lifelong, Continual and Incremental Learning}

Endowing robots with the ability to continuously acquire new capabilities is a key step toward adaptive and autonomous embodied agents. Unlike conventional task generalization, which typically evaluates transfer to a fixed set of unseen tasks, lifelong and incremental learning consider a non-stationary setting in which new tasks or skills arrive sequentially. The central challenge is therefore twofold: newly acquired knowledge should benefit future learning, while previously learned capabilities should be retained without repeated full retraining. In robotic manipulation, existing approaches can be broadly distinguished into lifelong or continual learning, which emphasizes knowledge retention and transfer across a sequence of tasks, and skill-incremental learning, which focuses more explicitly on expanding and updating an evolving repertoire of manipulation skills.

\noindent \textbf{Lifelong and Continual Learning.}
Lifelong and continual learning aims to retain and reuse knowledge as robots encounter a sequence of new manipulation tasks~\cite{yao2025think,roy2025m2distill,gao2021cril,lei2026dynamic}. A major direction is to organize manipulation knowledge into reusable modular components. LOTUS~\cite{wan2024lotus} continually discovers recurring skills from incoming task demonstrations, incrementally expands or updates a skill library, and composes these skills through a hierarchical meta-controller, enabling forward and backward transfer across sequential tasks. PPL~\cite{yao2025think} instead represents shared motion primitives as reusable and extensible prompts; pretrained primitive prompts are preserved while new prompts are introduced for incoming skills, allowing previous knowledge to facilitate subsequent learning without overwriting existing capabilities. Complementary approaches focus more directly on preserving learned representations. M2Distill~\cite{roy2025m2distill} employs multimodal knowledge distillation to maintain consistent representations across sequential tasks and mitigate catastrophic forgetting, while CRIL~\cite{gao2021cril} uses generative replay to preserve experience from previously learned behaviors. Together, these approaches show that continual robot learning increasingly combines modular skill reuse with explicit mechanisms for knowledge preservation, balancing forward transfer to new tasks against retention of previously acquired capabilities.

\noindent \textbf{Incremental Learning.}
In contrast, skill-incremental learning focuses more explicitly on settings in which robots progressively acquire or refine atomic manipulation skills while preserving their existing skill repertoire. iManip~\cite{zheng2025imanip} formalizes this setting with a skill-incremental benchmark built on RLBench~\cite{james2020rlbench} and introduces temporal replay together with an Extendable PerceiverIO, whose expandable action prompts accommodate new action primitives while mitigating catastrophic forgetting of previously learned skills. SIL-C~\cite{lee2025policy} further reveals that incremental skill acquisition introduces an additional challenge beyond forgetting: as the underlying skill library evolves, previously learned high-level policies may become incompatible with newly added or updated skills. To address this issue, SIL-C introduces a lazy-learning interface that dynamically aligns policy-level subtasks with the evolving skill space according to trajectory-distribution similarity, allowing new skills to be incorporated while maintaining compatibility with existing downstream policies. These approaches highlight two complementary aspects of skill-incremental manipulation: extending the policy representation to accommodate new primitives and maintaining a stable interface between an evolving skill library and previously learned task policies.

\subsubsection{Cross-Embodiment Generalization}

Cross-embodiment generalization refers to the ability to transfer manipulation knowledge across agents or robotic platforms with different visual appearances, kinematics, action spaces, morphologies, or sensing configurations. Rather than categorizing existing approaches according to specific algorithmic mechanisms, we organize this direction according to the relationship between the source and target embodiments. Specifically, existing studies can be broadly divided into human-to-robot generalization, cross-robot transfer between functionally similar embodiments, and heterogeneous cross-embodiment generalization across substantially different robot morphologies and action spaces.

\textbf{i) Human-to-Robot Generalization}

Human demonstrations provide a scalable source of manipulation experience, but transferring them to robots requires bridging substantial differences in appearance, sensing, morphology, and action spaces. Early approaches therefore seek embodiment-invariant representations or transferable intermediate signals that preserve task semantics while suppressing embodiment-specific motion details. XSkill~\cite{xu2023xskill} discovers shared skill prototypes from unlabeled human and robot manipulation videos and maps them to robot actions through a skill-conditioned diffusion policy. Human2Sim2Robot~\cite{lum2025crossing} instead extracts object trajectories and hand configurations from a single human RGB-D demonstration and converts them into embodiment-agnostic rewards and initialization priors for reinforcement learning in simulation. X-Sim~\cite{dan2025x} similarly avoids direct human-to-robot action mapping by reconstructing photorealistic simulation from human videos and using object motion as a transferable learning signal before distilling the resulting behavior into a real-world policy.

More recent methods explicitly align human and robot representations during policy learning. UniSkill~\cite{kim2025uniskill} learns embodiment-agnostic skill representations from large-scale unlabeled cross-embodiment videos, allowing skills extracted from human video prompts to guide robot policies trained with robot actions. ImMimic~\cite{liu2025immimic} maps retargeted human trajectories to robot trajectories and constructs intermediate domains through trajectory interpolation, reducing visual, morphological, and physical discrepancies during co-training. EgoBridge~\cite{punamiya2025egobridge} further formulates human-to-robot transfer as domain adaptation and aligns joint latent feature--action distributions using optimal transport guided by trajectory similarity. Together, these approaches illustrate a progression from indirect motion and object-centric transfer toward explicit human--robot representation alignment.

\textbf{ii) Similar-Embodiment Cross-Robot Generalization}

A second setting considers transfer between robot platforms that retain similar functional structures, such as serial manipulators equipped with comparable end effectors. Although these robots may differ in appearance, degrees of freedom, kinematics, or control dynamics, substantial policy knowledge can often be reused without learning entirely new behavioral abstractions. Mirage~\cite{chen2024mirage} enables zero-shot transfer between different robot arms and grippers by separately correcting perceptual and control discrepancies through cross-painting, state alignment, and dynamics compensation. Wang et al.~\cite{wang2024cross} instead project heterogeneous state and action spaces into a shared latent control space, enabling transfer across manipulators with mismatched kinematics. Modularity through Attention~\cite{zhou2023modularity} takes a complementary approach by decomposing language-conditioned manipulation policies into reusable functional components and robot-specific modules, so that only embodiment-dependent components need to be adapted. These studies show that when source and target robots remain functionally similar, transfer can often be achieved through selective alignment of perception, control, or modular policy components.

\textbf{iii) Heterogeneous Cross-Embodiment Generalization}

The most challenging setting considers substantially heterogeneous embodiments whose observation spaces, action dimensions, kinematic structures, or functional capabilities may differ significantly. One direction addresses this problem through large-scale heterogeneous co-training and shared policy representations. Pushing the Limits of Cross-Embodiment Learning~\cite{yang2024pushing} jointly learns from manipulation and navigation data spanning robotic arms, mobile bases, quadrupeds, and aerial robots, demonstrating positive transfer across markedly different embodiments. CrossFormer~\cite{doshi2024scaling} further represents heterogeneous observations and actions as variable-length token sequences processed by a shared Transformer, supporting manipulation, navigation, locomotion, and aviation within a unified policy. HPT~\cite{wang2024scaling} adopts a modular alternative by combining embodiment-specific input stems and output heads with a shared Transformer trunk, allowing heterogeneous visual and proprioceptive data to contribute to common policy pretraining.

A complementary direction constructs explicit structural, functional, or system-level bridges between different morphologies. AnyBimanual~\cite{lu2025anybimanual} transfers pretrained unimanual skills to bimanual manipulation by composing reusable skill representations and aligning arm-specific visual observations. LEGATO~\cite{seo2025legato} introduces a shared grasping tool that provides a common interaction interface across embodiments, allowing manipulation behaviors to be retargeted through embodiment-specific inverse kinematics. CEI~\cite{wu2026cei} explicitly models functional similarity between different robot arms and end effectors and aligns trajectories in 3D space, including transfer between parallel-jaw grippers and dexterous hands. At a higher abstraction level, RoboOS~\cite{tan2025roboos} organizes heterogeneous robot capabilities through standardized interfaces for perception, control, communication, and skill execution, enabling reusable high-level skills to operate across different robotic platforms. Together, these methods show that heterogeneous cross-embodiment transfer can be supported through shared policies, functional correspondence, common physical interfaces, and system-level abstraction.

\subsection{Agent}
\label{subsec:agent}

Recent advances in foundation models have motivated agent-based robotic manipulation systems that augment learned policies with higher-level reasoning, tool use, memory, verification, and adaptation. Rather than replacing low-level manipulation policies or VLA models, agents typically orchestrate these components within closed perception--reasoning--action loops, enabling long-horizon task decomposition, execution monitoring, failure recovery, and continual improvement. This paradigm extends language-based robotic planning toward persistent embodied systems that can adapt their behavior during deployment.

\noindent \textbf{Planning and Tool Orchestration.}
Early language-grounded systems establish the foundation for agent-based manipulation by connecting high-level reasoning with executable robot skills. SayCan~\cite{brohan2023can} grounds language-model planning in learned skill affordances, Code as Policies~\cite{liang2023code} synthesizes executable robot programs, Inner Monologue~\cite{huang2022inner} incorporates environmental feedback for closed-loop replanning, and VoxPoser~\cite{huang2023voxposer} grounds language instructions into spatial value maps. Recent agentic systems increasingly orchestrate learned policies and external tools. Agentic Robot~\cite{yang2025agentic} coordinates reasoning, VLA execution, and temporal verification for long-horizon manipulation, while VLA$^2$~\cite{zhao2025vla2} invokes external perception and retrieval tools to improve manipulation of unseen concepts. Guava~\cite{liu2026guava} studies agent workflows, action abstractions, and multimodal observations for embodied tool use, whereas Harness VLA~\cite{zhang2026harness} combines a frozen VLA with analytic primitives and execution memory to improve robustness beyond its original task distribution.

\noindent \textbf{Memory and Runtime Adaptation.}
Memory enables agents to accumulate execution experience and adapt without repeatedly modifying the underlying policy. SOMA~\cite{li2026soma} augments frozen VLAs with dual-memory retrieval, failure attribution, and agent-driven interventions for in-context adaptation. Long-Term Memory for VLA-based Agents~\cite{huang2026long} similarly combines hierarchical planning with persistent trajectory memory for long-horizon execution. Agentic-VLA~\cite{jin2026agentic} further integrates experience memory with adaptive reward synthesis and language-guided exploration for efficient online adaptation. These methods shift robot deployment from isolated policy execution toward persistent agents that reuse prior successes and failures to improve subsequent decisions.

\noindent \textbf{Policy Synthesis and Self-Improvement.}
Beyond runtime adaptation, recent agents increasingly modify the programs, data, or policies that govern robot behavior. ARCHITECT~\cite{chen2026few} formulates robot policy acquisition as interactive program synthesis, using language corrections and execution traces to build reusable skill libraries. ENPIRE~\cite{xiao2026enpire} establishes an autonomous real-world improvement loop in which agents execute, evaluate, diagnose, and improve robot policies, while RoboClaw~\cite{li2026roboclaw} unifies autonomous data collection, policy refinement, and long-horizon deployment through self-resetting manipulation loops and policy orchestration. Together, these approaches extend agents from policy consumers toward active participants in robot learning and improvement.

\noindent \textbf{Systemization and Multi-Agent Coordination.}
As agentic manipulation becomes more complex, recent work increasingly treats orchestration, verification, memory, and safety as system-level capabilities. PhyAgentOS~\cite{liu2026phyagentos} provides a runtime abstraction that decouples cognitive planning from physical execution while exposing scheduling, semantic verification, persistent memory, benchmarking, and safety as shared services across embodiments. Multi-agent approaches further distribute these responsibilities across specialized components. A Closed-Loop Multi-Agent Framework~\cite{he2026closed}, for example, separates planning, manipulation, and verification among dedicated agents and uses execution feedback to coordinate multiple robots during long-horizon manipulation. These developments suggest a transition from monolithic robot policies toward modular agent systems in which reasoning models, learned controllers, tools, memories, and physical embodiments interact through explicit closed-loop interfaces.

\subsection{Human--Robot Interaction and Collaboration}
\label{subsec:hri_collaboration}

Real-world manipulation increasingly requires robots to operate with humans rather than as isolated autonomous agents. This introduces challenges beyond conventional policy generalization: human goals may be implicit or change during execution, feedback may arrive through language, gaze, motion, or physical contact, and the appropriate division of autonomy may vary with uncertainty and user preference. Recent approaches therefore increasingly treat interaction itself as an information channel through which robots infer intent, receive corrections, align behavior, and coordinate collaborative actions.

\noindent \textbf{Human Intent Understanding and Proactive Assistance.}
A central challenge in human--robot collaboration is inferring human intent early enough for the robot to provide timely assistance. Pérez-D'Arpino and Shah~\cite{perez2015fast} predict human reaching targets from partial motion observations, enabling robots to anticipate intended objects during cooperative manipulation. More recent approaches exploit richer interaction signals: EDITH~\cite{lee2026hierarchical} combines speech with egocentric vision and gaze to infer human intent and generate grounded subtasks, while Sticky-Glance~\cite{lai2026sticky} stabilizes gaze-based object intent for continuous interaction. Physical feedback provides another channel, with TATIC~\cite{song2026tatic} interpreting brief physical corrections as both task-level intent and motion-level guidance. Recent methods further address dynamic or underspecified goals: I've Changed My Mind~\cite{ghose2025ve} adapts to changing human objectives, BALI~\cite{ghose2026open} jointly reasons over actions and language for open-ended goal inference, and NIABench~\cite{zhang2026assistance} studies when and how robots should proactively assist without interrupting ongoing human activity.

\noindent \textbf{Preference and Uncertainty Alignment.}
Beyond immediate corrections, effective collaboration requires robots to infer persistent human preferences and recognize when autonomous decisions are unreliable. Jain et al.~\cite{jain2015learning} establish an early framework for manipulation preference learning from online coactive feedback, allowing users to iteratively improve robot trajectories without providing explicit reward functions. KnowNo~\cite{ren2023robots} calibrates uncertainty in language-based robot planners and requests human assistance when ambiguity is high, while Text2Interaction~\cite{thumm2024text2interaction} translates natural-language preferences into task, motion, and safety constraints. RAPL~\cite{tian2024maximizing} learns visual rewards from sparse human preferences to align pretrained visuomotor policies with behavior. M2HRI~\cite{hasan2026m2hri} incorporates multimodal interaction and long-term memory, extending alignment from task-level preferences toward persistent human-centered adaptation.

\noindent \textbf{Learning Collaborative Behaviors.}
Beyond interpreting human feedback, robots must learn policies whose behavior is inherently coupled with other agents. GenH2R~\cite{wang2024genh2r} develops scalable simulation and imitation learning for generalizable human-to-robot handover, while generative simulation for physical HRI~\cite{wang2026generative} automatically synthesizes diverse human--robot interaction scenarios for policy training and sim-to-real transfer. Collaboration also extends to multiple robotic agents: Sequential Asymmetric Imitation~\cite{chen2026robots} learns physically coupled robot policies through staged imitation rather than synchronized multi-operator demonstrations. Together, these directions indicate a transition from robots that merely respond to explicit instructions toward collaborative agents that continuously infer human state, adapt autonomy, incorporate feedback, and coordinate their actions with humans and other robots.

\section{Applications}
\label{sec:applications}

\begin{figure}[!t]
\centering
\includegraphics[width=0.85\linewidth]{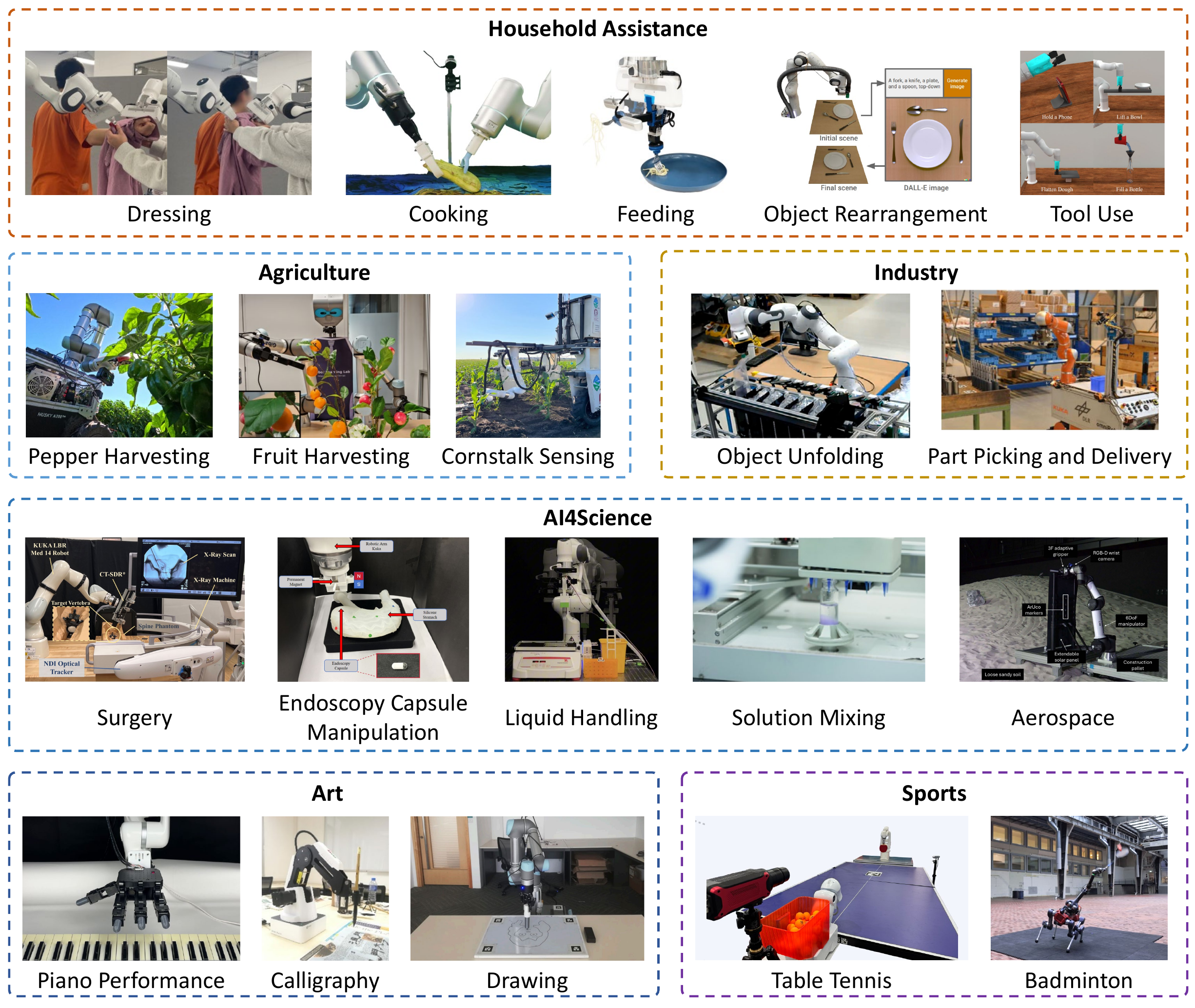}  
\caption{Overview of robotic manipulation applications across diverse domains. The figure is adapted from representative works in household assistance~\cite{zhao2025bimanual, liu2025forcemimic, sundaresan2023learning, chen2023predicting, kapelyukh2023dall, tang2025functo}, agriculture~\cite{kim2025autonomous, li2025enhanced, guri2024hefty}, industry~\cite{adetunji2025vision, domel2017toward, buvs2024deepcraft}, AI4Science~\cite{maroufi2025s3d, he2025capsdt, angers2025roboculture, mishra2025enhancing}, art~\cite{wang2023style, zeulner2025learning, wang2020robocodraw}, and sports~\cite{ma2025learning, wang2026spikepingpong}.
}
\label{fig: case_app}
\end{figure}

Robotics research plays a central role in advancing intelligent systems with tangible real-world impact. Over the past decades, robots have evolved from rigid, hard-coded tools executing predefined routines into perceptive agents capable of interpreting and interacting with their environments, with vision as a primary modality. Today, empowered by large language models and multimodal foundation models, the field is rapidly progressing toward embodied intelligence, where robots integrate perception, planning, and control to accomplish complex tasks in dynamic and uncertain settings~\cite{zhang2026embodied, firoozi2025foundation}. This paradigm shift requires not only precise mechanical design but also robust and generalizable decision-making that enables robots to adapt flexibly across diverse tasks and user instructions. These advances are fueling applications in domestic, healthcare, industrial, and scientific domains. To illustrate these developments, Figure~\ref{fig: case_app} presents representative application domains and typical tasks across sectors. The remainder of this section focuses on these application scenarios, outlining how robots are being deployed in practice across different fields.

\subsection{Household Assistance}
Robotic manipulation in household environments targets essential daily activities such as dressing, cooking, feeding, assisting, stowing, object rearrangement, and tool use. Dressing assistance has been studied through bimanual strategies and safe human-robot interaction with deformable garments~\cite{zhao2025bimanual,liu2025forcemimic,erickson2019multidimensional}. Cooking and feeding tasks emphasize contact-rich manipulation and adaptive policies for varied food types and configurations~\cite{liu2025forcemimic,liu2024adaptive,sundaresan2023learning}. Assistive care applications explore shared control, personalized physical interaction, and soft manipulators for elderly support~\cite{fu2025tasc,madan2025prioritouch,ansari2017towards}. In particular, PrioriTouch adapts whole-arm physical interaction to individual user contact preferences, highlighting the growing importance of personalized assistance. Stowing and object rearrangement tasks leverage behavior primitives, language-guided planning, and vision-language models to enable flexible scene organization~\cite{chen2023predicting,jin2025paca,chang2024lgmcts,ding2023task,kapelyukh2023dall}. Finally, tool manipulation has emerged as a critical subdomain, where function-centric imitation, language grounding, and generative design support skill transfer across diverse household tasks~\cite{lin2025robotsmith,tang2025functo,chen2025tool,ren2023leveraging}.

\subsection{Agriculture}
Agriculture is an important yet highly challenging application domain for robotic manipulation, where robots must operate in unstructured environments characterized by dense canopies, variable lighting, irregular crop shapes, and the need for delicate handling. Recent works demonstrate progress across multiple crop types and farming conditions. For instance, robotic systems have been deployed for fruit and pepper harvesting in outdoor fields, showing the feasibility of autonomous harvesting under occlusion and cluttered backgrounds~\cite{subedi2025find,kim2025autonomous,li2025enhanced}. Other efforts address the safe manipulation of fragile bio-products, such as irregular poultry carcasses, by integrating customized grippers with learned control strategies~\cite{davar2025chicgrasp}. Beyond harvesting, new controllers have been designed for navigating dense plant canopies and adapting to complex interactions with foliage~\cite{parayil2026rice}. Together, these studies highlight the potential of robotics to enhance productivity, safety, and precision in agriculture, paving the way toward scalable automation in one of the most labor-intensive sectors.

\subsection{Industry}
In industrial contexts, robotic manipulation plays a crucial role in automating complex and large-scale processes such as assembly, logistics, and material handling. Early efforts emphasized enabling autonomous mobile manipulation in structured factory settings, laying the foundation for robots to navigate and interact in dynamic production environments~\cite{domel2017toward}. To support system design and deployment, modeling frameworks such as RobotML have been developed for simulating and validating industrial manipulation scenarios~\cite{kchir2016robotml}. More recent advances demonstrate how intelligent systems can enhance manufacturing flexibility, with reinforcement learning enabling adaptive assembly strategies that improve efficiency and reduce manual intervention~\cite{li2022flexible}. At the same time, specialized applications have emerged to address challenging settings, such as vision-based manipulation of transparent plastic bags, which are common in industrial packaging and logistics but difficult for traditional perception systems~\cite{adetunji2025vision}. Collectively, these applications illustrate how robotic manipulation is transitioning from fixed, rigid automation pipelines toward more adaptive, perception-driven systems capable of handling diverse tasks in real-world industrial environments.
It should be noted, however, that most learning-driven approaches are still evaluated in simulation or laboratory setups rather than on fully deployed factory lines, as industrial deployment demands strict reliability and safety, where rule-based systems remain the prevailing choice.

\subsection{AI4Science}
Robotic manipulation is increasingly being applied as a powerful enabler for scientific discovery, where precision, repeatability, and autonomy are critical. In medical science, robotic systems are advancing surgical assistance and minimally invasive procedures, ranging from autonomous intubation~\cite{tian2025learning}, capsule robot navigation~\cite{he2025capsdt}, and robotic spinal fixation~\cite{maroufi2025s3d}, to language-conditioned surgical planning~\cite{kim2025surgical, kim2025srt} and high-fidelity training simulators such as SonoGym for ultrasound-guided surgery~\cite{ao2025sonogym}. Beyond medicine, robotics also supports biology and life sciences, with platforms like RoboCulture~\cite{angers2025roboculture} enabling automated biological experimentation and robotic micromanipulation systems providing real-time spatiotemporal assistance in microscopy~\cite{mori2024real}.
In chemistry, the recently proposed robotic AI chemist~\cite{song2025multiagent} demonstrates how multi-agent robotic systems can autonomously conduct chemical experiments on demand, accelerating discovery by integrating automation with intelligent decision-making.
In aerospace, manipulation technologies extend to high-stakes environments, such as space assembly and in-orbit trajectory learning for robotic arms~\cite{mishra2025enhancing, shyam2020imitation}. More broadly, simulation frameworks like LabUtopia~\cite{li2025labutopia} establish standardized environments for training and benchmarking embodied scientific agents. Collectively, these applications highlight how robotics is becoming an essential tool in AI4Science, accelerating progress in medicine, biology, aerospace, and beyond.

\subsection{Art}
Robotic manipulation has also found applications in the artistic domain, where tasks such as music performance, calligraphy, drawing, and co-creative design showcase the expressive and demonstrative potential of embodied agents. In music, RP1M provides a large-scale motion dataset enabling robots to perform piano playing with bi-manual dexterous hands, highlighting how robotic systems can be trained to execute highly coordinated and nuanced actions beyond traditional industrial tasks~\cite{zhao2025rp1m}. In visual art, robotic calligraphy demonstrates how imitation learning can support the reproduction of culturally significant practices by planning precise brush trajectories in three-dimensional space~\cite{xie2023end, wang2023style}, while RoboCoDraw integrates GAN-based style transfer with time-efficient path optimization to allow robots to generate stylized portrait drawings~\cite{wang2020robocodraw}. Beyond performance, robotics has also been integrated into design and architecture through co-intelligent processes, where robots learn to translate design intent into executable construction paths, expanding their role as creative collaborators~\cite{buvs2024deepcraft}. Together, these studies illustrate how robotics can extend from functional manipulation to artistic expression, providing both a testbed for fine-grained motor control and a platform for human–robot collaboration in creative fields.

\subsection{Sports}
Robotic manipulation has also entered the domain of sports, where dynamic, high-speed interactions demand precise perception–action coordination. Recent advances demonstrate robots learning to perform complex athletic skills, such as legged manipulators executing coordinated badminton movements that couple locomotion with dexterous striking~\cite{ma2025learning}. In table tennis, robots have been trained to handle fast-paced rallies, from early efforts in sample-efficient reinforcement learning for controlled ball exchanges~\cite{tebbe2021sample} to more recent systems like SpikePingpong, which leverages spike-based vision sensors for real-time, high-frequency striking with enhanced precision~\cite{wang2026spikepingpong}. These applications not only showcase robotics as a platform for pushing the limits of real-time control and multimodal perception but also offer new opportunities for human–robot interaction, training, and performance augmentation in competitive sports.

\section{Prospective Future Research Directions}

To advance robotic manipulation from controlled laboratory settings to open, dynamic, and real-world environments, the long-term goal is to build increasingly autonomous embodied systems that can perceive, reason, act, and improve through interaction. Achieving this goal requires more than scaling individual perception or control modules. Future robots must generalize across embodiments, learn from data far beyond conventional robot demonstrations, predict the consequences of their actions, continually acquire new capabilities, and remain reliable during complex physical interaction. We highlight five closely connected directions toward this goal: general-purpose and self-evolving robot brains, scalable embodied data, predictive world models, multimodal physical interaction, and safe autonomous operation.

\subsection{Building a General-Purpose and Self-Evolving Robot Brain}

\begin{quote}
\itshape
-- Core Challenge 1: From ``One Brain, Multiple Embodiments'' to Autonomous and Self-Evolving Robotic Agents.
\end{quote}

Robotics is gradually shifting from the paradigm of `one model per task'' toward general-purpose foundation models capable of supporting many tasks and embodiments. Beyond simply increasing model scale, the more ambitious objective is to build a reusable robotic `brain'' that can reason over complex tasks, coordinate specialized capabilities, adapt to different bodies, and continuously improve after deployment.

\noindent \textbf{General-Purpose Architecture Across Embodiments.}
A universal robotic model must operate across heterogeneous observation and action spaces, ranging from cameras with different viewpoints and resolutions to systems with or without tactile sensing, and from joint-level control of fixed manipulators to whole-body coordination of mobile and humanoid robots. This diversity makes direct parameter sharing difficult because the same high-level intention may correspond to substantially different low-level actions across embodiments. Future architectures therefore need to separate transferable task knowledge from embodiment-specific execution through morphology-aware representations, modular action interfaces, embodiment-conditioned policies, or shared latent action spaces. The ultimate objective is not merely to train one large policy on many robots, but to enable knowledge acquired on one embodiment to accelerate learning and execution on others.

\noindent \textbf{Agentic Planning and Long-Horizon Autonomy.}
As manipulation tasks become longer and more compositional, directly mapping an instruction to a fixed sequence of actions becomes increasingly fragile. Everyday activities such as ``making a cup of coffee'' may require task decomposition, tool selection, intermediate state tracking, recovery from unexpected outcomes, and coordination among multiple skills. Future robotic systems are therefore likely to evolve from monolithic policies toward agentic architectures that can reason over task progress, invoke specialized perception or manipulation skills, interact with external tools or memory, monitor execution, and replan when necessary. In this view, a robot is no longer merely an action generator but an embodied agent that actively manages the entire perception--reasoning--execution loop.

\noindent \textbf{Continual and Self-Evolving Learning.}
A truly autonomous robot should not remain fixed after deployment. Beyond preventing catastrophic forgetting when acquiring new skills, robots should achieve positive transfer, where new experience improves existing capabilities and enriches their understanding of the physical world. More importantly, future systems should be able to identify informative successes and failures during operation, convert them into reusable experience, and selectively update memories, skills, data, or model parameters. This creates a self-evolving loop of interaction, evaluation, learning, and redeployment. Active exploration, autonomous skill discovery, memory-based adaptation, and large-scale parallel experience collection may all contribute to such systems. A major challenge, however, is ensuring that self-improvement accumulates reliable knowledge rather than amplifying model errors or undesirable behaviors.

\noindent \textbf{Stable and Adaptive Motion Generation.}
High-level intelligence must ultimately be translated into physically feasible motion. Advanced policies should therefore generate smooth and dynamically consistent actions while adapting online to contact, disturbances, and embodiment-specific constraints. Rather than treating learned action generation and classical control as competing paradigms, future systems may increasingly combine learned policies with impedance control, model predictive control, or other feedback mechanisms. Such integration can provide the flexibility required for open-world manipulation while retaining the stability and compliance needed for precise and safety-critical physical interaction.

\noindent \textbf{Predictive World Models and Internal Simulation.}
A general-purpose robot brain should not only react to current observations but also anticipate how the environment may evolve under different actions. World models provide such predictive capability by learning compact representations of objects, spatial relations, robot states, and task progress, together with their action-conditioned dynamics. The key challenge is to move beyond visually plausible future prediction toward physically grounded models that capture contact, force transmission, object permanence, and causal interactions, particularly for deformable, articulated, and partially observed environments. More importantly, these predictions should directly support decision-making by enabling robots to compare possible futures, evaluate action consequences, and select plans before execution. Such internal simulation could bridge reactive policy learning with deliberate planning and provide a foundation for continual policy refinement through imagined experience.

\subsection{Scaling Robot Intelligence Beyond Robot Data}

\begin{quote}
\itshape
-- Core Challenge 2: Breaking the Robot-Data Bottleneck through Heterogeneous Embodied Experience.
\end{quote}

Modern robot learning is increasingly data-driven, yet high-quality robot interaction remains expensive to collect and difficult to scale. Compared with the vast corpora available to language and vision models, existing robot datasets cover only a small fraction of the diversity of objects, environments, embodiments, and interaction patterns encountered in the real world. Addressing this limitation requires both more efficient use of robot-generated experience and systematic exploitation of data that were not originally collected by robots.

\noindent \textbf{From Data Collection to a Data Flywheel.}
Real-world robot datasets remain fragmented across hardware platforms, action spaces, sensor configurations, and collection protocols. Their quality is also highly variable: successful demonstrations, suboptimal trajectories, exploratory behaviors, and outright failures may coexist within the same dataset. Simply increasing dataset size can therefore provide diminishing returns if low-value or redundant samples dominate training. Future research should move from passive data accumulation toward an efficient data flywheel in which current models help identify missing capabilities, collect informative experience, evaluate trajectory quality, and prioritize data for subsequent training. Data filtering, automatic annotation, uncertainty estimation, trajectory valuation, and failure mining will become increasingly important for extracting useful supervision from large-scale interaction logs.

\noindent \textbf{Learning from Robot-Free and Weakly Embodied Data.}
An especially promising direction is to expand robotic learning beyond conventional teleoperated robot demonstrations. Egocentric human videos contain rich information about object interaction, temporal structure, and task intent; Internet videos capture enormous diversity in objects and activities; portable interfaces such as UMI reduce the cost of collecting manipulation demonstrations without requiring a complete robot platform; and human motion or hand-object interaction datasets provide additional priors about physical behavior. These sources dramatically expand the potential scale of embodied experience, but they do not directly provide robot-compatible actions. The key challenge is therefore to recover action-relevant physical knowledge despite differences in embodiment, viewpoint, kinematics, and control interfaces. Cross-embodiment alignment, action inference, retargeting, and shared latent representations will be essential for transforming abundant human-centered data into useful robot supervision.

\noindent \textbf{Simulation, Synthetic Experience, and Sim-to-Real Transfer.}
Simulation remains another important source of scalable experience because it enables controlled variation, automatic annotation, and inexpensive exploration. However, the gap between simulated and real interaction remains particularly severe for contact-rich manipulation, where friction, compliance, collision, deformation, and sensor noise are difficult to reproduce accurately. Future work should therefore pursue both higher-fidelity simulation and methods that explicitly reduce dependence on perfect simulation, including domain randomization, system identification, adaptive sim-to-real transfer, generative environment synthesis, and hybrid training with small amounts of real-world experience. Differentiable simulation and learned simulators may further provide efficient mechanisms for optimizing behaviors and physical parameters. The broader objective is to combine real robot data, human-centered data, and synthetic experience rather than relying on any single source.

\subsection{Multimodal and Contact-Rich Physical Interaction}

\begin{quote}
\itshape
-- Core Challenge 4: From Visual Intelligence to Deep Physical Interaction.
\end{quote}

Robots interact with the world through a perception--action loop involving vision, proprioception, touch, force, audition, and other sensory signals. Although vision provides rich semantic and geometric information, many manipulation states cannot be reliably inferred from appearance alone. True physical intelligence therefore requires robots to reason from multimodal feedback and adapt their actions during interaction.

\noindent \textbf{Fusing a Broader Spectrum of Sensory Modalities.}
Consider how a human can locate a key and unlock a door in darkness using touch, or determine whether two parts have snapped together through force and sound. Future robots will similarly need to exploit tactile sensing to estimate contact location, texture, slip, and pressure; force sensing to regulate interaction; proprioception to track body configuration and dynamics; and audio to detect events that may be difficult to observe visually. The main challenge is not simply adding more sensors, but learning representations that align heterogeneous, asynchronous, and multi-rate signals while preserving information relevant to action. Effective multimodal fusion should also remain robust when individual modalities become noisy, delayed, or temporarily unavailable.

\noindent \textbf{Contact-Rich and Complex Manipulation.}
Many practically important tasks are dominated by contact rather than free-space motion. Insertion, wiping, polishing, assembly, tool use, cloth folding, cable organization, and fluid manipulation require continuous adaptation to interaction forces and uncertain object states. Small errors in geometry or force can rapidly turn into task failure. Future research therefore needs policies that jointly reason about geometry, contact, compliance, and dynamics rather than treating motion as purely kinematic trajectory generation. Combining multimodal perception with predictive physical models and adaptive feedback control will be particularly important for extending robotic manipulation from structured pick-and-place tasks to richer forms of physical interaction.

\subsection{Safety, Recovery, and Collaborative Autonomy}

\begin{quote}
\itshape
-- Core Challenge 5: Making Autonomous Manipulation Reliable in the Open World.
\end{quote}

As robots leave controlled laboratory environments and increasingly operate around humans, other robots, and valuable objects, reliability becomes as important as task success. Safety should therefore not be treated as an independent post-processing layer, but as a property that spans perception, decision-making, control, and learning.

\noindent \textbf{Intrinsic Safety and Self-Constrained Control.}
Future robots must protect both their surroundings and themselves. Excessive joint velocities, abrupt accelerations, unstable force outputs, or repeated operation near mechanical limits can damage hardware or create unsafe interaction. Next-generation systems should therefore explicitly reason about kinematic, dynamic, force, energy, and workspace constraints during action generation. Learning-based policies can provide adaptability, while classical mechanisms such as model predictive control, impedance control, barrier functions, and rule-based safety constraints can offer predictable safeguards. Hybrid architectures that combine these complementary strengths are likely to remain important for safety-critical deployment.

\noindent \textbf{Autonomous Fault Detection and Recovery.}
Real-world manipulation inevitably encounters failures, including missed grasps, unexpected contacts, sensor anomalies, actuator degradation, planning loops, and environmental changes. A robust system should therefore behave less like an open-loop executor and more like an organism with an ``immune system'': continuously monitoring its own state, recognizing anomalous outcomes, diagnosing likely causes, and selecting an appropriate recovery strategy. Depending on the situation, recovery may involve local replanning, retrying with a modified strategy, retreating to a safe configuration, switching to another skill, or requesting human assistance. Integrating anomaly detection, uncertainty estimation, causal diagnosis, and agentic replanning will be central to such capabilities.

\noindent \textbf{Human--Robot and Inter-Robot Collaboration.}
Future robots will increasingly operate as collaborative agents rather than isolated machines. Effective human--robot interaction requires understanding not only explicit language commands but also contextual cues such as human motion, gaze, object interaction, and task progress. A robot assisting with assembly, for example, should be able to infer when to provide a tool, stabilize an object, or yield control without requiring detailed instructions at every step. Similar challenges arise in multi-robot settings, where robots must share workspaces, anticipate one another's motion, negotiate task allocation, and avoid interference. Achieving such collaboration will require predictive interaction models, shared representations of task state, and explicit mechanisms for uncertainty, communication, and responsibility.

Ultimately, progress across these directions is tightly coupled. Diverse robot and human-centered data provide the experience from which general-purpose models can learn; world models provide predictive understanding of physical consequences; agentic systems organize reasoning, skills, and recovery over long horizons; multimodal sensing grounds these capabilities in physical interaction; and safety mechanisms constrain autonomous behavior during real-world deployment. The convergence of these components may gradually transform robotic manipulation from task-specific policy execution into continuously learning, predictive, and autonomous embodied intelligence.

\section{Conclusion}

This survey provides a comprehensive and systematic overview of robot manipulation, covering fundamental background knowledge, task-specific benchmarks, representative methods, critical bottlenecks, and real-world applications. Despite substantial progress, robotic manipulation remains far from achieving human-level versatility. Major open challenges persist, including the development of a unified “robot brain,” the resolution of data and perception bottlenecks, and the assurance of safety in human–robot collaboration. Bridging these gaps is essential for enabling learning-based robotic systems to move beyond controlled laboratory environments and into everyday life and diverse industries. We hope this survey will serve as both a roadmap for newcomers and a comprehensive reference for experienced researchers, fostering a unified understanding of robotic manipulation and inspiring future advances in embodied intelligence.

\section*{Acknowledgments}
This work was supported by the National Natural Science Foundation of China (Grant No. U21A20485).

\section*{Author Contributions}
\addcontentsline{toc}{section}{Author Contributions}
The contributions of all participating authors are summarized below, indicating the primary author responsible for each section and the supporting contributors. The detailed roles of each author are as follows:

{\setstretch{1.3} 
\begin{itemize}
    \item \textbf{Corresponding Authors:} Shanghang Zhang, Badong Chen
    \item \textbf{Project Lead:} Shuanghao Bai
    \item \textbf{Background:} Shuanghao Bai, Han Zhao
    \item \textbf{Benchmarks and Datasets:} Shuanghao Bai
    \item \textbf{Manipulation Tasks:} Shuanghao Bai
    \item \textbf{High-level Planning:} Zhide Zhong, Wei Zhao
    \item \textbf{Low-level Learning-based Action Modeling:}
        \begin{itemize}
            \item Learning Strategy: 
                \begin{itemize}
                     \item Reinforcement Learning: Han Zhao
                     \item All other subsections: Shuanghao Bai 
                \end{itemize}
            \item Input Modeling: 
                \begin{itemize}
                     \item Vision Action Models: Zhe Li
                     \item Vision-Language-Action Models: Yuheng Ji, Wenxuan Song, Jiayi Chen
                     \item Tactile-based Action Models: Wenxuan Song
                \end{itemize}
            \item Latent Learning: Wenxuan Song
            \item Policy Learning: Wenxuan Song, Jiayi Chen
        \end{itemize}
    \item \textbf{Challenges and Bottlenecks:} Jiayi Chen, Jin Yang
    \item \textbf{Applications:} Wanqi Zhou
    \item \textbf{Prospective Future Research Directions:} Pengxiang Ding, Shuanghao Bai
    \item \textbf{Other Contributions:}
        \begin{itemize}
            \item Figures: Shuanghao Bai, Wenxuan Song, Jiayi Chen, Yuheng Ji, Zhide Zhong, Jin Yang, Wanqi Zhou
            \item Review and Editing: Cheng Chi, Haoang Li, Chang Xu, Xiaolong Zheng, Donglin Wang, Shanghang Zhang, Badong Chen
        \end{itemize}
\end{itemize}
}

The second version of this survey was independently developed and substantially revised by Shuanghao Bai.

In addition, we distilled the core content and methodological framework of this survey into a concise version~\cite{bai2025embodied}.

We also welcome constructive feedback and suggestions from the broader research community, which will help us further improve this work and enhance its value as a resource for the field.

\bibliography{
refs/1-intro,
refs/2-hardware, refs/2-control_and_learning, refs/2-robotics_models,
refs/3-benchmarks,
refs/4-manipulation_tasks,
refs/5-high-level_planner, refs/5-affordance, refs/5-3d_representation, refs/5-video,
refs/6-rl, refs/6-il, refs/6-rl_il, refs/6-learn_with_auxilary_tasks, refs/6-vla, refs/6-va, refs/6-latent_learning, refs/6-policy_learning, refs/6-tactile_action,
refs/7-data, refs/7-generalization, refs/7-agent, refs/7-human-robot,
refs/8-application,
refs/9-future_research
}

\end{document}